\documentclass[letterpaper,12pt]{report}	% The documentclass must be ``report''.

\usepackage{styles/utformat}  		% Package style file for UT ETDs.

\usepackage{pdflscape}
\usepackage{layout}
\usepackage{graphicx}
\usepackage{amsmath}
\usepackage{verbatim}      	% Allows quoting source with commands.
\usepackage{styles/citesort}         	%
\usepackage{algorithm}
\usepackage{bibentry}
\usepackage{algorithmic}
\usepackage{xcolor}
\usepackage{markdown}
\usepackage{booktabs}
\usepackage{url}

\usepackage{multirow}
\usepackage{makecell}
\usepackage{tabularx}
\usepackage{bbm}% Allows good typesetting of web URLs.
\usepackage{mathtools}
\usepackage{xcolor}        % For colors
\usepackage{framed}        % For framing
\usepackage{changepage} 
\usepackage{listings}

\oneandonehalfspacing
\newcommand{\latexe}{{\LaTeX\kern.125em2%
                      \lower.5ex\hbox{$\varepsilon$}}}

\chardef\bslash=`\\	% \bslash makes a backslash (in tt fonts)
\makeatletter		% Starts section where @ is considered a letter
\def\square{\RIfM@\bgroup\else$\bgroup\aftergroup$\fi
  \vcenter{\hrule\hbox{\vrule\@height.6em\kern.6em\vrule}%
                                              \hrule}\egroup}
\makeatother		% Ends sections where @ is considered a letter.
\usepackage{tocloft}

\author{Yifeng Zhu}  	% Required

\address{yifeng.zhu@utexas.edu}  % Required

\title{Efficient Sensorimotor Learning For \\ Open-world Robot Manipulation} % Required

\supervisor[Yuke Zhu]
	{Peter Stone}

\committeemembers
	[Joydeep Biswas]
	[Shuran Song]
        {} 

\degree{Doctor of Philosophy}
\degreeabbr{PhD} 

\dissertation 

\usepackage[english]{babel}

\usepackage[left=3cm,right=3cm]{geometry}  % Adjust margin size as needed

\usepackage{amsmath, amssymb}
\usepackage{dsfont}
\usepackage{graphicx}
\usepackage{multirow}
\usepackage{booktabs}
\usepackage{tcolorbox}
\usepackage{subcaption}
\usepackage{tikz}
\usepackage{url}
\usepackage{wrapfig}
\usepackage{booktabs, tabularx, colortbl} 

\usepackage{chngcntr}
\usepackage{xcolor} % For defining custom colors
\definecolor{mycustomcolor}{RGB}{255,127,0} % Example: Orange
\usepackage[colorlinks=true, allcolors=mycustomcolor]{hyperref}
\usepackage[numbers]{natbib} % For bibliography

\newcommand{\highlight}[1]{\textcolor{magenta}{\textbf{#1}}}

\newcommand{\viola}{\textsc{VIOLA}}
\newcommand{\groot}{\textsc{GROOT}}
\newcommand{\buds}{\textsc{BUDS}}
\newcommand{\lotus}{\textsc{LOTUS}}

\newcommand{\orion}{\textsc{ORION}}
\newcommand{\okami}{\textsc{OKAMI}}

\newcommand{\robosuite}{\textsc{Robosuite}}
\newcommand{\libero}{\textsc{Libero}}

\newcommand{\bc}{BC}
\newcommand{\oreo}{OREO}
\newcommand{\bcrnn}{BC-RNN}
\newcommand{\patch}{\viola{}-Patch}

\newcommand{\mae}{\textsc{MAE-Policy}}

\newcommand{\lldm}{\textsc{LLDM}}
\newcommand{\liberohundred}{\textsc{LIBERO-100}}

\newcommand{\liberospatial}{\textsc{LIBERO-Spatial}}
\newcommand{\liberoobject}{\textsc{LIBERO-Object}}

\newcommand{\liberogoal}{\textsc{LIBERO-Goal}}
\newcommand{\liberofifty}{\textsc{LIBERO-50}}

\newcommand{\loosepar}{\looseness=-1}

\newcommand{\multitaskkitchen}{\texttt{Multitask-Kitchen}}

\newcommand{\bg}{\text{bg}}
\newcommand{\cam}{\text{cam}}
\newcommand{\obj}{\text{obj}}
\newcommand{\spatial}{\text{spatial}}

\newcommand{\er}{\textsc{ER}}
\newcommand{\mter}{\textsc{ER}}
\newcommand{\mtft}{\textsc{Sequential}}

\newcommand{\emphasize}[1]{\textbf{#1}}

\newcommand{\toy}{\texttt{Plush-toy-in-basket}}
\newcommand{\salt}{\texttt{Sprinkle-salt}}
\newcommand{\drawer}{\texttt{Close-the-drawer}}
\newcommand{\laptop}{\texttt{Close-the-laptop}}
\newcommand{\snacks}{\texttt{Place-snacks-on-plate}}

\newcommand{\sort}{\texttt{Sorting}}
\newcommand{\stack}{\texttt{Stacking}}
\newcommand{\kitchen}{\texttt{Kitchen}}

\newcommand{\platefork}{\texttt{Dining-PlateFork}}
\newcommand{\bowl}{\texttt{Dining-Bowl}}
\newcommand{\coffee}{\texttt{Make-Coffee}}

\newcommand{\canonical}{\texttt{Canonical}}
\newcommand{\placement}{\texttt{Placement}}

\newcommand{\distractionseasy}{\texttt{Background(Easy)}}
\newcommand{\distractionshard}{\texttt{Background(Hard)}}
\newcommand{\cameraeasy}{\texttt{Camera(Easy)}}
\newcommand{\camerahard}{\texttt{Camera(Hard)}}
\newcommand{\camera}{\texttt{Camera-Shift}}
\newcommand{\newobject}{\texttt{New-Object}}
\newcommand{\distracting}{\texttt{Background-Change}}

\newcommand{\tooluse}{\texttt{Tool-Use}}
\newcommand{\hammer}{\texttt{Hammer-Place}}
\newcommand{\multitask}{\texttt{Multitask-Kitchen}}
\newcommand{\budsrealrobot}{\texttt{Real-Kitchen}}
\newcommand{\budstrainmulti}{\textbf{Train (Multi)}}
\newcommand{\budstrainsingle}{\textbf{Train (Single)}}
\newcommand{\budstest}{\textbf{Test}}

\newcommand{\task}[1]{\texttt{Task-{#1}}}
\newcommand{\variant}[1]{\texttt{Variant-{#1}}}

\newcommand{\mutex}{\textsc{MUTEX}}

\newcommand{\target}{\text{target}}
\newcommand{\reference}{\text{ref}}

\newcommand{\mugcoaster}{\texttt{Mug-on-coaster}}
\newcommand{\simpleboat}{\texttt{Simple-boat-assembly}}
\newcommand{\chip}{\texttt{Chips-on-plate}}
\newcommand{\llama}{\texttt{Succulents-in-llama-vase}}
\newcommand{\rearrange}{\texttt{Rearrange-mug-box}}
\newcommand{\complexboat}{\texttt{Complex-boat-assembly}}
\newcommand{\breakfast}{\texttt{Prepare-breakfast}}

\newcommand{\bagging}{\texttt{Bagging}}

\newcommand{\colorsquare}[4][0.5em]{%
    \tikz[baseline=-0.5ex]{%
        \node[
            fill={rgb,255:red,#2; green,#3; blue,#4},
            inner sep=0pt,
            minimum size=#1,
            anchor=base
        ] {};}%
}

\newcommand{\mS}{\mathcal{S}}
\newcommand{\mA}{\mathcal{A}}
\newcommand{\mP}{\mathcal{P}}
\newcommand{\mR}{R}

\newcommand{\initstate}{\rho_{0}}
\newcommand{\initstatefunc}{\rho}

\newcommand{\mC}{\mathcal{C}}

\newcommand{\context}{c}
\newcommand{\ContextSpace}{\mC}
\newcommand{\CMDPMapping}{\phi_{\mathcal{M}}}

\newcommand{\mJ}{J}
\newcommand{\objectnum}{Q}

\newcommand{\feature}{\psi}
\newcommand{\Feature}{\Psi}

\newcommand{\tokenin}{y^{\text{in}}}
\newcommand{\tokenout}{y^{\text{out}}}
\newcommand{\Tokenin}{Y^{\text{in}}}
\newcommand{\Tokenout}{Y^{\text{out}}}

\newcommand{\hcbvar}{*}

\newcommand{\llstep}{l}
\newcommand{\LLTotalStep}{L_{\text{max}}}

\newcommand{\mG}{\mathcal{G}}

\newcommand{\languagegoal}{\context_{\text{lang}}}

\newcommand{\maxtasknum}{M}
\newcommand{\Task}[1]{T^{#1}}

\newcommand{\tasknum}{m}
\newcommand{\param}{\omega}

\newcommand{\skillpolicy}[1]{\pi^{(#1)}_{L}}
\newcommand{\metacontroller}[1]{\pi^{#1}_{H}}
\newcommand{\maxhorizon}{H_{\text{max}}}
\newcommand{\subgoaltime}{\Delta t_{\text{lookahead}}}
\newcommand{\historytime}{\Delta t_{\text{H}}}

\newcommand{\kfindex}{f}
\newcommand{\okamiplan}{\text{RefPlan}}
\newcommand{\KF}{F}

\newcommand{\segmentfinal}{t_{\text{segment}}}

\newcommand{\objectnode}{vo}

\newcommand{\kptraj}{\tau}

\newcommand{\tcp}{X}

\newcommand{\dataset}{D}

\newcommand{\tg}{t_g}

\begin{document}

\copyrightpage  % Produces the copyright page. Optional.

\commcertpage   % Produces the Committee Certification
			%   of Approved Version page (doctoral)
			%   or Signature page (masters).
			%		20 Mar 2002	cwm
                % Required.

\titlepage      % Produces the title page. Required.

\title{Efficient Sensorimotor Learning for \\ Open-world Robot Manipulation}
%\title{Sensorimotor Skissll Learning for Open-World \\ Robot Manipulation}

% Problem -> open-world manipulation
% methodology -> human-like learning (so that it can adapt to the human-centric environment)

% \author{Yifeng Zhu\\
% Department of Computer Science,\\
% The University of Texas at Austin,\\
% Austin, Texas, 78712\\
% yifeng.zhu@utexas.edu
% }
% \date{}
% \maketitle
\pagenumbering{roman}

\utabstract

% \begin{abstract}
In recent years, there has been growing interest in building general-purpose personal robots, driven by the promise that the robots can assist people with a large variety of everyday manipulation tasks. Such robots must adapt their skills to a wide range of completely new scenarios. This dissertation considers Open-world Robot Manipulation, a manipulation problem where a robot must generalize or quickly adapt to new objects, scenes, or tasks for which it has not been pre-programmed or pre-trained. This dissertation tackles the problem using a methodology of efficient sensorimotor learning. 

The key to enabling efficient sensorimotor learning lies in leveraging regular patterns that exist in limited amounts of demonstration data. These patterns, referred to as ``regularity,'' enable the data-efficient learning of generalizable manipulation skills. This dissertation offers a new perspective on formulating manipulation problems through the lens of regularity. Building upon this notion, we introduce three major contributions. First, we introduce methods that endow robots with object-centric priors, allowing them to learn generalizable, closed-loop sensorimotor policies from a small number of teleoperation demonstrations. Second, we introduce methods that constitute robots' spatial understanding, unlocking their ability to imitate manipulation skills from in-the-wild video observations. Last but not least, we introduce methods that enable robots to identify reusable skills from their past experiences, resulting in systems that can continually imitate multiple tasks in a sequential manner.

Altogether, the contributions of this dissertation help lay the groundwork for building general-purpose personal robots that can quickly adapt to new situations or tasks with low-cost data collection and interact easily with humans. By enabling robots to learn and generalize from limited data, this dissertation takes a step toward realizing the vision of intelligent robotic assistants that can be seamlessly integrated into everyday scenarios.

{
  \hypersetup{linkcolor=black}
  \listoffigures
}

{
  \hypersetup{linkcolor=black}
  \listoftables
}

\newpage

{
  \hypersetup{
    colorlinks=true,      % false: boxed links; true: colored links
    linkcolor=black,      % color of internal links
    citecolor=black,      % color of links to bibliography
    filecolor=black,      % color of file links
    urlcolor=black        % color of external links
  }
  \tableofcontents
}
\newpage

\clearpage
\pagenumbering{arabic}

\chapter{Introduction}
\label{chapter:intro}
For millennia, humans have dreamt of autonomous beings that assist with mundane daily tasks, a long-lasting theme from ancient Greek myths of self-operating automatons to Isaac Asimov's futuristic robots handling laborious duties. These stories reflect the innate desire of humans to create intelligent and autonomous robots that seamlessly integrate into our lives, alleviating the physical and cognitive burdens of everyday chores. The invention of the first programmable robot arm, Unimate, in the mid-20th century, marked a pivotal moment in human’s quest for automation. The presence of Unimate~\cite{UnimateUrL}, along with the first programmable computer, ENIAC~\cite{EniacUrl}, has fueled dreams of building general-purpose personal robots capable of accomplishing a large variety of tasks. Since then, people have envisioned a tangible future where personal robots carry out tasks from cleaning and cooking all the way to nursing home services, addressing the labor shortages worsened by aging populations. Nowadays, while the descendants of ENIAC, like personal computers and smartphones, have become ubiquitous, personal robots for household tasks like robot butlers remain out of reach.

The missing presence of general-purpose robots in our everyday environments is primarily rooted in a fundamental challenge: enabling robots to accomplish a large variety of manipulation tasks in open-world settings. Much research has studied manipulation under ``closed-world'' settings using pre-programmed solutions, where scenarios are constrained to those with known object geometries and locations, clean backgrounds, and predefined sets of tasks. However, this closed-world setting contrasts starkly with the reality where personal robots must operate. The endless variety of situations presents the complexities of robots' real-world deployment that are beyond the reach of existing approaches. 

To deploy robots in the real world, manipulators should be equipped with diverse sensorimotor skills that operate in open-world settings. Sensorimotor skills refer to computational programs that process sensory inputs and synthesize motor commands for purposeful physical movements like completing certain goals in manipulation tasks. In this setting, it is ideal for robots to quickly adapt their sensorimotor skills to new \textit{objects}, \textit{scenes}, or \textit{tasks} for which they haven't been pre-programmed or pre-trained, a problem we term \textbf{Open-world Robot Manipulation}. We use Open-world Robot Manipulation as a systematic framework for studying the fundamental challenges of deploying personal robots in our daily lives. Recent work has shown robots can operate in open-world settings by directly using pre-trained Large Language Models (LLMs) or Visual Language Models (VLMs)~\cite{ahn2022can, liang2023code, liu2024moka, nasiriany2024pivot} for decision making. However, these works abstract away the mapping from inputs to robot actions using predefined primitives, limiting the motion expressiveness that is critical to synthesizing diverse contact-rich behaviors. An alternative path explores training large models using vast amounts of data collected on robots~\cite{padalkar2023open, khazatsky2024droid}. Though it is intriguing that this approach tries to reproduce the scaling laws in training Large Language/Vision Models, robotics tends to operate in a low-data regime, making such a methodology suffer from high costs in hardware and operation, as well as substantial overhead for data re-collection upon any major hardware changes. Thus, scaling up data collection in robot manipulation induces high cost in the face of the countless situations inherent in Open-world Robot Manipulation, encouraging alternatives to tackle the problem. 

To tackle Open-world Robot Manipulation in a cost-effective way, a promising approach is for robots to rapidly acquire sensorimotor skills for new scenarios or tasks using limited, high-quality data from the real world. Efficient learning from limited data is key to developing robots that operate proficiently in everyday environments and can be easily configured by end-users, making the process as intuitive as using today's electronic devices. Robots that learn efficiently would allow users to specify behaviors through easy teleoperation or video demonstrations. This \textit{efficient sensorimotor learning} paradigm holds the promise of building personal robots that can easily learn new skills without significant time and effort from users. Equipped with learning mechanisms that can acquire new tasks efficiently instead of enumerating diverse situations \textit{a priori}, personal robots could become as affordable and widespread as today's PCs and smartphones.

To enable efficient sensorimotor learning in Open-world Robot Manipulation, we focus on developing methods that learn generalizable policies from a small amount of demonstration data (collected through teleoperation or video recording). However, open-world generalization from a few examples is nontrivial: Small demonstration datasets fail to cover sufficient distributions to approximate the countless situations presented in open-world scenarios. To address this challenge, we leverage our key insight: harnessing \textit{regularity} in the physical world that is present in the demonstration data. \textit{Regularity} refers to the physical world's regular and consistent patterns or structures that reveal invariant features across various situations. Regularity is common in the physical world and exploiting them can help robots learn, predict, and act effectively across the many different scenarios robots might encounter during deployment.

This dissertation provides a new perspective on learning-based robot manipulation by systematically tackling problems with the notion of regularity. Regularity has been a well-established concept in cognitive science used to study how humans learn in the physical world and generalize from a small number of examples. The regularities of the physical world are statistical, revealing regular and consistent patterns of phenomena regardless of individual variations, yet humans excel at identifying these regularities and developing intelligent behaviors based on them~\cite{zacks2001event, greff2020binding, lake2017building, spelke2007core,tversky2008motion}. 

This concept of regularity has been less systematically explored in the discipline of learning for robot manipulation. While the community has leveraged ``structures'' or injecting ``model priors'' when developing data-driven approaches, there has not been a systematic way of looking at the problem from the perspective of \textbf{regularity}. For example, objects that a robot perceives respond to actions in a predictable way despite the variations in visual appearance or spatial layouts; behaviors that a robot performs in a task might be reused in another task even if the goal is different. All these common-sense knowledge that seem obvious to humans are regularities in the real world that can afford robots' efficient learning. 

In this dissertation, inspired by cognitive science and recognizing that robots operate in the same physical world as humans, we attempt to wire robots' ``brains'' to leverage regularity so that robots can learn efficiently like humans, just as evolution has wired human brains to detect and extract the regularity of the physical world from sensory inputs across time and space. Building upon regularities as a key principle, we develop a suite of methods that enable robots to tackle open-world manipulation challenges through efficient sensorimotor learning.

In this dissertation, we will identify the regularities that our methods exploit for learning diverse manipulation skills from demonstrations. As we will elaborate in Chapter~\ref{chapter:bg}, our methods center around three critical statistical regularities for generalization in robot manipulation: \textit{object regularity}, \textit{spatial regularity}, and \textit{behavioral regularity}. However, it is challenging to leverage these statistical regularities--—they cannot be expressed in simple formulas or easily described by hand-crafted programs. Instead, capturing such regularities requires a common-sense understanding of semantics and space. Thanks to recent progress in computer vision and natural language processing, we can leverage foundation models for programs to detect and extract these statistical regularities. Foundation models~\cite{bommasani2021opportunities, ouyang2022training,kirillov2023segment, oquab2023dinov2}, a class of deep neural networks that are pre-trained on internet-scale datasets of images and texts, serve as the substrate for concept understanding and extracting semantics from real-world sensory inputs. In this dissertation, a key design principle in developing the methods is to integrate foundation models into workflows such that robots can achieve unprecedented performance and generalization with just a handful of demonstration data.

In summary, this dissertation studies the problem of Open-world Robot Manipulation with a methodology of data-efficient learning. By developing methods to enable efficient sensorimotor learning, we aim to answer the following research question: 

\begin{tcolorbox}[colback=white, colframe=black, boxrule=0.2mm, arc=0.2mm, boxsep=0.5mm]
How can robots exploit regularities in the physical world to efficiently learn generalizable manipulation policies?
\end{tcolorbox}

This dissertation answers the question in the following ways.

\begin{enumerate}
    \item Showing how a robot can learn a generalizable, closed-loop sensorimotor policy from a small number of teleoperation demonstrations;
    \item Showing how a robot can imitate manipulation skills from actionless video observations;
    \item Showing how a robot can continually imitate multiple tasks in a sequential manner.
\end{enumerate}

The contributions in this dissertation are largely based on leveraging the three major regularities in the physical world we identify (More details in Section~\ref{sec:bg:regularity}). These regularities provide priors for developing robot learning algorithms that efficiently learn generalizable policies from a small number of examples.

\section{Dissertation Overview}
\begin{figure}[ht!]
    \centering
    \includegraphics[width=0.8\linewidth]{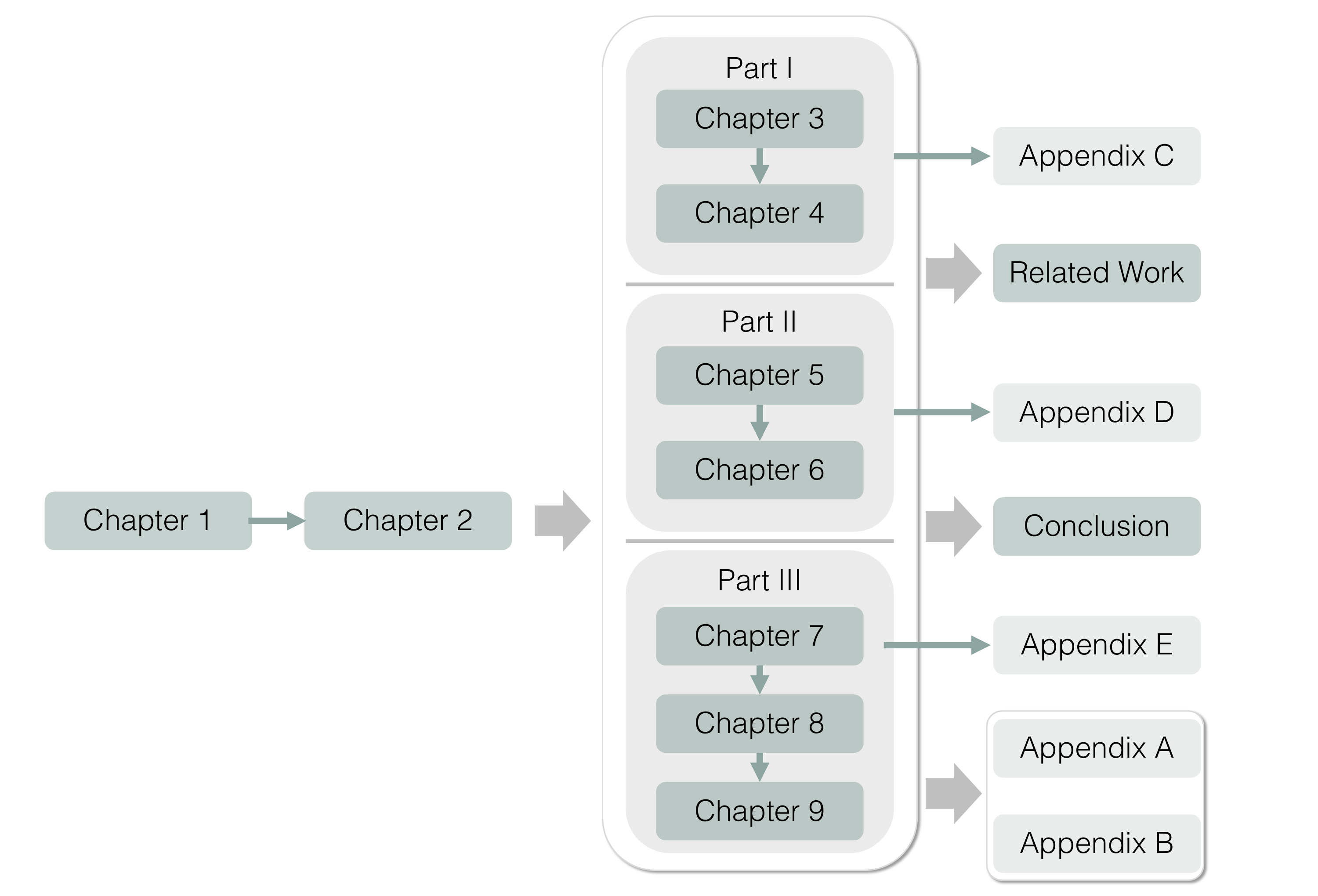}
    \caption[Overview of chapter depencencies.]{Overview of the chapter dependencies. An arrow connection means that one chapter (or a group of chapters) should be read before another. \textbf{The chapters from different parts can be read independently.} Large arrows are used when at least one of the connections is a group of chapters.}
    \label{fig:intro:chapter_dependencies}
\end{figure}

The rest of the dissertation is organized as follows. We first formulate the problem of Open-world Robot Manipulation and introduce necessary background in Chapter~\ref{chapter:bg}. Chapter~\ref{chapter:bg} is then followed by three groups of chapters: efficient imitation learning with object-centric priors (Chapters~\ref{chapter:viola} and~\ref{chapter:groot}),  imitation from in-the-wild video observations (Chapters~\ref{chapter:orion} and~\ref{chapter:okami}), and lifelong robot learning with skills (Chapters~\ref{chapter:buds},~\ref{chapter:lotus}, and~\ref{chapter:libero}). After these chapters, we provide a comprehensive literature review of related work in Chapter~\ref{chapter:related-works}. In the end, we summarize the contributions and takeaways from this dissertation in Chapter~\ref{chapter:conclusion}. While this dissertation is written to be read sequentially from beginning to end, readers do not necessarily need to follow the sequential order. To accommodate selective reading, we provide a visualization of chapter dependencies in Figure~\ref{fig:intro:chapter_dependencies}. 

Chapter~\ref{chapter:bg}---\ref{chapter:related-works} are laid out as follows.
\begin{enumerate}

\item In Chapter~\ref{chapter:bg}, we present a formulation of Open-world Robot Manipulation, as well as necessary background about policy learning and data collection that are vital for readers to understand the rest of the dissertation. We also formally define the three regularities that this dissertation focuses on: object regularity, spatial regularity, and behavioral regularity.

\item In Chapter~\ref{chapter:viola}, we introduce  \viola{}, an object-centric imitation learning framework that enables robots to learn closed-loop neural network policies from a small number of teleoperation demonstrations. In this chapter, we explain how \textit{object regularity} has motivated us to use vision foundation models to endow policies with object-centric priors. 
This work was published as: Yifeng Zhu, Abhishek Joshi, Peter Stone, Yuke Zhu. VIOLA: Imitation Learning for Vision-Based Manipulation with Object Proposal Priors. Conference on Robot Learning (CoRL 2022).

\item In Chapter~\ref{chapter:groot}, we introduce an object-centric 3D policy learning method, \groot{}. \groot{} enables robots to derive behavioral cloning policies trained with demonstrations collected under one setting while generalizing to backgrounds, camera angles, and object instances that vary greatly from the data collection setting. In this chapter, we explain how \textit{object regularity} has inspired us to derive the object-centric 3D representations (point clouds) as policy inputs, a key design to achieve generalizable policy learning.
This work was published as: Yifeng Zhu, Zhenyu Jiang, Peter Stone, Yuke Zhu. Learning Generalizable Manipulation Policies with Object-Centric 3D Representations. Conference on Robot Learning (CoRL 2023). 

\item In Chapter~\ref{chapter:orion}, we formulate the problem \textit{open-world imitation from observations}. To tackle our posed problem, we introduce a method, \orion{}, that allows a tabletop manipulator to imitate manipulation skills from single-video human demonstrations. In this chapter, we explain how \textit{spatial regularity} has inspired us to create a spatiotemporal abstraction of a human video, termed a manipulation plan, for the robot to synthesize its actions at test time. 
A preprint of this work is on arXiv: Yifeng Zhu, Arisrei Lim, Peter Stone, Yuke Zhu. Vision-based Manipulation from Single Human Video with Open-World Object Graphs. ArXiv.

\item In Chapter~\ref{chapter:okami}, we further study the problem \textit{open-world imitation from observations} on humanoid robots, where we demonstrate how humanoid robots can learn from single-video human demonstrations and acquire diverse manipulation skills including bimanual, dexterous behaviors. In this chapter, we explain how \okami{} exploits \textit{spatial regularity} while leveraging the morphological similarity between human demonstrators and humanoid robots. 
This work was published as: Jinhan Li, Yifeng Zhu, Yuqi Xie, Zhenyu Jiang, Mingyo Seo, Georgios Pavlakos, Yuke Zhu. OKAMI: Teaching Humanoid Robots Manipulation Skills through Single Video Imitation. Conference on Robot Learning (CoRL 2024). 

\item In Chapter~\ref{chapter:buds}, we introduce a bottom-up skill discovery method \buds{} that discovers reusable sensorimotor skills from unsegmented demonstration datasets. In this chapter, we also explain how \textit{behavioral regularity} has inspired us to discover reusable skills from demonstrations by identifying recurring temporal segments.  
This work was published as: Yifeng Zhu, Peter Stone, Yuke Zhu. Bottom-Up Skill Discovery from Unsegmented Demonstrations for Long-Horizon Robot Manipulation. International Conference on Robotics and Automation (RA-L 2022).

\item In Chapter~\ref{chapter:lotus}, we introduce a continual imitation learning method \lotus{}, which continually learns new tasks by maintaining a library of discovered skills. In this chapter, we also explain how \textit{behavioral regularity} has inspired us to identify recurring temporal segments using semantic features from vision foundation models, a key design to enable skill discovery in the lifelong learning setting.  
This work was published as: LOTUS: Continual Imitation Learning for Robot Manipulation Through Unsupervised Skill Discovery. International Conference on Robotics and Automation (ICRA 2024).

\item In Chapter~\ref{chapter:libero}, we present \libero{}, a lifelong robot learning benchmark that supports community-wide research on continual imitation learning. We present a procedural task generation pipeline that supports the programmatic generation of manipulation tasks, leading to our task suites in \libero{}. This benchmark is designed to support our experiments in Chapter~\ref{chapter:lotus}. 
This work was published as: Bo Liu, Yifeng Zhu, Chongkai Gao, Yihao Feng, Qiang Liu, Yuke Zhu, Peter Stone. LIBERO: Benchmarking Knowledge Transfer in Lifelong Robot Learning Conference and Workshop on Neural Information Processing Systems (NeurIPS 2023, Datasets and Benchmarks Track).

\item In Chapter~\ref{chapter:related-works}, we survey the literature that is relevant to the contributions of this dissertation.

\item In Chapter~\ref{chapter:conclusion}, we summarize this dissertation and discuss future directions. 

\item In Appendix~\ref{chapter:notation}, we summarize all the notation used in this dissertation.

\item In Appendix~\ref{chapter:acronym}, we summarize all the acronyms used in this dissertation. 

\item In Appendix~\ref{chapter:appendix_chapter_I}, we provide additional details about implementations and model designs in Part~\ref{part:I} (Chapters~\ref{chapter:viola} and~\ref{chapter:groot}).

\item In Appendix~\ref{chapter:appendix_chapter_II}, we provide additional details about implementations and model designs in Part~\ref{part:II} (Chapters~\ref{chapter:orion} and~\ref{chapter:okami}).

\item In Appendix~\ref{chapter:appendix_chapter_III}, we provide additional details about implementations and model designs in Part~\ref{part:III} (Chapters~\ref{chapter:buds},~\ref{chapter:lotus} and~\ref{chapter:libero}).

\end{enumerate}

\section{Contribution Overview}
\label{sec:intro-contributions}

This dissertation makes the following contributions to the robot learning literature:

\begin{itemize}
    \item An object-centric imitation learning framework for learning generalizable manipulation policies using behavioral cloning. (Chapter~\ref{chapter:viola})
    
    \item One of the first Transformer architectures in implementing closed-loop sensorimotor policies. (Chapter~\ref{chapter:viola})
        
    \item The first end-to-end closed-loop neural network policy that can make coffee autonomously. (Chapter~\ref{chapter:viola})
    
    \item The first object-centric 3D method for learning behavior cloning policies that generalize to unseen background changes, diverse camera perspectives, and new object instances. (Chapter~\ref{chapter:groot})
    
    \item Formulation of a newly posed problem, \textit{open-world imitation from observation}, for the community to study how robots learn from human videos \textit{without} extra effort in collecting meta-training data and assuming ground-truth object categories are unknown. (Chapter~\ref{chapter:orion})
    
    \item An algorithm for a tabletop manipulator to imitate from a single video recording of a human demonstration. (Chapter~\ref{chapter:orion})
    
    \item An algorithm for humanoid robots to imitate manipulation skills from single-video human demonstrations. This algorithm exploits the morphological similarity between humans and humanoids and enables robots to acquire diverse sensorimotor skills, including bimanual dexterous manipulation. (Chapter~\ref{chapter:okami})
    
    \item A bottom-up skill discovery approach learning a library of sensorimotor skills from unsegmented demonstration data. (Chapter~\ref{chapter:buds})
    
    \item A lifelong skill discovery algorithm that continually maintains a library of sensorimotor skills. (Chapter~\ref{chapter:lotus})
    
    \item A lifelong robot learning benchmark for evaluating lifelong learning algorithms of robot manipulation policies. Unlike the majority of robot learning benchmarks, our benchmark can include more tasks than existing task suites due to our new design of procedural generation pipeline, which can create new manipulation tasks programmatically. (Chapter~\ref{chapter:libero})
    
    % \item Support of diverse robot embodiments such as tabletop manipulators, mobile manipulators, and humanoid robots in a popular robot simulator Robosuite. (Chapter~\ref{chapter:robot_learning})
    
    % \item A real-time robot controller library with high-performance impedance controllers that support the execution of robot learning policies. (Chapter~\ref{chapter:robot_learning})
\end{itemize}

\newpage{}

\chapter{Problem Formulation and Background}
\label{chapter:bg}

We introduce the problem formulation of Open-world Robot Manipulation and all the necessary background for tackling the problem. This chapter is laid out as follows. We introduce our mathematical formulation of Open-world Robot Manipulation (Section~\ref{sec:bg:open-world-formulation}). Then we introduce the general policy formulation of sensorimotor skills (Section~\ref{sec:bg:policy-skills}), followed by a description of imitation learning algorithms and demonstration data collection used throughout the dissertation (Sections~\ref{sec:bg:imitation-learning} and~\ref{sec:bg:data-collection}). Then we describe \textit{object regularity}, \textit{spatial regularity}, and \textit{behavioral regularity} in robot manipulation that serve as the substrate for methods developed in this dissertation (Section~\ref{sec:bg:regularity}). Finally, we introduce the joint configurations and task-space control of the robots that are used throughout our real-robot experiments in this dissertation (Section~\ref{sec:bg:robot-bg}). For a full list of notations used in this chapter, please see Appendix~\ref{chapter:notation}.

\paragraph{Markov Decision Process For Vision-Based Robot Manipulation.} Before introducing Open-world Robot Manipulation, we first describe the common formulation of vision-based robot manipulation using Markov Decision Processes~\cite{puterman1990markov, mandlekar2020learning, wang2021generalization}. A vision-based robot manipulation task can be formulated as a finite-horizon Markov Decision Process (MDP), which is defined as a tuple: $<\mS, \mA, \mP, \mR, \maxhorizon, \initstate{}>$. $\mS$ is the state space of all robot's observations, $\mA$ is the action space of robot commands, $\mP(s_{t+1}|s_t, a_t)$ is the stochastic transition probability function that describes the dynamics of transitioning to any possible next states given the current state and actions, $\mR(s_t, a_t, s_{t+1})$ is the reward function that describes task goals, $\maxhorizon$ is the maximal horizon, and $\initstate{}$ is the initial state distribution.
The objective of tackling a task is to design or learn a policy $\pi$ that maximizes the expected return: $\max_{\pi} \mJ(\pi) = \mathbb{E}[\sum_{t=0}^{\maxhorizon}\mR(s_{t}, a_{t}, s_{t+1})]$. We consider a sparse-reward setting, which does not need additional reward-shaping functions that often require an extensive amount of time to design. In this setting, $\mR(s_{t}, a_{t}, s_{t+1})=1$ when a robot accomplishes the goal of a task, otherwise $0$. The objective in the sparse-reward setting is equivalent to expected task success rates.

\section{Open-world Robot Manipulation}
\label{sec:bg:open-world-formulation}
In this dissertation, we introduce \textit{Open-world Robot Manipulation}, a class of manipulation problems in open-world settings where a robot must quickly adapt to new objects, environments, or tasks for which the robot hasn’t been pre-programmed or pre-trained. In this problem, a robot is expected to perform a wide range of tasks that will be specified by humans. Solutions to the problem are policies that can generalize across diverse objects, scenes, and tasks. We use the Contextual Markov Decision Process (CMDP)~\cite{hallak2015cmdp} to model vision-based robot manipulation in the open-world setting. In the CMDP formulation, we treat human user specifications as ``contexts,'' separating them from the robot's own sensory observations. This separation models human inputs or specifications explicitly separated from the states in the robot's decision-making process. A context $\context$ specifies the task that a robot needs to complete, and $\context$ can take various forms, such as language instructions or demonstrations.

A CMDP formulation of an Open-world Robot Manipulation task is defined as a tuple $<\mS, \mA, \ContextSpace, \CMDPMapping(\context)>$, where $\ContextSpace$ is the context space of human specifications and $\CMDPMapping$ is a function that maps any context $\context \in \ContextSpace$ to a finite-horizon MDP $\CMDPMapping(\context)=<\mS, \mA, \mP^{\context}, \mR^{\context}, \maxhorizon, \initstate{}>$. In our problem setting, we assume the function mapping $\CMDPMapping$ to be surjective so that every considered task in an experiment can be clearly specified by a context $\context$. A task $\Task{\context}\equiv (\initstate^{\context}, \mR^{\context})$ is uniquely defined by the initial state distribution $\initstate{}$ and its reward function $\mR^{\context}$ that indicates if the task goal specified by $\context$ is achieved. Since this dissertation considers robot manipulation, whose dynamics are governed by the laws of physics, the transition probabilities are universal across all tasks. Therefore, while the original formulation of a CMDP includes context-specific transition probabilities $\mP^{\context}$, $\mP^{\context}$ is independent of the context variable and the superscript can be omitted. The solution to the CMDP is a policy that conditions on a specified task $\Task{\context}$, denoted as $\pi(\cdot;\Task{\context})$.

In Open-world Robot Manipulation, we categorize generalization scenarios into two settings: \textit{intra-task} and \textit{inter-task}. The intra-task setting involves adapting to diverse initial states within a task, while the inter-task setting involves learning across multiple tasks and quickly adapting to new tasks. The intra-task setting corresponds to the systematic generalization of a policy evaluated on a single task. The inter-task setting, on the other hand, is equivalent to multitask and lifelong robot learning, where a robot must learn and adapt to a sequence of tasks throughout its deployment.

By considering these two generalization settings, we unify policy generalization within the same task and across different tasks under the single framework of Open-world Robot Manipulation. This unified perspective enables the robotics community to address both single-task policy learning and multitask/lifelong robot learning within a coherent framework. In the rest of this section, we introduce each setting in detail. 

\paragraph{Systematic Generalization Over a Task.} In the intra-task setting, we focus on developing a robot policy with systematic generalization over a manipulation task. Systematic generalization evaluates how well a constructed or learned policy can generalize to conditions beyond those seen during training. This generalization includes both visual and spatial aspects. 

\begin{figure}[ht!]
    \centering
    \includegraphics[width=\linewidth]{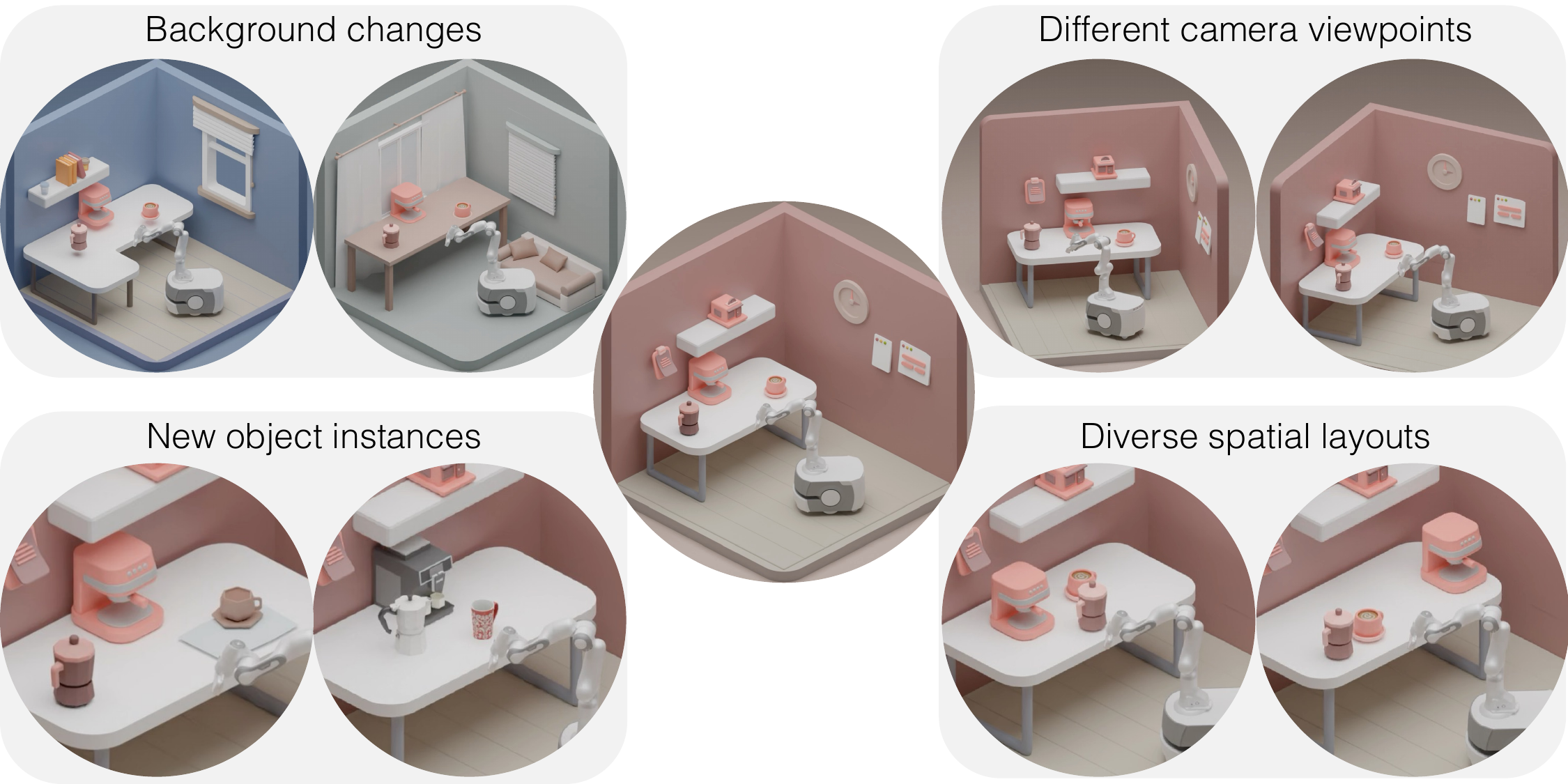}
    \caption{Intra-task generalization involves four dimensions of variations: background changes, different camera viewpoints, new object instances, and diverse spatial layouts.}
    \label{fig:bg:intra_task_generalization}
\end{figure}

In this dissertation, we identify four dimensions of policy generalization that can describe intra-task variations: background changes, different camera viewpoints, new object instances, and diverse spatial layouts. We explain each of the variations as follows. 1) \emph{Background changes}: Any visual changes in the scene, except for the foreground objects involved in a specific manipulation task, are considered background changes. 2) \emph{Different camera viewpoints}: Cameras during deployment can be mounted at angles different from the ones during training. 3) \emph{New object instances}: We consider new objects in the same category as those seen in the demonstration trajectories. These objects mainly differ in color or size while having similar geometry. 4) \emph{Diverse spatial layouts}: The spatial layouts of objects vary and extend beyond the location range of objects during the training stage but remain within the boundary of the robot's feasible workspace. Figure~\ref{fig:bg:intra_task_generalization} visualizes the four axes of variations. 

Formally, a policy $\pi$ needs to be learned to tackle a task $\Task{\context}$ specified by a context variable $\context \in \ContextSpace$. 
As we have mentioned above, $c$ can be specified through language instructions, demonstrations, or any other forms that humans use to show the robot how to perform a task~\cite{shah2023mutex}. 
The initial state distribution $\initstate^{\context}$ of a task $\Task{\context}$ is characterized by four axes of variation. The range of $\initstate^{\context}$ is determined by the span of features along these axes. We can express the relationship between the initial state distribution and these variations as $\initstate^{\context}=\initstatefunc(\feature_{\bg}, \feature_{\cam}, \feature_{\obj}, \feature_{\spatial})$. The features are typically specified implicitly based on configurations of the environmental setup in the initial states, and their ranges are determined by the possible configurations a robot might encounter. Specifically, $\feature_{\bg} \in \Feature_{\text{bg}}$, where $\Feature_{\text{bg}}$ is the space that covers possible background variations and lighting conditions in everyday environments. $\feature_{\cam} \in \Feature_{\text{cam}}$, where $\Feature_{\text{cam}}$ is the space that covers possible camera perspectives, including both camera locations and angles. $\feature_{\obj} \in \Feature_{\text{obj}}$, where $\Feature_{\text{obj}}$ is the space of features that covers the presence of possible instances of task-relevant objects. Finally, $\feature_{\spatial} \in \Feature_{\text{spatial}}$, where $\Feature_{\text{spatial}}$ is the space of features that covers the possible locations of the objects.
Note that the boundary of $\initstate$ is also confined to the feasible workspace of a robot.
The objective of learning a policy is to maximize the expected return of a policy over the initial state distribution $\initstate$ of a given task $\Task{\context}$:

\begin{equation}
\label{eq:singletask-objective}
\max_\pi~\mJ_{\text{intra}}(\pi ; \Task{\context}) = \mathop{\mathbb{E}}\limits_{s_t, a_t \sim \initstate^{\context}, \pi(\cdot;\Task{\context})} \bigg[\sum_{t=1}^{\maxhorizon} \mR^{\context}(s_t, a_t, s_{t+1}) \bigg].
\end{equation}

\paragraph{Multitask and Lifelong Learning.} The inter-task setting of Open-world Robot Manipulation is equivalent to the formulation of multitask and lifelong learning. 
The formulation is the foundation of the methods developed in Chapters~\ref{chapter:buds},~\ref{chapter:lotus}, and~\ref{chapter:libero}.

In multitask learning, a robot is assumed to have direct access to all the tasks that need to be learned. In lifelong learning, the robot can access only a subset of tasks at any given time.  This dissertation considers both cases in a single formulation as follows. Consider a robot sequentially learning $\maxtasknum$ tasks $\{\Task{c_{i}}\}_{i=0}^{\maxtasknum}$ in open-ended environments over $\LLTotalStep$ steps. The robot encounters the tasks in sequence $\Task{\context_{0:\tasknum_{0}}}$, $\Task{\context_{\tasknum_{0}:\tasknum_{1}}}$, \dots, $\Task{\context_{\tasknum_{\LLTotalStep-1}:\tasknum_{\LLTotalStep}}}$, where $0<\tasknum_{\llstep}<\maxtasknum (0\leq \llstep \leq \LLTotalStep)$ and $\tasknum_{\LLTotalStep}=\maxtasknum$. Here, we refer to $\Task{\context_\tasknum}$ as the $\tasknum$-th task. The formulation is equivalent to multitask learning when $\LLTotalStep=0$; otherwise, it represents lifelong learning when $\LLTotalStep > 0$. The goal of lifelong learning is to derive a policy $\pi$ that not only performs well on new tasks but also retains proficiency in previously learned tasks.

The lifelong learning process can be divided into two stages: a \textit{base task stage} for learning a multitask policy over $\Task{\context_{0:\tasknum_{0}}}$, and a \textit{lifelong task stage} where the robot sequentially learns all the other tasks $\Task{\context_{\tasknum_{0}:\tasknum_{1}}}$, \dots, $\Task{\context_{\tasknum_{\LLTotalStep-1}:\tasknum_{\LLTotalStep}}}$ throughout its continual deployment. The \textit{base task stage} is equivalent to the multitask learning setting.
Upon learning the $\tasknum$-th task $\Task{\context_{\tasknum}}$, the objective of finding a generalist policy is to optimize the overall expected return across all the previously learned tasks:
\begin{equation}
\label{eq:multitask-objective}
    \max_\pi~ \mathcal{J}_{\text{inter}}(\pi) = \frac{1}{\tasknum}\sum_{i=0}^{\tasknum} \bigg[ \mathcal{J}_{\text{intra}}(\pi ; \Task{\context_{i}}) \bigg].
\end{equation}

where $\mathcal{J}_{\text{intra}}(\pi ; \Task{c_{i}})$ is defined in Equation~\ref{eq:singletask-objective} that describes the expected return of a policy over a single task.

% \todo{Add a figure to explain the evaluation metrics for lifelong learning.}

\begin{figure}[ht!]
    \centering
    \includegraphics[width=1.0\linewidth]{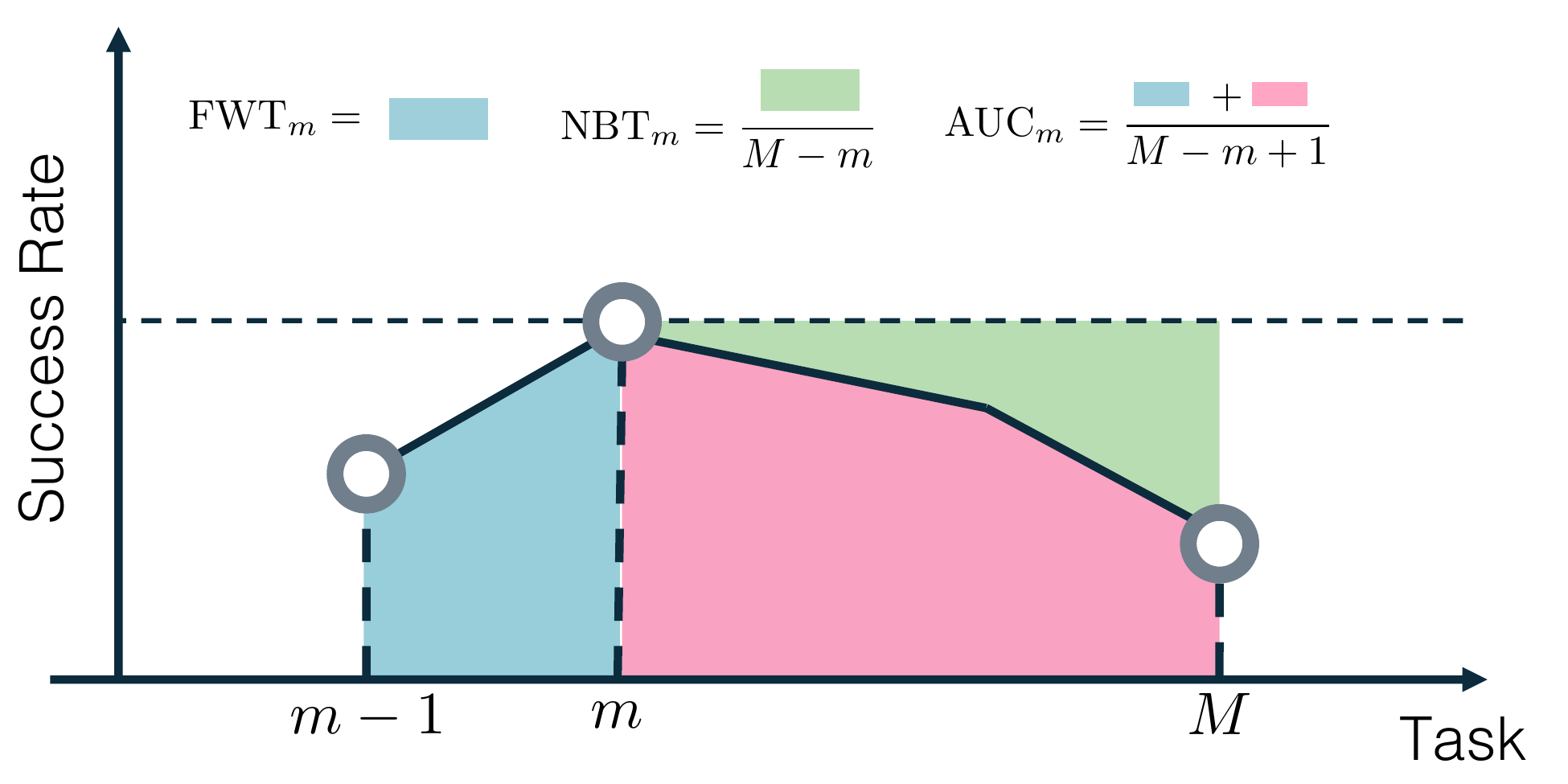}
    \caption[Visualization of evaluation metrics in lifelong robot learning.]{This figure visualizes the three lifelong long metrics evaluated over a task $m$, denoted as $\text{FWT}_{\tasknum}$, $\text{NBT}_{\tasknum}$, and $\text{AUC}_{\tasknum}$. They are computed as follows: $\text{FWT}_{\tasknum}=\bar{r}_{\tasknum}$, $\text{NBT}_{\tasknum}=\frac{1}{\maxtasknum - \tasknum}\sum_{i=\tasknum+1}^{\maxtasknum}(r_{\tasknum, \tasknum} - r_{i, \tasknum})$, and $\text{AUC}_{\tasknum}=\frac{1}{\maxtasknum - \tasknum + 1}(r_{\tasknum, \tasknum} + \sum_{i=\tasknum+1}^{\maxtasknum}r_{i, \tasknum})$. Here, $r_{i, j}$ denotes the agent's success rates on a task $j$ and $\bar{r}_{i}$ denotes the average of success rates of agents' intermediate checkpoint models when learning on a task $i$. FWT, NBT, and AUC measure
    the overall performance of a policy in the entire lifelong learning process (i.e., over $\maxtasknum$ tasks). They are computed as follows: $\text{FWT}=\sum_{\tasknum\in[\maxtasknum]}\frac{\text{FWT}_{\tasknum}}{\maxtasknum}$, $\text{NBT}=\sum_{\tasknum\in[\maxtasknum]}\frac{\text{NBT}_{\tasknum}}{\maxtasknum}$, and $\text{AUC}=\sum_{\tasknum\in[\maxtasknum]}\frac{\text{AUC}_{\tasknum}}{\maxtasknum}$.
     }
    \label{fig:bg:lifelong_metrics}
\end{figure}

\paragraph{Evaluation Metrics For Lifelong Learning.} We describe three standard metrics to evaluate policy performance in lifelong learning~\cite{lopez2017gradient,diaz2018don}: FWT (forward transfer), NBT (negative backward transfer), and AUC (area under the success rate curve). These three metrics are introduced in our published work at 
the Conference and Workshop on Neural Information Processing Systems Datasets and Benchmarks Track, 2023~\cite{liu2023libero}. This work is collaborative work with Bo Liu, Yuke Zhu, and Peter Stone, and my contribution is introduced in Chapter~\ref{chapter:libero}. Figure~\ref{fig:bg:lifelong_metrics} explains FWT, NBT, and AUC in detail. All three metrics are calculated in terms of success rates, where a higher FWT suggests quicker adaptation to new tasks, a lower NBT indicates better performance on past tasks, and a higher AUC means better average success rates across all tasks evaluated.

% Denote $r_{i, j}$ as the agent's success rates on task $j$ when it has just learned over the previous $i$ tasks. Denote $\bar{r}_{i}$ as the average of success rates of agents' intermediate checkpoint models when learning on task $i$. 
% These three metrics are defined as follows.  $\text{FWT}=\sum_{\tasknum\in[\maxtasknum]}\frac{\text{FWT}_{\tasknum}}{\maxtasknum}$, while $\text{FWT}_{\tasknum}=\bar{r}_{\tasknum}$, $\text{NBT}=\sum_{\tasknum\in[\maxtasknum]}\frac{\text{NBT}_{\tasknum}}{\maxtasknum}$, while $\text{NBT}_{\tasknum}=\frac{1}{\maxtasknum - \tasknum}\sum_{i=\tasknum+1}^{\maxtasknum}(r_{\tasknum, \tasknum} - r_{i, \tasknum})$, and $\text{AUC}=\sum_{\tasknum\in[\maxtasknum]}\frac{\text{AUC}_{\tasknum}}{\maxtasknum}$, while $\text{AUC}_{\tasknum}=\frac{1}{\maxtasknum - \tasknum + 1}(r_{\tasknum, \tasknum} + \sum_{i=\tasknum+1}^{\maxtasknum}r_{i, \tasknum})$.

\section{Sensorimotor Skills}
\label{sec:bg:policy-skills}

The sensorimotor skills of a robot are its computational programs that process sensory inputs and synthesize motor commands for purposeful movements such as completing goals in manipulation tasks. In this section, we introduce the policy formulation of sensorimotor skills used in the dissertation. 
We first introduce the general policy formulation and the notation and then describe the hierarchical formulation of visuomotor policies used in Chapters~\ref{chapter:buds} and \ref{chapter:lotus}. In this dissertation, we refer to the policies as \textit{sensorimotor policies} or \textit{visuomotor policies} interchangeability. We use the term visuomotor policies when we emphasize that images are the major input modality of the policies. 

\paragraph{Policy Formulation.} In Section~\ref{sec:bg:open-world-formulation}, we formulate Open-world Robot Manipulation as a CMDP, where its solution, a policy denoted by $\pi(\cdot;\Task{\context})$, models a sensorimotor skill. To explicitly denote inputs and outputs, we write the policy as either $a_{t}\sim \pi(\cdot|s_t, \context)$ or $\pi(a_t|s_{t}, \context)$. The policy can be implemented using either a neural network or an optimization program. When a policy is designated for solving a single task, we drop the context variable $\context$ for simplicity in notation.

\paragraph{Hierarchical Policy Formulation.} In practice, learning policies that directly map from state inputs to action outputs can be computationally prohibitive and subject to error compounding for long-horizon tasks. Also, policies easily suffer from catastrophic forgetting when learning over a sequence of tasks in the lifelong learning setting~\cite{kirkpatrick2017overcoming, van2024continual}: fine-tuned policies can easily overfit to new tasks and completely fail on the previous tasks. To make policy learning tractable and scalable, we factorize a robot policy into a two-level hierarchy: a high-level policy $\metacontroller{}$ and a library of $K$ low-level policies $\{\skillpolicy{i}\}_{i=1}^{K}$, where the low-level policies are commonly termed low-level skills. This factorization has benefits for both intra-task and inter-task generalization in Open-world Robot Manipulation. When a task involves multiple subgoals to complete, we can learn multiple low-level policies for achieving individual subgoals in a much shorter horizon and use a simpler high-level policy to stitch low-level policies together. When the robot needs to learn multiple tasks or adapt to new tasks sequentially, the factorization enables the robot to reuse its learned policies from previous tasks. The policies in such a two-level hierarchy can be expressed as follows.
\begin{equation}
\label{eq:hierarchical-policy}
    \pi(a_{t}|s_{t}, \context) = \sum_{i=1}^{K}\metacontroller{}(i, \param|s_{t}, \context)\mathds{1}(i=k)\skillpolicy{i}(a_{t}|s_{t}, \param)
\end{equation}

In this hierarchical policy formulation, $\param$ is the skill parameters of a low-level policy $\skillpolicy{i}$ that are agnostic to the task specified by $\context$. In this way, we can split a policy into two parts, one that reasons how to accomplish the specified task $\Task{\context}$ while reusing the low-level policies. The combinatorial complexity of reasoning over low-level skills becomes computationally feasible when the high-level policy $\metacontroller{}$ predicts a categorical distribution over the low-level policies. Equation~\ref{eq:hierarchical-policy} can be written in a more concise form as follows.
\begin{equation}
\label{eq:simple-hierarchical-policy}
    \pi(a_{t}|s_{t}, \context) =  \metacontroller{}(k, \param|s_{t}, \context)\skillpolicy{k}(a_{t}|s_{t}, \param)
\end{equation}
where $k\sim \text{Categorical}(K, \mathbf{p}(s_{t}, \context))$, $\mathbf{p}(s_t, \context)=(p_{1}(s_t, \context), p_{2}(s_t, \context), \dots, p_{K}(s_t, \context))$ represents the probability of selecting a low-level skill conditioning on $s_t$ and $\context$, and $\sum_{i=1}^{K}p_{i}(s_t, \context)=1$.\loosepar{}

For all our methods that use the hierarchical policy formulation, we assume a finite set of distinct skill policies $\{\skillpolicy{i}\}_{i=1}^{K}$, which we refer to as a skill library. Whenever we use a skill library, the selection of skills is categorical. When $K=1$, the hierarchical policy is equivalent to a single policy, as the prediction of the high-level policy $\metacontroller{}$ is always the only low-level policy available. Without loss of generality, all the robot policies in this dissertation can be uniformly represented in Equation~\ref{eq:simple-hierarchical-policy}, where $K\geq 1$.

\section{Imitation Learning of Robot Manipulation}
\label{sec:bg:imitation-learning}
In this section, we introduce the standard imitation learning algorithms used in this dissertation. 

\subsection{Behavioral Cloning (BC)}
\label{sec:2-bc}
Behavioral cloning is chosen as the imitation learning algorithm for training policies in Chapters~\ref{chapter:viola} and~\ref{chapter:groot}. Behavioral cloning can be applied when a human specification $\context$ is provided in the form of demonstrations. Let $D^\context = \{ \tau^\context_i \}_{i=1}^N$ denote $N$ demonstrations for task $\Task{\context}$. Each $\tau^\context_i = (o_0, a_0, o_1, a_1, \dots, o_{|\tau^{\context}_{i}|})$ where $|\tau^{\context}_{i}| \leq \maxhorizon$. Here, $o_t$ represents the robot's sensory input, including visual observations and proprioception. The visual observations are captured as either RGB or RGB-D images, while proprioception refers to the configuration of the robot's arm joints and gripper states. In practice, the observation $o_t$ is often non-Markovian. Therefore, we follow works in partially observable MDPs and represent $s_t$ by the aggregated history of observations, i.e., $s_t \equiv o_{\leq t} \triangleq (o_0, o_1, \dots, o_t) $s~\citep{hausknecht2015deep}. When learning $\pi$ using $D^\context$, behavioral cloning optimizes the policy with the following surrogate objective function instead of the expected task success rates: 

\begin{equation}
\label{eq:single-task-il-objective}
    \min_\pi~ J^{\text{BC}}_{\text{intra}}(\pi) = \mathop{\mathbb{E}}\limits_{s_t, a_t \sim D^\context} \bigg[ \sum_{t=1}^{\maxhorizon} \mathcal{L}_{\text{BC}}\big(\pi(s_t; \Task{\context}), a_t\big)\bigg]\
\end{equation}

where $\mathcal{L}_{\text{BC}}$ is a supervised learning loss, e.g., the negative log-likelihood loss. Furthermore, a behavioral cloning policy in robot manipulation is often modeled as a Gaussian Mixture Model (GMM)~\cite{reynolds2009gaussian, mandlekar2021matters}. The loss function to optimize such a policy can be written as follows:
% (with a slight abuse of notation, as we use $K_{\text{mix}}$ instead of the conventional $K$ to avoid conflict with other sections where $K$ denotes the number of low-level skills):
\begin{equation}
\label{eq:gmm-loss}
\begin{split}
\mathcal{L}_{\text{GMM}}(\theta) = & -\mathbb{E}_{\tau\sim D^c} \log \left( \sum_{k=1}^{K_{\text{mix}}} \eta_k \mathcal{N}(\tau | \mu_k(\theta), \sigma_k(\theta)) \right), \\ & \text{where } 0 \leq \eta_k \leq 1, \sum_{k=1}^{K_{\text{mix}}} \eta_k = 1
\end{split}
\end{equation}
where $\theta$ is the learnable parameters of a GMM policy model, and $K_{\text{mix}}$ represents the number of modes in the GMM.

\subsection{Hierarchical Behavioral Cloning (HBC)}
\label{sec:bg:hbc}

Section~\ref{sec:bg:policy-skills} introduces the hierarchical policy formulation that models a sensorimotor skill. Training a hierarchical policy using behavioral cloning, also known as hierarchical behavioral cloning~\cite{le2018hierarchical}, is vital to policy learning in Chapters~\ref{chapter:buds} and \ref{chapter:lotus}. 

In this section, we introduce the hierarchical behavioral cloning algorithm. Hierarchical behavioral cloning aims to train policies that take the hierarchical form. Without loss of generality, we consider a general hierarchical policy formulation (Note that this is a simplified formulation compared to the hierarchical formulation in Section~\ref{sec:bg:policy-skills}): $\pi(a_t|s_t)=\metacontroller{}(\hcbvar(t)|s_t)\pi_{L}(a_t|\hcbvar(t); s_t)$. Here, $\hcbvar(t)$ represents intermediate variables that are predicted by a high-level policy $\metacontroller{}$ (also known as a meta-controller in the literature~\cite{le2018hierarchical}) and subsequently used as inputs to the low-level policy $\pi_{L}$. A concrete example of $\hcbvar(t)$ in our hierarchical policy formulation of sensorimotor skills is the tuple $(k, \param)$.

Implementing a hierarchical behavioral cloning algorithm requires augmenting demonstrations with ``pseudo-labels'' of $\hcbvar(t)$. Specifically, an augmented demonstration dataset $\bar{D}^{\context}=\{\bar{\tau}^{\context}_{i}\}_{i=1}^{N}$ must be created from $D^\context$, where each augmented trajectory takes the form $\bar{\tau}^{\context}_{i}=(o_0, a_0, \hcbvar(0), o_1, a_1, \hcbvar(1), \dots, o_{|\bar{\tau}^{\context}_{i}|})$.

With the augmented demonstration datasets, learning a hierarchical policy through hierarchical behavioral cloning is done by training $\metacontroller{}$ and every low-level policy $\skillpolicy{i}$ separately through behavioral cloning, using the labels from the augmented dataset $\bar{D}^{\context}$.

\subsection{Generalization of Imitation Learning Policies}
\label{sec:bg:generalization-il-policy}

\paragraph{Systematic Generalization for Imitation Learning Policies.} In Section~\ref{sec:bg:open-world-formulation}, we have introduced the notion of systematic generalization in Open-world Robot Manipulation. When we consider imitation learning algorithms, the systematic generalization of a policy is equivalent to the generalization of a policy beyond the distributions covered by demonstrations. Concretely, we quantify the systematic generalization of an imitation learning policy as follows. The demonstration dataset $D^{\context}$ of a task $\Task{\context}$ specifies an initial state distribution covered during training, which we denote as $\initstate^{D^{\context}}$. 
% which can be expressed as $\initstate^{D^{\context}}=\rho(\feature^{D^{\context}}_{\bg}, \feature^{D^{\context}}_{\cam}, \feature^{D^{\context}}_{\obj}, \feature^{D^{\context}}_{\spatial})$. Here, $\feature^{D^{\context}}_{\bg}$, $\feature^{D^{\context}}_{\cam}$, $\feature^{D^{\context}}_{\obj}$, and $\feature^{D^{\context}}_{\spatial}$ represent specific conditions of the environment covered in $D^{\context}$. 
When learning policies through imitation learning, we quantify the systematic generalization of a policy by computing the expected task success rates over $\initstate^{\context} \setminus \initstate^{D^{\context}}$. 

\paragraph{Lifelong Imitation Learning.} 
\label{sec:bg:lifelong-imitation-learning}
In the context of sparse-reward settings for lifelong learning, we consider a practical scenario where a user provides a small demonstration dataset $D^{\context_{\tasknum}}$ whenever a robot needs to learn task $\Task{\context_{\tasknum}}$ in the sequence. We refer to this problem as lifelong imitation learning. This dissertation also uses behavioral cloning in lifelong learning settings. Upon learning the $\tasknum$-th task during the lifelong task stage, we assume that $\{D^{\context_i}\}_{i < \tasknum}$ are not fully available when learning $\Task{\context_{\tasknum}}$. The policy is learned with the following surrogate objective function:
\begin{equation}
\label{eq:lifelong-il-objective}
 \min_\pi~ \mathcal{J}^{\text{BC}}_{\text{inter}}(\pi) = \frac{1}{\tasknum}\sum_{i=0}^{\tasknum} \mathop{\mathbb{E}}\limits_{s_t, a_t \sim D^{\context_i}} \bigg[ \sum_{t=0}^{\maxhorizon} \mathcal{L}_{\text{BC}}\big(\pi(s_t; \Task{\context_i}), a_t\big)\bigg]\,,
\end{equation}

\section{Demonstration Data Collection}
\label{sec:bg:data-collection}

\begin{figure}[ht!]
    \centering
    \includegraphics[width=1.0\linewidth]{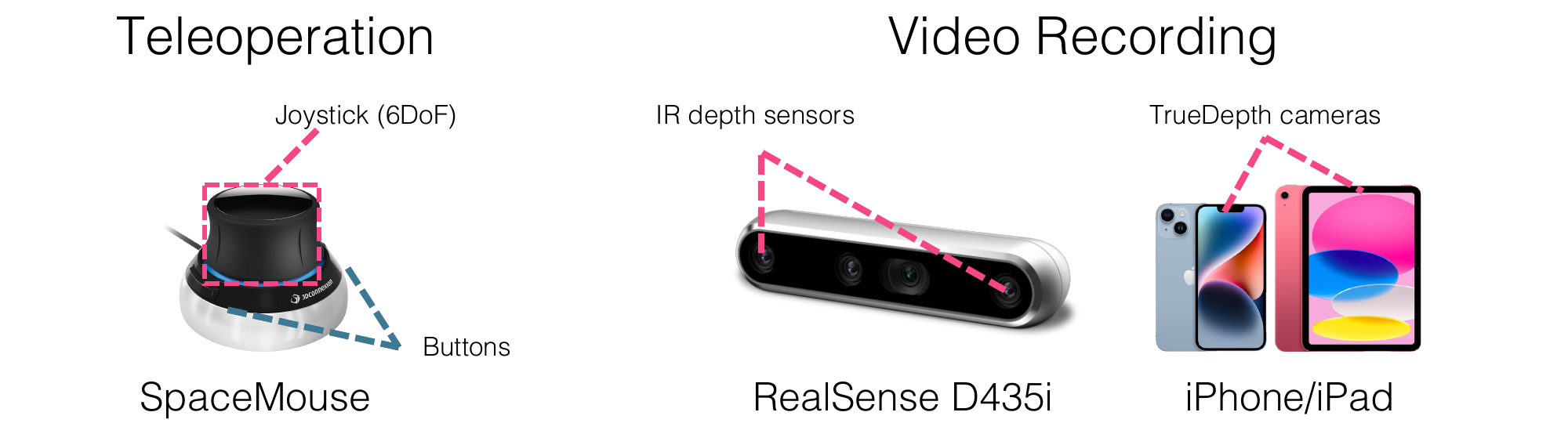}
    \caption[Visualization of devices used for collecting demonstrations.]{We show the devices used for collecting demonstrations through either teleoperation or video recording.}
    \label{fig:bg:devices}
\end{figure}

In this dissertation, we assume demonstrations for imitation learning are provided either using teleoperation devices (SpaceMouses) or video recording devices (cameras, iPhones). We visualize the devices in Figure~\ref{fig:bg:devices}. In this dissertation, demonstrations take the form of teleoperation data in  Chapters~\ref{chapter:viola}, \ref{chapter:groot}, \ref{chapter:buds}, and \ref{chapter:lotus}. Video data is used in Chapters~\ref{chapter:orion} and~\ref{chapter:okami}.

\paragraph{Teleoperation.} Teleoperation has been a well-established standard for humans to control robots remotely without manually dragging them around (known as kinesthetic teaching). Teleoperation allows humans to control robots' low-level, detailed actions while the robot records all the observations and actions as if it were performing the tasks on its own, collecting high-quality demonstration data without a demonstrator occluding large portions of visual observations. We can collect a demonstration trajectory in the form of ${(o_{t}, a_{t})}_{t=1:\maxhorizon}$, where $\maxhorizon$ is the timestep when a demonstrator specifies the termination of an episode. To collect high-quality data for training policies, a demonstrator typically terminates an episode when a manipulation task is successfully completed.

All the teleoperation data in this dissertation is collected using a Spacemouse. We choose the SpaceMouse for the following reasons: 1) They are easy to use, as they share similar characteristics as normal mouses, allowing a person to place a SpaceMouse on the table without holding it in the air all the time (like holding a VR joystick); 2) The control degree-of-freedom (DoF) of the joystick in a SpaceMouse is six, making it easy to map a control command to a 6-DoF task-space control of an end effector. Also, the buttons on the sides provide an easy interface to open/close the gripper; 3) These devices only require simple software settings to connect to normal PCs.

\paragraph{Video Recording.} Videos are another form of data that are easy to collect and capture daily interactions without any robot action labels. People nowadays use cameras or smartphones daily to record videos of themselves doing things like cooking, sorting, and cleaning. Such data is crowdsourced on online platforms such as YouTube and has become a rich, readily available source of daily activities. These data capture human knowledge about contact-rich interactions that are useful for robots to learn manipulation. While there are a variety of cameras available, we specifically focus on the following devices for recording videos: RealSense D435i cameras and iPhones/iPads. RealSense D435i is a camera type that is commonly used in robot systems nowadays, and iPhones/iPads are among the most widely used daily devices. Using videos recorded by iPhones/iPads also promises a near future when everyone can record a video with his or her device and teach robots new manipulation skills. In this dissertation, we use RGB-D videos as input data for robots to learn from visual observations. All these cameras provide depth recordings: RealSense D435i cameras have built-in infrared (IR) depth sensors, while the latest iPhones/iPads have TrueDepth cameras that can record depth images.

\section{Regularity in Robot Manipulation}
\label{sec:bg:regularity}
In Chapter~\ref{chapter:intro}, we introduced the notion of \textit{regularity} as a fundamental concept for designing efficient sensorimotor learning methods. We define regularity in the physical world as the presence of regular, consistent patterns and structures that reveal invariant features across various situations.

% \todo{Address the comment}
% While regularity has been implicitly exploited throughout the robotics literature, researchers have traditionally approached these concepts through the lens of ``structures'' or ``model priors.'' Regularity offers a more systematic framework than other notions like ``structures'' or ``model priors,'' as it explicitly encompasses consistent and regular patterns that persist across different scenarios.

We use several instances to illustrate that regularities have been leveraged widely in robotics and machine learning, even though these instances have not been examined explicitly under the perspective of regularity. In SLAM (Simultaneous Localization and Mapping), researchers formulate problems as factor graphs with sparse connections~\cite{dellaert2017factor}, enabling real-time solutions that would be computationally intractable with dense connections. This approach inherently leverages a key regularity: the sparse dependency patterns between visual features and robot observations, as only certain landmarks are visible from each robot pose. Beyond robotics, Large Language Models in natural language processing parse input sentences into sequences of substrings based on statistical frequencies in linguistic data---effectively exploiting the regularity that certain word combinations consistently appear together across different texts. Similarly, computer vision researchers improve vision model performance on ImageNet~\cite{deng2009imagenet} by normalizing image inputs using the mean and variance of dataset-wide RGB distributions. The insight of image normalization stems from the regularity that pixel values in natural images consistently follow characteristic statistical distributions. Therefore, image normalization can effectively reduce the negative impact of distribution biases on model performance.

 These examples illustrate how researchers have been leveraging regularities in their algorithms or models, whether consciously or unconsciously. 
 In this dissertation, we explicitly leverage regularities to develop algorithms and models for learning generalizable manipulation policies.

In the remainder of this section, we describe the three major regularities mentioned in Chapter~\ref{chapter:intro}: \textit{object regularity}, \textit{spatial regularity}, and \textit{behavioral regularity}. Exploiting these three regularities in the physical world serves as the premise for the methods proposed in Chapters~\ref{chapter:viola}---\ref{chapter:lotus}. We begin by presenting the definitions of each regularity, accompanied by concrete illustrations. We then describe how each regularity connects to specific parts of this dissertation. In individual chapters, we motivate our methods by highlighting their connections to regularity so that readers can understand how this perspective shapes the development of our methods. Note that these three regularities are by no means mutually exclusive, nor do they completely cover all patterns and structures in the physical world. They are important regularities for developing efficient sensorimotor learning methods considered in this dissertation.

% \todo{Mention a little bit how demonstrations play a role in capturing the regularities of the physical world.}

It is important to note that our contribution does not lie in the proposition of these regularities---they are inherent properties of the physical world, which the field has tapped into. Instead, we provide a holistic perspective on learning-based robot manipulation through the lens of regularities. We envision that this perspective will guide the development of future methods in a principled manner, whether they are based on classical motion planning or large robot foundation models.

\subsection{Object Regularity}
\label{sec:bg:object_regularity}

Interactions with objects fundamentally shape how humans engage with the physical world. Humans are often unaware of their sophisticated understanding of objects because object reasoning is deeply ingrained in their physical interactions with the world.

\begin{figure}[ht!]
    \centering
    \includegraphics[width=1.0\linewidth]{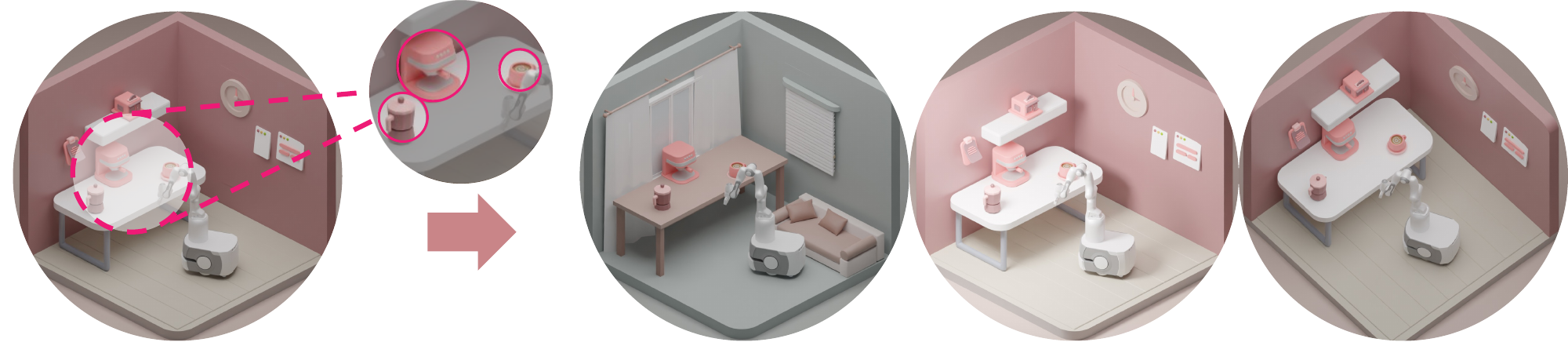}
    \caption[Object regularity illustration.]{We use an example to illustrate \textit{object regularity}. Objects of interest from the left scene appear and behave regularly despite visual variations, no matter how the background, lighting, and camera angle change.}
    \label{fig:bg:object_regularity}
\end{figure}

The advanced ability of humans to reason about object interaction is attributed to the existence of object regularity. Here, we formally define object regularity as the regular and consistent patterns in which objects appear and behave within the physical world. Under this broad definition, we are particularly interested in the following two key aspects: 1) Despite variations in visual appearance, such as changes in lighting, background, and camera perspectives, the inherent semantics of an object remain unchanged. 2) Functionality persists across objects within the same category. These invariant properties implied by object regularity highlight the importance of leveraging object priors in visual observations for learning robot manipulation. Rather than treating images as mere collections of pixels or uniform patches, we can decompose images into regions of objects. In this way, robots can exploit the rich semantic information even with limited demonstration data. Moreover, designing methods based on object regularity is critical for achieving intra-task generalization of sensorimotor skills. Figure~\ref{fig:bg:object_regularity} visualizes a concrete example to illustrate object regularity. Exploiting object regularity serves as a cornerstone for developing efficient sensorimotor learning methods mentioned in Chapters~\ref{chapter:viola} and \ref{chapter:groot}. Studying how robots exploit object regularity also has profound implications for building object reasoning capability for intelligent robots.

\subsection{Spatial Regularity}
\label{sec:bg:spatial_regularity}
Spatial understanding of the world is critical for an intelligent agent, whether human or robotic, to comprehend tasks and interact with the environment effectively. Humans can form such an understanding because \textit{spatial regularity} exists in the physical world, and humans excel at exploiting such a regularity. We hypothesize that for robots to acquire such an advanced level of understanding, they also need to exploit spatial regularity.

\begin{figure}[ht!]
    \centering
    \includegraphics[width=1.0\linewidth]{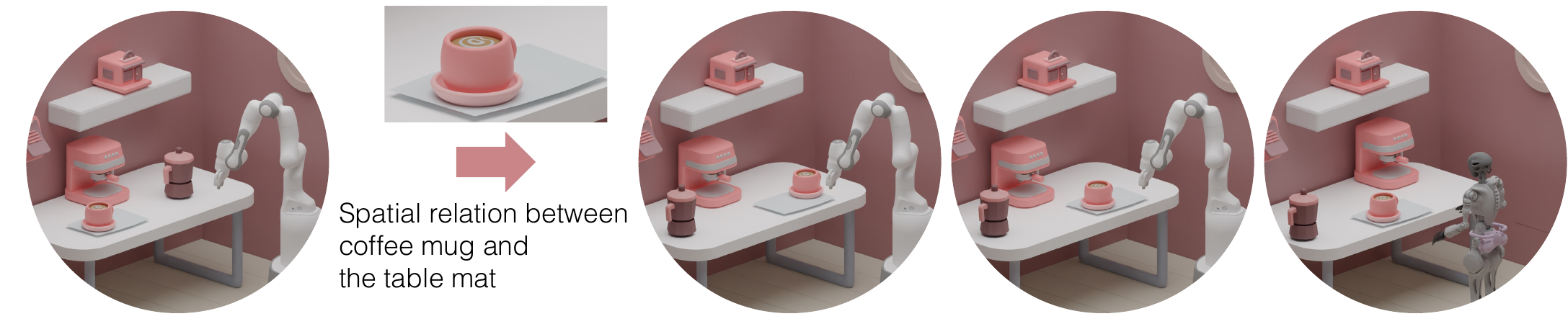}
    \caption[Spatial regularity illustration.]{We use an example to illustrate \textit{spatial regularity}. In this example, we consider a task goal of having a coffee mug on top of a table mat. Despite variations in object locations or different robot embodiments, the task goal is achieved as long as the spatial relation between the coffee mug and the table mat satisfies the goal.}
    \label{fig:bg:spatial_regularity}
\end{figure}

Formally, we define spatial regularity as follows. Certain patterns exist in \textit{spatial relationships} between objects and manipulators that regularly determine successful task execution, with such patterns remaining invariant across different manipulator embodiments. Concretely, this definition implies that the successful completion of a task is governed by the topology or semantics of object layouts that fulfill specific goals, in spite of changes in task-irrelevant features or the operator's identity (be it a human, a tabletop manipulator, or a humanoid robot). Figure~\ref{fig:bg:spatial_regularity} shows an example to illustrate spatial regularity.

Spatial regularity enables learning from visual observations without needing ground-truth action labels. An intelligent agent, either a person or a robot, can interpret the invariant spatial relations between objects of interest while ignoring the location variations of irrelevant objects. It can also reason about actions that move objects around and reproduce certain spatial relations between objects of interest that have the same semantics or topology as the spatial relations present in the visual observations. 

By focusing on the invariant patterns in spatial relationships that determine task success, we can create intelligent robot autonomy that effectively learns from visual observations. Such autonomy makes robots learn from human videos, a data source that is easier to obtain than teleoperation data. Spatial regularity is the premise for methods developed in Chapters~\ref{chapter:orion} and \ref{chapter:okami}. 

% Developing methods that exploit spatial regularity is essential for deepening our understanding of how to construct robust spatial understanding in intelligent robot autonomy.  

\subsection{Behavioral Regularity}
\label{sec:bg:behavioral_regularity}

While imitation learning from demonstration trajectories can capture the low-level motions of manipulation tasks, it often fails to preserve the semantic meaning behind these movements---the underlying purpose and structure that makes a sequence of actions meaningful. Instead of viewing manipulation purely through the lens of continuous trajectories, we can gain deeper insights by considering behaviors as fundamental building blocks that combine to create more sophisticated manipulation capabilities.

\begin{figure}[ht!]
    \centering
    \includegraphics[width=1.0\linewidth]{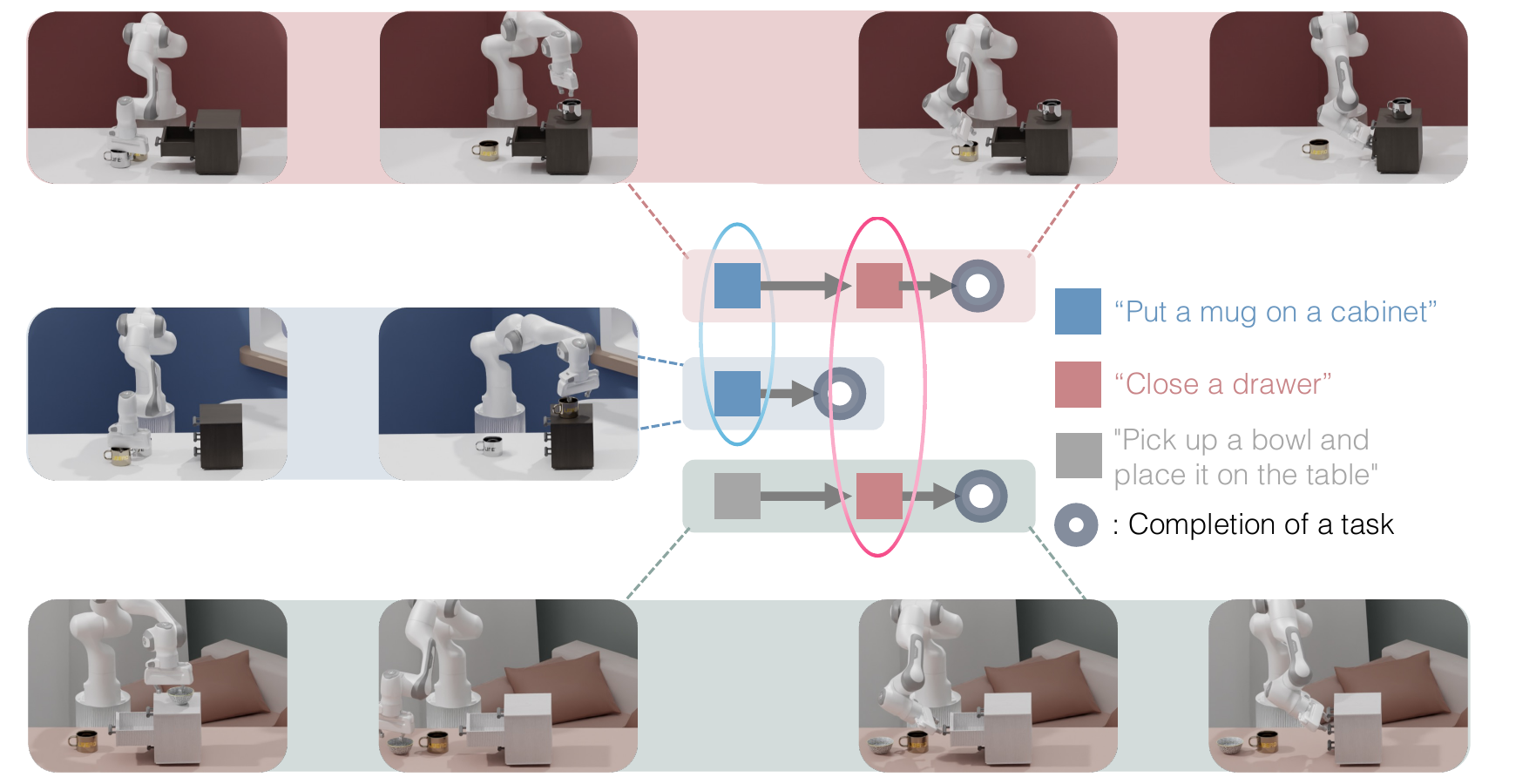}
    \caption[Behavioral regularity illustration.]{We use an example to illustrate \textit{behavioral regularity}. The figure shows three different tasks. Behaviors in each task can be decomposed into primitives, abstracted in the central diagram. Squares of the same color refer to a recurring primitive. (\colorsquare{166}{166}{166} corresponds to a distinct primitive behavior). This central diagram shows that primitive behaviors can appear across tasks. We highlight the recurring primitives by circling out the corresponding squares in the central diagram. In this figure, we also annotate the recurring primitives with language descriptions to explain which behavior each primitive corresponds to.}
    \label{fig:bg:behavioral_regularity}
\end{figure}

We formally define behavioral regularity as the principle that manipulation behaviors can be decomposed into recurring primitives that regularly appear across various tasks.
Behavioral regularity suggests that because of common substructures shared across daily manipulation tasks, we can break trajectories down into recurring segments and identify reusable components rather than treating each new task as an entirely unique problem.
Figure~\ref{fig:bg:behavioral_regularity} shows an illustrative example to explain \textit{behavioral regularity}.

% Consider common household chores as illustrative examples. When setting a table, a robot must execute a series of pick-and-place operations with dishes and utensils. Similarly, cleaning up a bedroom involves picking and placing toys, clothes, and other items in their designated locations. In the kitchen, cooking tasks require manipulating various objects and appliances. Throughout these scenarios, we observe recurring primitive actions: picking up and placing cups in cupboards, moving toys into storage bins, opening and closing drawers for utensils, or operating microwave doors. Across these diverse tasks, we can identify how primitive behaviors are shared: the fundamental action of grasping an object remains consistent, whether handling a cup or a toy, while the basic mechanics of interacting with articulated objects apply similarly to both drawer handles and microwave doors.

Leveraging behavioral regularity in policy learning reduces the amount of training data required while simultaneously learning more tasks than baseline methods that don't exploit behavioral regularity. As a result, robots can continually learn to solve new tasks efficiently.

Moreover, developing methods that exploit behavioral regularity has implications for building intelligent robot autonomy. We can explore and understand how to design procedural memory for a robot~\cite{souza2012processing}, a critical mechanism for the robot to reuse its past experiences for learning in new situations and enhance existing abilities through new experiences. Understanding how to design procedural memory for robots is crucial for achieving inter-task generalization in Open-world Robot Manipulation (Section~\ref{sec:bg:open-world-formulation}) and enabling lifelong robot learning in general. Exploiting behavioral regularity serves as the fundamental premise for the methods developed in Chapters~\ref{chapter:buds},~\ref{chapter:lotus}, and \ref{chapter:libero}.

\section{Robot Joint Configuration and Task-space Control}
\label{sec:bg:robot-bg}

\begin{figure}[ht!]
    \centering
    \includegraphics[width=\linewidth]{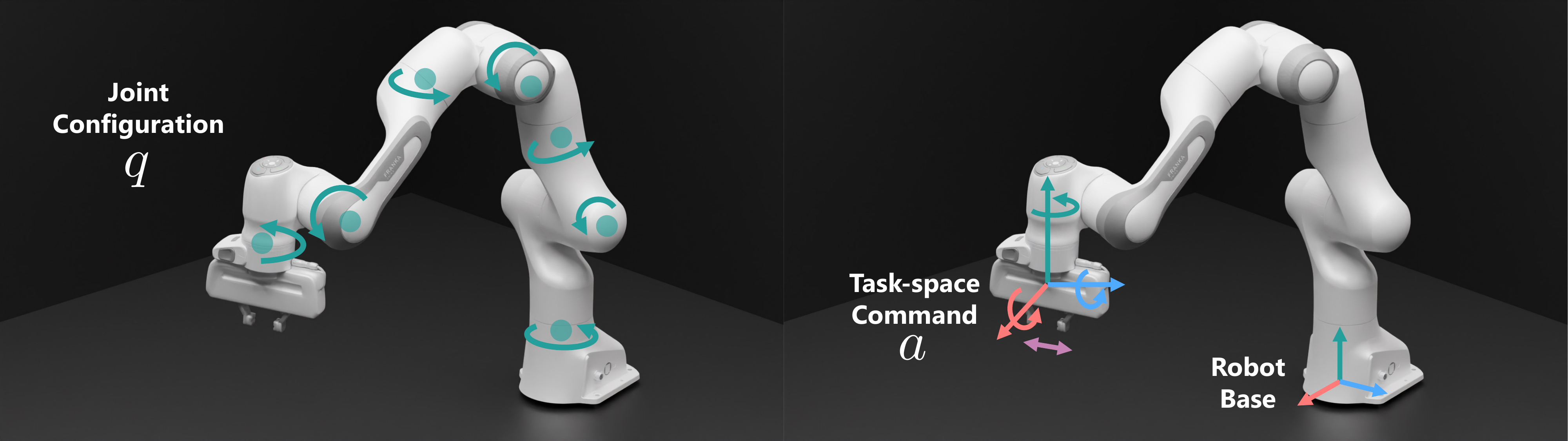}
    \caption[Visualization of a Franka Emika Panda robot used in experiments.]{This figure shows joint configurations, task space commands, and the base frame for task-space commands for a Franka Emika Panda arm.}
    \label{fig:background-panda}
\end{figure}

\begin{figure}[ht!]
    \centering
    \includegraphics[width=\linewidth]{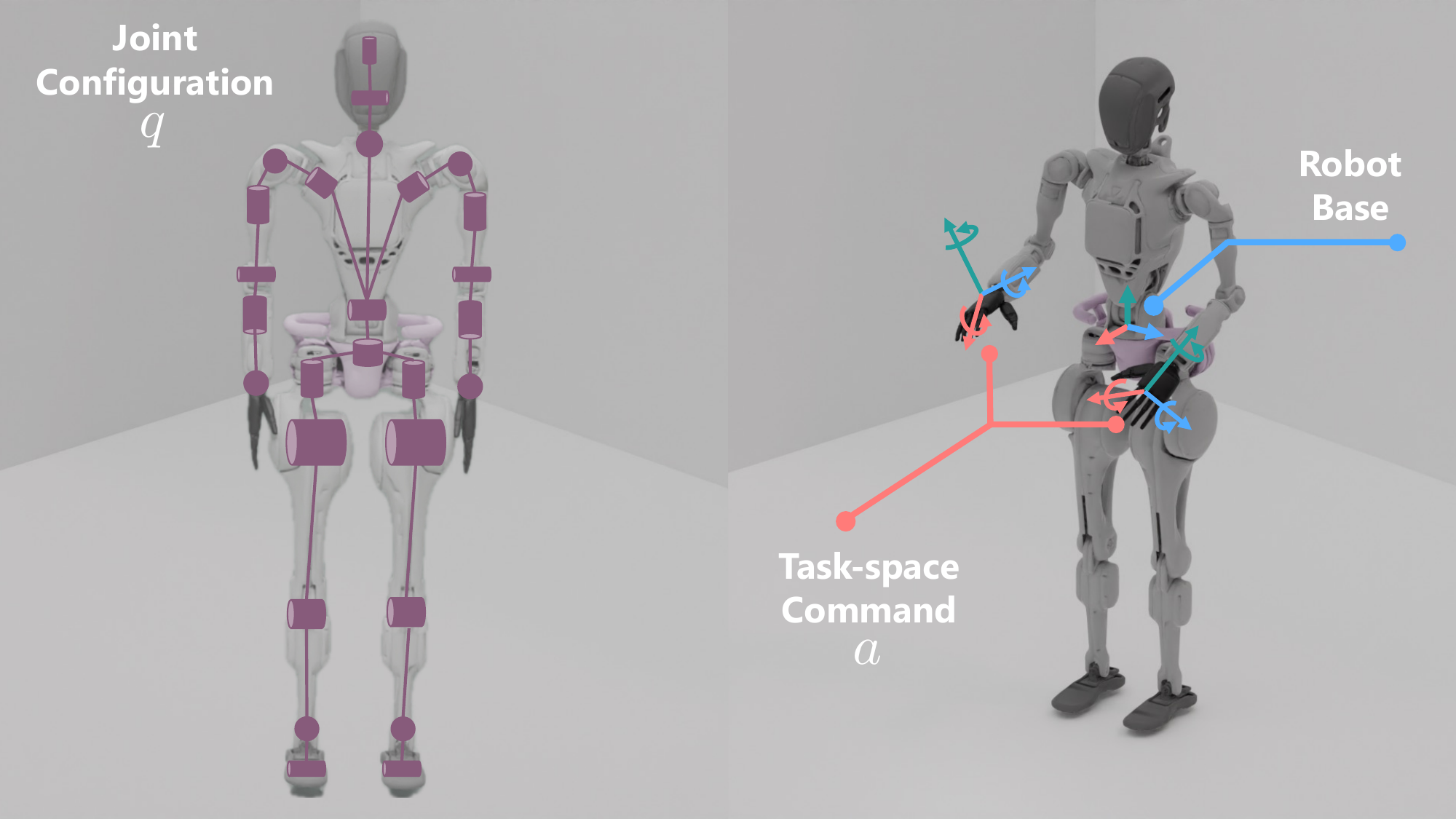}
    \caption[Visualization of a Fourier GR1 humanoid robot used in experiments.]{This figure shows joint configurations, task space commands, and the base frame for task-space commands for a Fourier-GR1 humanoid robot.}
    \label{fig:background-gr1}
\end{figure}

This dissertation uses two types of robots in real-robot experiments: Franka Emika Panda\footnote{\url{https://franka.de/}} and Fourier-GR1 humanoid\footnote{\url{https://fourierintelligence.com/gr1/}}. 
Figures~\ref{fig:background-panda} and \ref{fig:background-gr1} visualize these two types of robots. A Panda robot is equipped with a parallel-jaw gripper, while a GR1 robot is equipped with a pair of InspireHand\footnote{\url{https://inspire-robots.store/collections/the-dexterous-hands}} dexterous hands. The Panda robot is used for experiments in Chapters~\ref{chapter:viola},~\ref{chapter:groot},~\ref{chapter:orion},~\ref{chapter:buds}, and~\ref{chapter:lotus}. The GR1 robot is used for experiments in Chapter~\ref{chapter:okami}.
This section introduces their joint configurations and the task-space controllers in the following paragraphs.

\paragraph{Robot Joint Configurations.} We describe the joint configurations of the Franka Emika Panda robot and the Fourier-GR1 humanoid robot. Figures~\ref{fig:background-panda} and \ref{fig:background-gr1} visualize the joints of both robots except the ones on end-effectors for clarity. The joint configuration constitutes a robots' proprioceptive state, consisting of joint angle readings from the robots' joint encoders, as well as the joint angle readings from either the gripper or dexterous hands. The joints are denoted as $q \in \mathbb{R}^{n}$, where n is the number of joints that can be controlled. A Franka Emika Panda robot with a parallel jaw gripper has $n=8$ degrees of freedom in total. A Fourier-GR1 humanoid robot mounted with two dexterous hands has $n=44$ degrees of freedom in total.

\paragraph{Task-space Control.} Throughout this dissertation, we choose the task-space control command of a robot as a policy action $a$. For a robot to execute $a$ on the hardware, $a$ is converted into a sequence of high-frequency control signals that actuate the robot through designated firmware. Task-space control commands are computed based on the robot's end-effectors' locations, with the coordinate frames chosen to be the robot base as visualized in Figures~\ref{fig:background-panda} and \ref{fig:background-gr1}. This dissertation implements two types of controllers for executing task-space control commands: In Chapters~\ref{chapter:viola},~\ref{chapter:groot},~\ref{chapter:buds}, and \ref{chapter:lotus} where robot teleoperation data is provided, a task-space force controller, Operational Space Controller (OSC), is implemented~\cite{khatib1987unified}. OSC converts the task-space commands directly into joint torques, enabling the robot to accomplish specific tasks with compliant motions. For Chapters~\ref{chapter:orion} and \ref{chapter:okami} where no teleoperation data is provided, we implement a controller based on inverse kinematics: the implemented controller first takes in task-space control commands and uses an inverse kinematics solver to obtain desired joint configurations~\cite{pink2024,mink2024}; then a joint impedance control law is implemented to compute the joint torques that actuate the robot to reach the desired joint configurations~\cite{zhu2022deoxys}. The implementation of the joint impedance controller allows a robot to exhibit behaviors as compliant as an OSC.

\section{Summary}
\label{sec:bg:summary}
In this chapter, we have introduced the formulation of Open-world Robot Manipulation and sensorimotor skills. We have described the imitation learning algorithms that are used in the following chapters, as well as the data collection process of demonstrations. We have also explained the regularities that are exploited in the following chapters(Chapters~\ref{chapter:viola}---\ref{chapter:libero}). In the end, we have introduced the two types of robots used in real-robot experiments, focusing on their joint configurations and task-space control. 

\newpage{}
\part{Efficient Imitation Learning with Object-centric Priors}
\label{part:I}
\chapter{Imitation Learning with Object Proposal Priors}
\label{chapter:viola}

Vision-based manipulation is a critical ability for autonomous robots to interact with everyday environments. It requires robots to understand the unstructured world through visual perception to determine intelligent behaviors. In recent years, deep imitation learning~\cite{mandlekar2020learning,mandlekar2020iris, zhang2018deep} has emerged as a powerful approach to training visuomotor policies on diverse offline data, particularly demonstrations collected via teleoperation. The success of deep imitation learning stems from the effectiveness of training over-parameterized neural networks end-to-end with supervised learning objectives. These models excel at mapping raw visual observations to motor actions without manual engineering. While deep imitation learning methods often distinguish themselves from reinforcement learning counterparts in their scalability to long-horizon tasks, a large body of recent work has pointed out that imitation learning methods lack robustness to covariate shifts and environmental perturbations~\cite{ross2011reduction, park2021object, de2019causal, wen2020fighting, codevilla2019exploring,zhang2016query,kostrikov2019imitation}. End-to-end visuomotor policies tend to falsely associate actions with task-irrelevant visual factors, leading to poor generalization in new situations.

In this chapter, we introduce how to achieve efficient deep imitation learning while deriving generalizable manipulation policies. The key to solving this problem is to leverage \textit{object regularity} as in Section~\ref{sec:bg:object_regularity}, making robots always attend to objects of interest despite visual variations that differ from demonstration datasets. Specifically, we propose a solution that endows imitation learning algorithms with awareness about objects to improve their efficacy and robustness in vision-based manipulation tasks. As cognitive science studies suggest, explaining a visual scene as objects and interactions between the objects helps humans learn fast and make accurate predictions~\cite{greff2020binding, lake2017building, spelke2007core}. Inspired by these findings, we hypothesize that decomposing a visual scene into factorized representations of objects would enable robots to reason about the manipulation workspace in a modular fashion and improve policy generalization.

To this end, we develop an \textit{object-centric imitation learning} approach, which infuses object regularity into the model architecture of visuomotor policies. Training policies with this regularity makes it easy for the model to focus on task-relevant visual cues while discarding spurious dependencies.

The first and foremost challenge of such an object-centric approach is to determine what constitutes an \textit{object} and how objects are represented. The definitions of objects are often fluid and task-dependent for manipulation tasks. This chapter studies the notions of objects operationally and considers objects as disentangled visual concepts that can inform the robot's decision-making. Prior works have explored learning visuomotor policies with awareness of objects, but they are limited to simple control domains~\cite{park2021object}, or single object manipulation~\cite{florence2019self}; or require costly annotations for object detection~\cite{sieb2020graph}. We are motivated by the recent advances in visual recognition, in particular, image models for generating object proposals, which are localized bounding boxes on 2D images~\cite{he2017mask, ren2015faster}. These object proposals capture the regularity of ``objectness'' across appearance variations and object categories. They have served as intermediate representations for downstream vision tasks, such as object detection and instance segmentation~\cite{cai2018cascade, zhou2019objects, zhou2022detecting}. In this chapter, we investigate using object proposals from a pre-trained vision model as object-centric priors for learning visuomotor policies in manipulation.

\section{\viola{}}
\label{sec:viola:method}

\begin{figure}[t!]
    \centering
    \includegraphics[width=1.0\linewidth]{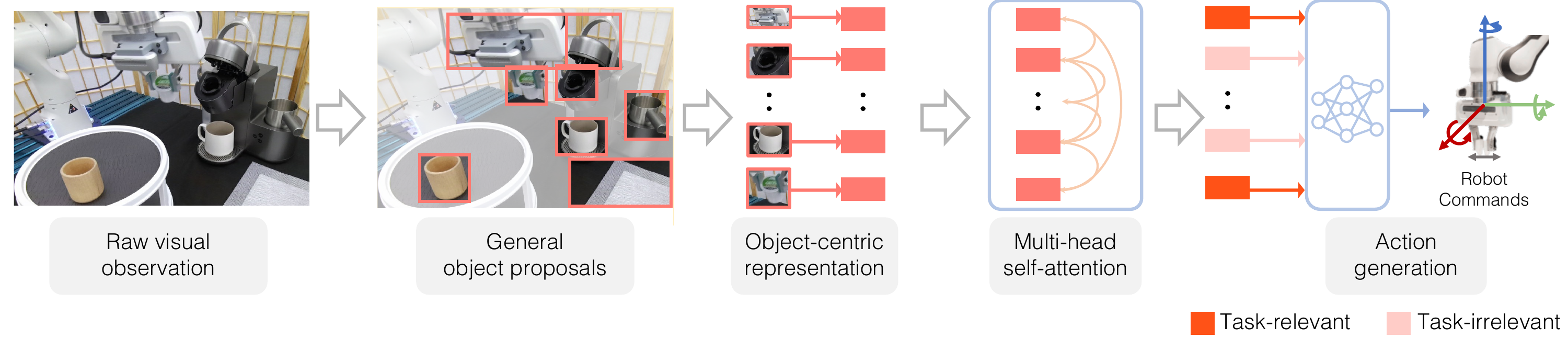}
    \caption[\viola{} overview.]{\textbf{\viola{} Overview.} \viola{} first obtains a set of general object proposals from raw visual observations. It extracts object features from the proposals to build the object-centric representation. The transformer-based policy uses multi-head self-attention to reason over the representation and identify task-relevant regions for action generation.}
    \label{fig:viola:overview}
\end{figure}

We introduce \viola{} (\emphasize{V}isuomotor \emphasize{I}mitation via \emphasize{O}bject-centric \emphasize{L}e\emphasize{A}rning), an object-centric imitation learning model to train closed-loop visuomotor policies for robot manipulation. Our work in this chapter was published at the 6th Annual Conference on Robot Learning~\cite{zhu2022viola}. 
The high-level overview of the method is illustrated in Figure~\ref{fig:viola:overview}. \viola{} first uses a pre-trained Region Proposal Network (RPN)~\cite{ren2015faster} to get a set of general object proposals from raw visual observations. We extract features from each proposal region to build the factorized object-centric representations of the visual scene. These object-centric representations are converted into a set of discrete tokens and subsequently processed by a transformer encoder~\cite{vaswani2017attention}. When trained on supervised imitation learning objectives, the transformer encoder learns to focus on task-relevant regions while ignoring the irrelevant visual factors for decision-making through a multi-head self-attention mechanism.

\subsection{Object-centric Representation}
\label{sec:viola:method:object-centric-repr}
\vspace{-1mm}
This section describes how to build the object-centric representation, denoted as $z_t$. \viola{} first obtains general object proposals using a pre-trained RPN. Then, it computes region features from proposals and obtains a per-step feature $h_t$ using the region features and three context features. In the end, \viola{} builds $z_t$ through the temporal concatenation of per-step features from a history of observations.\loosepar{}

\paragraph{General Object Proposals.} At each time step $t$, we generate object proposals from the workspace image. We use a pre-trained RPN, which takes an RGB image of the workspace as input and outputs a number of bounding boxes localized over the image, each bounding box being an object proposal. 
We select the top $\objectnum$ proposals based on the highest confidence scores from RPN predictions, which indicate regions that contain objects with the highest likelihood. The intuition of using a pre-trained RPN is that it captures the regularity of ``objectness'' in RGB images through the supervision of natural image datasets. Zhou et al.~introduced Detic, whose RPN trained on an 80-class training dataset can be generalized to a 2000-class testing dataset~\cite{zhou2022detecting}. Our preliminary study has suggested that Detic's generalization ability also holds for localizing objects on raw images in our simulation and real-world tasks despite the domain gaps. Given the superior performance of pre-trained Detic RPN, we choose it to localize regions with objects in all our experiments.

\paragraph{Region Features.} For our policy to reason over objects and their spatial relations, we need to identify what objects each region contains and where these regions are from the top $K$ proposals. To encode this information, we design a \textit{visual feature} and a \textit{positional feature} for each region. To extract the visual feature from a region, we learn a spatial feature map by encoding the workspace image with the ResNet-18 module~\cite{he2016deep} and extract the visual feature using the ROI Align operation~\cite{he2017mask}. We use a learned spatial feature map as the visual features rather than from a pre-trained feature pyramid network in object detection models because we share the same intermediate objective of localizing objects as object detectors but different final objectives for the downstream tasks---pre-trained feature pyramid networks are optimized for visual recognition tasks, but we need actionable visual features that are informative for continuous control. Such a design choice is supported by our ablation study in Section~\ref{sec:viola:results}. For positional features, we encode the coordinates of bounding box corners on images using sinusoidal positional encoding~\cite{vaswani2017attention} (See implementation details in Appendix~\ref{ablation_sec:viola:model}). We flatten each visual feature and add it to the positional feature of the same region to obtain a region feature.

\paragraph{Context Features.} We extract the region features for the policies to reason over individual objects. However, they are insufficient for decision-making in vision-based manipulation tasks, so we introduce three context features to assist decision-making.
As regions only encode local information, we add a \textit{global context feature} to capture the current task stage from observation. The global context feature is derived by computing Spatial Softmax over the spatial feature map of the workspace image~\cite{levine2016end}.
During manipulation, the robot's gripper often occludes objects in the workspace view, so we add an \textit{eye-in-hand feature} from the eye-in-hand images to encode the information about occluded objects. We also encode the robot's measurements of its joints and the gripper into \textit{proprioceptive features} for the policy to generate precise actions based on the robot's states. We aggregate the context and region features at each time step $t$ into a set, referred to as the per-step feature $h_t$.

\paragraph{Temporal Composition.} We build object-centric representation $z_t$ through the temporal composition of $h_t$ that captures temporal dependencies and dynamic changes of object states. Building policies over a sequence of past observations rather than the most recent observation has been shown effective by prior work~\cite{mandlekar2021matters,mandlekar2020iris}. In our method, the temporal composition also increases the recall rate of object proposals over image observations, making the policy more robust to detection failures than using a single image.
Concretely, $z_t$ is built from  a set of per-step features $\{h_{t-i}\}_{i=0}^{\historytime}$ from the last $\historytime+1$ time steps. To preserve the temporal ordering of per-step features, we add sinusoidal position encoding of temporal positions $\{PE_{i}\}_{i=0}^{\historytime}$ to the per-step features, resulting in our object-centric representation $z_t = \{h_{t-i}\oplus PE_{i}\}_{i=0}^{\historytime}$ (Details of sinusoidal position encoding are provided in Appendix~\ref{ablation_sec:viola:implementation}). Our ablation study in Section~\ref{sec:viola:results} shows the importance of temporal positional encoding that retains the temporal ordering of features in $z_t$.\loosepar{}

\begin{figure}[t!]
    \centering
    \includegraphics[width=1.0\linewidth]{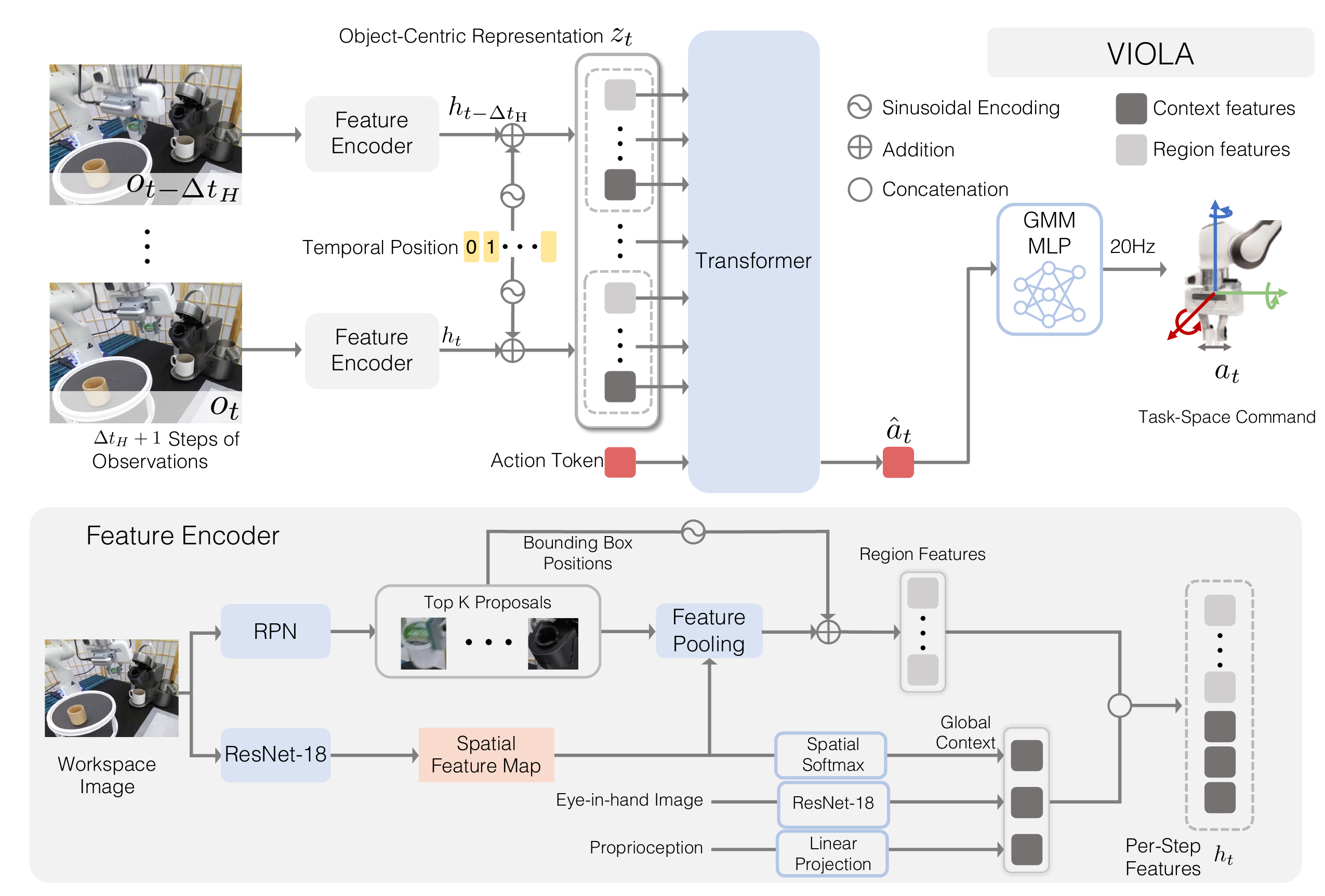}
    \caption[Model architecture of \viola{} policies.]{\textbf{\viola{} Model Architecture.} At time $t$, \viola{} computes the per-step features $h_t$ using the top $\objectnum$ object proposals. Then, it constructs the object-centric representation $z_t$ by composing per-step features from the last $\historytime+1$ time-step observations along with their temporal positional encodings. The transformer encoder reasons over $z_t$ to output a latent vector of the action token, $\hat{a}_{t}$, which is passed through a multi-layer perceptron (MLP) to generate robot actions.\loosepar{}}
    \label{fig:viola:model}
\end{figure}

\subsection{Transformer-based Policy}
\label{sec:viola:method:transformer-policy}

We desire a policy focusing on task-relevant region features in $z_t$ to generate actions. Regions that associate with task-relevant objects facilitate the accurate prediction of actions, while regions with task-irrelevant objects are likely to confound the policies. We seek to use a transformer~\cite{vaswani2017attention} as the policy backbone, which can reason over objects and their relations using its self-attention mechanism. The core of a transformer is an encoder layer, which consists of a multi-head self-attention block (MHSA), a layer-normalization function~\cite{ba2016layer}, and a Feed-Forward Network (FFN) consisting of fully-connected layers. A transformer encoder layer takes as input a sequence of $n$ latent vectors $(\tokenin_1, \dots, \tokenin_n)$ (also called \textit{tokens}) and outputs a sequence of $n$ latent vectors $(\tokenout_1, \dots, \tokenout_n)$. An MHSA block consists of multiple self-attention sub-blocks in parallel, which computes attention weights over all the tokens and a weighted sum of input token values. Through an MHSA block, a token that corresponds to a task-relevant region is assigned higher attention weights than a token that corresponds to a task-irrelevant region. Each transformer encoder outputs $\Tokenout=\text{FFN}(\text{LayerNorm}(\text{MSHA}(\Tokenin))$, where each row of $\Tokenout$ is an output latent vectors $\tokenout_i$ that corresponds to $\tokenin_i$. Our transformer-based policy is a stack of multiple transformer encoder layers, which allows for a higher degree of expressiveness over input tokens compared to a single layer. 

For our policy, we tokenize our object-centric representation $z_t$, treating each region and context feature vector as an input token. To make action generation attend more to task-relevant region features than task-irrelevant ones, we append a learnable token, action token, to the input sequence of a transformer. The action token design resembles the specific class tokens of transformer-based models in natural language understanding tasks~\cite{devlin2018bert} or visual recognition tasks~\cite{kostrikov2020image}, in which the specific class tokens are used for outputting latent vectors for downstream task predictions. Similarly, we can get the output latent vector $\hat{a}_{t}$ from the action token, which learns to attend to task-relevant regions through training supervision. In the end, we pass $\hat{a}_{t}$ through a two-layered fully-connected network, followed by a GMM (Gaussian Mixture Model) output head, which has been shown effective to capture the diverse multimodal behaviors in demonstration data~\cite{wang2020critic, mandlekar2021matters}.\loosepar{}

\section{Experiments}
\label{sec:viola:experiments}

\begin{figure}[t!]    
    \includegraphics[width=1.0\linewidth]{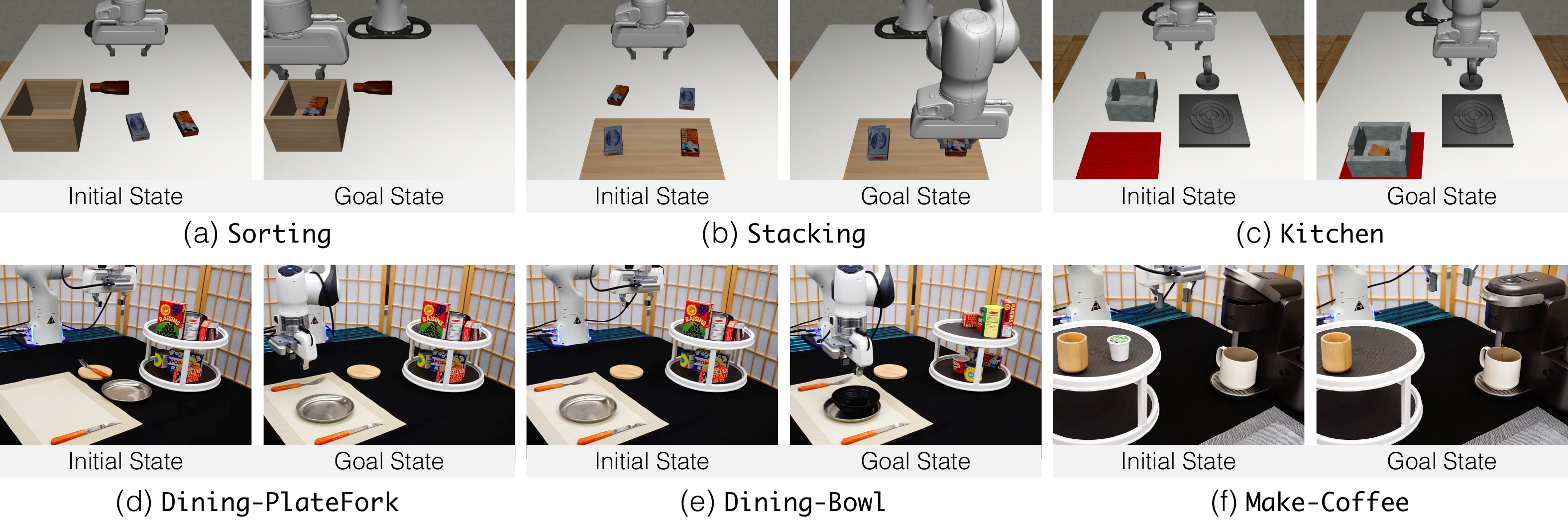}
    \caption[Visualization of real-world tasks in \viola{} experiments.]{Visualization of the initial and goal configurations for real-world tasks. }
    \label{fig:viola:tasks}
\end{figure}

We design our experiments to answer the following questions: 1) How well does \viola{} perform against state-of-the-art end-to-end imitation learning algorithms? 2) How does it take advantage of object-centric representations? 3) What design choices are essential for good performance? 4) Is \viola{} practical for real-world deployment? 5) How do we decide the number of object proposals extracted from each image observation?

\subsection{Experimental Setup} 
\label{sec:viola:experiment-setup}
\paragraph{Task Details.} We conduct quantitative evaluations in simulation and real-world tasks to validate our approach. We design simulation tasks using Robosuite simulation~\cite{zhu2020robosuite} and use the tasks for quantitative comparisons between \viola{} and baselines. We also validate our design choices through ablation studies in simulation. We design three simulation tasks, \sort{}, \stack{}, \kitchen{}, and three real-world tasks, \platefork{}, \bowl{}, \coffee{}. These six tasks cover a wide range of manipulation behaviors that include prehensile and non-prehensile motions. For the \sort{} task, the robot needs to pick up two boxes sequentially and place them together in the sorting bin. For the \stack{} task, the robot needs to stack the same type of boxes in a designated region. For the \kitchen{} task, the robot needs to place the bread in the pot and serve it after putting it on the stove, and turn off the stove after serving. In \platefork{}, the robot needs to push the plate into the target region and place the fork next to the plate. The \bowl{} task requires the robot to push the turntable around, pick up the bowl, and place it on the plate. In \coffee{}, the robot is tasked with an entire coffee-making procedure, from opening up the K-cup holder to pushing the button to activate the espresso machine. 

We visualize their initial and goal configurations in Figure~\ref{fig:viola:tasks}. We design these tasks to understand if object-centric priors benefit policy generalization along two axes of task characteristics: large placement variations of objects and multi-stage long-horizon execution. For the first axis, we design \sort{} and \stack{} to have large initial ranges of object placements. For the second axis, we use \kitchen{}, a multi-stage long-horizon manipulation task. The real-world tasks are designed to resemble everyday activities: \platefork{} and \bowl{} for dining table arrangement, \coffee{} for espresso coffee making.

 We use the Franka Emika Panda arm in all tasks. We choose Operational Space Controller~\cite{khatib1987unified} for end-effector control, a binary command control for parallel-gripper, both operating at $20$Hz. We use Kinect Azure as the workspace camera and Intel Realsense D435i as the eye-in-hand camera, capturing the workspace images and eye-in-hand images, respectively.

\paragraph{Data Collection.} We use the SpaceMouse to collect 100 teleoperated demonstration trajectories for each simulation task and 50 for each real-world task. We apply color augmentation~\cite{he2019bag, jangir2022look} to the collected images to increase the visual diversity in datasets, making learned policy robust to visual cue variations due to surrounding lighting conditions.\loosepar{}

\paragraph{Evaluation Setup.} To systematically evaluate the efficacy and robustness of the learned policies, we design the following testing setups in simulation: 1) \canonical{}: all the objects and sensor configurations follow the same distribution as seen in demonstrations; and 2) Three testing variants, namely \placement, \distracting, and \camera, which are part of the intra-task generalization as mentioned in Section~\ref{sec:bg:generalization-il-policy}. These three variants are designed in the following way: location shifting in initial object placements (\placement{}), distractor objects present in the scene (\distracting{}), and jittering in camera pose (\camera{}).

% The design principle of testing variants is to keep the task semantics the same as in the canonical setup. At the same time, we identify the three challenging axes of variations for learning-based manipulation, namely changes in initial object placements (\placement{}), distractor objects presented in the scene (\distracting{}), and camera pose jitters (\camera{}).

The success rates of policies are computed based on the evaluation rollouts. In the simulation, the success is determined by whether the object states in the simulator meet the pre-programmed goal function. For example, in \sort{} task, the goal function is programmed to check if the two boxes are both in the bin, and the simulation environment decides a rollout is successful only if the goal function returns true value. In the real world, the success of a rollout is determined by whether the objects are in the goal configuration, as shown in the given demonstrations.

\subsection{Results}
\label{sec:viola:results}

We answer question (1) by quantitatively comparing \viola{} with the following baselines:
\begin{itemize}
    \item \textbf{\bc}: behavioral cloning that conditions on current observations.
    \item \textbf{\oreo}~\cite{park2021object}: a behavioral cloning method that uses object-aware discrete codes~\cite{van2017neural}. The discrete codes are trained via a self-superivsed VQ-VAE model.
    \item \textbf{\bcrnn}~\cite{mandlekar2021matters, dinyari2020learning, florence2019self, rahmatizadeh2018vision}: a behavioral cloning method that uses a temporal sequence of past observations with recurrent neural networks.
    \item \textbf{\patch}: a variant of \viola{}, where we use a regular grid of image patches as inputs to the policy rather than proposals. The use of patches resembles the image patches used in ViT~\cite{dosovitskiy2020image}.
\end{itemize}

We randomly generate 100 initial configurations and repeat evaluation runs with three different random seeds. We evaluate all policies on the same set of pre-generated initial configurations for a fair comparison.\loosepar{}

\makeatletter\def\@captype{table}
\begin{table}
\centering
  \resizebox{1.0\linewidth}{!}{  
  \begin{tabular}{llccccc}
    \toprule
    \textbf{Tasks} & \textbf{Variants} & \textbf{BC~\cite{finn2016deep}} & \textbf{OREO~\cite{park2021object}} &  \textbf{BC-RNN}~\cite{mandlekar2021matters}  & \textbf{\viola{}-Patch} & \textbf{\viola}\\
    \midrule
    \sort~ & {\canonical} & {25.1 $\pm$ 1.6}   & {38.7 $\pm$ 0.5} & {62.8 $\pm$ 0.9} & {71.2 $\pm$ 1.0} & {\textbf{87.6} $\pm$ 1.1} \\
    {} & {\placement} & {1.9 $\pm$ 0.7}   & {8.3$\pm$1.7} & {11.7 $\pm$ 1.0} & {48.5 $\pm$ 2.2} & {\textbf{68.3} $\pm$ 1.5} \\
     {} & {\distracting} & {14.5 $\pm$ 1.9} & {26.0$\pm$ 11.4} & {46.7 $\pm$ 6.5} & {58.6 $\pm$ 4.2} & {\textbf{74.4} $\pm$ 5.7} \\
    {} & {\camera} & {6.1 $\pm$ 0.3}   & {16.3$\pm$ 3.3} & {9.6 $\pm$ 0.4} & {34.6 $\pm$ 1.4} & {\textbf{50.7} $\pm$ 0.6}  \\
    \midrule
    \stack~ & {\canonical} & {14.9 $\pm$ 1.1} & {13.3 $\pm$ 2.0} & {27.8 $\pm$ 0.8} & {71.2 $\pm$ 1.0} & {\textbf{71.3} $\pm$ 1.0}  \\
    {} & {\placement} & {1.6 $\pm$ 0.3}   & {0.3$\pm$ 0.5} & {0.0 $\pm$ 0.0} & {28.4 $\pm$ 1.9} & {\textbf{46.7} $\pm$ 0.2} \\
     {} & {\distracting} & {5.6 $\pm$ 1.4}   & {5.6 $\pm$ 4.5} & {14.4 $\pm$ 3.2} & {\textbf{41.4} $\pm$ 5.1} & {38.6 $\pm$ 2.8} \\
    {} & {\camera} & {2.1 $\pm$ 0.3}   & {0.6 $\pm$ 0.9} & {1.0 $\pm$ 0.0} & {17.6 $\pm$ 1.0} & {\textbf{29.3} $\pm$ 1.7} \\
    \midrule
    \kitchen{} & {\canonical} & {0.0 $\pm$ 0.0}   & {0.0 $\pm$ 0.0} & {0.0 $\pm$ 0.0} & {5.8 $\pm$ 0.6} & {\textbf{84.2} $\pm$ 1.3} \\
     {} & {\placement} & {0.0 $\pm$ 0.0}   & {0.0 $\pm$ 0.0} & {0.0 $\pm$ 0.0} & {3.1 $\pm$ 0.6} & {\textbf{58.4} $\pm$ 1.1} \\
     {} & {\distracting} & {0.0 $\pm$ 0.0}   & {0.0 $\pm$ 0.0} & {0.0 $\pm$ 0.0} & {10.7 $\pm$ 0.7} & {\textbf{73.2} $\pm$ 6.2} \\
    {} & {\camera} & {0.0 $\pm$ 0.0}   & {0.0 $\pm$ 0.0} & {0.0 $\pm$ 0.0} & {2.6 $\pm$ 0.3} & {\textbf{41.2} $\pm$ 1.7} \\
    \bottomrule
  \end{tabular}
  }
\caption[Evaluation of \viola{} policies in simulation.]{\label{tab:viola:single-task-results} Success rates (\%) in simulation setups averaged over 100 initializations with repeated runs of three random seeds.}  
\end{table}

Table~\ref{tab:viola:single-task-results} shows that \viola{} outperforms the most competitive baseline \bcrnn{} by $50.8\%$ success rates in \canonical{} and $44.1\%$ in three testing variants.
 \viola{}'s superior performance shows the advantage of object-centric representations for visuomotor imitation. 
 % \oreo{}'s performance implies that learning object-aware discrete codes via unsupervised learning does not consistently improve performance  simple \bc{} for all tasks. 
 % The comparisons between \bc{}, \oreo{}, and the other methods using temporal windows show the importance of temporal modeling for these algorithms to achieve high performances and robustness in complex vision-based manipulation domains.
 Table~\ref{tab:viola:single-task-results} shows that \patch{} has comparable performance to \viola{} in partial evaluation setups. This result suggests that the transformer backbone can attend to patch regions that cover task-relevant visual cues, even though the patch division is agnostic to objects. Nonetheless, \viola{} still performs better than \patch{}, especially in the long-horizon task \kitchen{}, showing that reasoning over regions with complete coverage of objects is critical for the success of \viola{}. This comparison result highlights the importance of leveraging object regularity.\loosepar{}

\begin{table}[ht!]
\centering
  \resizebox{1.0\linewidth}{!}{
  \begin{tabular}{lcccccccc}
    \toprule
    \textbf{Models} & \canonical & \placement & \distracting & \camera \\
    \midrule
Base & {0.0 $\pm$ 0.0} & {0.0 $\pm$ 0.0} & {0.0 $\pm$ 0.0} & {0.0 $\pm$ 0.0} \\
+ Temporal Observation & $\uparrow$ 69.5 $\pm$ 2.5 & $\uparrow$ 25.1 $\pm$ $\uparrow$ 2.2 & $\uparrow$ 37.7 $\pm$ 3.4 & $\uparrow$ 39.9 $\pm$ 1.5 \\
+ Temporal Positional Encoding & $\uparrow$ 74.0 $\pm$ 0.8 & $\uparrow$ 48.3 $\pm$ 1.9 & $\uparrow$ 48.3 $\pm$ 3.3 & $\uparrow$ 50.9 $\pm$ 1.5 \\
+ Region Visual Features &  72.8 $\pm$ 0.9 & $\downarrow$ 37.7 $\pm$ 1.2 & $\uparrow$ 54.6 $\pm$ 4.6 & 49.2 $\pm$ 1.6  \\
+ Region Positional Features & $\uparrow$ 80.2 $\pm$ 2.9 & 38.6 $\pm$ 0.3 & $\uparrow$ 62.0 $\pm$ 5.7 & $\downarrow$ 46.5 $\pm$ 1.9 \\
+ Random Erasing (=\viola{}) & {$\uparrow$ 87.6 $\pm$ 1.1} & {$\uparrow$ 68.3 $\pm$ 1.5} & {$\uparrow$ 74.4$\pm$5.7} & {$\uparrow$ 50.7$\pm$0.6}\\
    \bottomrule
    \end{tabular}
    }
    \caption[Ablation study of \viola{} policy designs.]{\label{tab:viola:main-ablation-results} The effect of \viola{} model designs on success rates ($\%$) in \sort{} task. Changes larger than 2$\%$ are annotated with $\uparrow$ / $\downarrow$ to indicate performance increase or decrease compared with the result from the previous row.}
\end{table}

\paragraph{Ablation Studies.} To answer questions (2) and (3), we use ablation studies to validate our model's design and show how it takes advantage of the object regularity. Table~\ref{tab:viola:main-ablation-results} quantifies the effects of the transformer backbone, the use of temporal windows, object proposal regions, and our data augmentation technique. We start with our base ablation model, a transformer model that only takes the current observations as input. Second, we add the temporal window of observations as in \bcrnn{}. Results show that the use of temporal observation is key to unleashing the power of the transformer architecture, making the performance of this ablated version comparable to \bcrnn{}. However, this model does not encode temporal ordering because the transformer model is invariant to input permutation. In the next row, we add temporal positional encoding~\cite{vaswani2017attention} to the input sequence, which produces a method that outperforms the top baselines. \loosepar{}

In the next two rows, we procedurally add visual and positional features of regions to prove that our model \textit{does} exploit the object regularity. Results show that visual features alone only improve our model's robustness to \distracting{}. When using both visual and positional features, the model performs better in \canonical{} and \distracting{}, but worse in \placement{} and \camera{}. We hypothesize that this ablated version overfits the locations of proposal boxes in demonstrations; therefore, it generalizes worse in \placement{} and \camera{} where the distribution of object locations on 2D images shifts from demonstrations. To mitigate such overfitting, we use Random Erasing~\cite{zhong2020random} to achieve the highest success rates in all evaluation setups.

\paragraph{Real Robot Evaluation.} We compare \viola{} against the baseline, \bcrnn{}, on all three real-world tasks, answering question (4). Note that \bcrnn{} was the SOTA baseline when \viola{} was published in 2022. We evaluate in 10 different initial configurations and repeat three times in each configuration to evaluate the policies.
To mitigate potential human bias introduced by setting up objects before each trial, we use the A/B testing paradigm for evaluation. In A/B testing, we reset the environment and randomly choose the next policy to execute. The entire evaluation process for one initialization is repeated until all trials are completed.

\begin{wrapfigure}{r}{7cm}
\centering
\includegraphics[width=\linewidth,trim=0cm 0cm 0cm 0cm,clip]{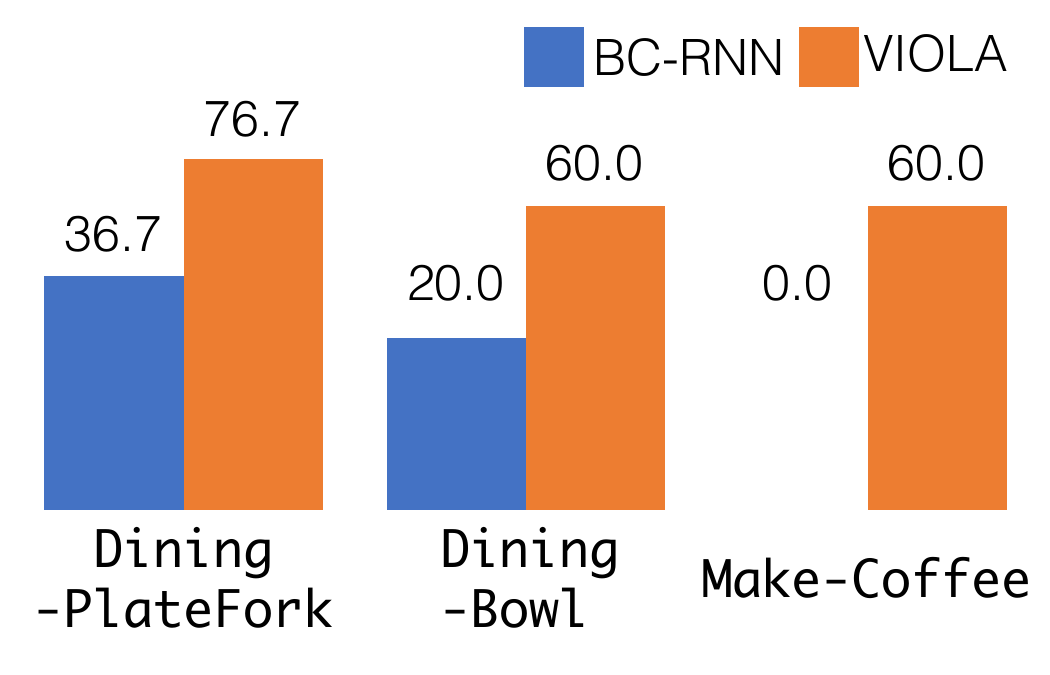}
        \caption[Real robot evaluation of \viola{}.]{\label{tab:viola:real-robot-results} Success rates (\%) in real robot tasks.}  
\end{wrapfigure}

\begin{figure}[t!]
    \centering
    \includegraphics[width=1.0\linewidth]{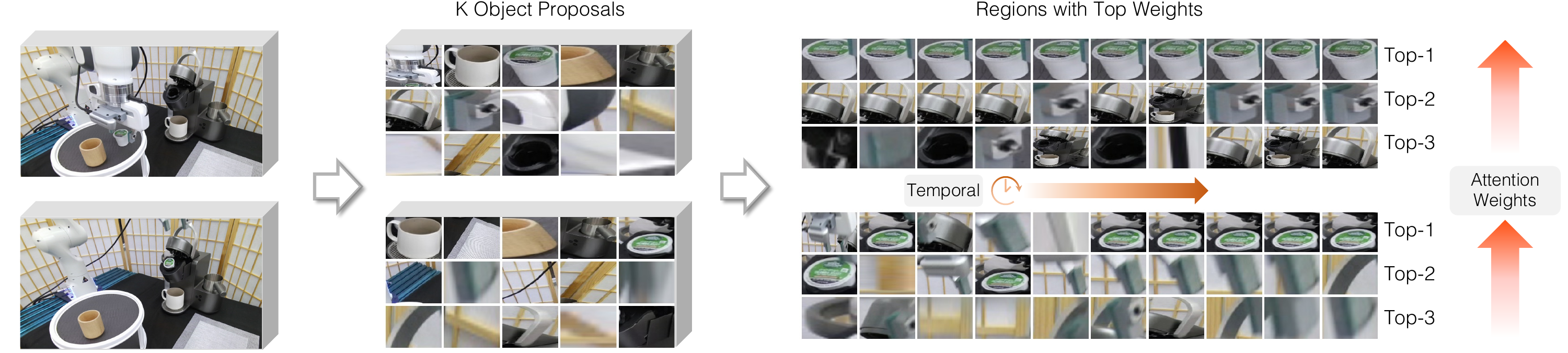}
    \caption[Visualization of important proposal regions weighted by transformer attention.]{Visualization of top-3 regions weighted most by transformer attention. When the robot is to grasp the K-cup, \viola{}'s top attention is over the K-cup across observations while it takes the robot fingers and coffee machine into account to help spatial reasoning. When the robot is to close the coffee machine, \viola{}'s top-1 attention is all over K-cups, the robot gripper, and the coffee machine, and it reasons over these regions to generate actions.}
    \label{fig:viola:attention-visualization}
\end{figure}

The quantitative evaluation in Figure~\ref{tab:viola:real-robot-results} shows that \viola{} outperforms \bcrnn{} by $46.7\%$ success rate on average. Qualitatively, we observe that the \viola{} policy can robustly grasp K-cups or open the coffee machine in the \coffee{} task, while the \bcrnn{} policy tends to reach the wrong positions where grasps are missed or to release the gripper mistakenly. We also qualitatively visualize the attention of \viola{} in Figure~\ref{fig:viola:attention-visualization} by showing the top 3 regions with the highest attention weights at each timestep of temporal observation. The figure shows that when the robot is to grasp the K-cup, \viola{}'s top attention is over the K-cup. At the same time, it also considers the robot fingers and coffee machine to facilitate spatial reasoning. When the robot is to close the coffee machine, \viola{}'s top-1 attention spreads over the K-cup, robot gripper, and the coffee machine, and the policy generates actions for closing through spatial reasoning over these regions. \loosepar{}

\paragraph{Choice of $\objectnum$ in simulation.} We answer question (5) with the help of simulation. While simulation images have a huge domain gap from the real world, we find that RPN can still localize objects on simulated images given its good priors of ``objectness.'' Here, we show the preliminary experiment that justified the choice of $\objectnum=20$ in simulation experiments. We quantitatively compute the coverage of object proposals from RPN over objects in the scene (including robots) and we evaluate the coverage with recall rates of object proposals. The recall rates in the simulation are easy to compute as we can easily access ground-truth object bounding boxes in the simulation. We iterate $\objectnum$ from $5$ to $40$ with an interval of $5$. We compute recall rates of object proposals by setting a hyperparameter IoU$=0.5$ (Intersection over Union). This hyperparameter means that we consider an object covered if the IoU of a proposal bounding box and the ground-truth bounding box is larger than $0.5$. The results are provided in Table~\ref{tab:viola:recall-rates}. We can see that when $\objectnum=20$, we have recall rates that are larger than $70\%$, which we consider as large coverage. Moreover, we can see that when $\objectnum$ is larger than 20, the marginal increase of recall rates is only $1\%$ for every $5$ more object proposals. So, we choose $\objectnum=20$ in our simulation experiments. 

\begin{table}[t]
\centering
  \begin{tabular}{lcccccccc}
    \toprule
    \textbf{$\objectnum$} & $5$ & $10$ & $15$ & $20$ & $25$ & $30$ & $35$ & $40$ \\
    \midrule
   \sort{} & 59.0& 69.0& 72.0& \textbf{74.0} & 75.0& 76.0& 77.0& 77.0
\\
   \stack{} & 67.0& 74.0& 77.0& \textbf{79.0} & 80.0& 81.0& 82.0& 83.0
 \\
   \kitchen{} & 72.0& 81.0& 83.0& \textbf{84.0} & 85.0& 86.0& 86.0& 87.0
\\
    \bottomrule
    \end{tabular}
    \caption[Recall rates of object proposals using the pre-trained RPN.]{\label{tab:viola:recall-rates} Recall rates (\%) of object proposals from pre-trained RPN on all simulation environments with IoU$=0.5$.}
\end{table}

\section{Summary}
\label{sec:viola:discussion}
In this chapter, we have presented \viola{}, an object-centric imitation learning approach to learning closed-loop visuomotor policies for robot manipulation. Our approach uses general object proposals to build the object-centric representation. We design a transformer-based policy to identify task-specific relevant regions for action generation. The results show the superior performances of \viola{} compared to state-of-the-art baselines in both simulation and the real world. We also validate our model designs through ablation studies, showing how each model component impacts policy performance. \viola{} serves as a general framework to develop sensorimotor learning based on \textit{object regularity}, obtaining generalizable behavioral cloning policies from a small amount of demonstration data. 

\paragraph{Limitations and Future Work.} In \viola{}, we have only considered using RGB images in the input data, which lacks the 3D information of objects. The missing 3D information prevents \viola{} from excluding the background visual elements in the input, which would limit the generalization ability of \viola{} when facing large visual variations, such as visual appearance change in table texture. Also, while \viola{} has achieved great success in learning robust visuomotor policies, the object-centric representation can be used in policies generally beyond imitation learning. One future direction is to use this object-centric representation in reinforcement learning policies. 

In the following chapter, we extend \viola{} to using object-centric 3D representation based on depth images and vision foundation models that are more powerful at localizing objects than the Detic RPN. With our improved representation, we can derive behavioral cloning policies that are robust to the large variety of visual distractions, such as backgrounds and camera viewpoints.

% GROOT

\chapter{Imitation Learning with Object-centric 3D Representations}
\label{chapter:groot}

In the last chapter, we introduced \viola{}, an object-centric imitation learning framework to learn sensorimotor skills for complex manipulation tasks.
However, \viola{} still faces difficulties in generalizing beyond the training environments from which the demonstration data is collected. Applying these methods to real-world tasks often entails time-consuming data collection and model re-training for each setting. As a result, imitation learning policies trained on a limited set of demonstrations are commonly evaluated in the same workspace with strictly controlled conditions, including fixed backgrounds and carefully positioned cameras that cannot be moved throughout the entire experiment.

In this chapter, we focus on training manipulation policies that are more generalizable than \viola{} while using a limited number of demonstrations. We argue that object-centric representations must be endowed with 3D-awareness. Object-centric 3D representations are vital for training imitation learning policies that handle visual variations naturally encountered in real-world deployments. The 3D-aware property lifts spatial reasoning from the 2D plane to a unified reference frame of 3D coordinates. This property improves the representation's spatial invariance against changes in camera viewpoints, capturing the regularity of objects better than 2D representations. By leveraging the 3D-aware property in object-centric imitation learning, we exploit the object regularity in our policy design to achieve generalizable manipulation policies.

\begin{figure}
    \centering
    \includegraphics[width=1.0\linewidth]{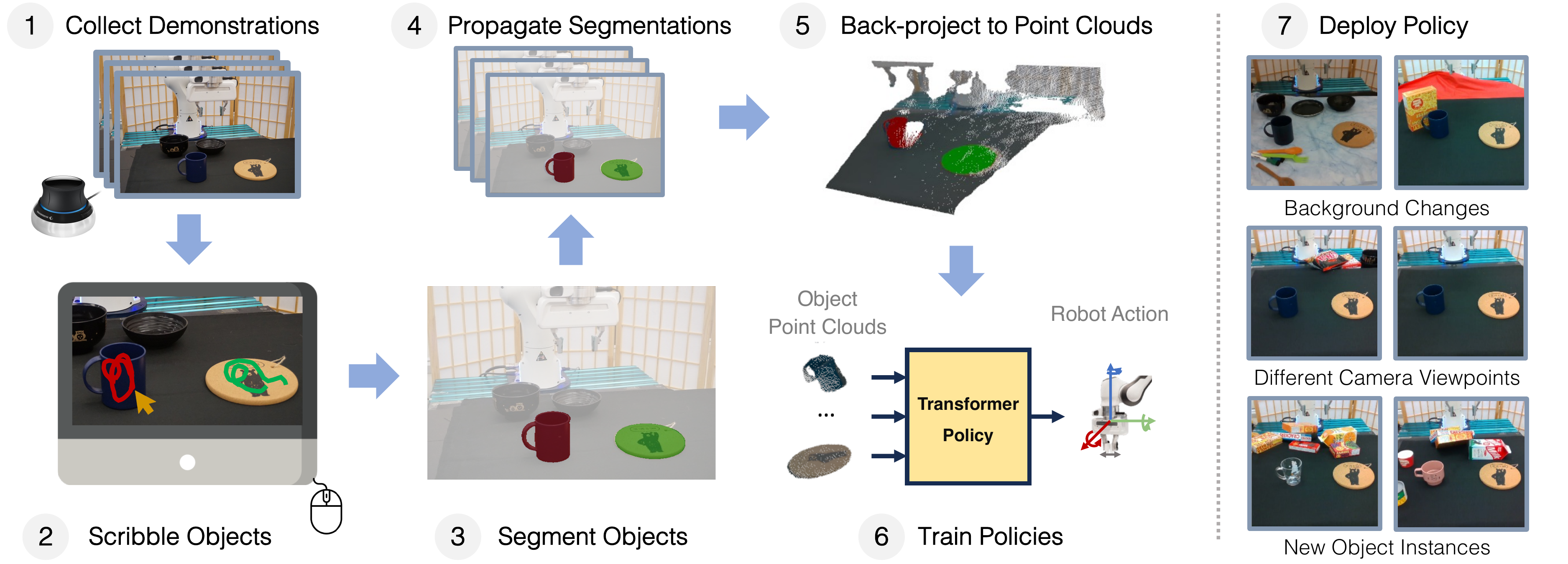}
    \caption[\groot{} overview.]{\textbf{\groot{} overview.} \groot{} learns closed-loop visuomotor policies from demonstrations under a single setup, and generalizes to new setups with unseen conditions, namely different visual distractions, changed camera angles, and new objects. }
    \label{fig:groot:overview}
\end{figure}

\section{GROOT}
\label{sec:4-overview}

\begin{figure}[t]
    \centering
    \includegraphics[width=1.0\linewidth, trim=0cm 0cm 0cm 0cm,clip]{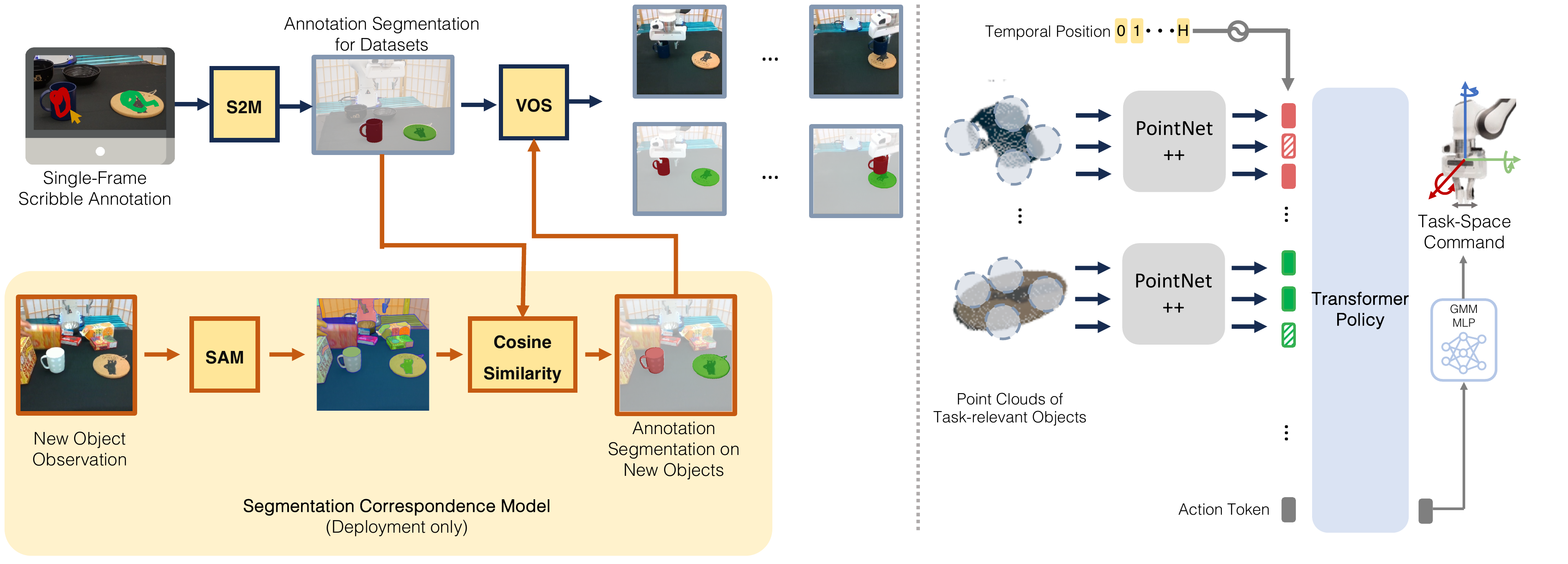}
    \vspace{-5mm}
    \caption[\groot{} model architecture.]{\textbf{\groot{} Model Architecture.} \groot{} leverages an interactive segmentation model, S2M, to obtain a single-frame annotation from demonstrators. Then a Video Object Segmentation model, XMem, propagates segmentation masks across time frames. The object masks are then back-projected into point clouds, and a transformer-based policy processes the point clouds to output actions. During deployment, \groot{} uses a segmentation correspondence model based on an open-vocabulary segmentation model (SAM) and a pretrained semantic feature model (DINOv2) to allow generalization to new objects.}
    \label{fig:groot:model}
\end{figure}

We present our method \groot{} (\emphasize{G}eneralizable \emphasize{RO}bot Manipulation Policies for Visu\emphasize{O}motor Con\emphasize{T}rol), an imitation learning method for learning generalizable policies of vision-based manipulation. Our work in this chapter was published at the 7th Annual Conference on Robot Learning, 2023~\cite{zhu2023groot}. 
Our core idea is to factorize raw RGB-D images observed from a calibrated camera into segmented point clouds of task-relevant objects, forming object-centric 3D representations for policy learning. Figure~\ref{fig:groot:model} illustrates the overall workflow of our method. In the rest of this section, we first introduce our object-centric 3D representations. Then, we describe our transformer-based policy. Finally, we introduce a new segmentation correspondence model that allows our policies to generalize to new object instances.

\subsection{Object-centric 3D Representations}
\label{sec:4-object-centric-3d}

Our main goal is to build representations with both object-centric and 3D-aware properties that allow policies to generalize beyond the training conditions. In the following, we describe the three major steps of building the object-centric 3D representations: 1) scribble annotation on images from demonstrations to identify task-relevant objects, 2) tracking the objects at the segmentation level to keep models task-focused and exclude the irrelevant visual factors; 3) back-projecting object segmentation to 3D point clouds that is invariant to camera view changes.

\paragraph{Scribble Annotation of Task-relevant Objects.}  
In order to build representations that consistently attend to task-relevant objects, the first step is to inform our model of what they are in image observations. Prior work of unsupervised object discovery requires extensive data while being limited to toy domains~\cite{locatello2020object}.  Instead, we adopt a semi-supervised approach that annotates objects on a single image with an instance segmentation mask after demonstrations are collected. To operationalize the semi-supervised approach, we leverage an interactive segmentation model, S2M~\cite{cheng2021modular}, to allow demonstrators to scribble on an image frame with a few mouse clicks, specifying the foreground objects involved in a specific manipulation task, namely the task-relevant objects. Notably, the annotation only needs to be done on \textit{a single image}, inducing negligible efforts from a demonstrator. Once we have the scribbled annotation, we can use a pretrained vision model to track object segmentation over images in the rest of the datasets. 

For scribble annotations, we select the first frame from a demonstration trajectory where task-relevant objects are visible for the demonstrator to annotate. We choose segmentation to localize objects for two reasons. First, segmentation outlines objects at the pixel levels, allowing a demonstrator to annotate objects with fine-grained contours. Second, segmentation can effectively exclude irrelevant visual features as opposed to less flexible representations such as proposal boxes as in Chapter~\ref{chapter:viola}, improving the robustness of policies.\loosepar{}

\paragraph{Tracking Task-relevant Objects Over Time.}
The first step of scribble annotation localizes task-relevant objects in a single frame. To keep policies focused on these objects while being robust to large visual variations, we need to track them across temporal frames.
We employ a pretrained Video Object Segmentation (VOS) model, XMem~\cite{cheng2022xmem}, that takes in the single-frame instance mask from the scribble annotation step and returns a sequence of instance masks that track objects across all the demonstration frames. For all trajectories in the demonstrations, we can apply XMem to propagate the annotation across trajectories by simply adding the annotated frame at the beginning of the trajectories. In this way, we can annotate the 2D segmentation of task-relevant objects on all images in the datasets. During deployment, we treat the annotated frame as the initial observation and subsequently stream the temporal observations to XMem. In this way, \groot{} can segment the objects from image observations in the dataset and back-project them into point clouds for policy inference.
\vspace{-0.1in}

\paragraph{Backprojection to Point Clouds.}
Tracking objects across time frames gives us 2D segmentation masks. However, objects in 2D images are tailored to the specific camera viewpoint. Henceforth, they are not robust to camera view changes. To overcome this issue, we need 3D-aware representations that do not overfit to particular camera angles. Concretely, we back-project the segmented objects from 2D images to point clouds using depth images~\cite{zhou2018open3d}. The point clouds can encode object geometry and allow spatial reasoning over object loations. Point clouds also allow easy generalization to new objects, which we describe later in Section~\ref{sec:4-nocs}. To represent the point clouds that are independent of camera locations, we transform the point clouds into a predetermined reference frame through SE(3) transformation based on the known camera extrinsic~\cite{lynch2017modern}. We choose the robot base as the reference frame of coordinates, which is a fixed-base frame of coordinates that does not change with joint configurations or camera setups. Consequently, the transformed point clouds are spatially invariant to changes in camera views, ruling out the potential of significant distribution shifts introduced by viewpoint changes. \loosepar{}

\subsection{Policy Design}
\label{sec:4-policy-design}

In order to harness our object-centric representations, we design a policy architecture, which first encodes the point clouds into a discrete set of tokens and processes them with a transformer-based architecture. \groot{} first divides each point cloud into clusters of points in the same procedure as in Point-MAE~\cite{pang2022masked}. 
The division encourages policies to attend to the local geometry of point clouds, which can be further improved by performing random masking over the input. Our ablation study in Section~\ref{sec:groot:results} confirms that random masking improves the robustness of policies when facing large camera changes. Before performing random masking, we pass each point cloud cluster through a shared PointNet~\cite{qi2017pointnet++} to compactly represent object geometry and spatial information in latent vectors, also referred to as tokens, making the training and inference of the transformer module more efficient than directly processing point clouds in the transformer.

Random masking prevents policies from overfitting to global features of point clouds such that the representations remain robust when partial point clouds are missing due to realistic noise in sensing or large shifts in camera viewpoints. These masked tokens are passed through a transformer-based architecture that processes object-centric representations. This transformer architecture is the same as the one used in Chapter~\ref{chapter:viola}. The same as in \viola{}, we pass a $\historytime+1$-timestep sequence of tokens, allowing the policy to keep track of the temporal information and mitigate the partial observability issue (\textit{e.g.}, a single image does not inform the velocities) and bypassing the non-Markovian property of gripper actions (\emph{e.g.}, a single image cannot differentiate ongoing actions of opening versus closing). Because input to transformers is permutation-invariant, we add sinusoidal positional encoding~\cite{vaswani2017attention} to the input tokens based on their positions in the temporal sequence (More details in Appendix~\ref{ablation_sec:groot:implementation}). The action generation of the transformer-based policy also follows the design in VIOLA, as we have described in Chapter~\ref{chapter:viola}.

% The action generation of the transformer architecture follows the design of prior work, where an action token is appended to the input sequence, and the transformer outputs the corresponding latent vector for performing downstream tasks. Such designs allow the output latent vector to implicitly encode the object-centric and 3D information from the input sequence through the self-attention mechanisms. The output latent vector is subsequently processed through a Gaussian-Mixture-Moel-based multi-layer perception (MLP)~\cite{mandlekar2021matters, bishop1994mixture}, outputting a distribution of robot action. During training, the network is optimized over a negative log-likelihood objective to mimic the distribution of actions from demonstrations. At test time, the robot samples action from the distribution at every decision-making step, performing closed-loop manipulation at $20$ Hz.

\subsection{New Object Generalization via Segmentation Correspondence}
\label{sec:4-nocs}
To generalize to new object instances during deployment, we also introduce a new segmentation correspondence model that propagates instance segmentation masks to new object instances without additional human annotations. This model is built based on two vision foundation models, SAM~\cite{kirillov2023segment} and DINOv2~\cite{oquab2023dinov2}. SAM is an open-vocabulary segmentation model that takes an RGB image as input and outputs a single-channel image that consists of a set of object masks. DINOv2 is a pretrained semantic feature model that takes an RGB image as input and outputs a latent vector that is a semantic embedding.

The segmentation correspondence model first uses SAM to generate a set of object masks from visual observations of deployment environments. Then, the model identifies the object masks semantically closest to the ones among the training objects. The closest match is determined by DINOv2. DINOv2 is optimized to map similar concepts to embeddings that have closer distances in a latent feature space than different concepts, making it possible to find the closest match by computing the cosine similarity scores between DINOv2 features of objects. In our model, we first compute the DINOv2 feature of every segmentation mask from SAM and every object mask in the annotated image. For each pair of SAM segmentation masks and annotated masks, we find the closest matching SAM masks by deciding the mask with the highest cosine similarity score. For every annotated object mask, we select a mask from SAM outputs with the highest similarity score, resulting in a new instance segmentation. Consequently, this segmentation correspondence model can propagate the object masks from training data to images from the test scenes that contain previously unseen objects. Once we have the segmentation of new objects, we can back-project the segmentation masks of the objects into point clouds. The point clouds of new objects are then used in the subsequent networks in \groot{} policies. \groot{} policies can generalize to the new instances due to the similarity between the point clouds of the new instances and the training objects. As shown in the right part of Figure~\ref{fig:groot:overview}, this correspondence model allows the policy to manipulate different mugs even though it only trains on a single mug during training.

\section{Experiments}
\label{sec:groot:experiments}
We conduct experiments to answer the following questions: 1) Are object-centric 3D representations in \groot{} better at generalization compared to baseline methods? 2) What design choices are essential for good performance of \groot{} policies? 3) Does \groot{} generalize to real-world settings in the face of background changes, different camera angles, and new object instances? 

% 4) Is our choice of point transformer encoder for encoding point clouds? 

\subsection{Experimental Setup}
\label{sec:groot:setup}

We use both simulation and real-robot experiments to evaluate \groot{} policies. 

\paragraph{Task Designs.} We use three simulation tasks based on Robosuite~\cite{zhu2020robosuite}. These three simulation tasks are namely ``Put The Moka Pot on The Stove,'' ``Put The Frying Pan on The stove,'' and ``Reposition The Yellow and White Mug.''  For real robot experiments, we design five tasks, namely ``Pick Place Cup,'' ``Stamp The Paper,'' ``Take The Mug,'' ``Put The Mug On The Coaster,'' and ``Roll The Stamp,'' which involve behaviors from pick-and-place to non-prehensile motions such as rolling the stamp. We describe the success conditions of these five tasks in Appendix~\ref{ablation_sec:groot:implementation}.

\paragraph{Data Collection.} We use a SpaceMouse to collect 50 human-teleoperated demonstrations for every real-world task. We also collect 50 demonstration using the SpaceMouse for every simulation task. 

\begin{figure}
  \centering
  \includegraphics[width=0.9\textwidth,trim={1\% 1\% 1\% 1\%},clip]{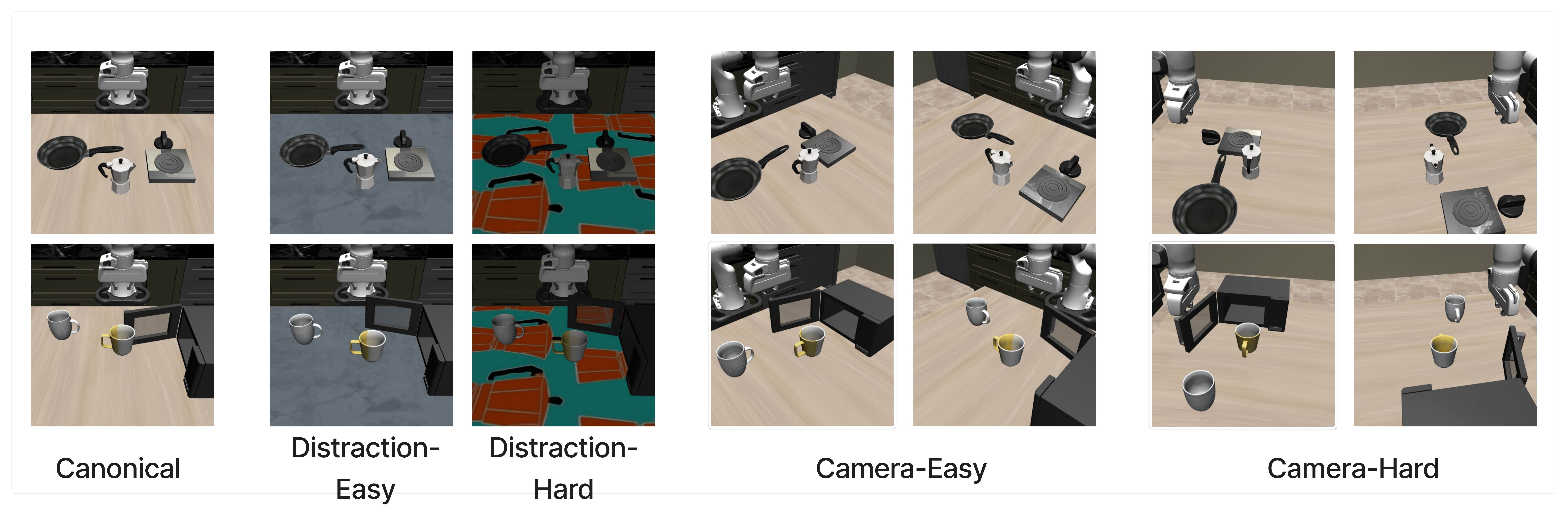}
  \caption[Visualization of simulation tasks in \groot{} experiments.]{Visualization of simulation tasks, for \canonical{}, \distractionseasy{}, \distractionshard{}, \cameraeasy{}, and \camerahard{}.}
  \label{fig:groot:sim-tasks}
\end{figure}

\paragraph{Generalization Settings in Simulation.} 
In our experiments, we have the following testing setups that evaluate the generalization abilities of our policies: 1) \canonical{}: All the objects and sensors are initialized in the same distributions as in demonstrations; 2) Three variants for generalization tests: \distracting{}, \camera{}, and \newobject{}. In the simulation, because of the limited availability of object assets, we mainly focus on generalizing different background changes and camera angles. Specifically, we consider two levels of difficulty for each generalization test in simulation, denoted as \distractionseasy{}, \distractionshard{}, \cameraeasy{}, and \camerahard{}. Figure~\ref{fig:groot:sim-tasks} visualizes these variations. Note that ``Put the moka pot on the stove'' and ``Put the frying pan on the stove'' share the same initial distributions, so we only visualize one of the tasks for showing the initialization settings. \distractionshard{} is the hard level as we changed both the lighting conditions and added the table cloth that has object patterns. \camerahard{} is harder than \cameraeasy{}, as the cameras are rotated with 40 more degrees. Such a large change in camera viewpoints results in a very different perspective on objects. With these settings, we evaluate the visual aspects of intra-task generalization as we have described in Section~\ref{sec:bg:generalization-il-policy}.

% Challenges the generalization abilities of policies.

\begin{figure}[h]
    \centering
    \includegraphics[width=\linewidth]{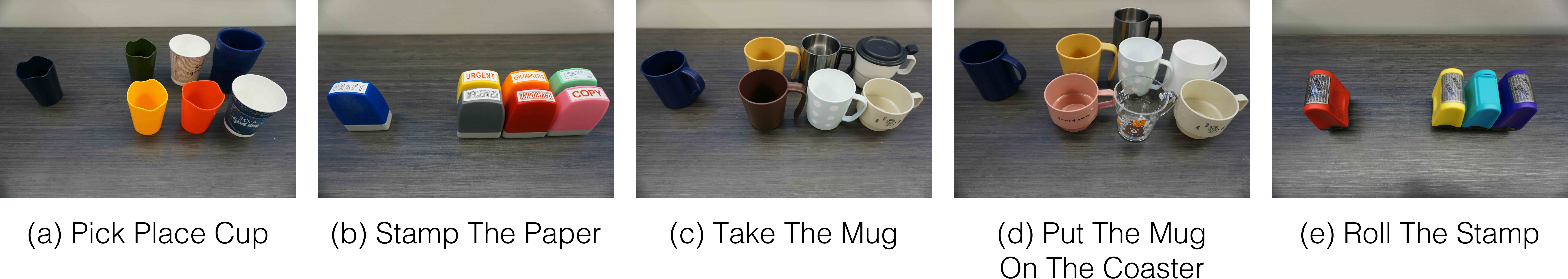}
    \caption[Visualization of object instances for evaluation of new object generalization.]{Visualization of objects used in real-robot experiments. In each image, the single object on the \textbf{left} side is used during data collection, and \textbf{all} the objects on the right side are \textit{unseen} during training. They are used for the evaluation of new object generalization in each task. }
    \label{fig:groot:new-objects}
\end{figure}

\paragraph{Evaluation Metric.} We quantify the policy performance by evaluating their task success rates. In the simulation, we evaluate all the models for 20 episodes and average the results over models trained with three random seeds. For real-robot experiments, we evaluate 10 episodes for each task in the settings of \canonical{} and \camera{}. For \distracting{}, we evaluate 30 episodes that cover a diverse range of table colors, lighting conditions, and distracting objects. For \newobject{}, we evaluate our policies on each new object instance three times with varied initial locations. Figure~\ref{fig:groot:new-objects} visualizes the set of new objects used in experiments. For policy evaluation, we limit the decision horizons to $600$ timesteps.

\paragraph{New Object Generalization.} In Figure~\ref{fig:groot:new-objects}, we show pictures of seen/unseen objects. We evaluate policies on multiple objects despite each policy being trained on a single object. These new objects require policies to be able to generalize across variations in color and geometry.

\subsection{Results}
\label{sec:groot:results}

\begin{table}[h!]
    \centering
    \resizebox{1.0\textwidth}{!}{%
\begin{tabular}{ll|cccc}
\toprule
Tasks & Evaluation Setting & \textsc{BC-RNN} & \textsc{VIOLA} & \textsc{MAE} & \groot{}\\
\midrule
Put The Moka Pot on The Stove & \canonical{} & 56.7 $\pm$ 15.5 & 61.7 $\pm$ 27.2 & 73.3 $\pm$ 6.2 & \textbf{80.0}$\pm$ 8.2  \\
{} & \distractionseasy{} & 11.7 $\pm$ 16.5 & \textbf{77.5}$\pm$ 17.5 & 56.7 $\pm$ 29.5 & 63.3 $\pm$ 13.1  \\
{} & \distractionshard{} & 0.0 $\pm$ 0.0 & 0.0 $\pm$ 0.0 & 31.7 $\pm$ 16.5 & \textbf{63.3}$\pm$ 8.5  \\
{} & \cameraeasy{} & 0.0 $\pm$ 0.0 & 0.0 $\pm$ 0.0 & 45.0 $\pm$ 20.4 & \textbf{81.7}$\pm$ 9.4  \\
{} & \camerahard{} & 0.0 $\pm$ 0.0 & 0.0 $\pm$ 0.0 & 26.7 $\pm$ 9.4 & \textbf{61.7}$\pm$ 9.4  \\
\midrule
Put The Frying Pan on The Stove & \canonical{} & 51.7 $\pm$ 6.2 & \textbf{90.0}$\pm$ 4.1 & 81.7 $\pm$ 4.7 & 65.0 $\pm$ 10.8  \\
{} & \distractionseasy{} & 10.0 $\pm$ 10.8 & 62.5 $\pm$ 2.5 & \textbf{73.3}$\pm$ 16.5 & 61.7 $\pm$ 11.8  \\
{} & \distractionshard{} & 0.0 $\pm$ 0.0 & 6.7 $\pm$ 4.7 & 52.5 $\pm$ 7.5 & \textbf{58.3}$\pm$ 11.8  \\
{} & \cameraeasy{} & 0.0 $\pm$ 0.0 & 1.7 $\pm$ 2.4 & \textbf{76.7}$\pm$ 20.5 & 71.7 $\pm$ 12.5  \\
{} & \camerahard{} & 0.0 $\pm$ 0.0 & 1.7 $\pm$ 2.4 & 65.0 $\pm$ 7.1 & \textbf{73.3}$\pm$ 8.5  \\
\midrule
Reposition The Yellow and White Mug & \canonical{} & 63.3 $\pm$ 16.5 & \textbf{85.0}$\pm$ 8.2 & 66.7 $\pm$ 8.5 & 73.3 $\pm$ 2.4  \\
{} & \distractionseasy{} & 53.3 $\pm$ 30.9 & 81.7 $\pm$ 6.2 & 38.3 $\pm$ 12.5 & \textbf{81.7}$\pm$ 12.5  \\
{} & \distractionshard{} & 0.0 $\pm$ 0.0 & 48.3 $\pm$ 31.7 & 6.7 $\pm$ 9.4 & \textbf{83.3}$\pm$ 6.2  \\
{} & \cameraeasy{} & 0.0 $\pm$ 0.0 & 3.3 $\pm$ 4.7 & 55.0 $\pm$ 16.3 & \textbf{61.7}$\pm$ 18.9  \\
{} & \camerahard{} & 0.0 $\pm$ 0.0 & 5.0 $\pm$ 4.1 & 8.3 $\pm$ 8.5 & \textbf{38.3}$\pm$ 12.5  \\
\bottomrule
\end{tabular}
    }
    \vspace{5pt}
    \caption[Quantitative evaluation of \groot{} policy generalization in simulation.]{Quantitative evaluation of policy generalization in simulation. The reported success rates (\%) are averaged over three random seeds. }
    \label{tab:groot:simulation}
\end{table}

We answer question (1) by comparing \groot{} policies in the three simulation tasks against three baselines that also learn closed-loop visuomotor policies: 1) \bcrnn{}~\cite{mandlekar2021matters, florence2019self} that directly conditions on raw perception, 2) \viola{} from Chapter~\ref{chapter:viola} that uses object proposal priors, and 3) \mae{}~\cite{seo2023multi} that uses random masking on image patches for learning policies. \bcrnn{} is a baseline to present the lowest bound of performance we can achieve in the canonical settings of the simulation tasks. Implementation details of the baselines can be found in Appendix~\ref{ablation_sec:groot:implementation}.

Table~\ref{tab:groot:simulation} presents the full simulation results that show the effectiveness of our representations in learning robust policies. As the table shows, while prior works such as \viola{} excel at the canonical settings in some tasks, \groot{} significantly outperforms baseline methods in most of the generalization tests by an amount of $22\%$ success rates compared to the best baseline, \mae{}. The comparison between \viola{} and \groot{} suggests that while task-agnostic object proposal priors are good at handling easy background changes, such priors significantly fail when the background changes are adversarial, as in the \distractionshard{} case. This result supports the choice of our representations that directly attend to task-relevant objects, excluding task-irrelevant visual factors. The comparison between \groot{} and \mae{} shows that random masking generally improves the policy robustness across different variations. However, naively dividing observations into patches impedes the policies from achieving the performance of \groot{} in hard levels of generalization tests. This comparison further validates the object-centric priors used in our policy learning that effectively exploit object regularity. \loosepar{}

Furthermore, in some simulation and real-robot tasks, policies are better in some generalization tests than in canonical setups. We posit that it results from the efficacy of the object-centric 3D representations in \groot{}, where visual variations in image observations cease to be the only limiting factor for policy generalization. In particular, policies trained over the supervised learning objective of behavioral cloning (BC) do not necessarily achieve the best task success rates, and the BC objective is only a \textit{surrogate} function of maximizing task success rates. As a result, our policies perform better in various generalization tests against visual variations. As the primary focus of \groot{} is to develop visuomotor policies that generalize to visual variations beyond the one in training conditions, we leave for future work to improve other aspects of the visuomotor policy learning.\loosepar{}

\begin{figure}
  \centering
  \begin{minipage}[b]{0.49\textwidth}
    \centering
    \includegraphics[width=\textwidth]{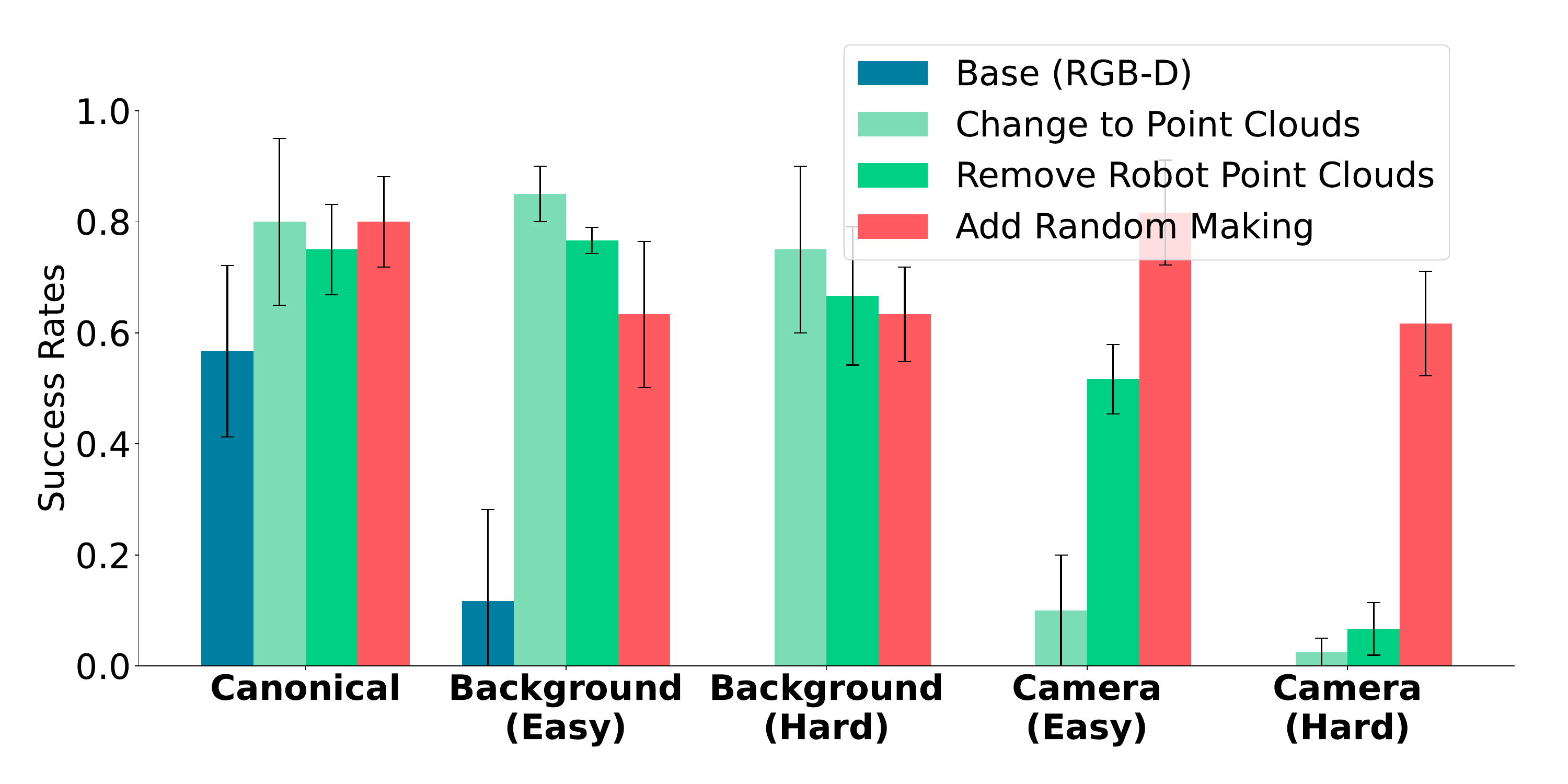}
    \caption[Ablation study of \groot design choices.]{Changes in success rates (\%) with design choices for \groot{}.}
    \label{fig:groot:ablation}
  \end{minipage}
  \hfill
  \begin{minipage}[b]{0.49\textwidth}
    \centering
    \includegraphics[width=\textwidth]{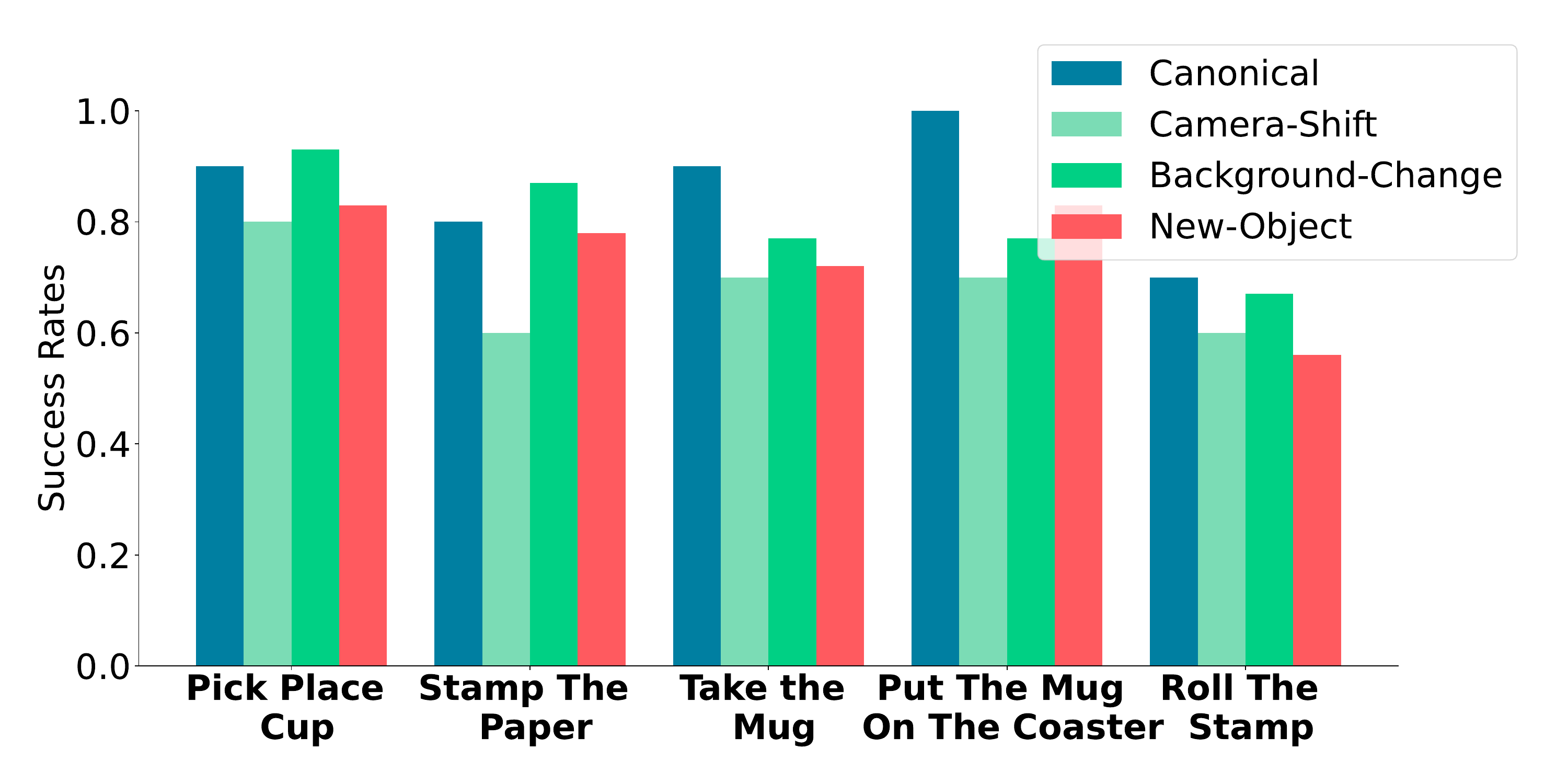}
    \caption[Evaluation of \groot{} policies in real-robot tasks.]{Success rates (\%) of \groot{} in real-robot tasks.}
    \label{fig:groot:real-robot}
  \end{minipage}
\end{figure}

\paragraph{Important Design Choices.} We present an ablation study on the task ``Put the moka pot on the stove'' to answer question (2), showing how our model takes advantage of the object-centric 3D representations. Figure~\ref{fig:groot:ablation} visualizes the change in performance when using point clouds, removing point clouds of the robots, and using random masking on point clouds for training policies. We start with the base ablation model that naively takes in RGB-D observations. This base model fails to generalize in all four test settings. To mitigate irrelevant visual factors, we adopt an ablation model that only takes point clouds of objects and the robot as inputs, enhancing generalization for different background changes and validating the effectiveness of our object-centric priors. Despite these improvements, the model struggles with camera view changes due to the inclusion of unseen robot parts in the observations. To address these distribution shifts, we exclude the robot point clouds from observations, which can be done using robot URDF models known \textit{a priori}. This design change improves the performance under minor camera view shifting yet struggles with substantial view changes. To accommodate large view changes, we incorporate random masking into the point cloud representation. Consequently, we achieve a substantial performance boost in handling camera view changes, albeit at a modest reduction in background change generalization capability.

\paragraph{Real Robot Experiments.} We answer question (3) by presenting results in Figure~\ref{fig:groot:real-robot}. The table shows that our policies successfully generalize to all three variations with high success rates despite the challenging settings. Our real-robot experiments validate the effectiveness of our approach in the experiments. Note that the policies still fail to achieve perfect success rates. Most of the failures come from missing the grasp by a few millimeters. This is within our expectation as the workspace cameras do not have enough resolution to account for this level of location errors, and we do not include the eye-in-hand cameras in our real robot experiments. Eye-in-hand cameras have been proven effective in learning robust grasping~\cite{hsu2022vision}. We expect a performance boost by adding the eye-in-hand camera for precise grasping, but this hardware setup is orthogonal to our research topic, and we leave this for future work.

\section{Summary}
\label{sec:groot:discussion}

In this chapter, we have introduced \groot{}, an imitation learning method for learning closed-loop manipulation policies based on object-centric 3D representations. Just like \viola{}, \groot{} is developed based on our understanding of \textit{object regularity} and is implemented using powerful vision foundation models that leverage object regularity operationally in behavioral cloning policies. \groot{} leverages object segmentation and point clouds to build effective representations that allow policies to generalize to background changes, different camera angles, and new object instances. Our results show that \groot{} outperforms state-of-the-art baselines in terms of its generalization abilities, and we validate our method in five real robot tasks, showcasing the superior generalization capabilities of \groot{} in settings beyond the training conditions.

\paragraph{Limitations and Future Work.} In this chapter, we have assumed only one instance of each task-relevant object is present in a task. When multiple instances of the same object category are present, the segmentation correspondence model would consider all of them as candidate objects, causing ambiguities that \groot{} cannot resolve. One possible future direction is to explore language-conditioned policies with spatial specifications to resolve the ambiguities.

Also, while \groot{} focuses on achieving wide generalization across visual variations, our formulation assumes the robot manipulator must remain the same. An interesting extension would be to study how to generalize the behavioral cloning policies on manipulators with different system identifications (robots of the same types but with different physical parameters) or different morphologies (e.g., trained on a tabletop manipulator and tested on mobile manipulators or even humanoids). Such policy transfer would require a unified design of action space and need to consider the robots' physical parameters, which we leave for future work.

In summary, we have introduced our methods that efficiently exploit object regularity in learning generalizable manipulation policies in both this chapter and the previous chapter. These two chapters constitute Part~\ref{part:I} of this dissertation. The following chapter marks the start of Part~\ref{part:II} of this dissertation, where we introduce our methods based on the concept of spatial regularity, explaining how robots can imitate manipulation skills from in-the-wild video observations.

\newpage{}
\part{Imitating From In-the-wild Video Observations}
\label{part:II}
\chapter{Open-world Video Imitation}
\label{chapter:orion}

So far, we described how to learn generalizable policies from teleoperated demonstrations. However, teleoperation demands high physical efforts that are laborious for users to teach robots, and it requires expertise that challenges people without any robotics background. A promising alternative is teaching robots through human videos of manipulation behaviors that can be recorded in everyday scenarios. These methods have great potential to tap into the readily available source of Internet videos that cover a wide distribution of human activities, paving the way for scaling up skill learning.

Prior work on learning from human videos has focused on pre-training representations and value functions~\cite{chen2021learning, nair2022r3m, ma2022vip, wang2023mimicplay, xiong2021learning}. However, these methods do not explicitly estimate the object states and the object interactions in 3D space where robot motions are defined. In contrast, our goal is for a robot to imitate a task robustly in the ``open world.'' In this setting, a robot is expected to operate under different visual and spatial conditions from the video demonstration, without knowing the ground-truth object states or the demonstrated behaviors beforehand. Notably, these videos are actionless, as they do not contain any labels of robot actions. Such actionless videos are equivalent to state-only demonstrations used in the problem of ``Imitation from Observation''\cite{torabi2021imitation}. To tackle the aforementioned setting, we pose a new problem named \textit{open-world imitation from observation}, formulated in Section~\ref{sec:orion:open_world}.\loosepar{}

The key to tackling the problem is to equip robots with spatial understanding. The spatial understanding allows a robot to interpret the invariant spatial relations between objects of interest while ignoring the location variations of irrelevant objects, exploiting spatial regularity as we motivated in Section~\ref{fig:bg:spatial_regularity}. 

Developing a method that can exploit spatial regularity is only possible due to the recent advances in vision foundation models~\cite{kirillov2023segment, oquab2023dinov2}. Pre-trained on Internet-scale visual data, these models excel at understanding open-vocabulary visual concepts. The vision foundation models can help localize task-relevant objects in videos without known object categories or access to physical states. Consequently, we can model the important spatial relations between the localized objects and the manipulator, whether human or robotic. Then, we can exploit \textit{spatial regularity}, mentioned in Section~\ref{fig:bg:spatial_regularity}, in deriving policies by identifying the consistent patterns in the spatial relations that regularly determine the completion of task goals.

Our method in this chapter marks the first step toward tackling the problem of open-world imitation from observation. We first introduce our formulation of the new problem and then describe our proposed method. A preprint version of our work in this chapter is on arXiv~\cite{zhu2024orion}.
%, where a robot imitates how to interact with objects given \textit{a single video only} while deployed in environments with different visual backgrounds and unseen spatial configurations. 

% In this chapter, we consider using RGB-D video demonstrations where a person manipulates a small set of task-relevant objects with their single \textit{hand}, recorded with a stationary camera. These videos are actionless or state-only, as they do not come with any ground-truth action labels for the robot.

\section{Open-world Imitation From Observation}
\label{sec:orion:open_world}

We consider the general problem of a robot learning a vision-based manipulation task from a single video demonstration in an open-world setting. In this problem, the robot is not pre-programmed to have access to either ground-truth categories or locations of objects in the video.

We formulate a vision-based manipulation task following our CMDP formulation with the sparse-reward setting in Section~\ref{sec:bg:open-world-formulation}. The robot does not have direct access to the ground-truth task reward or the physical states of objects. The task context is specified through a single \textit{actionless} video $V$. $V$ serves as a \textit{state-only} demonstration, consisting of a video stream that captures a demonstrator performing a manipulation task. Each frame in $V$ is an image in either RGB or RGB-D format.
The use of actionless videos as state-only demonstrations aligns with the problem of ``imitation from observation''\cite{torabi2021imitation}. Notably, our formulation requires a robot to generalize in the open-world setting instead of the training environments. Therefore, we refer to our problem as \textit{open-world imitation from observation}.

Open-world imitation from observation is a general problem that focuses on how robots learn from in-the-wild video observations. A solution to the problem can take any form and is not limited to end-to-end neural network policies. However, a solution to this complete newly posed problem is beyond the scope of our work or any existing work. In this dissertation, we make certain assumptions and focus on solving a restricted version of the problem. In Section~\ref{sec:orion:problem_assumptions} and Section~\ref{sec:okami:problem_assumptions} from the next chapter, we introduce the assumptions made before introducing our methods. By explicitly stating the assumptions, we provide a clear problem scope that each of our methods can tackle.
  
% In this dissertation, we describe our efforts to tackle the problem with certain assumptions being made. 

% Later in this chapter and the next chapter, we describe the assumptions made before introducing our methods. By explicitly stating the assumptions, we provide a clear problem scope that each of our methods.

\section{\orion{}}
\label{sec:orion:method}

\begin{figure}
    \centering
    \includegraphics[width=1.0\linewidth]{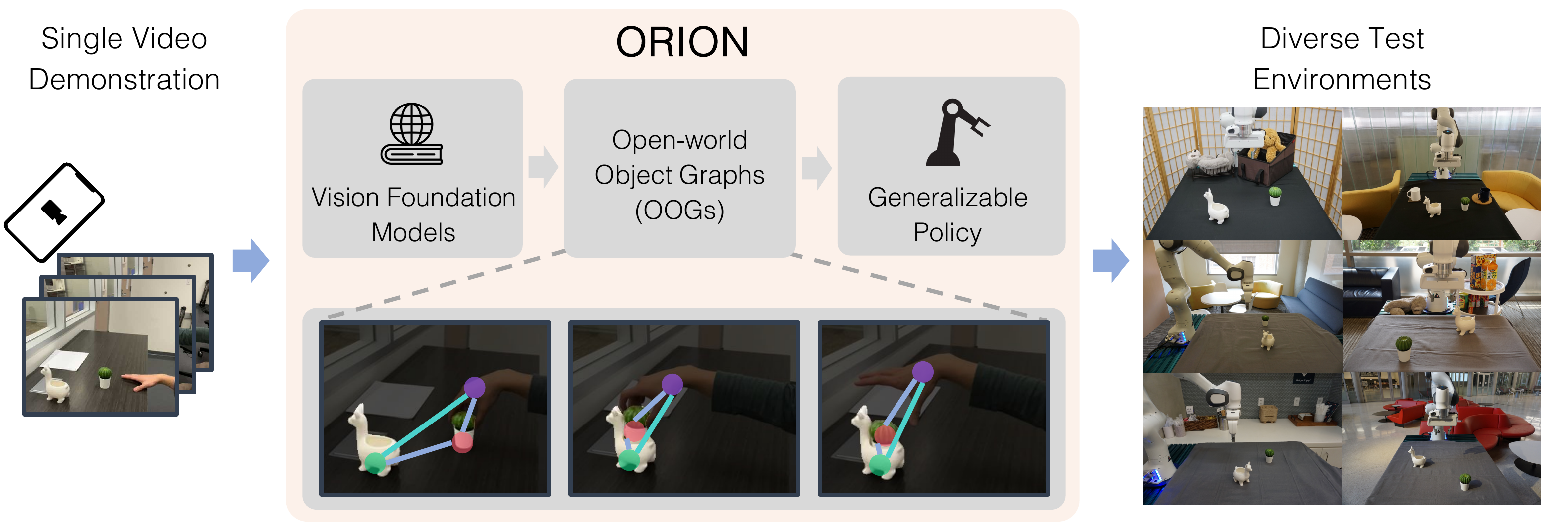}
    \caption[\orion{} overview.]{\textbf{Overview of \orion{}.} \orion{} tackles the problem of imitating manipulation from single human video demonstrations. \orion{} first extracts a sequence of Open-World Object Graphs (OOGs), where each OOG models a keyframe state with task-relevant objects and hand information. Then \orion{} leverages the OOG sequence to construct a manipulation policy that generalizes across varied initial conditions, specifically in four aspects: visual background, camera shifts, spatial layouts, and novel instances from the same object categories. }
    \label{fig:orion:overview}
\end{figure}

We introduce~\orion{} (\emphasize{O}pen-wo\emphasize{R}ld video \emphasize{I}mitati\emphasize{ON}), an algorithm that allows a robot to mimic how to perform a manipulation task given a single human video, $V$. To effectively construct a policy $\pi$ from $V$, \orion{} employs a learning objective based on an object-centric prior. The goal is to create a policy $\pi$ that directs the robot to move objects along 3D trajectories that mimic the directional and curvature patterns observed in $V$, relative to the objects' initial and final positions. This objective is based on the observation that objects are likely to achieve target configurations by moving along trajectories similar to those in $V$. Key to \orion{} is generating a manipulation plan from $V$, which serves as the spatiotemporal abstraction of the video that guides the robot to perform a task. A plan is a sequence of keyframes, where each keyframe specifies an initial state or a subgoal state in $V$ and is abstracted in an object-centric representation.\loosepar{}

We first describe the assumptions made in this chapter for tackling the problem of \textit{open-world imitation from observation}. Then, we introduce our formulation of the object-centric representation of a state, Open-world Object Graph (OOG), used in \orion{}. Based on our proposed representation, we describe the algorithm that derives a robot policy that imitates the human video to complete the task.\loosepar{}

\subsection{Problem Assumptions}
\label{sec:orion:problem_assumptions}

\begin{figure}[ht!]
    \centering
    \includegraphics[width=1.0\linewidth, trim=0cm 0cm 0cm 0cm,clip]{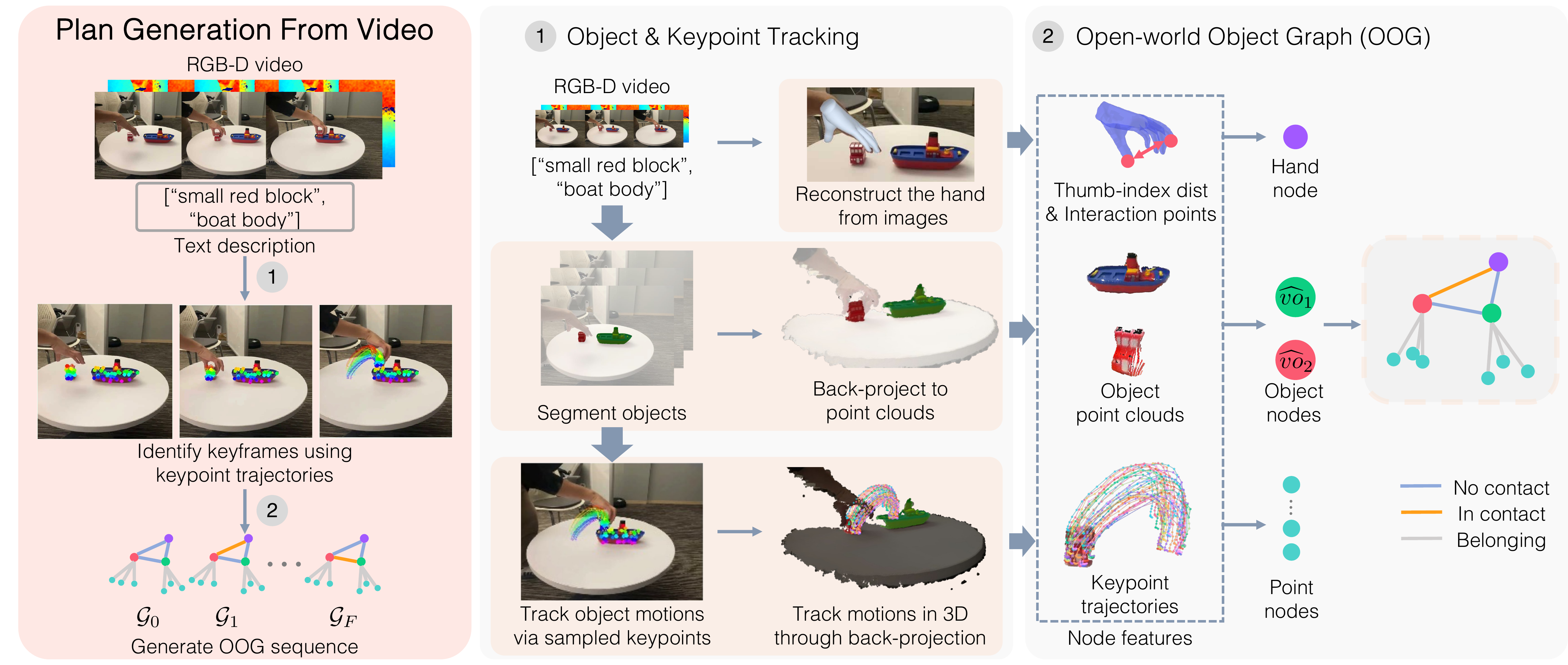}
    \caption[Plan generation in \orion{} method.]{\textbf{Plan Generation.} \orion{} generates a manipulation plan from a given video $V$ for subsequent action synthesis. \orion{} first tracks the objects and keypoints across the video frames. Then, \textit{keyframes} are identified based on the velocity statistics of the keypoint trajectories. \orion{} generates an  Open-world Object Graph (OOG) for every keyframe, resulting in a sequence of OOGs that serves as the spatiotemporal abstraction of $V$. A detailed description of data structure of an OOG can be found in Table~\ref{tab:oog_data_structure}.}
    \label{fig:orion:model-part1}
\end{figure}

\orion{} is our first method to tackle the problem of \textit{open-world imitation from observation}. As mentioned in Section~\ref{sec:orion:open_world}, we describe the following assumptions upon which \orion{} is built:

\begin{itemize}
    \item The demonstration video $V$ captures a person manipulating task-relevant objects with a single \textit{hand}.
    \item The robot can access common sense knowledge through foundation models pre-trained on internet-scale data.
    \item To address inherent video ambiguities arising from irrelevant objects or unclear user specifications (such as whether an object's color is task-relevant), each $V$ is accompanied by a complete list of English descriptions of task-relevant objects. The descriptions include user-specified feature descriptions, such as color, which uniquely define the object instances in $V$. This list is represented in a comma-separated format. For example, ``[\texttt{small red block}, \texttt{boat body}]'' is used for the task shown in Figure~\ref{fig:orion:model-part1}.
\end{itemize}

In the following, we introduce \orion{} which leverages these assumptions.

\subsection{Open-world Object Graph}
\label{sec:orion:oog}

At the core of our approach is a graph-based, object-centric representation. This representation is known as Open-world Object Graphs (OOGs). OOGs use open-world vision models to represent visual scenes with task-relevant objects and the hand. As a result, they naturally exclude distracting factors in visual data. Additionally, they localize the task-relevant objects regardless of their spatial locations (See Section~\ref{sec:orion:plan-generation}).

We denote an OOG as $\mG$. At the high level, each object node corresponds to a task-relevant object from the result of open-world vision models. Every object node comes with node features consisting of colored 3D point clouds derived from RGB-D observations. This node feature indicates what and where objects are and represents their geometry information. Additionally, to inform the robot where to interact with objects (e.g., where to grasp), we introduce the specialized \textit{hand node}, which stores the interaction cues such as contact points and the grip status (open or closed) that can be directly mapped to the robot end-effector during execution. At the low level, each point node corresponds to a keypoint that belongs to a task-relevant object. Every point node has a feature that represents the keypoint's trajectory in 3D space. Trajectories of the keypoints explicitly models how an object should be moved during a manipulation task. By motion features of a point node in $G_\kfindex$, we mean a 3D trajectory of a keypoint between keyframes $\kfindex$ and $\kfindex+1$ (Here, we use $\kfindex$ to index a keyframe). 

In an OOG, all the object and hand nodes are fully connected, reflecting real-world spatial relationships. Additionally, the edges are augmented with a binary attribute, indicating if two entities are in contact to represent their pairwise contact relations. The two entities can be either two objects or an object and the hand. This attribute allows our developed algorithm to check the set of contact relations that are satisfied, retrieving the matched OOG from the generated plan (see Section~\ref{sec:orion:plan-generation}). The low-level point nodes are connected to their respective object node, indicating a belonging relationship. Table~\ref{tab:oog_data_structure} summarizes the variables to explain the data structure of an OOG. This table is intended for easy reproducibility of the proposed method.

\begin{table*}[htbp]
    \centering
    \begin{tabular}{@{} m{3cm} m{3cm} m{7cm} @{}}
        \toprule
        \textbf{Node/Edge} & \textbf{Type} & \textbf{Attributes} \\
        \midrule
        $\mG.vo_i$ & Object Node & 3D point cloud of an object. \\
        $\mG.vh$ & Hand Node & Hand mesh and locations of the thumb and index finger. \\
        $\mG.vp_{ij}$ & Point Node & A trajectory of a TAP keypoint between two keyframes, recorded in xyz positions. \\
        \midrule
        $\mG.eo_{ik}$ & Object-Object Edge & A binary value of contact or not. \\
        $\mG.eh_{i}$ & Object-Hand Edge & A binary value of contact or not. \\
        $\mG.ep_{ij}$ & Object-Point Edge & The presence of an edge represents the belonging relation, and no specific feature is attached.\\
    \bottomrule
    \vspace{0.5cm}
    \end{tabular}
    \caption[Data Structure of an Open-world Object Graphs.]{Data Structure of an OOG. For a given OOG, $\mG=(\mathcal{V}, \mathcal{E})$, it has $\mathcal{V}=\{\mG.vo_i\} \cup \{\mG.vh\} \cup \{\mG.vp_{ij}\}$, and $\mathcal{E}=\{\mG.eo_{ik}\} \cup \{\mG.eh_{i}\} \cup \{\mG.ep_{ij}\}$.  An illustration of this data structure can be found in the right part of Figure~\ref{fig:orion:model-part1}.
    }
    \label{tab:oog_data_structure}
\end{table*}

\subsection{Manipulation Plan Generation From Human Video}
\label{sec:orion:plan-generation}

We describe the first part of \orion{} (see Figure~\ref{fig:orion:model-part1}), where our method automatically annotates the video and generates a manipulation plan from a given human video, $V$. In this chapter, we refer to a manipulation plan as a spatiotemporal abstraction of $V$ that centers around the object states and their motions over time. This abstraction encodes all the important information that describes how a task should be completed. Our core insight is that we can cost-effectively model a task with object locations at some keyframe states where the set of satisfied contact relations is changed and abstract a majority of intermediate states into 3D motions of objects. Concretely, a plan is represented as a sequence of OOGs, $\{ \mG_\kfindex \}_{\kfindex=0}^{\KF}$ which corresponds to $\KF + 1$ keyframes identified from $V$, with $\mG_0$ representing the initial state. 

\paragraph{Tracking task-relevant objects.} \orion{} first localizes task-relevant objects in the video $V$. Given $V$ and the list of object descriptions mentioned in Section~\ref{sec:orion:problem_assumptions}, 
\orion{} uses an off-the-shelf vision model, Grounded-SAM~\cite{liu2023grounding}, to annotate video frames with segmentation masks of the task-relevant objects. Specifically, Grounded-SAM takes an image and a list of English descriptions of objects as inputs and outputs a single-channel image, i.e., a segmentation mask. In practice, Grounded-SAM is computationally demanding, so we reduce the computation by exploiting object permanence to track the objects. Specifically, \orion{} annotates the first video frame with Grounded-SAM, and then propagates the segmentation to the rest of the video frames using a Video Object Segmentation Model, Cutie~\cite{cheng2023putting}. Cutie takes a video sequence and the segmentation mask of the first frame as inputs and outputs a sequence of segmentation masks, where each image corresponds to a video frame. 
\loosepar{}

\paragraph{Discovering keyframes.} After annotating the locations of task-relevant objects, we track their motions across the video to discover the keyframes based on the velocity statistics of object motions. This design is based on the observation that changes in object contact relations, e.g., transitioning from free space motion to grasping an object, are often accompanied by sudden changes in object motions. However, keeping full track of object point motion using techniques like optical flow estimation requires heavy computation. The tracking quality is susceptible to slight hand occlusions involved in the observations. We use a Track-Any-Point (TAP) model, namely CoTracker~\cite{karaev2023cotracker}, that can track a subset of points in a long-term video with explicit occlusion modeling. A TAP model takes a video and sampled keypoints in the first video frame as inputs and outputs a set of the keypoints' trajectories that track the movement of keypoints across the video. TAP models have been successfully applied to track object motions in robot manipulation~\cite{vecerik2023robotap, wen2022you}. To use CoTracker in our method, we first sample keypoints within the object segmentation of the first frame and track the trajectories across the video. The changes in velocity statistics are straightforward to detect based on the TAP trajectories, where we discover the keyframes using a standard unsupervised changepoint detection algorithm~\cite{killick2012optimal}.

\paragraph{Generating OOGs from $V$.} Once \orion{} discovers the keyframes, it generates an OOG at each keyframe to model the state of task-relevant objects and the human hand. The creation of OOG nodes can reuse the results from the annotation process: for object nodes, the point clouds for node features are obtained by back-projecting the object segmentation with depth data. Each point node corresponds to a sampled keypoint, and its motion feature, 3D trajectory, is back-projected from the 2D trajectory of the sampled keypoint using depth data. Additionally, hand information is required to specify the interaction points with task-relevant objects and the grip status to be mapped to the robot gripper. We use a hand-reconstruction model, HaMeR~\cite{pavlakos2023reconstructing}, which gives a reconstructed hand mesh that pinpoints the hand locations at each keyframe. The distances between the fingertips of the mesh help determine the grip status, i.e., whether it is open or closed. 

With all the node information, \orion{} establishes the edge connections between nodes in OOGs, representing contact relations. Since all object and hand locations are computed in the camera frame while the camera extrinsic of $V$ is unknown, there is ambiguity when deciding the spatial relations between objects. We exploit the assumption of tabletop manipulation, where a table is always present with its normal direction aligned with the z-axis of the world coordinate system. Under this tabletop assumption, \orion{} estimates the transformation matrix of the table plane and transforms all the point cloud features in OOGs to align with the xy plane of the world coordinate (Full details appear in Appendix~\ref{ablation_sec:orion:implementation}). Then, the contact relations in each state can be determined based on the spatial relations and the computed distances between point clouds. The relations allow \orion{} to match the test-time observations with a keyframe state from the plan and subsequently decide which object to manipulate (see Section~\ref{sec:orion:action-synthesis}). In the end, \orion{} generates a complete OOG for each discovered keyframe.

\begin{figure}
    \centering
    \includegraphics[width=1.0\linewidth, trim=0cm 0cm 0cm 0cm,clip]{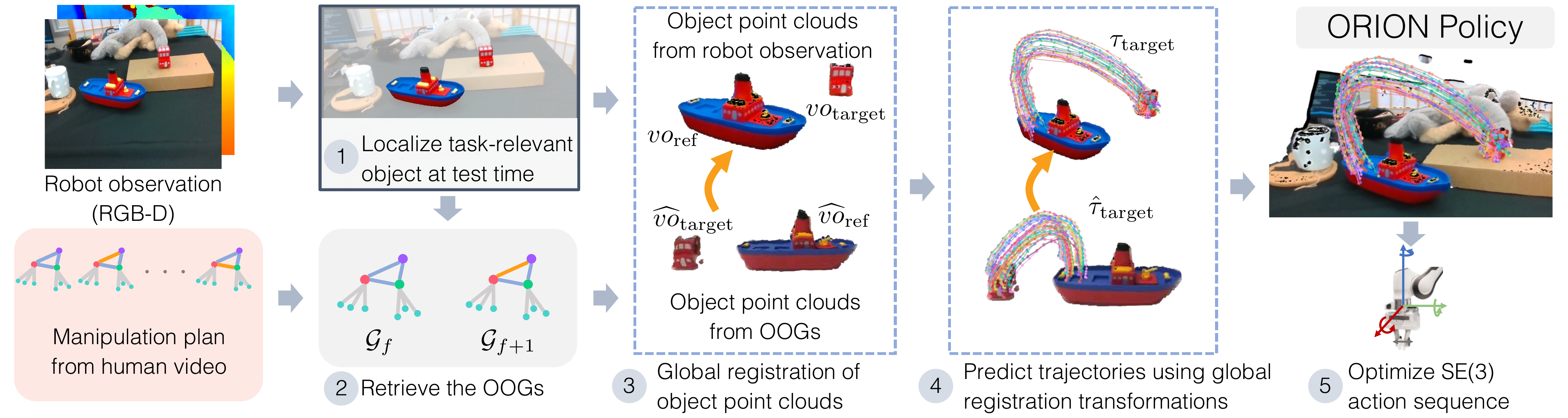}
    \caption[Action synthesis in \orion{} method.]{\textbf{Action Synthesis.} \orion{} first localizes task-relevant objects at test time and retrieves the matched OOG from the generated manipulation plan. Then \orion{} uses the retrieved OOGs to predict the object motions by first computing global registration of object point clouds and then transforming the observed keypoint trajectories from video into the workspace. The predicted trajectories are then used to optimize the SE(3) action sequence of the robot end effector, which is subsequently used to command the robot.}
    \label{fig:orion:model-part2}
\end{figure}

\subsection{Action Synthesis from Manipulation Plans}
\label{sec:orion:action-synthesis}
Given a manipulation plan, \orion{} derives a policy that synthesizes actions, detailed in Figure~\ref{fig:orion:model-part2}. The manipulation policy is derived based on the aforementioned learning objective that aims to achieve object motion similarity. The action synthesis comprises three major steps: 1) identify a keyframe from the plan that matches the current observation, 2) predict object motions, and 3) use the predictions to optimize the robot actions for the robot controller to execute. The policy repeats these three steps until a task is completed or fails based on the success conditions of the task. The success conditions of evaluated tasks are detailed in Section~\ref{sec:orion:experiments}.

\paragraph{Retrieving OOGs From The Plan.} \orion{} identifies the keyframe and retrieves OOGs to help decide what next actions to take. At test-time, \orion{} localizes objects in the new observations and estimates contact relations using the same vision pipeline as described in Section~\ref{sec:orion:plan-generation}. Then \orion{} retrieves the OOG that has the same set of contact relations as the current state, allowing us to identify a pair $(\mG_{\kfindex}, \mG_{\kfindex+1})$, where $\mG_{\kfindex}$ is the retrieved graph and $\mG_{\kfindex+1}$ the graph of the next keyframe. This pair of graphs provides sufficient information to decide which object to manipulate next, termed the \textit{target object}. We denote its point cloud at keyframe $\kfindex$ as $\widehat{\objectnode}_{\target}$, and its keypoint trajectories as $\hat{\kptraj}_{\target}$. We denote nodes from human videos with a hat to differentiate them from the nodes of robot rollout that do not have the hat. 
A target object is the one in motion due to manipulation between two keyframes, and it is determined by computing the average velocity per object using motion features in $\mG_{\kfindex}$. At the same time, another object called the \emph{reference object}, is involved in changing contact state relations from $\mG_{\kfindex}$ to $\mG_{\kfindex+1}$ and serves as a spatial reference for the target object's movement. We use the point cloud of the reference object at \textit{next keyframe} $\kfindex+1$, as the reference object might have location changes due to object interactions, and using the updated information from the next keyframe gives us an accurate prediction of the trajectories. Once the target and reference objects are determined, we can localize the corresponding objects in the new observations, and their point clouds are denoted as  $\objectnode_{\target}$ and $\objectnode_{\reference}$, respectively.

\paragraph{Predicting Object Motions.} Given the target and reference objects from keyframes $\kfindex$ and $\kfindex+1$, we predict the motion of the target object in the current state by warping the keypoint trajectories estimated from $V$. To warp the trajectories, we first identify the initial and goal locations of keypoints in the new configuration using information from the OOG pair. We use global registration of point clouds to align $\widehat{\objectnode}_{\target}$ with $\objectnode_{\target}$ and $\widehat{\objectnode}_{\reference}$ with $\objectnode_{\reference}$~\cite{choi2015robust} , giving us two transformations to compute the new starting and goal positions of target object keypoints conditioned on where the reference object is. Then we normalize $\hat{\kptraj}_{\target}$ with its starting and goal locations, obtaining $\hat{\kptraj}^{\text{norm}}_{\target}$. $\hat{\kptraj}^{\text{norm}}_{\target}$ only contains the directional and curvature patterns that are independent of the absolute locations of the initial and the goal keypoints. Then, we scale $\hat{\kptraj}^{\text{norm}}_{\target}$ back to the workspace coordinate frame using the new starting and goal locations, resulting in new keypoint trajectories of the target object $\kptraj_{\target}$.

\paragraph{Optimizing Robot Actions.} Once we obtain $\kptraj_{\target}$, we optimize for a sequence of SE(3) transformations that guide the robot end-effector to move. The SE(3) transformations are optimized to align the keypoint locations from the previous frames to the next frames along the predicted trajectories: 
\begin{equation}
    \min_{\tcp_0, \tcp_1, \dots,  \tcp_{t_{(\kfindex+1)} - t_{\kfindex}}} \sum_{i=0}^{t_{(\kfindex+1)} - t_{\kfindex}}(\kptraj_{\target}({i+1}) - \tcp_{i}\kptraj_{\target}({i}))
    \label{eq:orion:optim}
\end{equation}
where $\kptraj_{\target}(i)$ $(0 \leq i \leq t_{(\kfindex+1)} - t_{\kfindex})$ represents the keypoint locations at timestep $i$ along the trajectory. This optimization process naturally allows generalizations over spatial variations, as the action sequence always conditions on a new location instead of overfitting to absolute locations. To further specify where the gripper should interact with the object and whether it should be open or closed, we augment the resulting SE(3) sequence with the interaction information stored in the hand node. We implement a low-level controller that combines inverse kinematics (IK) and joint impedance control to follow the end-effector trajectory $\tcp_0, \tcp_1, \dots,  \tcp_{t_{(\kfindex+1)} - t_{\kfindex}}$ precisely and compliantly. 

The resulting \orion{} policy is robust to visual variations due to the use of open-world vision models. It also generalizes to different spatial locations because we choose to both represent object locations in object-centric frames and optimize actions independent of specific positions. \loosepar{}

\section{Experiments}
\label{sec:orion:experiments}

In this section, we design experiments to answer the following questions on the effectiveness of \orion{} and its important design choices. 1) Is \orion{} effective at constructing manipulation policies given a single human video in the open-world setting? 2) To what extent does the object-centric abstraction improve the policy performance? 3) Is it critical to model the object motions with keypoints and the TAP formulation? 4) Does \orion{} retain the policy performance given videos taken in different conditions? 5) Does \orion{} effectively scale to long-horizon manipulation tasks?

\subsection{Experimental Setup}
\label{sec:orion:experimental-setup}

\begin{figure}[ht!]
    \centering
    \includegraphics[width=\linewidth]{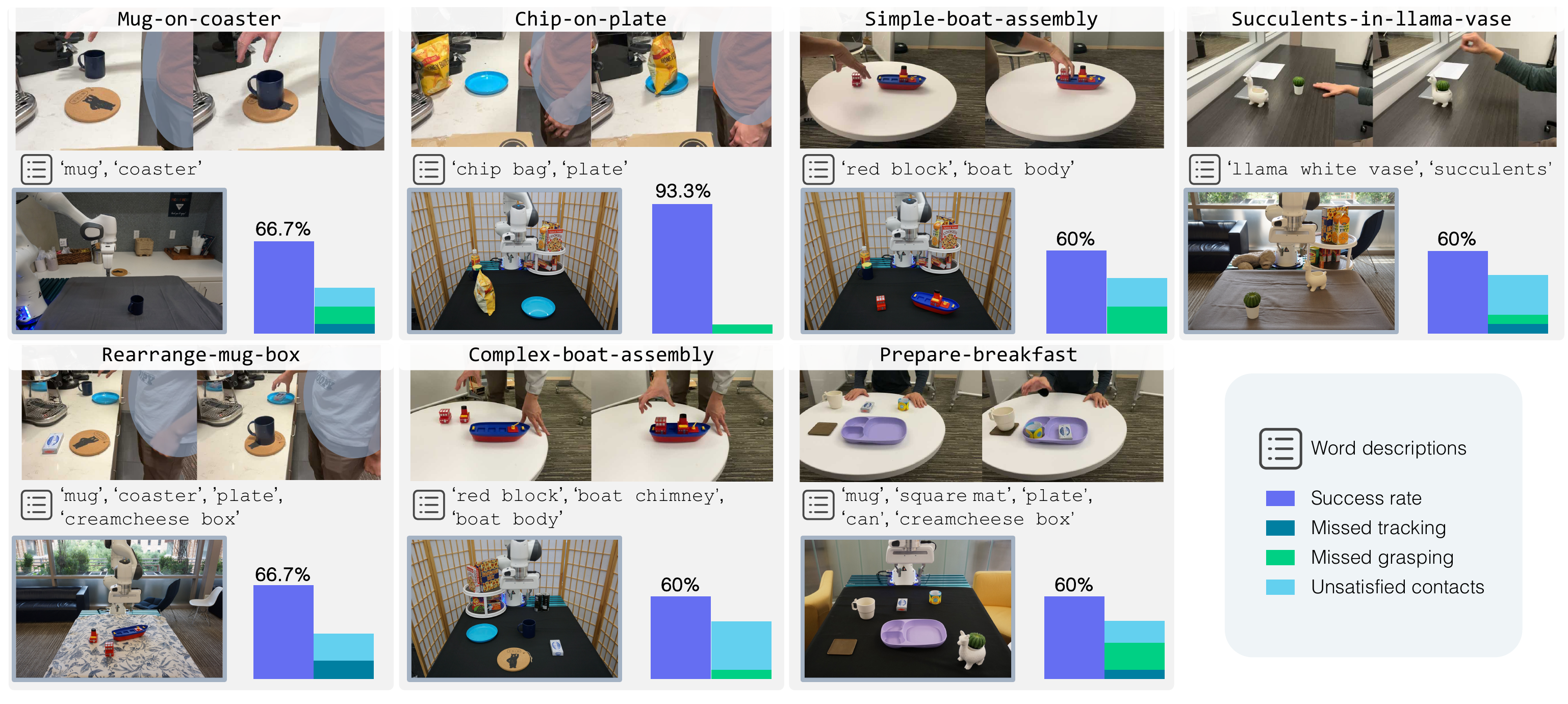}
    \caption[Evaluation tasks in \orion{} experiments.]{Visualization of evaluation tasks. Each block corresponds to one task. The top of a block illustrates the initial and final frames of a human video. Below the top images is the list of word descriptions provided along with the video. The lower part of a block shows two things side-by-side: an example image of the robot's initial state in the task and the evaluation of \orion{}, including the success rates and the failure rates separated into three failure modes.}
    \label{fig:tasks}
\end{figure}

We design experiments to test our method's efficacy. These experiments involve providing the robot with videos captured in everyday scenarios. These scenarios naturally include visual backgrounds and camera setups that differ from the robot's setup. Specifically, we record an RGB-D video of a person performing each of the seven tasks in everyday scenarios like an office or a kitchen. We use an iPad for recording, which comes with a TrueDepth Camera, and we fix it on a camera stand. During test time, the robot receives visual data through a single RGB-D camera, Intel Realsense435, and executes its policy in its space to evaluate policies. A Franka Emika Panda robot is used in all the experiments.\loosepar{}

\paragraph{Task Descriptions.} We design the following seven tasks to evaluate the policy performance: 1) \mugcoaster{}: placing a mug on the coaster; 2) \simpleboat{}: putting a small red block on a toy boat; 3) \chip{}: placing a bag of chips on the plate; 4) \llama{}: inserting succulents into the llama vase; 5) \rearrange{}: placing a mug on a coaster and placing a cream cheese box on a plate consecutively; 6) \complexboat{}: placing both a small red block and a chimney-like part on top of a boat. 7) \breakfast{}: placing a mug on a coaster and putting a food box and a can on the plate. 
The first four are ``short-horizon'' tasks, and the last three are ``long-horizon'' tasks. Here, ``short-horizon'' refers to tasks that only require one contact relation between two objects, while ``long-horizon'' refers to those that require more than one contact relation. The full description of the success conditions for each task is provided in Appendix~\ref{ablation_sec:orion:tasks}.

In practice, we record the success and failure of a rollout as follows. If \orion{} matches the observed state with the final OOG from a plan, we mark a trial as success as long as we observe that the object state indeed satisfies the success condition of a task as described above. Otherwise, if the robot generates dangerous actions (bumping into the table) or does not achieve the desired subgoal after executing the computed trajectory, we consider the rollout as a failure, and we manually record the failure.

\paragraph{Evaluation Protocol.} As described in the experimental setup, the videos naturally include various visual backgrounds and camera perspectives that are significantly different from the robot workspace. Therefore, we only need to intentionally vary two dimensions before evaluating each trial of robot execution, namely the spatial layouts and the new object instances. 
Furthermore, the new object generalizations are included in the tasks \mugcoaster{} and \chip{} as mugs and chip bags have many similar instances. As for the other five tasks, no novel objects are involved, but we extensively vary the spatial layouts of task-relevant objects for evaluation. Task performance is measured as the success rate averaged over 15 real-world trials. Aside from success rates, we also categorize the failed executions into three types: \textit{Missed tracking} of objects due to failure of the vision models, \textit{Missed grasping} of objects during execution, and \textit{Unsatisifed contacts} where the target object configurations are not achieved for reasons other than the previous two failure types, such as placing an object at a wrong location, or placing an object in an unstable way such that the object falls over.

\paragraph{Baselines.} To understand the model capacity and validate our design choices, we compare \orion{} with baselines. Since no prior work exists that matches the exact setting of \orion{}, we adopt the most important components from prior works and treat them as baselines to compare against \orion{}. Specifically, we implement the following two baselines: 1) \textsc{Hand-motion-imitation}~\cite{wang2023mimicplay, bharadhwaj2023zero}, which is a baseline that predicts robot actions by learning from the hand trajectories. The rest of the parts remain the same as \orion{}. We use this baseline to show whether it is critical to compute actions centering around objects. 2) \textsc{Dense-Correspondence}~\cite{ko2023learning, heppert2024ditto}, which is a baseline that replace the TAP model in \orion{} with a dense correspondence model, optical flows. This baseline is used to evaluate whether our choice of TAP model is beneficial. For this ablation study, we conduct experiments on \mugcoaster{} and \simpleboat{} to validate our model design, covering the distribution of common daily objects and assembly manipulation that requires precise control.

\begin{figure}
\centering
\begin{minipage}{\linewidth}
\centering
\includegraphics[width=\linewidth]{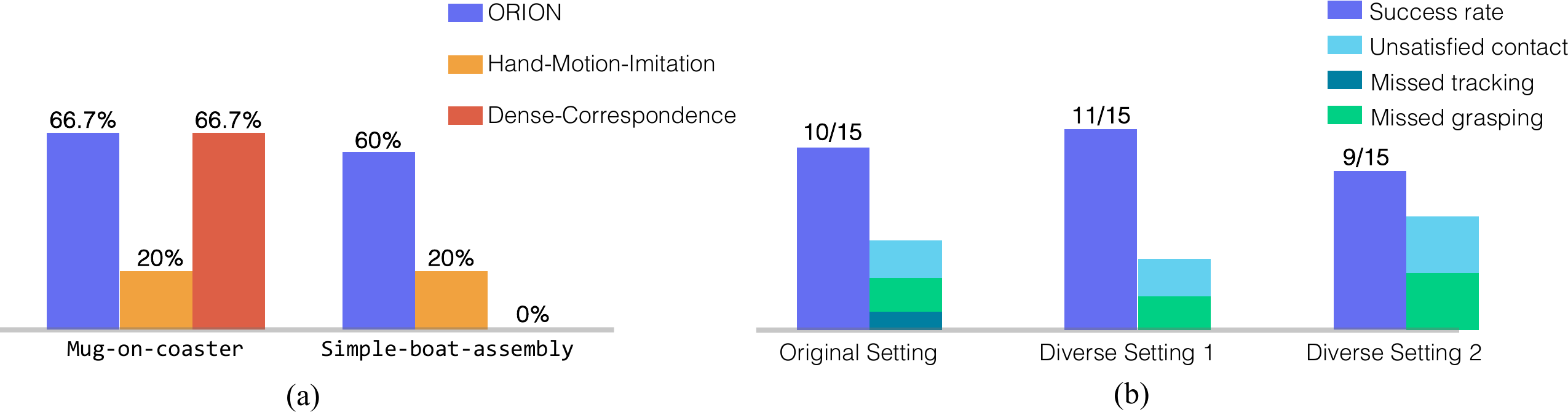}
\caption[Evaluation of \orion{} policies.]{\textbf{Experimental Evaluation of \orion{} Policies. } (a) Experimental comparison between \orion{} and the two baselines, namely \textsc{Hand-Motion-Imitation} and \textsc{Dense-Correspondence}. (b) Ablation study on using videos of the same task recorded in three different settings. We select the task \mugcoaster{} for conducting this ablation. We show the number of successful trials out of 15 total trials on the bar plots for each setting. Figure~\ref{fig:orion:supp-video} visualizes the different settings in this experiment.}
\label{fig:orion:ablation}
\end{minipage}
\end{figure}

\subsection{Results}
\label{sec:results}

Our evaluations are presented in Figures~\ref{fig:tasks} and~\ref{fig:orion:ablation}.
% , and \todo{~\ref{}}. 
We answer question (1) by showing the successful deployment of the \orion{} policies, while no other methods are designed to be able to operate in our setting. Furthermore, \orion{} yields an average of $69.3\%$ success rates, which validates our model design in learning from a single human video in the open-world setting.

We then answer question (2), showing the comparison results in Figure~\ref{fig:orion:ablation} against the baseline, \textsc{Hand-motion-imitation}. The baseline yields low success rates in both tasks. Concretely, \textsc{Hand-motion-imitation} typically succeeds in trials where the initial spatial layouts are similar to the one in $V$. Its major failure mode is not reaching the target object configuration, e.g., misplacing the mug on the table while not being on top of the coaster. These results imply that learning from human hand motion from $V$ results in poor generalization abilities of policies, supporting the design choice of \orion{}, which focuses on object-centric information. \loosepar{}

We further answer question (3) by comparing the performance between \orion{} and the baseline, \textsc{Dense-Correspondence}, shown in Figure~\ref{fig:orion:ablation} (a). We observe that the optical flow baseline performs worse on \simpleboat{} compared to \mugcoaster{}. With our further investigation, we find that the optical flow baseline discovers keyframes in the middle of smooth transitions as opposed to changes in object contact relations, resulting in a manipulation plan that leads to the synthesis of completely wrong actions. This finding further supports our choice of using TAP keypoints to discover the keyframes instead of optical flows. \loosepar{}

\begin{figure*}[htp]
    \centering
    \includegraphics[width=1.0\linewidth, trim=0cm 0cm 0cm 0cm,clip]{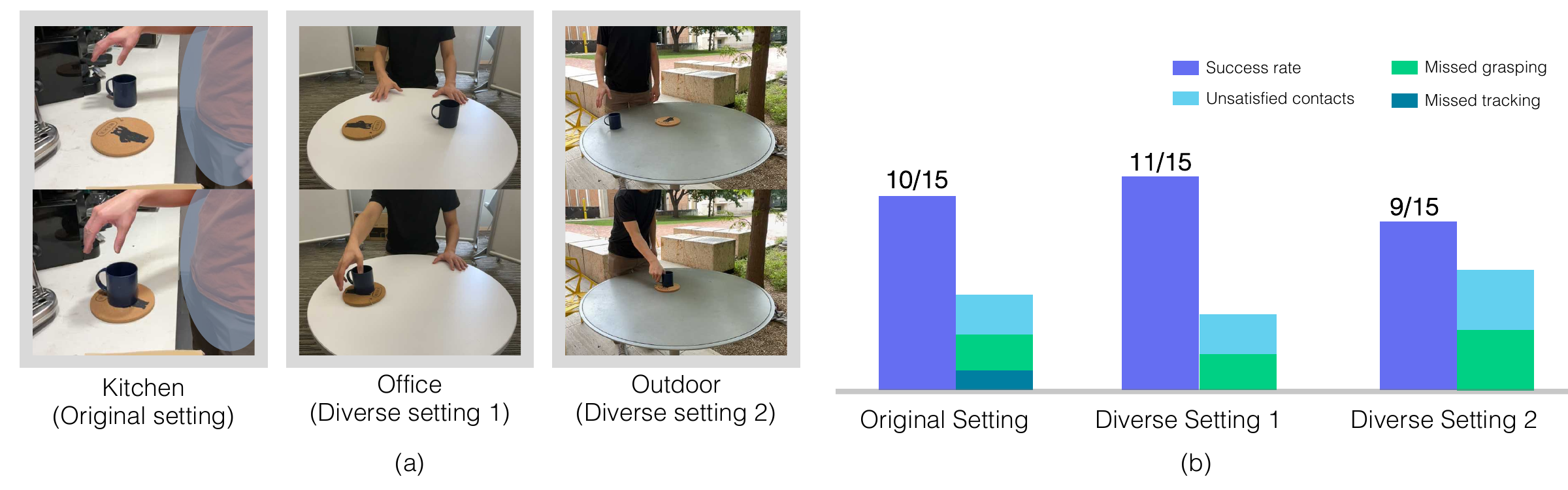}
    \caption[Visualization of three videos of the same task and the comparison result.]{(a) Visualization of initial and final frames of the three videos of the task~\mugcoaster{}, recorded under different conditions. (b) The performance plot of \orion{} using different videos.}
    \label{fig:orion:supp-video}
\end{figure*}

To answer question (4), we use the task \mugcoaster{} and conduct controlled experiments on it. Specifically, we construct policies from two additional videos of the same task, recorded in different environments with different spatial layouts. Then, we compare the two policies against the original one and test them using the same set of evaluation conditions. Figure~\ref{fig:orion:supp-video} shows the three videos taken in very different scenarios: kitchen, office, and outdoor. The video taken in the kitchen scenario is used in the major quantitative evaluation, termed ``Original setting.'' The other two settings are termed ``Diverse setting 1'' and ``Diverse setting 2.'' These three videos inherently involve varied visual scenes and camera perspectives. We conduct an ablation study where we compare policies imitated from these three videos. The result of the ablation study is shown in Figure~\ref{fig:orion:ablation}. The results show that there is no statistically significant difference in the performance, demonstrating that \orion{} is robust to videos taken under very different visual conditions. 

To answer question (5), we show that \orion{} is effective in scaling to long-horizon tasks. This conclusion is supported by the performance among the pairs of \mugcoaster{} versus \rearrange{} and \simpleboat{} versus \complexboat{}. In these two pairs, both the short-horizon tasks are subgoals of their long-horizon counterparts, yet we do not see any performance drop between the two. This result indicates that \orion{} excels at scaling to long-horizon tasks without a significant drop in policy performance. \loosepar{}

\section{Summary}
\label{sec:orion:discussion}

In this chapter, we introduced open-world imitation from observation, where robots learn to manipulate objects by watching a single video demonstration in the \textit{open-world setting}. 
By constructing spatiotemporal abstractions from video, \orion{} effectively leverages spatial regularities to transfer skills between robots and humans.

\paragraph{Limitations and Future Work.} \orion{} assumes RGB-D videos as inputs, which limits its applicability to lots of readily available videos that are in the format of monocular RGB. One promising future direction involves extending the system to learn manipulation policies from RGB images with dynamic camera views. 
Also, \orion{} only imitates kinematic motions from videos, falling short of imitating forceful interactions from observations~\cite{holladay2024robust}. Another direction is to explore forceful manipulation tasks beyond purely kinematic motions, such as turning screws.

\orion{} establishes object correspondence between demonstration and execution through global registration of point clouds. While effective, this geometric-only approach may face challenges with symmetrical objects where texture information becomes crucial for avoiding ambiguity. Future work could integrate both semantic and geometric object properties to establish more robust correspondence than using global registration.

This chapter has focused on learning from videos that capture a single human hand demonstrating skills and transferring the skills to a single-arm tabletop manipulator equipped with a parallel-jaw gripper. In the next chapter, we will propose another framework to enable \textit{humanoid robots} to imitate human videos. We will show that our single-video imitation methodology is general and can be applied to complex scenarios involving bimanual, dexterous manipulation.\loosepar{}

\definecolor{formalshade}{rgb}{0.85, 0.95, 0.95}
\newenvironment{formal}{%
  \def\FrameCommand{%
    \hspace{0.1pt}%
    {\color{darkgray}\vrule width 2pt}%
    {\color{formalshade}\vrule width 4pt}%
    \colorbox{formalshade}%
  }%
  \MakeFramed{\advance\hsize-\width\FrameRestore}%
  \noindent\hspace{-4.55pt}%
  \begin{adjustwidth}{}{5pt}%
  \vspace{-7pt}\vspace{0.1pt}%
}
{%
  \vspace{0.1pt}\end{adjustwidth}\endMakeFramed%
}

\chapter{Object-aware Kinematic Retargeting for Humanoid Manipulation Imitation}
\label{chapter:okami}

In the last chapter, we introduced the problem of \textit{open-world imitation from observation} and proposed \orion{} to tackle the problem for tabletop manipulators. In this chapter, we continue to develop an algorithm that exploits spatial regularity while learning from video observations, but with a focus on humanoid robots. Our proposed method endows humanoids with the ability to imitate manipulation tasks by watching single-video human demonstrations. Humanoid robots have an anthropomorphic physique that makes them easy to deploy in human-centric environments. With recent advances in hardware design and increased commercial availability, humanoid robots are emerging as a promising platform to co-exist with humans in our living and working spaces.

Enabling humanoids to imitate from single videos presents a significant challenge. The video does not have action labels, yet the robot has to learn to perform tasks in new situations beyond what's demonstrated in the video. As described in Chapter~\ref{chapter:orion}, \orion{} synthesizes the future object motion trajectories and uses them to synthesize robot actions through SE(3) optimization. However, such a process is computationally prohibitive for humanoid robots due to their high degrees of freedom and joint redundancy~\cite{asfour2003human}. In this chapter, we leverage the fact that humans and humanoids share similar kinematic structures, making it feasible to directly retarget human motions to robots~\cite{Gleicher1998RetargettingMT, darvish2019whole}. Prior retargeting techniques focus on free-space body motions~\cite{cheng2024expressive, nakaoka2005task, hu2014online, choi2020nonparametric, demircan2010human}, lacking the contextual awareness of objects and interactions needed for manipulation. To address this shortcoming, we introduce the concept of ``object-aware retargeting.'' By incorporating object contextual information into the retargeting process, the resulting humanoid motions can be efficiently adapted to the locations of objects in open-ended environments.

\begin{figure}[t!]
    \centering
    \includegraphics[width=1.0\linewidth]{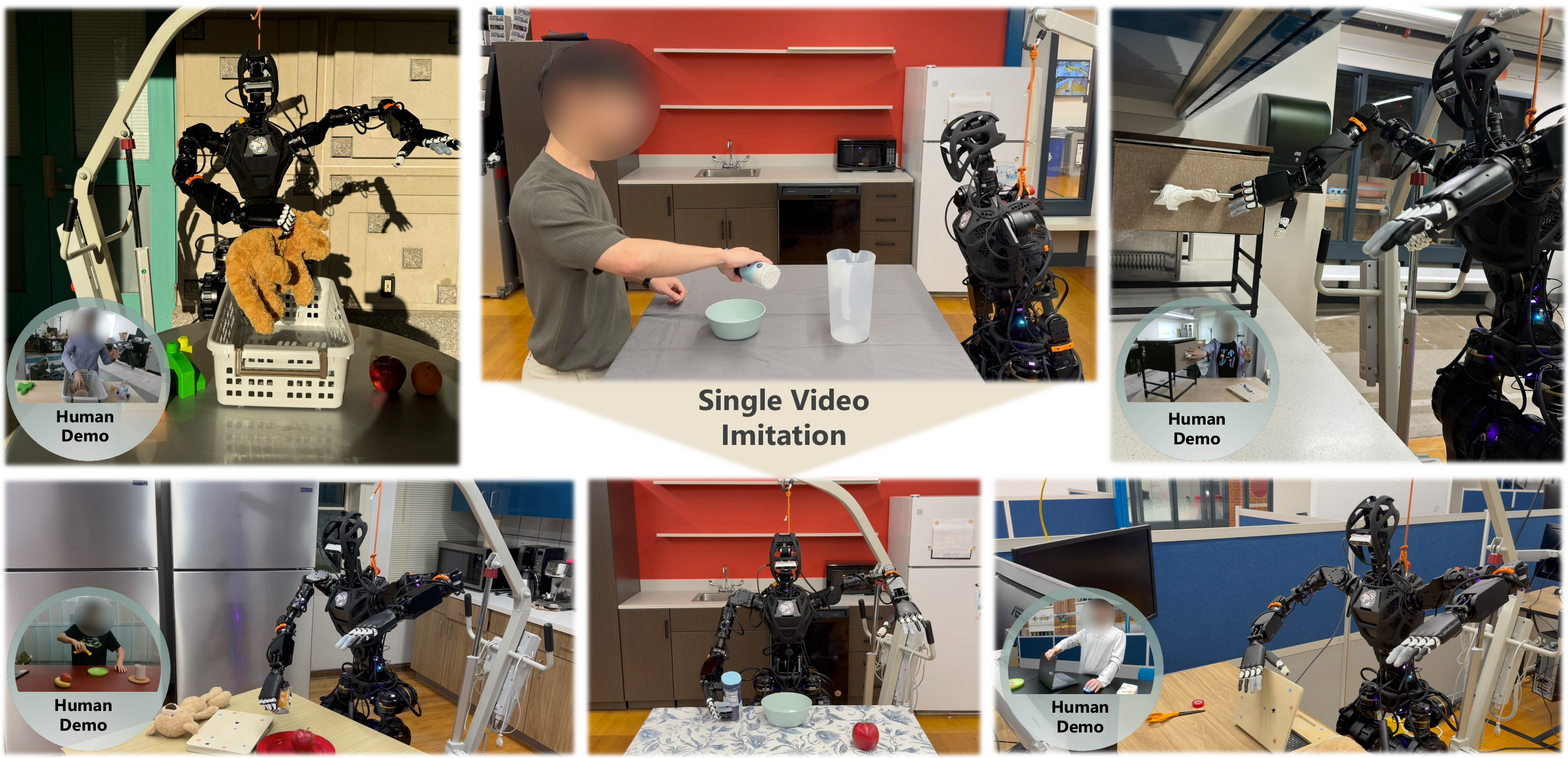}
    \caption[\okami{} overview.]{This chapter focuses on enabling a human user to teach the humanoid robot how to perform a task by providing a single-video demonstration.}
    \label{fig:okami:overview}
\end{figure}

\section{\okami{}}
\label{sec:okami:method}
\begin{figure}[ht!]
    \centering
    \includegraphics[width=1.0\linewidth]{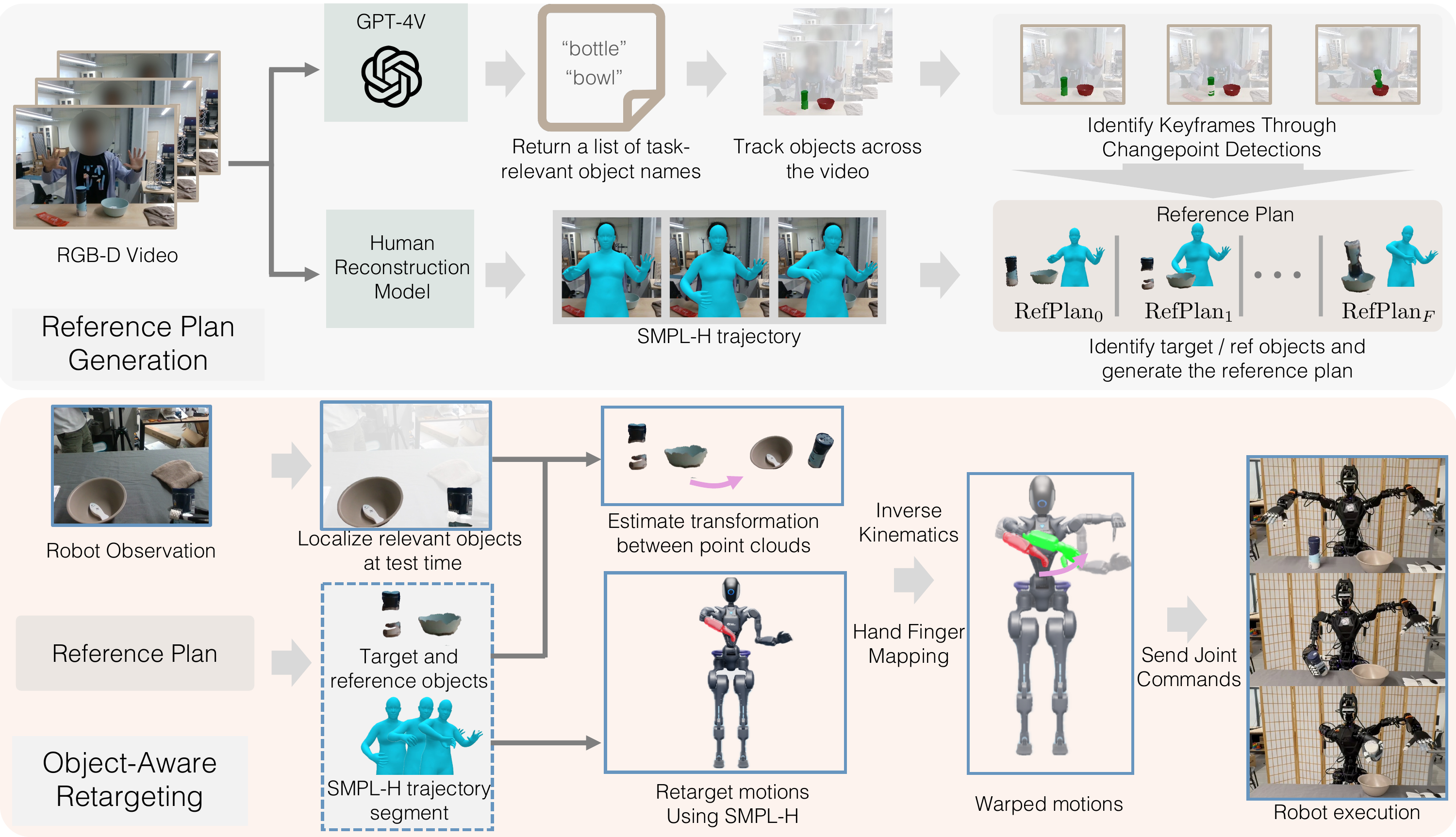}
    \caption[\okami{} model overview.]{\textbf{\okami{} Model Overview.} \okami{} is a two-staged method that enables a humanoid robot to imitate a manipulation task from a single human video. In the first stage, \okami{} generates a reference plan using GPT-4V and large vision models for subsequent manipulation. In the second stage, \okami{} follows the reference plan, retargeting the observed human motions onto the humanoid with object awareness. The retargeted motions are converted into a sequence of robot joint commands for the robot to follow.}
    \label{fig:okami:method-overview}
\end{figure}

In this chapter, we introduce \okami{} (\emphasize{O}bject-aware \emphasize{K}inematic ret\emphasize{A}rgeting for humanoid \emphasize{M}anipulation \emphasize{I}mitation), a two-staged method that tackles open-world imitation from observation for humanoid robots. \okami{} first generates a \textit{reference plan} using the object locations and reconstructed human motions from a given RGB-D video. Then, it retargets the human motion trajectories to the humanoid robot while adapting the trajectories based on new locations of the objects. Figure~\ref{fig:okami:method-overview} illustrates the whole pipeline. Our work in this chapter was published at the 8th Annual Conference on Robot Learning, 2024~\cite{li2024okami}.

\subsection{Problem Assumptions}
\label{sec:okami:problem_assumptions}

Section~\ref{sec:orion:problem_assumptions} described the assumptions that constrain the scope of the problem \orion{} can tackle. Likewise, we also introduce the assumptions made in this chapter for developing \okami{}, outlining the scope to which \okami{} is applicable.

The input of \okami{} is an RGB-D video of a human demonstration, the same format as the videos used in \orion{}. Additionally, we make two assumptions about $V$. 

\begin{itemize}
    \item Every image frame in $V$ needs to capture the human upper body and both hands. 
    \item The camera view of shooting $V$ is static throughout the recording. \textit{Unlike} \orion{}, \okami{} does not assume the availability of the list of task-relevant object names, and the task-relevant objects are identified using an off-the-shelf Vision-Language Model (VLM) that recognizes the names of task-relevant objects. 
\end{itemize}

\subsection{Reference Plan Generation}
\label{sec:okami:reference-plan}

To enable object-aware retargeting, \okami{} first needs to generate a reference plan for the humanoid robot to follow. To this end, \okami{} needs to understand what objects are involved and how humans move the objects in the demonstrated task, which are described first before we introduce the plan generation step. 

\paragraph{Identify and Localize Task-Relevant Objects.} 
Imitating a manipulation task requires the robot to understand what objects to interact with in order to complete the task. However, identifying task-relevant objects from pure images is a nontrivial challenge. While prior works use an unsupervised approach to identify the objects~\cite{caelles20192019, huang2023out}, they often assume simple visual backgrounds. Other alternatives require additional text inputs from humans, inducing extra annotation cost from the user~\cite{zhu2023groot, stone2023open}. Instead, we observe that most objects in everyday tasks are covered by commonsense knowledge. State-of-the-art Vision-Language Models (VLMs) such as GPT4V have internalized such knowledge through pre-training on Internet-scale data. Based on this observation, we leverage the power of GPT4V to identify task-relevant objects directly from the video demonstration $V$. Concretely, \okami{} samples the RGB image frames from $V$ and prompts GPT4V with the concatenated image of the sampled frames (We describe the details of text prompt we use to query the object names from GPT4V in Appendix~\ref{supp:okami:prompt}). GPT4V returns a list of texts that describe the names of task-relevant objects in $V$, in the same format as the English descriptions we use in \orion{}. Subsequently, \okami{} uses Grounded-SAM~\cite{liu2023grounding} with the list of object names to segment the objects on the first frame of $V$, and then track their locations across the entire video by propagating the first frame segmentation throughout the images using Cutie~\cite{cheng2023putting}. 
In the end, \okami{} localizes the task-relevant objects in $V$, which paves the way for all subsequent steps.\loosepar{}

\paragraph{Reconstruct Human Motions.} Retargeting human motions to the humanoid has the potential to generate feasible actions for humanoids due to their human-like embodiments. However, the video demonstration $V$ does not come with annotations on the human motions. To fill in the gap of missing data, we use a pre-trained vision model that can reconstruct 3D human models from in-the-wild videos (More details about training human reconstruction model are provided in Appendix~\ref{supp:okami:recon}). The model outputs a sequence of SMPL-H (Skinned Multi-Person Linear Model with Hands) features~\cite{pavlakos2019expressive}
SMPL-H is a low-dimensional feature descriptor that compactly captures the poses of the human body and hand poses. From the trajectory of SMPL-H models, we obtain the estimated full-body poses across the video. The estimated poses include locations of body joints in the task space with respect to the human pelvis, and hand poses in joint configurations that describe how a hand interacts with an object. With the SMPL-H trajectories, \okami{} is able to retarget the human motions to the humanoids, which is explained in Section~\ref{sec:okami:oar}. One advantage of using SMPL-H representation is that it captures human body poses while being invariant across humans with different demographic features, such as body sizes. Therefore, SMPL-H representation makes it easy to retarget the motions of humans in different body sizes to the humanoid robot.

\paragraph{Generate a Plan From $V$.} From the previous two steps, \okami{} localizes the task-relevant objects and estimates human motions from the video. However, naively warping the entire human motion trajectory based on object locations dooms to fail during deployment. Instead, \okami{} needs to identify the subgoals in $V$ such that we can warp segment of trajectories conditioning on the location of the object that is associated with a subgoal. The subgoals are identified as keyframes in $V$ using changepoint detection, the same as in \orion{} (See Section~\ref{sec:orion:plan-generation}). For each subgoal, we identify a target object (in motion due to manipulation) and a reference object (serving as a spatial reference for the target object's movements through either contact
or non-contact relations). For example, in a pouring task, the container is relevant to the task but never touched by the hand or the cup. The target object is determined based on the averaged keypoint velocities per object, while the reference object is identified through geometric heuristics or semantic relations predicted by GPT-4V (More implementation details can be found in Appendix~\ref{appendix:okami:plan-generation})

With subgoals and associated objects determined, we generate a reference plan $\{\okamiplan_{0}, \okamiplan_{1}, \dots, \okamiplan_{\KF}\}$ with $\KF+1$ steps. Each step $\okamiplan_{\kfindex}$ corresponds to a keyframe and includes the point clouds of the target object, the reference object, and the trajectory of the SMPL-H poses between two keyframes. A reference object may not be included in cases like grasping an object. Point clouds are obtained by back-projecting segmented objects from RGB images using depth images~\cite{zhou2018open3d}.\loosepar{}

\subsection{Object-aware Retargeting}
\label{sec:okami:oar}

Given a reference plan from the video demonstration, \okami{} enables the humanoid robot to imitate the task in $V$. The robot follows each step $\okamiplan_{\kfindex}$ in the plan by localizing task-relevant objects and retargeting the SMPL-H trajectory segment onto the humanoid. The retargeted trajectories are then converted into joint commands through inverse kinematics. This process repeats until all the steps are executed, and success is evaluated based on task-specific conditions (see Appendix~\ref{ablation_sec:okami:tasks}).

\paragraph{Localize Objects at Test Time.} 
 To execute the plan in the test-time environment, \okami{} must localize the task-relevant objects in the robot's observations, extracting 3D point clouds to track object locations. By attending to task-relevant objects, \okami{} policies generalize across various visual conditions, including different backgrounds or the presence of novel instances of task-relevant objects.

\paragraph{Retargeting Human Motions to the Humanoid.} The key aspect of \textit{object-awareness} is adapting motions to new object locations. After localizing the objects, we employ a factorized retargeting process that synthesizes arm and hand motions separately. \okami{} first adapts the arm motions to the object locations so that the fingers of the hands are placed within the object-centric coordinate frame. Then \okami{} only needs to retarget fingers in the joint configuration to mimic how the demonstrator interacts with objects with their hands. \loosepar{}

Concretely, we first map human body motions to the task space of the humanoid, scaling and adjusting trajectories to account for differences in size and proportion. \okami{} then warps the retargeted trajectory so that the robot's arm reaches the new object locations (See implementation details in Appendix~\ref{appendix:okami:warping}). We consider two cases in trajectory warping: the relational state between target and reference objects might be unchanged or might be changed. We adjust the warping based on the two cases accordingly. In the first case, we only warp the trajectory based on the target object locations. In the second case, the trajectory is warped based on the reference object location.

After warping, we use inverse kinematics to compute a sequence of joint configurations for the arms while balancing the weights of position and rotation targets in inverse kinematics computation to maintain natural postures. Simultaneously, we retarget the human hand poses to the robot's finger joints, allowing the robot to perform fine-grained manipulations (See implementation details in Appendix~\ref{appendix:okami:retarget}). In the end, we obtain a full-body joint configuration trajectory for execution. Since arm motion retargeting is affine, the retargeting process naturally scales and adjusts motions from demonstrators with varied demographic characteristics. By adapting arm trajectories to object locations and retargeting hand poses separately, \okami{} achieves generalization across various spatial layouts.\loosepar{}

\section{Experiments}
\label{sec:okami:experiments}

\begin{figure}[ht!]
    \centering
    \begin{minipage}{\linewidth}
    \centering
    \includegraphics[width=\linewidth]{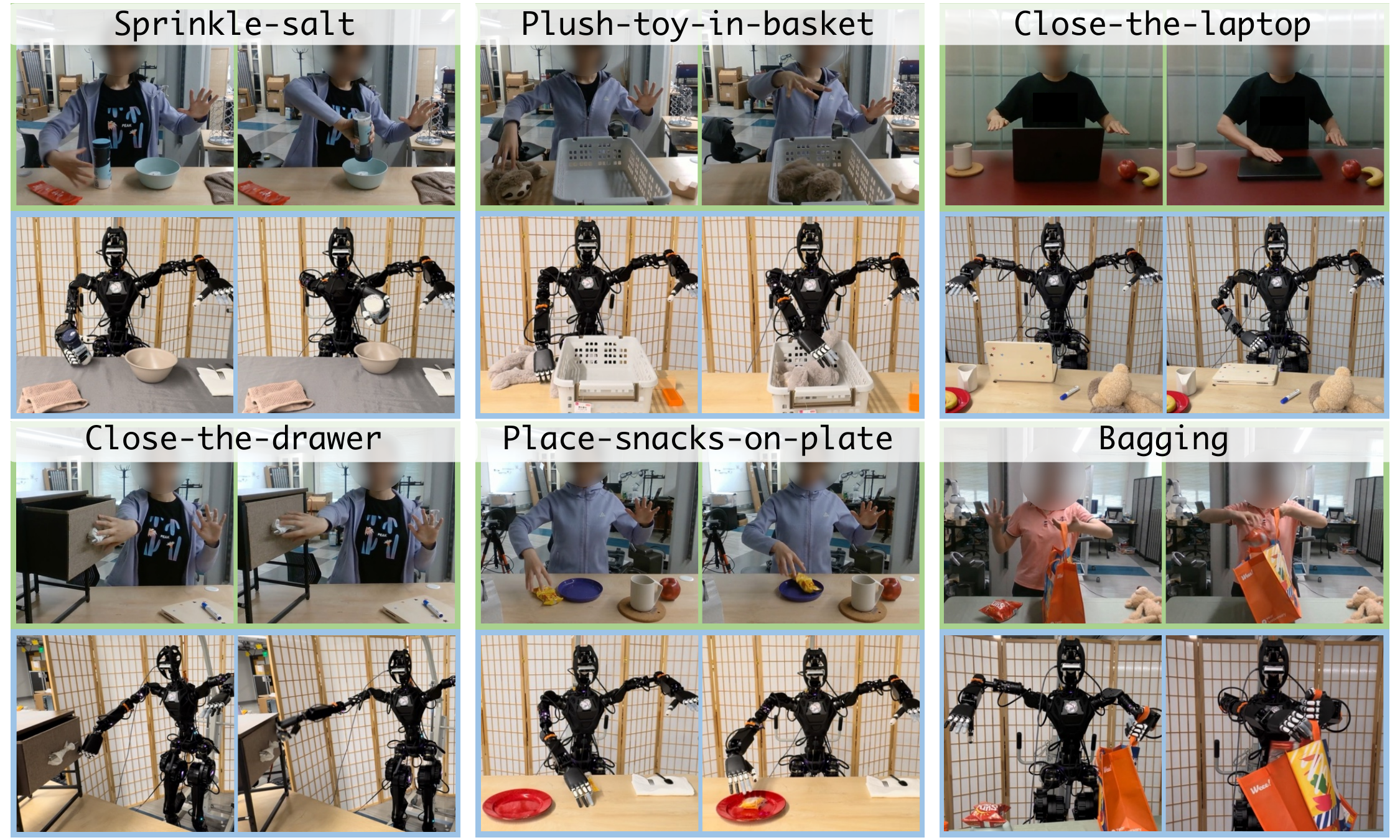}
    \caption[Visualization of video demonstrations and robot evaluation.]{Visualization of initial and final frames of both human demonstrations and robot rollouts for all tasks.}
    \label{fig:okami:experiment-overview}
    \end{minipage}
\end{figure}
    
Our experiments are designed to answer the following research questions: (1) Is \okami{} effective for enabling a humanoid robot to imitate diverse manipulation tasks by watching single videos of human demonstrations? (2) Is it critical in \okami{} to retarget demonstrators' body motions to the humanoid instead of retargeting based on object locations? 3) Can \okami{} retain its performances consistently on videos demonstrated by humans of diverse demographics? 4) Can the rollouts generated by \okami{} be used for training closed-loop visuomotor policies? \loosepar{}

\subsection{Experimental Setup}
    \label{sec:okami:experimental-setup}
    \paragraph{Tasks.} Here, we describe the tasks we choose. 
    1) \toy{}: placing a plush toy in the basket; 
    2) \salt{}: sprinkling a bit of salt into the bowl; 
    3) \drawer{}: pushing the drawer in to close it;
    4) \laptop{}: closing the lid of the laptop;
    5) \snacks{}: placing a bag of snacks on the plate. 6) \bagging{}: placing a chip bag into a shopping bag. The details of the success conditions of each task are provided in Appendix~\ref{ablation_sec:okami:tasks}. We select these six tasks that cover a diverse range of manipulation behaviors. \salt{} is the task that covers pouring behavior. \toy{} and \snacks{} require pick-and-place behaviors. \drawer{} and \laptop{} require the humanoid to interact with articulated objects, a common type of interaction in daily environments. \bagging{} involves dexterous, bimanual manipulation and includes multiple subgoals. While we mainly focus on real robot experiments, we also implement \salt{} and \drawer{} in simulation using RoboSuite~\cite{zhu2020robosuite} for easy reproducibility of \okami{}. More implementation details about the simulation tasks can be found in Appendix~\ref{ablation_sec:okami:simulation}.\loosepar{}

\paragraph{Hardware Setup.} We use Fourier-GR1 as the real robot hardware evaluation. We have provided the details of this type of humanoid robot in Section~\ref{sec:bg:robot-bg}. For both video recording and robot camera observation, we use an Intel RealSense D435i camera. In all our experiments, we use a joint position controller that operates at 400Hz. To avoid jerky movements, the inverse kinematic solver returns a sequence of commands at 40Hz and interpolates the commands to 400Hz.

\paragraph{Evaluation Protocol.} We run 12 trials for each task. The locations of the objects are initialized within a region that is visible in the robot camera's view and is reachable by the humanoid arms. The tasks are evaluated on a tabletop workspace with multiple objects, including both task-relevant objects and other distracting objects. Further, we test new object generalization on \snacks{}, \toy{}, and \salt{} tasks, changing the plate, snack bag, plush toy, and bowl to unseen instances from the same object category.

\paragraph{Baselines.} We compare our result with the baseline~\orion{} introduced in Chapter~\ref{chapter:orion}. Since \orion{} is proposed for parallel-jaw grippers, it is not directly applicable to our experiments, and we adopt it with minimal modifications. We first estimate the palm trajectory using the SMPL-H trajectories and then warp the trajectory conditioning on the new object locations. The subsequent process of inverse kinematics uses the warped trajectory for computing robot joint configurations. 
    
\subsection{Results}
\label{sec:okami:results}
\begin{figure}[ht!]
    \centering
    \includegraphics[width=\linewidth]{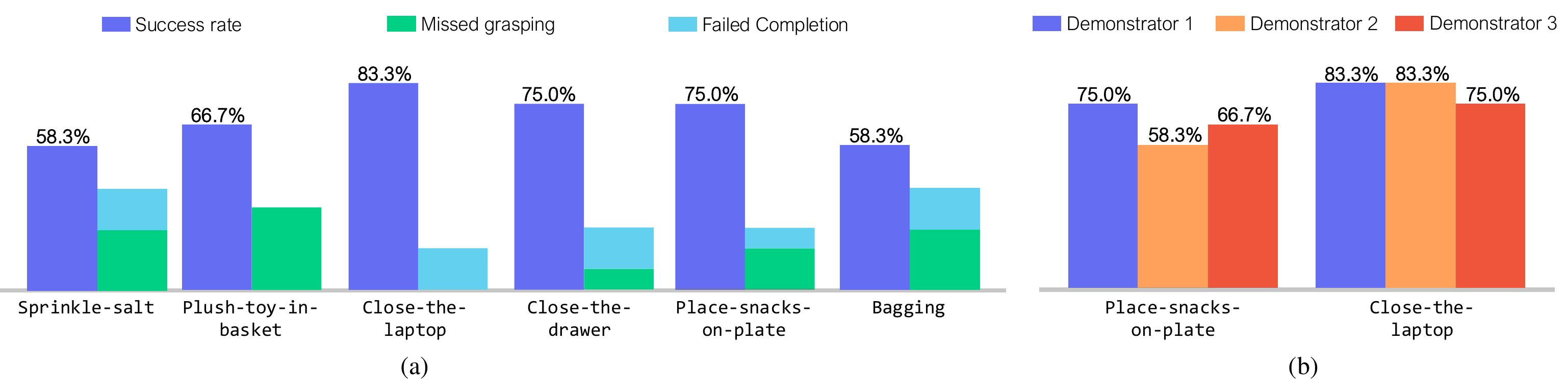}
    \caption[Experimental evaluation of \okami{} policies.]{\textbf{Experimental Evaluation of \okami{} Policies.} (a) Evaluation of \okami{} over all six tasks, including the success rates and the quantification of failed trials, separated by failure mode. (b) Evaluation of \okami{} using videos from different demonstrations. Demonstrator 1 is the main person recording videos for all evaluations in (a).\loosepar{}}
    \vspace{-0.3cm}
    \label{fig:okami:result}
\end{figure}

To answer question (1), we evaluate the policies of \okami{} across all the tasks, covering diverse behaviors such as daily pick-place, pouring, and manipulation of articulated objects. The results are presented in Figure~\ref{fig:okami:result} (a). In our experiment, we randomly initialize the object locations so that the robot must adapt its motions to complete the tasks successfully. This result shows the effectiveness of \okami{} in generalizing over different visual and spatial conditions.

 To answer question (2), we compare \okami{} against \orion{} on two representative tasks, \snacks{} and \laptop{}. In the comparison experiment, \okami{} differs from \orion{} in that \orion{} does not condition on the human body poses. \okami{} achieves 75.0\% and 83.3\% success rates, respectively, while \orion{} only achieves 0.0\% and 41.2\%, respectively. Additionally, we compare \okami{} against \orion{} on the two simulated versions of \salt{} and \drawer{} tasks. In simulation, \okami{} achieves 82.0\% and 84.0\% success rates in two tasks while \orion{} only achieves 0.0\% and 10.0\% (Table~\ref{tab:okami:sim_results} shows the full result of simulation evaluation in Appendix~\ref{ablation_sec:okami:simulation}). Most failures of \orion{} policies are due to failing to approach objects with reliable grasping poses. For example, in \snacks{} task, \orion{} tries to grasp the snack from the sides instead of the top-down grasp in the human video, failing to rotate the wrist fully to achieve behaviors such as pouring. Such failures are due to the fact that \orion{} ignores the embodiment information, thus falling short in performance compared to \okami{}. The superior performance of \okami{} suggests the importance of leveraging human body motions when using humanoids for imitating human videos.

\begin{figure}[t]
\centering
    \includegraphics[width=\linewidth]{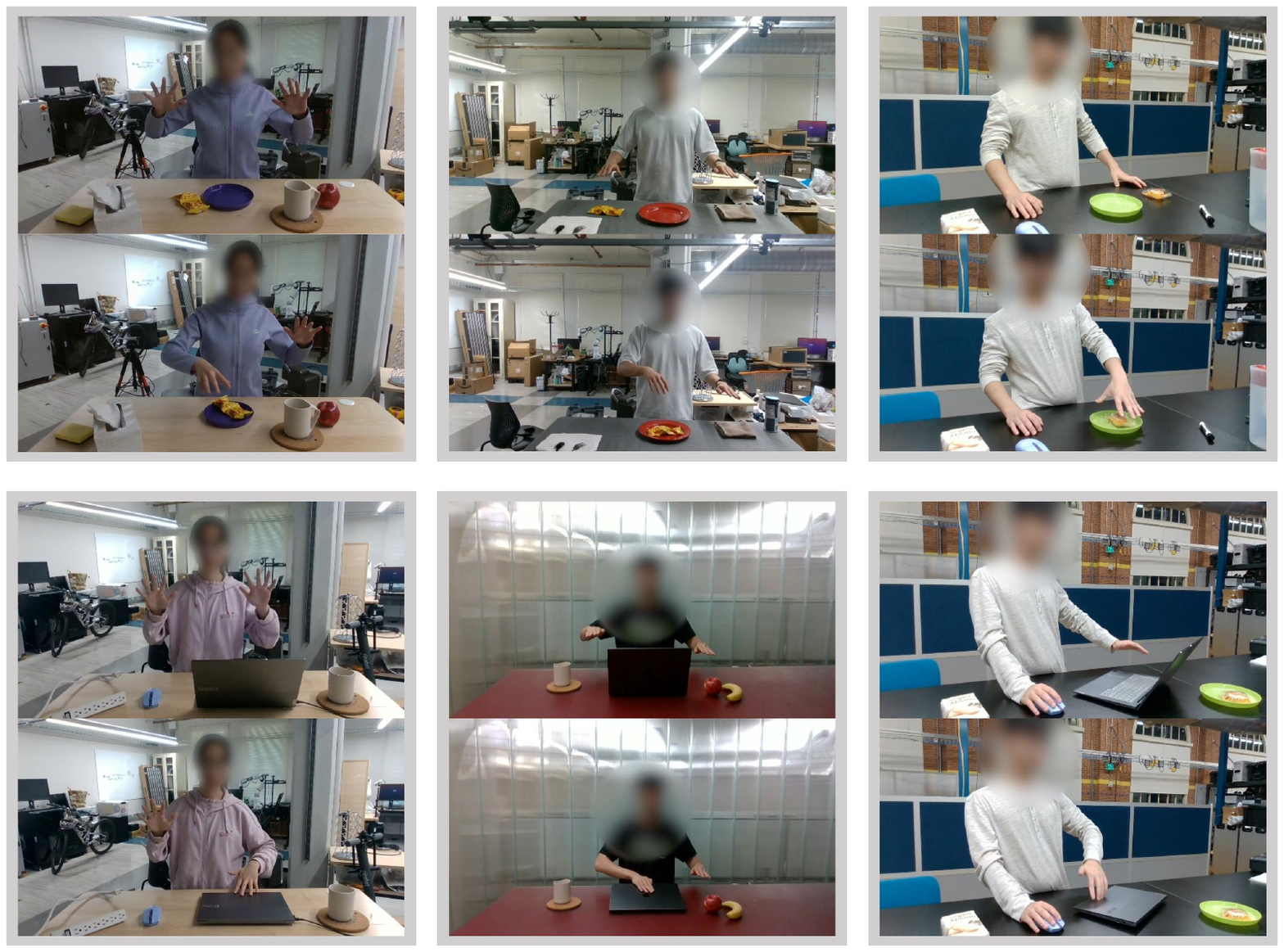}
    \caption[Visualization of videos collected by different human demonstrators.]{The initial and end frames of videos performed by different human demonstrators. The top row is \snacks{} task, and the bottom row is \laptop{} task.}
    \label{fig:okami:human_demonstrators}
\end{figure}

To answer question (3), we conduct a controlled experiment of recording videos of different demonstrators and test if \okami{} policies maintain strong performance across the video inputs. The same as the previous experiment, we evaluate \okami{} on the \snacks{} and \laptop{} tasks. Figure~\ref{fig:okami:human_demonstrators} shows the screenshots of three different human demonstrators performing \snacks{} and \laptop{} tasks. The results are presented in Figure~\ref{fig:okami:result} (b). We show that for the task \laptop{}, there is no statistical significance in performance change. As for task \snacks{}, while the evaluation maintains above 50\%, the worst policy performance is 16.7\% lower than the best policy performance. After looking into the video recording, we find that the motion of demonstrator 2 is relatively faster than the other two demonstrators and faster motions create a noisy estimation of motion when doing human model reconstruction. Overall, \okami{} can maintain reasonably good performance given videos from different demonstrators, but there is room for improvements on our vision pipeline to handle such variety.

\subsection{Rollout Distillation} 
\label{sec:okami:rollout}
We address question (4) by training neural-network-based visuomotor policies using \okami{} rollouts. We first run \okami{} over randomly initialized object layouts to generate multiple rollouts and collect a dataset of successful trajectories while discarding the failed ones. We train neural network policies on this dataset through a behavioral cloning algorithm. Since smooth execution is critical for humanoid manipulation, we implement behavioral cloning with ACT~\cite{zhao2023learning}, which predicts smooth actions via its temporal ensemble design. (More implementation details in Appendix~\ref{chapter:appendix_chapter_II}). We train visuomotor policies for \salt{} and \bagging{}. Figure~\ref{fig:okami:visuomotor_result} shows the success rates of these policies, demonstrating that successful \okami{} rollouts are effective data sources for training. We also show that the learned policies improve as more rollouts are collected. These results hold the promise of scaling up data collection for learning humanoid manipulation skills without laborious teleoperation.\loosepar{} 

\paragraph{Visuomotor Policy Details.} We choose ACT~\cite{zhao2023learning} in our experiments for behavioral cloning, an algorithm that has been shown effective in learning humanoid manipulation policies~\cite{cheng2024open}. Notably, we choose pretrained DinoV2~\cite{oquab2023dinov2} as the visual backbone of a policy. 
The policy takes a single RGB image and 26-dimension joint positions as input and outputs the action of the 26-dimension absolute joint position for the robot to reach. 
 In Table~\ref{tab:okami:act_hyperparams}, we list the hyparameters used in ACT implementation.

\begin{minipage}{0.45\linewidth}
    \begin{figure}[H]
     \raggedright
        \includegraphics[width=0.9\linewidth]{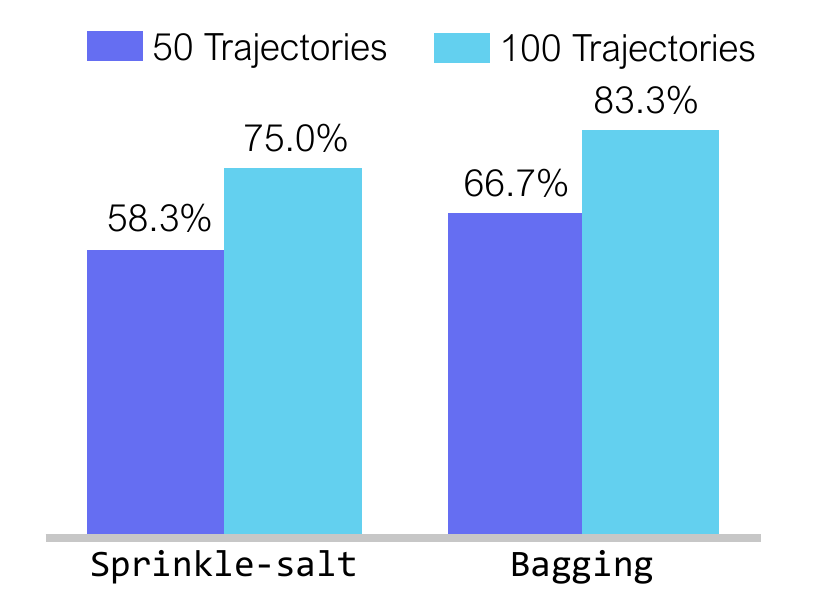}
        \caption[Evaluation of closed-loop visuomotor policies trained from collected rollouts.]{Success rates (\%) of learned visuomotor policies on \salt{} and \bagging{} using 50 and 100 trajectories, respectively.}
        \label{fig:okami:visuomotor_result}
    \end{figure}
\end{minipage}%
\hfill
\begin{minipage}{0.45\textwidth}
    \begin{table}[H]
        \centering
        \begin{tabular}{cc}
            \toprule
            KL weight & 10\\
            chunk size  & 60\\
            hidden dimension    & 512\\
            batch size & 45\\
            feedforward dimension & 3200\\
            epochs & 25000\\
            learning rate & 5e-5\\
            temporal weighting & 0.01\\
            \bottomrule
        \end{tabular}
        \caption[Hyperparameters in ACT implementation.]{The hyperparameters used in ACT.}
        \label{tab:okami:act_hyperparams}
    \end{table}
\end{minipage}

\section{Summary}
\label{sec:okami:discussion}
In this chapter, we introduced \okami{} which enables a humanoid robot to imitate a single RGB-D human video demonstration. At the core of \okami{} is object-aware retargeting, which retargets human motions onto the humanoid robot and adapts the motions to target object locations. \okami{} consists of two stages to realize object-aware retargeting. The first stage generates a reference manipulation plan from the video. The second stage retargets the human motions onto humanoids while considering the new object locations. Our experiments demonstrate the systematic generalization of \okami{} policies (i.e., the intra-task generalization as introduced in Section~\ref{sec:bg:open-world-formulation}). \okami{} enables efficient collection of manipulation data based on a single human video demonstration. \okami{}-based data collection significantly reduces the human cost for policy training compared to that required by teleoperation. This chapter shows how a humanoid robot can leverage \textit{spatial regularity} to learn from in-the-wild video observations while taking advantage of its embodiment similarity to humans.

\paragraph{Limitations and Future Work.} The current focus of \okami{} is on upper body motion retargeting of humanoid robots, particularly for manipulation tasks within tabletop workspaces. A promising future direction is to include lower body retargeting that enables locomotion behaviors during video imitation. To enable full-body loco-manipulation, a whole-body motion controller needs to be implemented as opposed to the joint position controller used in \okami{}. Additionally, we rely on RGB-D videos in \okami{}, which limits us from using in-the-wild Internet videos recorded in RGB. Extending \okami{} to use web videos will be another promising direction for future work. Finally, the current implementation of retargeting has limited robustness against large variations in object shapes. A future improvement would be to improve the vision models to endow the robot with a general understanding of how to interact with a class of objects despite their large shape changes. \loosepar{}

In summary, this chapter and the previous chapter constitute Part~\ref{part:II} of this dissertation. The following chapter marks the start of Part~\ref{part:III}, in which we tackle multitask and lifelong robot learning. In the next part, we will introduce our contributions that exploit behavioral regularity for achieving inter-task generalization in Open-world Robot Manipulation.

\part{Lifelong Robot Learning with Skills}
\label{part:III}

\chapter{Bottom-up Skill Discovery From Demonstrations}
\label{chapter:buds}

The next three chapters focus on how robots can continually imitate multiple tasks in a sequential manner, allowing robots to scale up the number of tasks they can learn. Our major idea is to leverage the behavioral regularity described in Section~\ref{sec:bg:behavioral_regularity}. The existence of behavioral regularity reveals the compositional structures of manipulation tasks, where subcomponents are shared across tasks. Temporal abstraction offers a powerful framework to model the compositional structures of manipulation tasks~\cite{sutton1999between,Precup00temporalabstraction}. These temporal abstractions correspond to purposeful behaviors that serve as the basic building blocks for synthesizing temporally extended behaviors. 

A major challenge is how to automate the discovery of reusable skills and build a diverse skill library without costly manual engineering, especially when we consider real-world sensory data in the state representation. Several paradigms have been explored for automating skill acquisition, including the \textit{options} framework~\cite{vezhnevets2017feudal,fox2017multi,gregor2016variational,konidaris2009skill,bagaria2019option,kumar2018expanding} and unsupervised skill discovery based on information-theoretic metrics~\cite{eysenbach2018diversity,hausman2018learning,sharma2019dynamics}. While these methods have shown success in simulation environments, they face challenges in real-world applications due to high sample complexity and reliance on ground-truth physical states. Learning from demonstrations offers an alternative by reducing the exploration burden. Instead of costly manual annotation of demonstration segments~\cite{hovland1996skill}, our approach discovers skills from unsegmented demonstrations without temporal labels. Prior works on learning from unsegmented demonstration use various techniques, including Bayesian inference~\cite{konidaris2009skill,niekum2012learning}, generative modeling~\cite{shankar2020learning,tanneberg2021skid,kipf2018compositional}, and dynamic programming~\cite{shiarlis2018taco}. However, these methods struggle with high-dimensional sensor data. 

In this chapter, we develop a hierarchical approach to tackling real-world robot manipulation based on skill discovery from demonstrations. Specifically, we consider a multitask learning setting in this chapter. Later in the next chapter, we will extend our method to the general lifelong learning setting. Our work in this chapter was published in IEEE Robotics and Automation Letters, 2022~\cite{zhu2022buds}.

\begin{figure}
    \centering
    \includegraphics[width=1.0\linewidth]{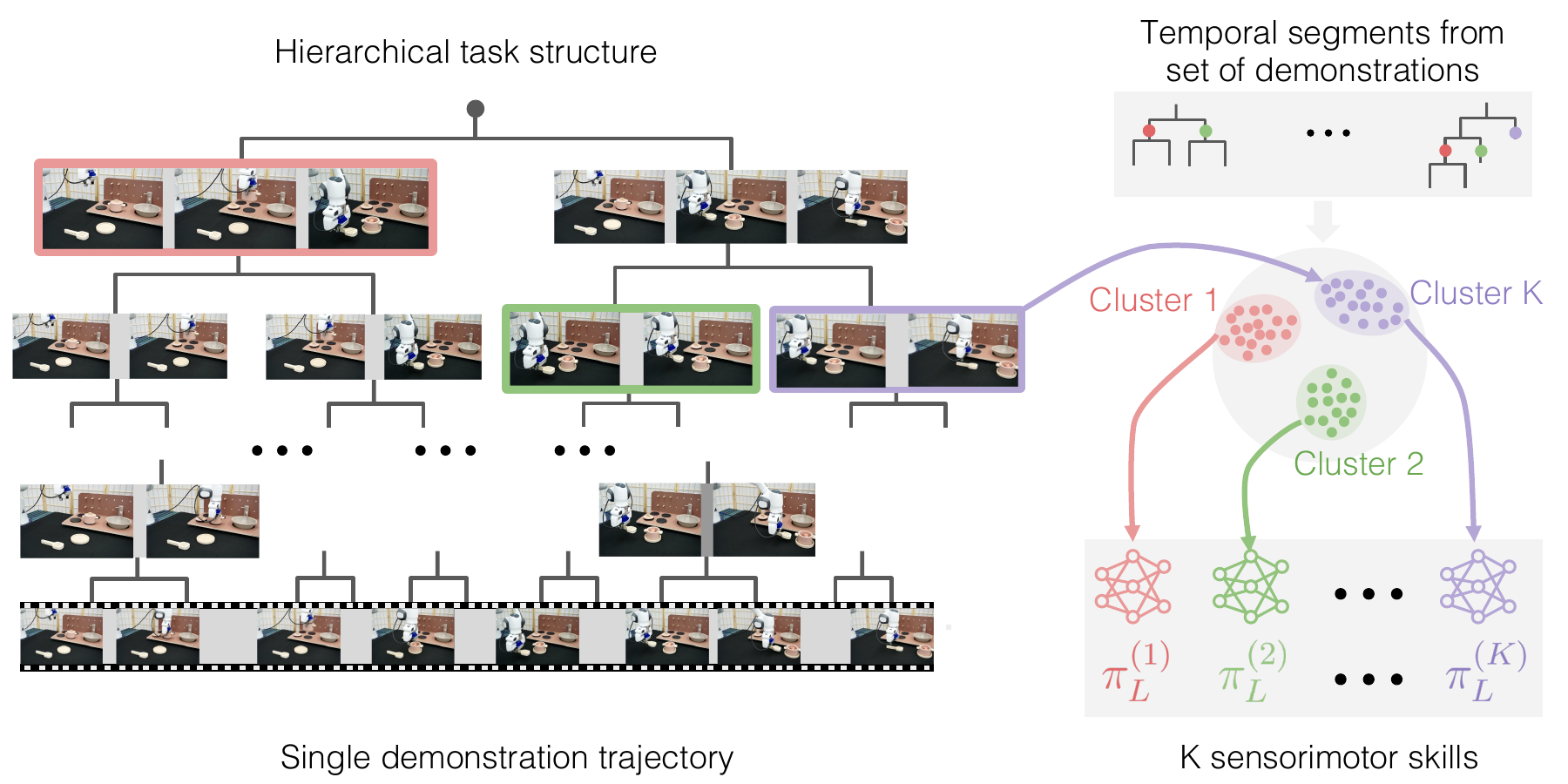}
    \caption[\buds{} overview.]{\textbf{\buds{} Overview.} \buds{} constructs hierarchical task structures of demonstration sequences in a bottom-up manner, from which mid-level temporal segments are discovered for discovering and learning sensorimotor skills.}
    \label{fig:buds:buds-overview}
\end{figure}

\section{BUDS}
\label{sec:buds:method}

We introduce \buds{} (\emphasize{B}uttom-\emphasize{U}p \emphasize{D}iscovery of sensorimotor \emphasize{S}kills), shown in Figure~\ref{fig:buds:buds-overview}. \buds{} starts with an unsupervised clustering-based segmentation model that extracts a library of sensorimotor skills from teleoperated demonstrations. Each skill is modeled as a goal-conditioned sensorimotor policy that operates on raw images and robot proprioception. \buds{} further learns a high-level meta-controller that selects a skill and predicts the subgoal for the skill to achieve at any given state. Both the skills and the meta-controller are trained with imitation learning. Together, it presents a scalable framework for solving complex manipulation tasks from raw sensory inputs, amendable to real-robot deployment.

Four key properties of \buds{} are crucial for its effectiveness: 1) it uses bottom-up agglomerative clustering to build hierarchical task structures from demonstrations. These hierarchical representations offer flexibility for the imitation learner to determine the proper granularity of the temporal segments. 2) \buds{} segments the demonstrations based on multi-sensory cues, including multi-view images and proprioceptive features. It takes advantage of the statistical patterns across multiple sensor modalities to produce more coherent and compositional task structures than using a single modality. 3) \buds{} extracts skills from demonstrations on multiple tasks that achieve different manipulation goals, facilitating knowledge sharing across tasks and improving the reusability of the discovered skills. 4) we train our goal-conditioned skills with the hierarchical behavioral cloning algorithm as introduced in Section~\ref{sec:bg:hbc}.

% , producing perceptually grounded yet versatile skills for composition.

In the rest of the section, we first formalize the problem of skill discovery from demonstrations, and then present the two key steps of our approach: 1) skill segmentation with hierarchical agglomerative clustering on unsegmented demonstrations, and 2) learning low-level skill policies and meta-controllers with hierarchical behavioral cloning.

\subsection{Skill Discovery From Demonstrations}
\label{sec:buds:skill_discovery_formulation}

We formalize the framework of skill discovery from demonstrations. This framework aims to learn $\maxtasknum$ tasks $\{\Task{\context_{i}}\}_{i=1}^{\maxtasknum}$ through the implementation of a hierarchical policy, whose low-level skills are identified and learned from demonstrations. Based on the background introduced in Section~\ref{sec:orion:open_world}, we use the CMDP formulation to model the manipulation tasks, wherein all the tasks are specified through contexts in the form of demonstrations. The contexts additionally include task ID labels or language descriptions so that we can differentiate demonstration trajectories that come from different tasks.

The solution to all considered tasks takes the form of a hierarchical policy, expressed in Equation~\ref{eq:hierarchical-policy} from Section~\ref{sec:bg:policy-skills}:

\begin{equation*}
     \pi(a_{t}|s_{t}, \context) = \sum_{i=1}^{K}\metacontroller{}(i, \param|s_{t}, \context)\mathds{1}(i=k)\skillpolicy{i}(a_{t}|s_{t}, \param)
\end{equation*}

In this equation, $k$ represents the skill index, while $\param$ denotes skill parameters that can be either a vector of physical states or latent states. Demonstrations are provided across all $\maxtasknum$ tasks. We denote the demonstration datasets  for the $\tasknum$-th task as $\dataset^{\tasknum}$, with the aggregation of all task demonstrations represented as $\dataset=\bigcup_{\tasknum=1}^\maxtasknum \dataset^{\tasknum}$.  For notational simplicity, we denote $\dataset^{\context_{\tasknum}}$ as $\dataset^{\tasknum}$, where we directly use the superscript $\tasknum$ to reference the task $\Task{\context_{\tasknum}}$ rather than explicitly noting the context.

Within this framework, $\dataset$ is separated into $K$ partitions $\{\tilde{\dataset}_1, \ldots, \tilde{\dataset}_K\}$. Each partition corresponds to a specific low-level skill, enabling the learning of a skill policy $\skillpolicy{k}$ from its associated partitioned dataset $\tilde{\dataset}_{k}$. The effective algorithmic design within this framework can provide insights in how to leverage behavioral regularity---specifically, how tasks can be effectively modeled in a modular fashion and how to identify the recurring behaviors.

In this chapter, we introduce our bottom-up approach, \buds{}, as a solution under the framework of skill discovery from demonstrations. Our approach specifically considers hierarchical policy learning through hierarchical behavioral cloning, as outlined in Section~\ref{sec:bg:hbc}. In the next chapter, we will discuss how to extend the skill discovery formulation to lifelong learning settings (See Section~\ref{fig:lotus:overview}).

\subsection{Bottom-up Discovery With Clustering}
\label{sec:buds:hiearrchical-clustering}

We present how to split the aggregated dataset $D$ into $K$ partitions, which we use to train the $K$ low-level skills. Our objective is to cluster similar temporal segments from multitask demonstrations into the same skill, easing the burden for the downstream imitation learning algorithm. To this end, we first learn per-state representations based on multi-sensory cues, which we use to form a hierarchical task structure of each demonstration trajectory $\tau^{\tasknum}_{i}$ with bottom-up agglomerative clustering. We then identify the recurring temporal segments across the entire demonstration dataset via spectral clustering. The whole process is fully unsupervised without additional manual annotations beyond the demonstrations.

\paragraph{Learning Multi-sensory State Representations.}
Our approach learns a latent representation per state in the demonstrations. It is inspired by research in event perception~\cite{zacks2001event}, which addresses the importance of correlation statistics presented in multiple sensory modalities for event segmentation. \buds{} learns the representations from multi-modal sensory data to capture their statistical patterns. Our method follows Lee et al.~\cite{lee2020making}, which learns a joint latent feature of all modalities by fusing feature embeddings from individual modalities (multi-view images and proprioception) with the Product of Experts~\cite{hinton2002training}. The feature is optimized over an adapted version of evidence lower bound loss~\cite{kingma2013auto}, which reconstructs the current sensor inputs. The reconstruction is different from the one from Lee et al. which is optimized for reconstructing the next states. Our design choice differs from the prior work, which focuses on directly using fused input representation in policy learning. To that end, the prior work needs to encode future state information while \buds{} focuses on learning the statistical patterns of multi-sensory data at the current state. By learning the joint representation, it captures the congruence underlying multi-sensory observations while retaining the information necessary to decode $s_t$.

% We denote $h_t$ as the latent vector computed from $s_t$.

\paragraph{Discovering Temporal Segments.} 
\buds{} uses the per-state representations to effectively group states in temporal proximity to build a hierarchical representation of a demonstration sequence. The strength of a hierarchical representation, as opposed to flat segmentation, is the flexibility to decide the segmentation granularity for imitation learning. Here, we use hierarchical agglomerative clustering, where in each step, we combine two adjacent temporal segments into one based on similarity until all segments are combined into the entire demonstration sequence. This process produces a tree of segments. To reduce the tree depth, we start with the bottom-level elements that contain temporal segments of a demonstration of $10$ steps.
The clustering process selects two adjacent segments that are most similar to each other among all pairs of adjacent segments, and merges them into one longer segment. 
The similarity between two segments is computed according to the $\ell_2$ distance between their \textit{segment features}, defined to be the average of latent features of all states in each segment. The process is repeated until we have only one segment left for each $\tau^{\tasknum}_{i}$. We discover a collection of intermediate segments, which we term as \textit{temporal segments}, from the formed hierarchies. This concept is inspired by the concept of Mid-level Action Elements~\cite{lan2015action} in the action recognition literature. The way we determine the temporal segments is by breadth-first searching from the root node of the hierarchy. During the search, we stop on one branch if the length of the intermediate segment is not longer than a given threshold of minimum length. The breadth-first search terminates when the number of segments at the lowest level of each branch exceeds a given threshold, and each segment at these lowest levels in $\tau^{\tasknum}_{i}$ represents a temporal segment.\loosepar{}

\paragraph{Partitioning Skill Datasets.}
After we have a set of temporal segments for every $\tau^{\tasknum}_{i}$, we aggregate them from all demonstrations into one set, and apply another clustering process to group them into $K$ partitions $\{\tilde{\dataset}_{1}, \ldots, \tilde{\dataset}_{K}\}$, and we use each partition $\tilde{\dataset}_{k}$ to train the skill $\skillpolicy{k}$. Training the skills on datasets from multiple tasks can improve the reusability of the skills as they are trained on diverse data. We use spectral clustering~\cite{von2007tutorial} with RBF kernel on the features of temporal segments for this partitioning step. The feature of a segment is computed as the concatenation of representations of the first, middle (or several frames in the middle), and last states of the segment. The number of keyframes chosen in the middle of a segment can vary based on the average length of demonstrations. The spectral clustering step results in $K$ datasets of temporal segments for skill learning.

% In practice, we set the maximum number of clusters in spectral clustering and merge any classified skill into an adjacent skill if its average length is below a threshold. The number of remaining classes is denoted as $K$, which is the final number of skills we partition demonstrations into. 

\begin{figure}[t]
    \centering
    \includegraphics[width=1.0\linewidth, trim=0cm 0cm 0cm 0cm,clip]{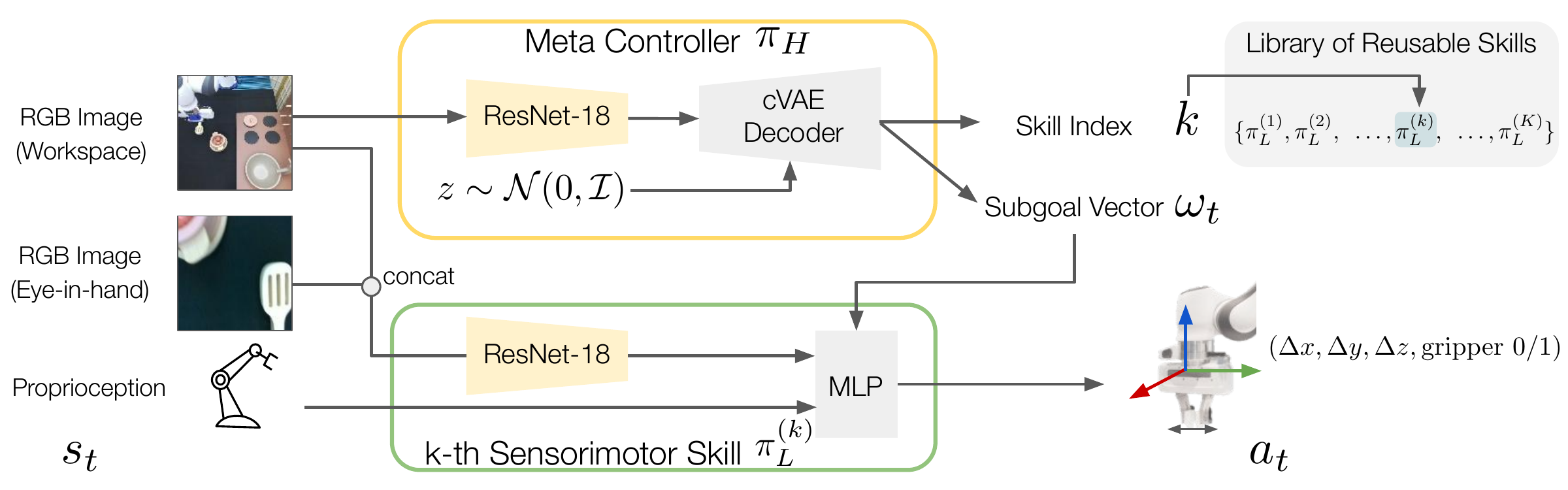}
    \caption[Hierarchical Visuomotor Policy in \buds{}.]{\textbf{Hierarchical Visuomotor Policy in \buds{}.} Given an observation image of the workspace, the meta-controller selects the skill index and generates the latent subgoal vector $\param_t$. Then, the selected sensorimotor skill generates action $a_t$ (end-effector displacements and gripper open/close) conditioned on observed images, proprioception, and $\param_t$.\loosepar{}}
    \label{fig:buds:hierarchical-policy}
\end{figure}

\subsection{Policy Learning with Hierarchical Behavioral Cloning}
\label{sec:buds:policy}

We use the obtained $K$ datasets of temporal segments from the segmentation step to train our hierarchical policies using a hierarchical behavioral cloning algorithm, including two parts: 1) skill learning with goal-conditioned imitation and 2) skill composition with a meta-controller. Figure~\ref{fig:buds:hierarchical-policy} visualizes the model structure of the hierarchical policy. 

\paragraph{Skill Learning with Goal-Conditioned Imitation.} 
We train each skill $\skillpolicy{k}$ on the corresponding dataset $\tilde{\dataset}_{k}~\forall k\in\{1, \dots, K\}$. Every skill $\skillpolicy{k}(a_t|s_t, \param_t)$ takes a sensor observation $s_t$ and a subgoal vector $\param_t$ as input, and produces a robot's motor action $a_t$. 
% By conditioning $\skillpolicy{k}$ on a subgoal vector, we enable the meta-controller to invoke the skills and specify the subgoals that these skills should achieve.
Instead of defining the subgoals in the original sensor space, which is typically high-dimensional, we instead learn a latent space of subgoals, where a subgoal state $s_{\tg}$ is mapped to a low-dimensional feature vector $\param_t$. For each state $s_t$, we define its subgoal as the future state either $\subgoaltime$ steps ahead of $s_t$ in the demonstration or the last state of a skill segment if it reaches the end of the segment from $s_t$ within $\subgoaltime$ steps. We define a subgoal as a look-ahead state with a constant number of future steps, rather than the final goal state of the task. This definition exploits temporal abstraction and reduces the computational burden of individual skills, as they only need to reach short-horizon subgoals without reasoning about long-term goals.
% The reason we define a subgoal as a look-ahead state of a constant number of steps in the future, as opposed to the final goal state of the task, is to exploit the temporal abstraction and reduce the computational burden of individual skills---skills only need to reach short-horizon subgoals, without the need for reasoning about long-term goals.
Concretely, we train such a goal-conditioned skill on skill segments in $\tilde{\dataset}_{k}$. For each  $\skillpolicy{k}$, we sample $(s_{t}, a_{t}, s_{\tg}) \sim \tilde{\dataset}_{k}$ where $\tg=\min{(t+\subgoaltime, \segmentfinal)}$ ($\segmentfinal$ is the last timestep of end of the segment),
and we generate a latent subgoal vector $\param_{t}$ for the subgoal of $s_t$, $s_{\tg}$, using a subgoal encoder $\param_{t}=\text{Encoder}_{k}(s_{\tg})$, where $\text{Encoder}_{k}(\cdot)$ is a ResNet-18 backboned network jointly trained with the policy $\skillpolicy{k}$. 

\paragraph{Skill Composition with A Meta-Controller.}
Now that we have a set of skills, a meta-controller is needed to decide which skill to use given $s_t$ and specify the desired subgoal for it to reach. We train a task-specific meta-controller $\metacontroller{}$ for each task. Given the current state, $\metacontroller{}$ outputs an index $k\in\{1, \ldots, K\}$ to select the $k$-th skill, along with a subgoal vector $\param_t$ on which the selected skill is conditioned. 
As teleoperated demonstrations are diverse by nature, the same state could lead to various subgoals in demonstration sequences (e.g., grasping different points of an object, or pushing an object at different contact points). Thus, the meta-controller needs to learn distributions of subgoals from demonstration data of a task $\dataset^{\tasknum}$, and we choose conditional Variational Autoencoder (cVAE)~\cite{kingma2013auto}. To obtain the training data of skill indices and subgoal vectors, we sample $(s_t, s_{\tg})\sim \dataset^{\tasknum}$. From the clustering step we have the correspondence between a skill index $k$ and state $s_t$, while from the skill learning step, we can generate a per-state subgoal vector $\param_{t}=\text{Encoder}_{k}(s_{\tg})$. The meta-controller $\metacontroller{}$ is trained to generate a skill index $k$ and a subgoal vector $\param_{t}$ conditioned on state $s_t$. During the evaluation, the controller generates the skill index and the subgoal vector conditioned on the current state $s_t$ and a latent vector sampled from a prior distribution, normal distribution $\mathcal{N}(0, \mathcal{I})$. The controller for evaluation is typically chosen to operate at a lower frequency than skills so that it can avoid switching among skills too frequently.

Training a meta-controller follows the same cVAE training convention in prior works~\cite{mandlekar2020iris, mandlekar2020learning}, which minimizes an ELBO loss on demonstration data. To obtain the training supervision for skill indices and subgoal vectors, we augment the demonstrations with results from the clustering and skill learning steps. The training labels of skill indices are derived from cluster assignments, while the latent subgoal vectors are computed from demonstration states using encoders that were jointly trained with skill policies.

\section{Experiments}
\label{sec:buds:experiments}

We design experiments to answer the following questions: 1) Is the hierarchical policy in \buds{} more effective than a monolithic policy? 2) Is the bottom-up discovery process critical to the quality of discovered skills? 3) Can \buds{} effectively discover reusable skills from multitask demonstrations?  4) Is \buds{} practical for real-robot deployment?  5) Can \buds{} discover skills that are semantically meaningful?

\subsection{Experimental Setup}
\label{sec:buds:experiments:setup}

We perform baseline comparisons and model analysis in simulation environments developed with the Robosuite framework~\cite{zhu2020robosuite} and present quantitative results on real hardware.

\begin{figure}[h]
\centering
\begin{minipage}[t]{0.17\linewidth}
        \includegraphics[width=\linewidth,trim=0cm 0cm 0cm 0cm,clip]{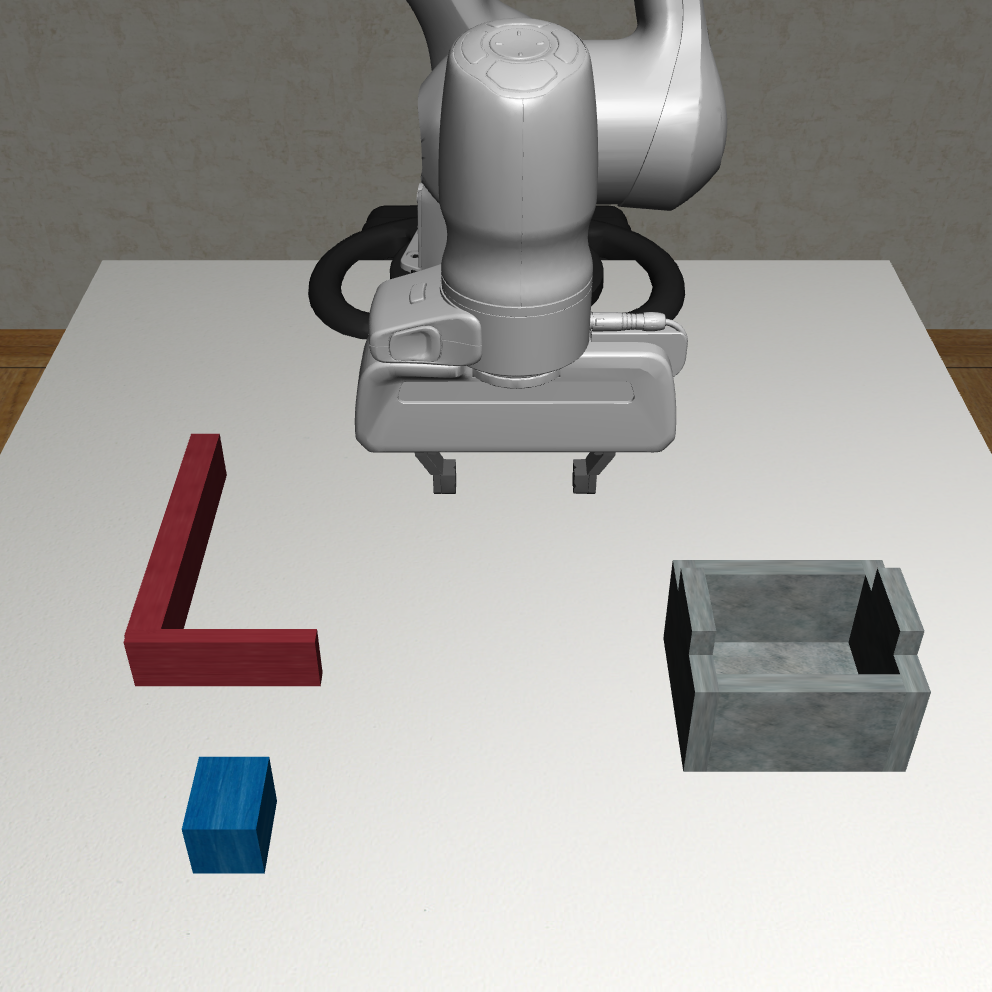}
        \vspace{-5mm}
     \subcaption{}
      \end{minipage}
      \hfill
 \begin{minipage}[t]{0.17\linewidth}
        \includegraphics[width=\linewidth,trim=0cm 0cm 0cm 0cm,clip]{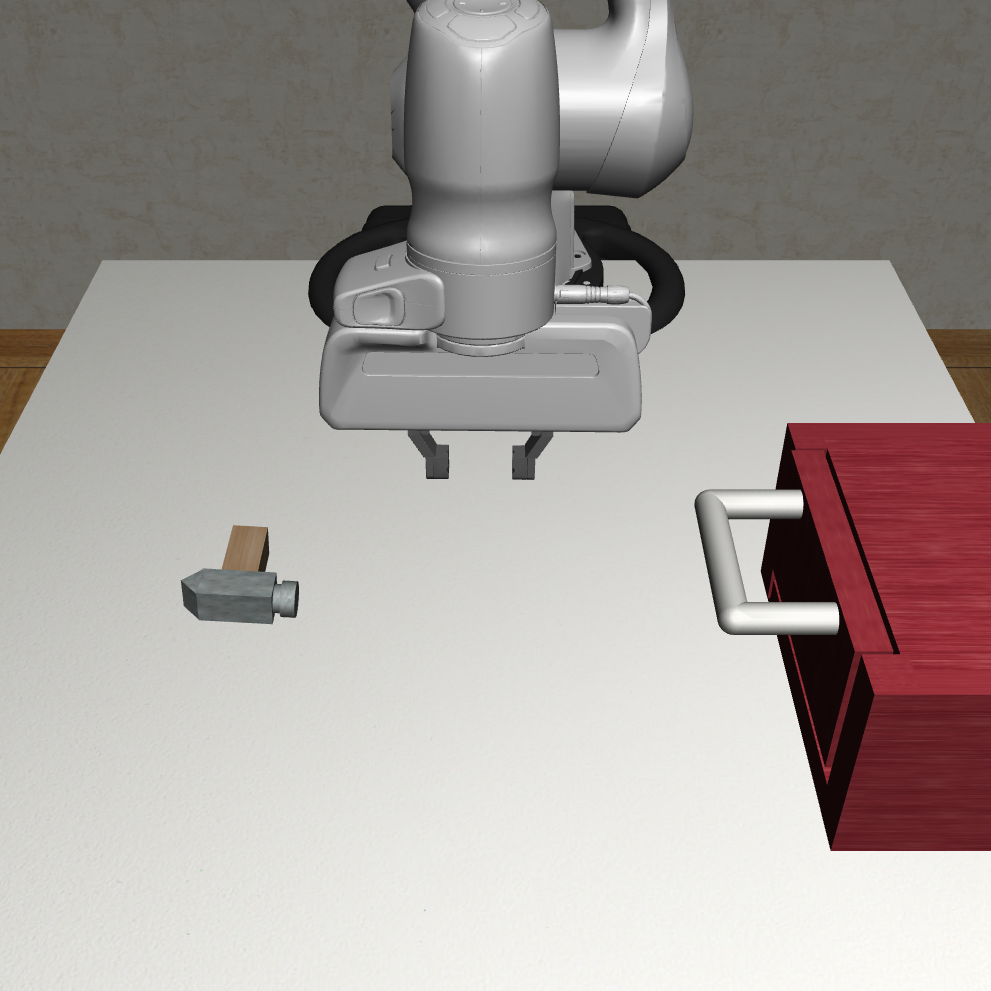}
        \vspace{-5mm}        
        \subcaption{}
      \end{minipage}
      \hfill
     \begin{minipage}[t]{0.17\linewidth}
        \includegraphics[width=\linewidth,trim=0cm 0cm 0cm 0cm,clip]{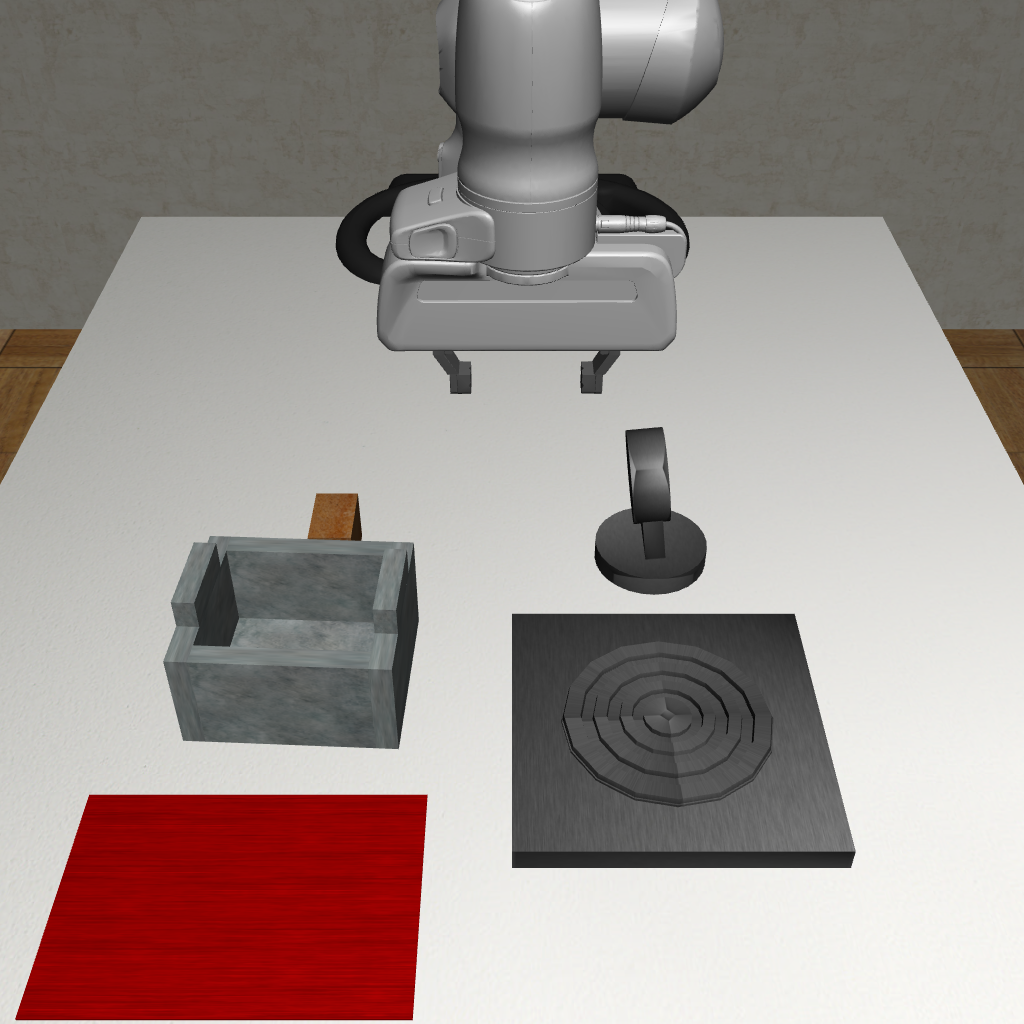}
        \vspace{-5mm}        
    \subcaption{}
      \end{minipage}   
      \hfill
     \begin{minipage}[t]{0.17\linewidth}
        \includegraphics[width=\linewidth,trim=0cm 0cm 0cm 0cm,clip]{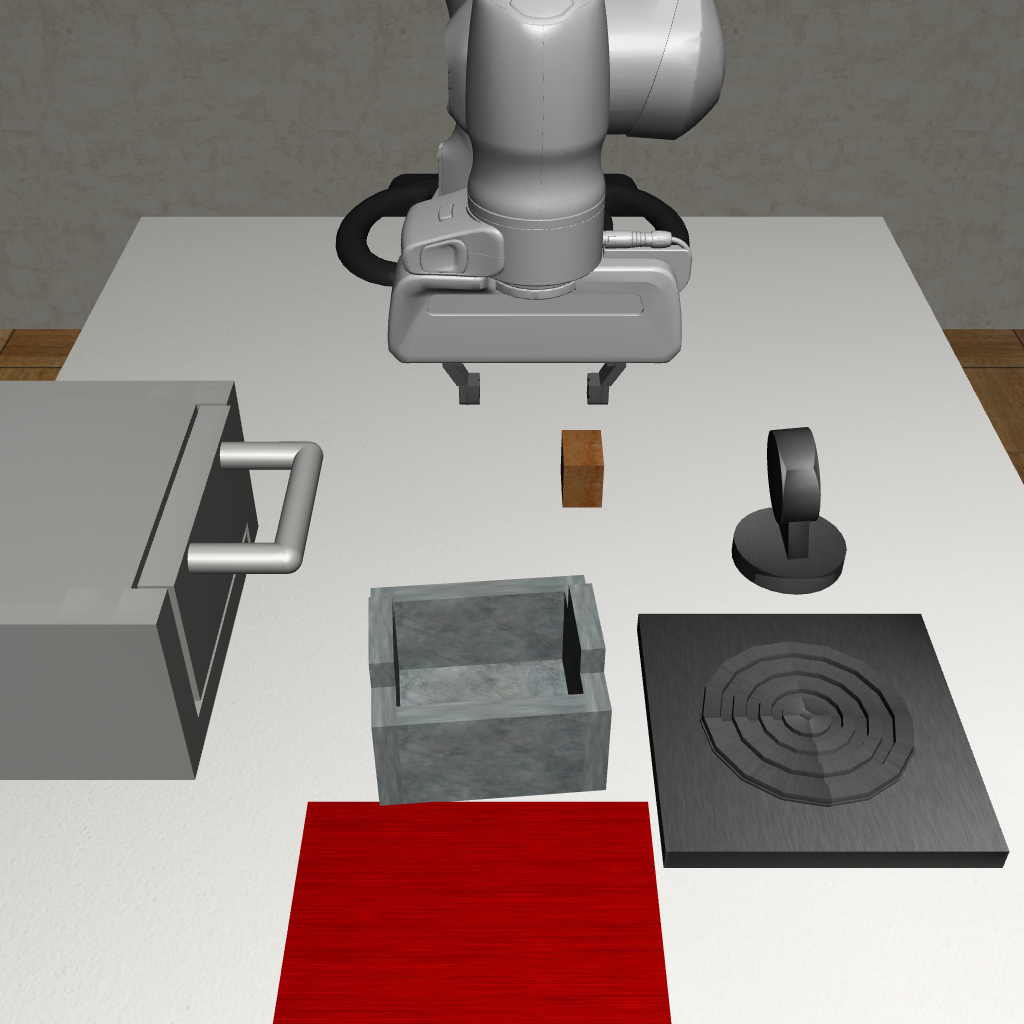}
        \vspace{-5mm}        
     \subcaption{}   
      \end{minipage}        
      \hfill
     \begin{minipage}[t]{0.17\linewidth}
        \includegraphics[width=\linewidth,trim=0cm 0cm 0cm 0cm,clip]{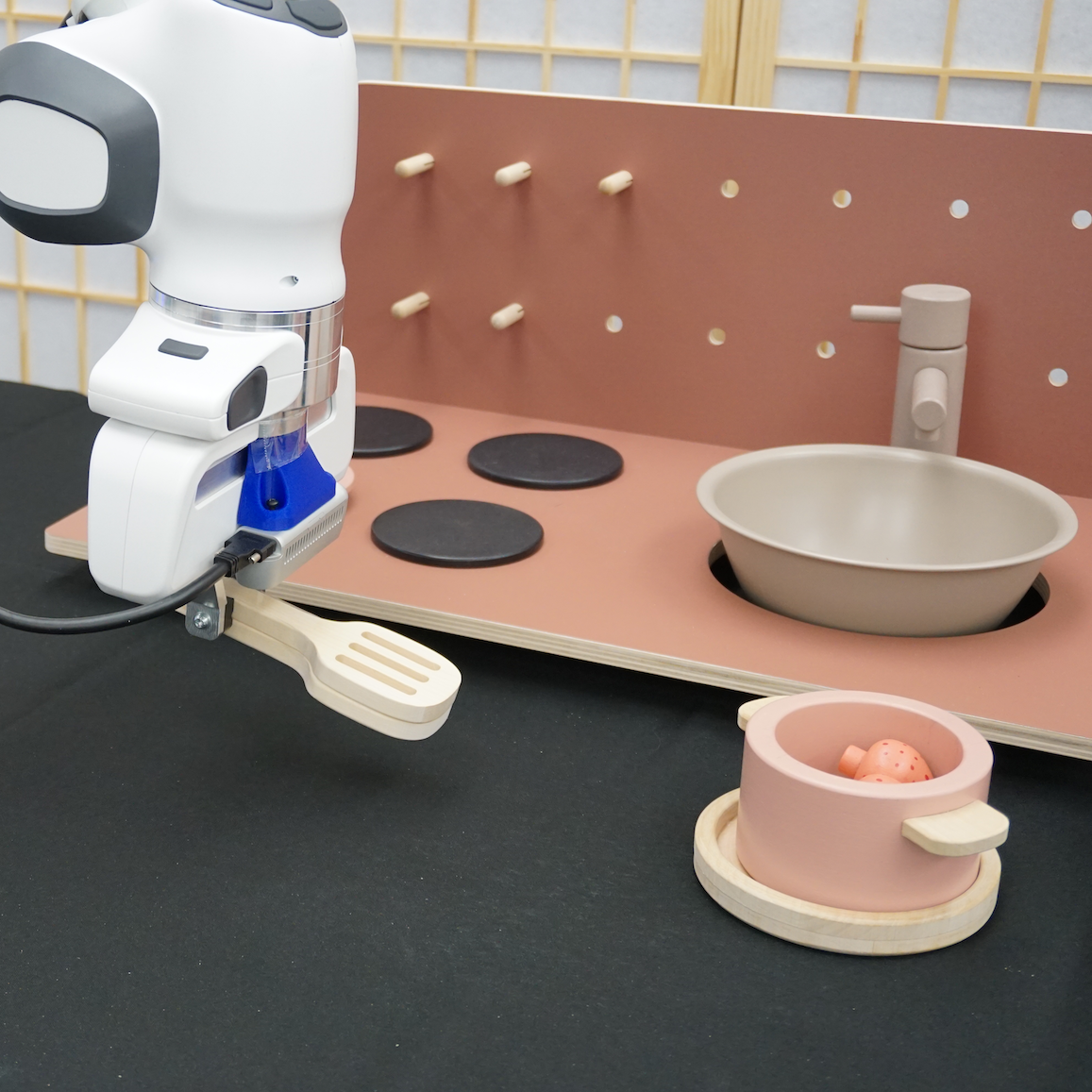}
        \vspace{-5mm}         
     \subcaption{}   
      \label{fig:real}
      \end{minipage}         
     \caption[Visualization of evaluated tasks in \buds{} experiments.]{Visualization of the four simulation tasks and one real robot task used in our experiments.
    (a)~\tooluse{}; (b)~\hammer{}; (c)~\kitchen{}; (d)~\multitask{}; (e)~\budsrealrobot{}.
    }
    \label{fig:buds:experiment}
\end{figure}

Figure~\ref{fig:buds:experiment} visualizes all the tasks. The first three single-task environments, \tooluse, \hammer, and \kitchen,~are designed primarily for baseline comparisons and ablation studies. The multitask domain \multitask{}~is designed for investigating the quality and reusability of skills discovered from multitask demonstrations. The \budsrealrobot{} task is designed for real-world validation and deployment. We provide detailed task descriptions below. For all the experiments, we use the Franka Emika Panda arm with a position-based Operational Space Controller and a binary command for controlling the parallel-jaw gripper. The meta-controller runs at $4$Hz, and the low-level skill policies run at $20$Hz. We provide brief descriptions of our evaluation tasks in the following and we provide more details about tasks in Appdendix~\ref{ablation_sec:buds:tasks}.

\begin{itemize}

\item \textbf{Single tasks.} The three single tasks require prolonged environmental interactions and encompass a broad range of prehensile and nonprehensile behaviors, including tool use and manipulation of articulated objects.
In \tooluse{}, the goal is to place a cube into a metal pot. Since the cube is initially beyond the robot's reach, the robot must first grasp an L-shaped tool to pull the cube closer. After retrieving the cube, the robot needs to set aside the tool, pick up the cube, and place it into the pot.
For \hammer{}, the goal is to place a small, hard-to-grasp hammer in a drawer and close it. This requires the robot to open the drawer, place the hammer inside, and close the drawer.
In \kitchen{}, the most complex of the three tasks, the goal is to cook and serve a simple dish in the serving region while managing the stove. This task requires a sequence of subtasks: turning on the stove, placing the pot on the stove, adding ingredients to the pot, moving the pot to the table, pushing it to the serving region (marked as a red region on the table), and finally turning off the stove.

\item \textbf{\multitaskkitchen{}.} The multitask domain comprises three tasks (\task{1}, \task{2}, and \task{3}), each with distinct goals.
For \task{1}, the goal state requires the drawer to be closed, the stove turned on, the cube placed in the pot, and the pot positioned on the stove. For \task{2}, the goal state requires the drawer to be closed, the stove turned on, the cube placed in the pot, and the pot moved to the serving region. For \task{3}, the goal state requires the drawer to be closed, the stove turned off, the cube placed in the pot, and the pot moved to the serving region.
To evaluate skill reusability, we design three variants (\variant{1}, \variant{2}, \variant{3}) for each task with different initial configurations, detailed in Figure~\ref{fig:buds:multitask-configs}. These varying initial configurations necessitate different combinations of subtasks, allowing us to examine whether skills learned in a subset of task variants can be effectively reused in new variants.
% \item \textbf{\multitaskkitchen{}.} The multitask domain includes three tasks (referred to as \task{1}, \task{2}, and \task{3}) that are distinct from each other by their goals. The goal state of \task{1} entails the drawer closed, the stove on, the cube in the pot, and the pot placed on the stove. The goal of \task{2} entails the drawer closed, the stove on, the cube in the pot, and the pot in the serving region. The goal of \task{3} entails the drawer closed, the stove off, the cube in the pot, and the pot in the serving region. To study the reusability of our skills, we design three variants for each task based on their initial configurations, which we refer to as \variant{1}, \variant{2}, \variant{3}. We describe all task variants in detail in Figure~\ref{fig:buds:multitask-configs}. Different initial configurations require solving different combinations of subtasks. Therefore, we examine whether skills learned in a subset of task variants can be reused in new variants.

\item \textbf{\budsrealrobot{}.}  The task requires versatile behaviors, including grasping, pushing, and tool use. The robot needs to remove the lid of the pot, place the pot on the plate, pick up the tool, use the tool to push the pot along with the plate to the side of the table, and put down the tool in the end. We use the workspace camera (Kinect Azure) and the eye-in-hand camera (Intel Realsense D435i) for capturing RGB images.

\end{itemize}

\paragraph{Evaluation Horizons.} The maximum number of steps we set during evaluation varies based on the task difficulty. We set $1500$ for \tooluse{}, \hammer{}, $5000$ for \kitchen{}, $2500$ for tasks in \multitask{}, and $5000$ for \budsrealrobot{}.

\paragraph{Data Collection.} We collect demonstrations through teleoperation with a SpaceMouse. We collect $100$ demonstrations for each of the three single tasks (less than 30 minutes for each task), $120$ demonstrations for each task ($40$ for each of the three task variants) in \multitaskkitchen{}, and $50$ demonstrations for \budsrealrobot{}. Each demonstration consists of a sequence of sensory observations (images from the workspace and the eye-in-hand cameras, proprioception) and actions (the end-effector displacement and the gripper open/close).

\subsection{Results}
\label{sec:buds:experiments:results}
For all simulation experiments, we evaluate \buds{} and baseline methods in each task for 100 trials with random initial configurations (repeated with five random seeds). We use the success rate over trials as the evaluation metric, and a trial is considered successful if the task goal is reached within the maximum number of steps. \loosepar{}

\paragraph{Single Task Experiments.} We compare \buds{} with imitation learning baselines on the single-task environments. We answer question (1) by examining the efficacy of hierarchical modeling for long-horizon tasks. Specifically, we compare \buds{} with a Behavioral Cloning (BC) baseline~\cite{zhang2018deep}, which trains a monolithic neural network policy on the demonstrations. We also answer question (2) by examining our bottom-up clustering-based segmentation method. Specifically, we compare \buds{} with a second baseline that uses a classical Changepoint Detection (CP) algorithm~\cite{niekum2015online} to temporally segment the demonstrations while keeping the rest of the model design identical to ours.

Table~\ref{tab:buds:single-task-results} reports the quantitative results. BUDS outperforms both baselines for all three tasks, by over 20\% on average. The comparison between BC and \buds{} shows that while BC is able to solve short-horizon task reasonably well, it suffers a significant performance drop in longer tasks, such as \kitchen. In contrast, \buds{} breaks down a long-horizon task with skill abstraction, leading to a consistent high performance across tasks of varying lengths.  The comparison between CP and \buds{} suggests that the quality of skill segmentation significantly impacts the final performance. Qualitatively, we find that the CP baseline fails to produce coherent segmentation results across different demonstrations, hindering the efficacy of policy learning.

% We observe two major failure modes in \buds{}: 1) Incorrect selection of skills due to out-of-distribution states, 2) Manipulation failures due to imprecise grasps. We quantify the failure modes in the \kitchen{} task with five repeated runs. Failures due to the first mode take up $12.3\% \pm 2.9\%$ of the evaluation trials, and failures due to the second one take up $9.0\% \pm 3.6\%$. Such failure modes can be alleviated by using structured input representations as mentioned in Chapters~\ref{chapter:viola} and~\ref{chapter:groot}, which is a good future extension. 

% Both failure types pertain to the fundamental limitations of imitation learning on small offline datasets. We believe the model performance could be improved with large-scale training and online robot experiences. We leave it for future work.

\begin{table}[t]
\centering
%\vspace{-2mm}
  \begin{tabular}{lccc}
    \toprule
    \textbf{Environments} & \textbf{BC~\cite{zhang2018deep}} & \textbf{CP~\cite{niekum2015online}} & \textbf{\buds{} (Ours)}\\
    \midrule
    \tooluse~ & 54.0 $\pm$ 6.3 & 36.8 $\pm$ 5.1 & \textbf{58.6} $\pm$ 3.1  \\
    \hammer~   & 47.8 $\pm$ 3.7 & 60.4 $\pm$ 4.5 & \textbf{68.6} $\pm$ 5.7 \\
    \kitchen~  & 24.4 $\pm$ 5.3 &  23.4 $\pm$ 3.4   & \textbf{72.0} $\pm$ 4.0\\
    \bottomrule
  \end{tabular}
\caption[Evaluation of \buds{} in single-task experiments.]{\label{tab:buds:single-task-results} Success rates (\%) of tasks in single-task experiments.}
\end{table}

\paragraph{Comparison to Hierarchical Imitation Learning Algorithms.} BUDS shares the same principle with prior works on hierarchical imitation learning, and we choose GTI~\cite{mandlekar2020learning} for our baseline comparison.
% including IRIS~\cite{mandlekar2020iris}, GTI~\cite{mandlekar2020learning}, and RPL~\cite{gupta2020relay}.
One notable distinction is that this baseline considers a single low-level skill policy rather than a library of skills. 

To compare BUDS with GTI, we evaluate our method with varying numbers of skills through a parameter sweep on the number of clusters $K$ in the spectral clustering step. In the special case when \buds{} has only a single skill ($K=1$), our method is equivalent to a variant of GTI without the image reconstruction term.
Table~\ref{tab:buds:varying-K} reports the results in the \kitchen{} task. We observe that the number of skills has a salient impact on model performance. Intuitively, when $K$ is too small, each skill will have difficulty dealing with diverse subgoals and various visual observations. When $K$ is too large, each skill has fewer data points to train on, as the dataset is fragmented into smaller partitions. The peak performance is observed with $K=6$ skills, which is the value we use for the main experiments.

The GTI variant with a single skill ($K=1$) fails to achieve non-zero task success. We also implemented the original GTI with the image reconstruction term, but observed no significant change in performance. After analyzing the qualitative behaviors of the GTI policy, we find that it works fine if the initial state is close to the task goal, but it cannot handle initial states that are further away. For quantitative evidence, we conduct an additional evaluation with the \tooluse{} task, where we reset the robot to the state when it has already fetched the cube and placed the tool down. To complete the task, the robot only needs to pick up the cube and place it in the pot. In this shorter subtask, the GTI variant and BUDS achieve $63.0\%$ and $60.3\%$ success rates, respectively. In comparison, they only achieve $0.0\%$ and $58.6\%$ success rates when starting from the original initial states (Table~\ref{tab:buds:single-task-results}). These results imply that GTI does not generalize well to the long-horizon tasks compared to \buds{}.

\begin{table}[t!]
\centering
\makeatletter\def\@captype{table}
  \resizebox{1.0\linewidth}{!}{  
  \begin{tabular}{lccccc}
    \toprule
    \textbf{} & $K=1$ (GTI~\cite{mandlekar2020learning}) & $K=3$ & $K=6$ & $K=9$ & $K=11$ \\
    \midrule
    \kitchen~ & $0.0\pm0.0$ & $24.2\pm3.6$ & $72.0\pm4.0$ & $60.6\pm6.53$ & $44.6\pm3.38$  \\
    \bottomrule
  \end{tabular}
  }
  \caption[Evaluation on \kitchen{} task with varying number of discovered skills.]{\label{tab:buds:varying-K} Success rates (\%) in \kitchen{} with varying numbers of skills.}  
  \end{table}

\paragraph{Learning from Multitask Demonstrations.}
\label{sec:buds:multitask}
We investigate if \buds{} is effective in learning from multitask demonstrations, answering question (3).~\buds{} discovers $K=8$ skills in the \multitaskkitchen{} domain, and we examine the skills from two aspects: \textit{quality} and \textit{reusability}. For quality, we examine if skills learned from multitask demonstrations are better than those from individual tasks. For reusability, we examine if skills can be composed to solve new task variants that require different subtask combinations. 

We evaluate three settings: 1) \budstrainmulti: the skills are discovered and trained on the multitask demonstrations, and the meta-controller is trained for each task respectively; 2) \budstrainsingle: the skills are discovered from demonstrations of each individual task; so is the meta-controller; and 3) \budstest: the skills are trained on demonstrations of \variant{1} and \variant{2} and the meta-controller is trained on \variant{3}. Table~\ref{tab:buds:multitask} presents the evaluation results. The comparisons between \budstrainmulti{} and \budstrainsingle{} indicate that skills learned across multitask demonstrations improve the average task performance by $8\%$ compared to those learned on demonstrations of individual tasks. We hypothesize that the performance gains are rooted in our method's ability to augment the training data of each skill with recurring patterns from other tasks' demonstrations. The results on \budstest{} show that we can effectively reuse the skills to solve the new task variants that require different combinations of the skills by solely training a new meta-controller to invoke the pre-defined skills.

% We also observe the low performance of \variant{3} on \budstest{} because it has more subtasks than its training counterparts, and the execution failure of each skill compounds, leading to a low success rate. 

% We observe that low performance of \task{3} on \budstest{} because \task{3} contains more subtasks than its training counterparts. Additionally, since the execution failures of individual skills accumulate, this leads to a low overall success rate.

\begin{table}[t]
\centering
\makeatletter\def\@captype{table}
  \begin{tabular}{lccc}% {lllll}
    \toprule
    \textbf{} & \budstrainmulti{} & \budstrainsingle{} & \budstest{}\\
    \midrule
    \task{1} &  $70.2 \pm 2.2$ & $52.6 \pm 5.6$ & $59.0 \pm 6.4$ \\
    \task{2} & $59.8 \pm 6.4$  & $60.8 \pm 1.9$ &  $55.3 \pm 3.3$  \\
    \task{3} &$75.0 \pm 2.0$ & $67.6 \pm 1.8$  & $28.4  \pm 1.5$ \\    
    \bottomrule
  \end{tabular}
\caption[Evaluation of \buds{} in multitask experiments.]{\label{tab:buds:multitask} Success rates (\%) in \multitask{}.}   
\end{table}

\begin{figure}
    \centering
    \begin{minipage}[t]{1.0\linewidth}
    \includegraphics[width=\linewidth]{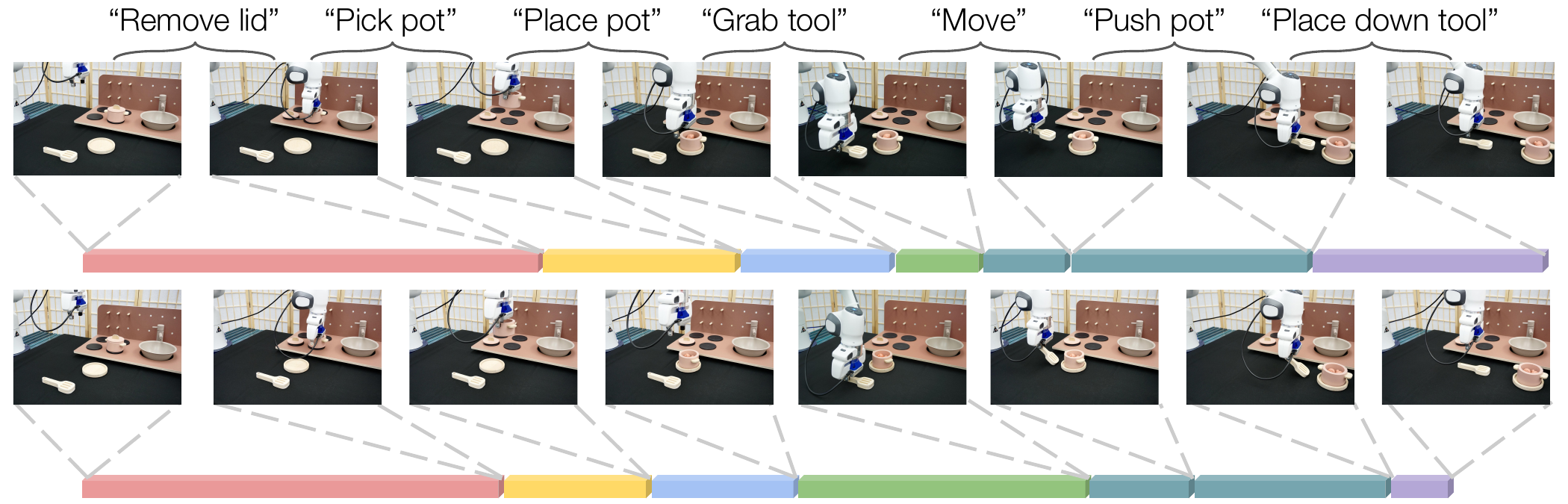}
    \end{minipage}
    \caption[\buds{} skill segmentation visualization.]{\buds{} skill segmentation visualization.}
    \label{fig:buds-segmentation-result}
\end{figure}

\paragraph{Real Robot Experiments.} To answer question (4), we perform evaluations in the \budsrealrobot{} task to validate the practicality of \buds{} for solving real-world manipulation tasks. Quantitatively, we evaluate $50$ trials on varying initial configurations, achieving a $56 \%$ success rate. The performance is on par with the performance of our simulation evaluations, showing that \buds~generalizes well to real-world data and physical hardware. We also evaluate the most competitive baseline CP model on the real robot, which only achieved a $18 \%$ success rate. A consistent failure mode of this baseline is that the robot fails to place the pot correctly on the plate. To answer question (5), we also qualitatively visualize the temporal segments of two demonstration sequences collected for the \budsrealrobot{} task in Figure~\ref{fig:buds-segmentation-result}. While our clustering-based segmentation algorithm is fully unsupervised, our qualitative inspection identifies consistent segments that can be interpreted with semantic meanings.

\paragraph{Varying Numbers of Skill Segments.} We present how the number of skills affects the performance of a task. We segment demonstrations of \kitchen{} task into different numbers of skills. Different numbers of skills can be discovered by varying the maximum number of clusters in the spectral clustering step. We show in Table~\ref{tab:buds:varying-K} that the number of skills matters to the performance, and just using a single skill under the hierarchical imitation learning framework does not lead to any performance for the \kitchen{} task.

\begin{figure}
\centering
  \begin{minipage}[t]{1.0\textwidth}
\makeatletter\def\@captype{figure}
  \includegraphics[width=1.0\linewidth,trim=0cm 0cm 0cm 0cm,clip]{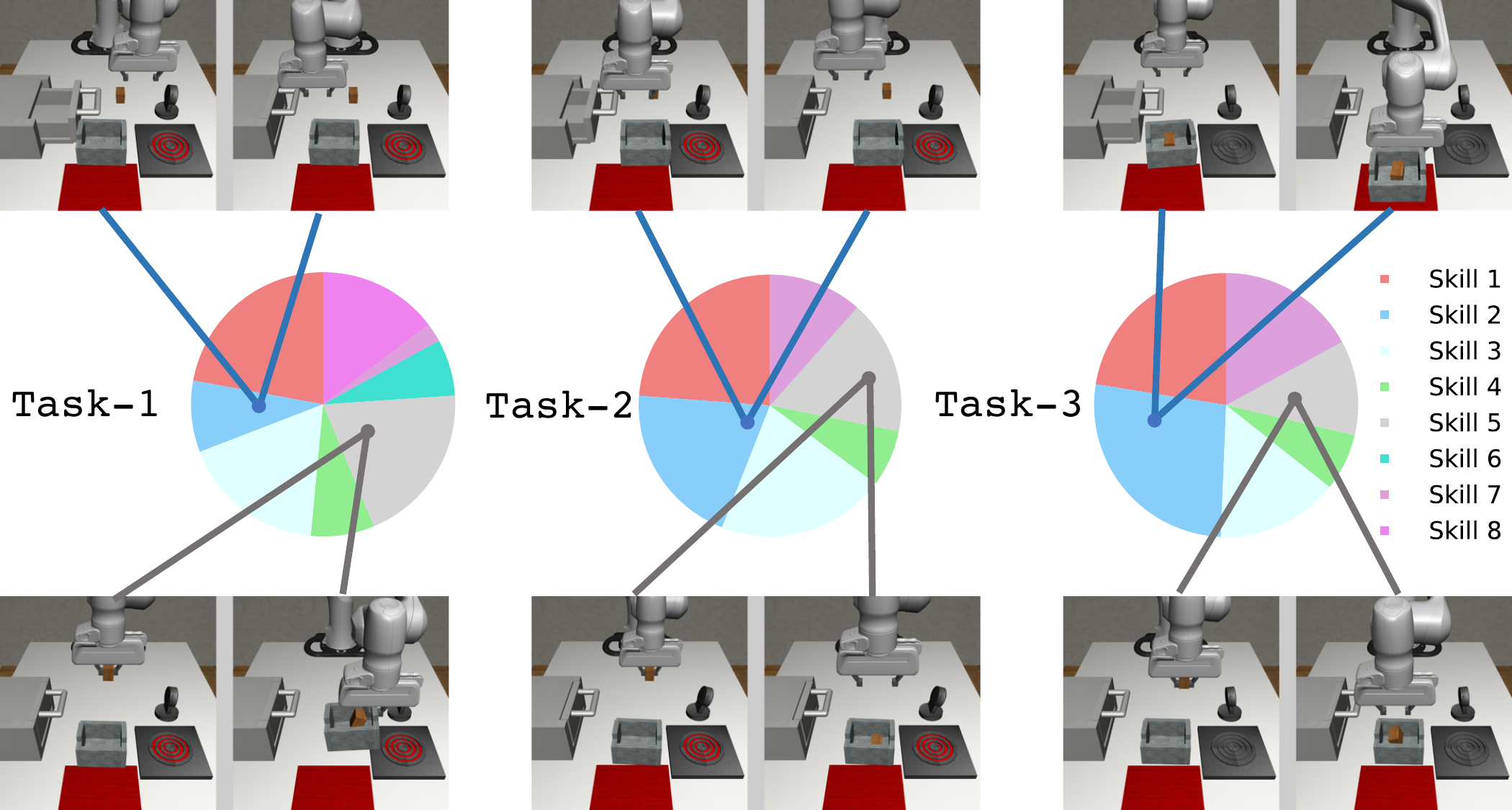}
  \caption[Percentage of discovered skills in each task in multitask experiments.]{We visualize the percentage of skills in each task. We also show three sequences of skill segments corresponding to Skill $2$ and Skill $5$. While segments of the same skill have some differences across the tasks, the overall semantic event the skills correspond to is consistent.}   
  \label{fig:buds:skill-percentage}
  \end{minipage}
\end{figure}

\paragraph{Reusable Skills in \multitask{} Domain.} In Figure~\ref{fig:buds:skill-percentage}, we use pie charts to show the percentage of the $8$ discovered skills in each of the \multitask{} tasks (We omit the number in the figure for clarity). As mentioned in Section~\ref{sec:buds:experiments}, most of the skills are indeed shared and reused across different tasks. We also visualize sequences of two skills across all three tasks in the figure. As we can see from the visualization, similar skills with slightly different visual features (closing the drawer, but the stove is either on or off) are clustered in one skill, and such variations in training data can improve the policy robustness to visual variance in the inputs.

\section{Summary}

In this chapter, we introduced \buds{}, a hierarchical approach to tackling vision-based manipulation by discovering sensorimotor skills from unsegmented demonstrations. \buds{} identifies recurring patterns from multitask demonstrations based on multi-sensory cues. Then, it trains the skills on the recurring temporal segments with imitation learning and designs a meta-controller to compose these skills for completing tasks. The results show the effectiveness of \buds{} in simulation and on real hardware. We also examine the impacts of different model designs through ablation studies.

\paragraph{Limitations and Future Work.} In this chapter, we assumed that \buds{} learns from expert demonstrations. However, data might not always be of high quality. One promising future direction is to enable skill discovery from suboptimal demonstrations and improve the discovered skills through interaction. Another potential extension of \buds{} is to stitch low-level skills together in sequences different from the demonstration orders. This ability to stitch skills in novel combinations could enable new task generalizations without requiring additional demonstration data.

While many interesting directions can be pursued based on \buds{}, this dissertation focuses on building on top of \buds{} to achieve continual learning of robot policies in the lifelong learning setting. \buds{} builds the groundwork for us to exploit behavioral regularity given demonstration data from multiple tasks. Building on top of \buds{}, we describe in the next chapter how to perform bottom-up skill discovery from demonstrations in a lifelong learning setting, developing robots that continually learn to discover skills and generalize across both previous and future tasks.

\chapter{Lifelong Imitation Learning through Skill Discovery}
\label{chapter:lotus}

In the previous chapter, we introduced \buds{} which discovers reusable manipulation skills from demonstrations. \buds{} is considered in a multitask setting, where a fixed set of tasks is given. However, deploying robots in the open world necessitates continual learning in ever-changing environments. Imagine you bring home a new appliance---your future home robot is unlikely to have seen it before and must quickly learn to operate it. Such a scenario pinpoints the importance of lifelong learning capabilities~\cite{thrun1995lifelong}, with which a robot can continually learn and adapt its behaviors over time. Lifelong robot learning is particularly challenging as it involves continual adaptation under distribution shifts throughout a robot's lifespan.

In this chapter, we design a \textit{lifelong imitation learning} method for robot manipulation. We aim to develop a practical algorithm that learns over a sequence of new tasks a physical robot may encounter. We continue our insight from the previous chapter to discover skills based on \textit{behavioral regularity}. Because of the existence of behavioral regularity, we can design an algorithm to identify recurring temporal segments from demonstrations in the lifelong learning setting. Our algorithm builds up a continually growing skill library by adding new skills and updating old ones. Our method efficiently learns a policy that leverages skills from past experiences, enabling both improved performance on new tasks (\textit{forward transfer}) and maintained performance on the previously learned tasks (\textit{backward transfer}). Our work in this chapter was published at the IEEE International Conference on Robotics and Automation, 2024~\cite{wan2023lotus}.

\section{\lotus{}}
\label{sec:lotus:method}

\begin{figure}[t]
    \centering
    \includegraphics[width=1.0\linewidth]{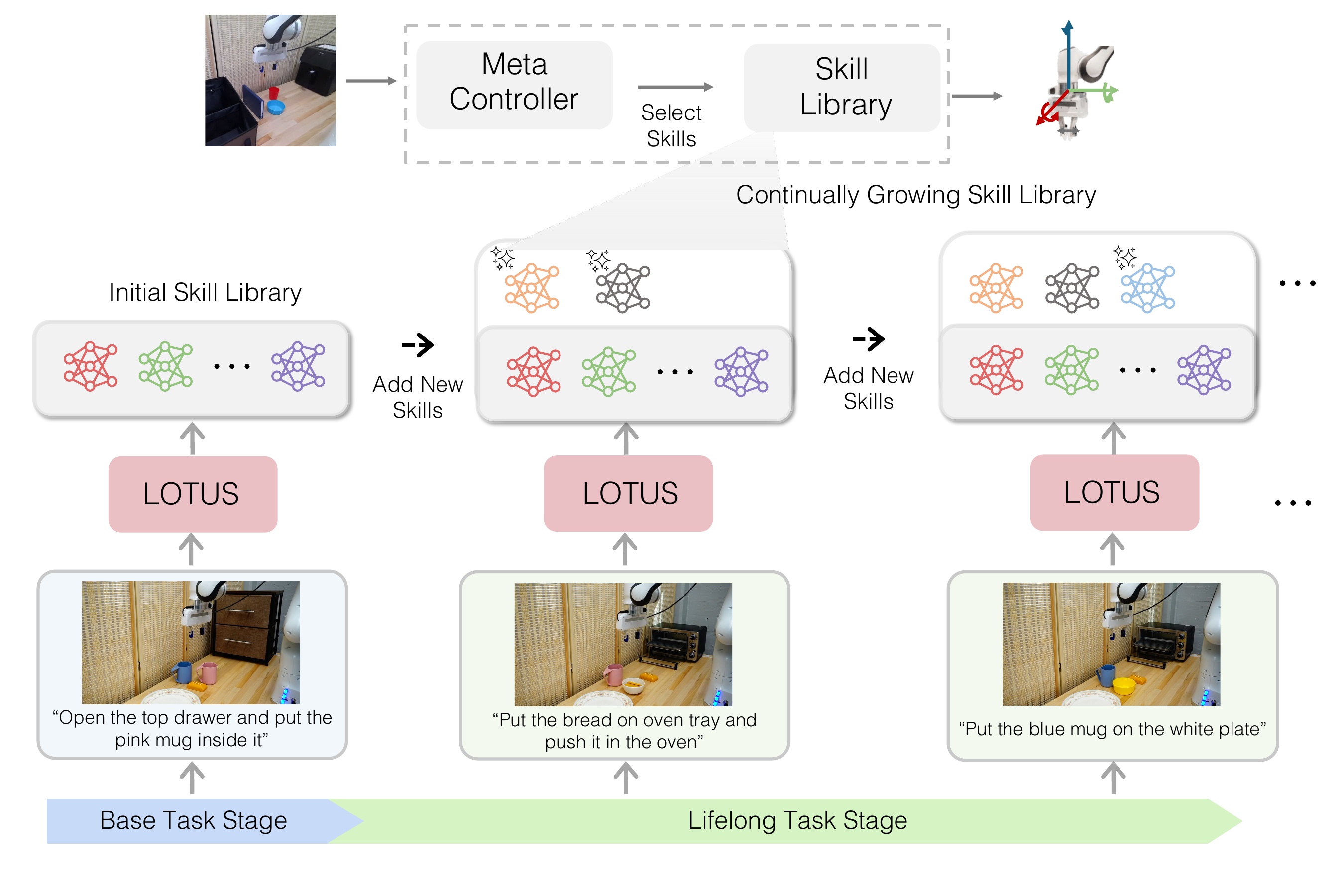}
    \caption[\lotus{} overview.]{\textbf{\lotus{} Overview.} \lotus{} is a lifelong imitation learning algorithm through continual skill discovery. \lotus{} starts from the base task stage, where it builds an initial library of sensorimotor skills. In the subsequent lifelong task stage, \lotus{} continuously discovers new skills from a stream of incoming tasks and adds them to its skill library. A high-level meta-controller composes skills from the library to solve new manipulation tasks. We mark the newly acquired skills at each lifelong learning step with \raisebox{-0.1cm}{\includegraphics[width=0.5cm]{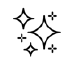}}.}
    \label{fig:lotus:overview}
\end{figure}

We present \lotus{} (\emphasize{L}ifel\emphasize{O}ng knowledge \emphasize{T}ransfer \emphasize{U}sing \emphasize{S}kills), a hierarchical imitation learning method for robot manipulation in a lifelong learning setting. The key to \lotus{} is building an ever-increasing library of skill policies through \textit{continual skill discovery}.  Throughout the lifelong learning process, \lotus{} continually adds new skills for learning new tasks while refining existing skills without catastrophic forgetting.  Moreover, a meta-controller is trained with hierarchical imitation learning to harness the skill library. Figure~\ref{fig:lotus:overview} shows the overview of \lotus{}. As we introduced in Section~\ref{sec:bg:open-world-formulation}, the lifelong learning process can be divided into two stages, namely \textit{base task stage} and \textit{lifelong task stage}. We first explain the formulation of hierarchical imitation learning with continual skill discovery. 
Then, we describe the two processes in \lotus{}: continual skill discovery with open-world perception and hierarchical imitation learning with Experience Replay (\er{}). In this chapter, we follow the notations of lifelong robot learning in Section~\ref{sec:bg:open-world-formulation}. We suggest readers review Section~\ref{sec:bg:open-world-formulation} before proceeding with this chapter.

\begin{figure*}[ht!]
    \centering
    \includegraphics[width=1.\linewidth, trim=0cm 0cm 0cm 0cm,clip]{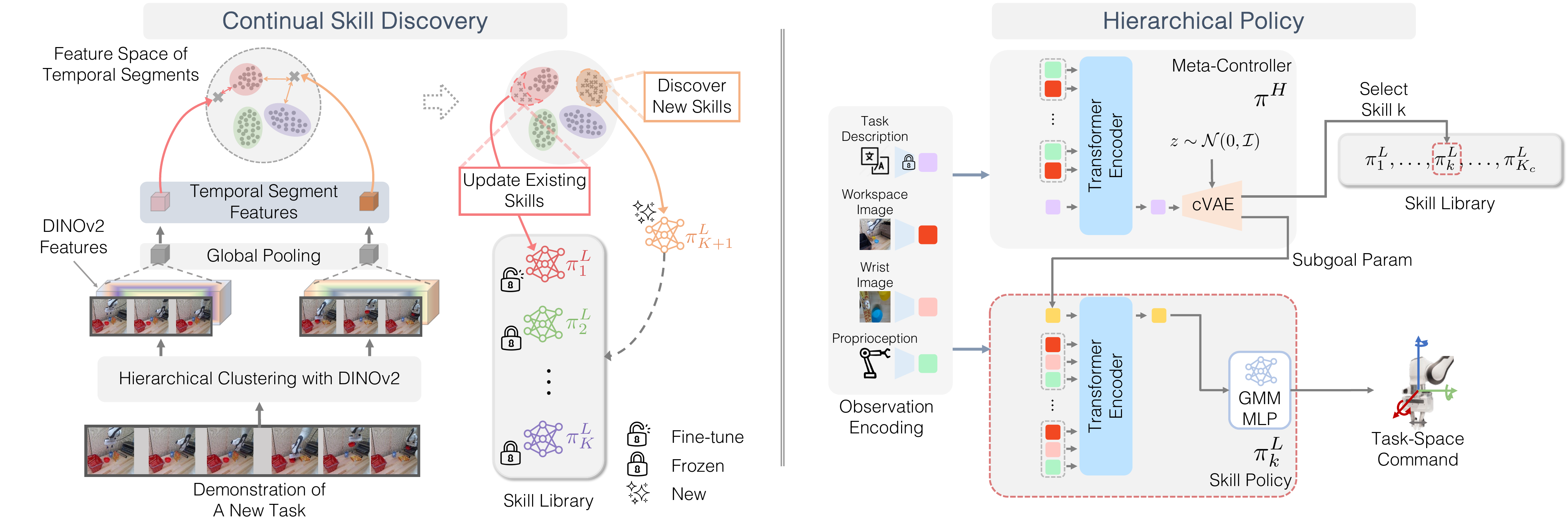}
    \vspace{-1mm}
    \caption[\lotus{} method overview.]{\textbf{Method Overview.} \lotus{} consists of two processes: continual skill discovery with open-world perception
    and hierarchical policy learning with the skill library. For continual skill discovery, we obtain temporal segments from demonstrations using hierarchical clustering with DINOv2 features. We then incrementally cluster the temporal segments into partitions to either update existing skills or learn new skills. For the hierarchical policy, a meta-controller $\metacontroller{}$ selects a skill by predicting an index $k$ and specifies the subgoal parameters for the selected skill $\skillpolicy{k}$ to achieve. Note that because the input to a transformer is permutation invariant, we also add sinusoidal positional encoding to input tokens to inform transformers of the temporal order of input tokens. We omit this information in the figure for clarity.}
    \label{fig:lotus:method-overview}
\end{figure*}

\paragraph{Hierarchical Imitation Learning With Continual Skill Discovery.}~\lotus{} aims to learn a hierarchical policy in the lifelong imitation learning setting. \lotus{} learns the hierarchical policy over a sequence of tasks that comes with demonstrations, and the hierarchical policy takes the same form as the one used in \buds{} (See Section~\ref{sec:buds:skill_discovery_formulation}). Key to \lotus{} is the skill discovery from demonstrations in the lifelong learning setting, termed continual skill discovery, where a growing library of skills is maintained. 

Continual skill discovery is performed as follows. At the beginning of the lifelong learning process (i.e., lifelong learning step $\llstep=0$), we use the skill discovery from demonstrations that is the same as in \buds{}, obtaining an initial library of low-level skills. This skill discovery process gives us $K_0$ number of low-level skills.  At a lifelong learning step $\llstep>0$, demonstrations are split into maximally $K_{\llstep}$ partitions, denoted as $\{\tilde{\dataset}_{k}\}_{k=1}^{K_{\llstep}}$. Each partition $\tilde{\dataset}_{k}$ is used to fine-tune an existing skill policy $k$ when $k \leq K_{\llstep-1}$ or train a new skill policy if $K_{\llstep-1} < k \leq K_{\llstep}$. If a partition $k$ is empty at step $\llstep$, we do not update $\skillpolicy{k}$. Then, the hierarchical policy can be trained using a hierarchical imitation learning algorithm with the partitioned data. Note that when learning at a step $\llstep>0$, demonstrations of tasks before the lifelong learning step $\llstep$ are not fully available. To tackle this challenge, we use memory buffers to store exemplar data from the previous tasks. 

\subsection{Continual Skill Discovery With Open-world Perception}
\label{sec:lotus:continual-discovery}
For continually discovering new skill policies $\{\skillpolicy{k}\}_{k=1}^{K}$, it is crucial to split demonstrations $D^{\tasknum}$ of a task $\Task{\tasknum}$ into partitions $\{\tilde{\dataset}_{k}\}_{k=1}^{K_{\llstep}}$. Here, $\tilde{\dataset}_{k}$ consists of training data for a skill policy $k$. The key to curating partitions for training skill policies is to identify the recurring temporal segments in the demonstration of new tasks. Our method, \lotus{}, first uses a bottom-up hierarchical clustering approach to temporally segment demonstrations and incrementally cluster temporal segments into partitions.

\paragraph{Temporal Segmentation with Open-World Vision Model.} To recognize recurring patterns for obtaining the partitions, \lotus{} first needs to identify the coherent temporal segments from demonstrations based on visual similarity~\cite{zhu2022buds}. We apply hierarchical agglomeration clustering on each demonstration~\cite{krishnan2017transition} following \buds{}, which breaks a demonstration trajectory into a sequence of disjoint temporal segments in a bottom-up manner. \loosepar{}

The primary challenge of applying hierarchical clustering in the lifelong learning setting is to consistently identify the semantic similarity between temporally adjacent segments in a non-stationary data distribution of lifelong learning.
We use DINOv2~\cite{oquab2023dinov2}, an open-world vision model that can output consistent semantic features of images from open-ended scenarios, allowing~\lotus{} to reliably measure the semantic similarity between temporally adjacent segments. Specifically, DINOv2 encodes an image into a latent feature vector that is correlated with visual semantics. To incorporate temporal information, we aggregate DINOv2 features of all frames in the segment using a global pooling operation. Then, the semantic similarity between consecutive temporal segments is quantified using the cosine similarity scores between the pooling features.

\paragraph{Incremental Clustering for Skill Partitions.}
The identified temporal segments from demonstrations pave the way for subsequent steps of identifying recurring patterns across tasks and clustering them into partitioned datasets that are used to train skill policies. We introduce incremental clustering, which clusters temporal segments into either existing or new partitions.
In the \emph{base task stage},~\lotus{} uses spectral clustering~\cite{rousseeuw1987silhouettes} to first partition the demonstrations into $K_0$ skills.~\lotus{} determines the value of $K_0$ using the Silhouette method~\cite{rousseeuw1987silhouettes}, which quantifies the score of how well data points match with their clusters on a scale of $-1$ to $1$. We sweep through the integer values of $K_0$ to find the one with the highest Silhouette value. In the subsequent \emph{lifelong task stages}
~\lotus{} continually groups new temporal segments into partitions. \lotus{} either adds a segment to an existing partition or creates a new partition used to train a new skill to learn new behaviors.

At the lifelong learning step $\llstep$,~\lotus{} calculates the Silhouette value between new temporal segments and the previous $K_{\llstep-1}$ partitions to determine the new partitions. Segments with Silhouette values above a threshold are grouped with the partition having the highest value, indicating high similarity between the temporal segment and existing skill. In contrast, segments with values below the threshold are assigned to a new partition. This process is repeated serially for all segments. Our preliminary results indicate that the clustering is robust and tolerates a range of threshold values.
After merging new segments into partitions,~\lotus{} clusters demonstrations $D^{\tasknum}$ into partitions $\{\tilde{\dataset}_{k}\}_{k=1}^{K_{\llstep}}$, each corresponding to an existing or new skill policy. \loosepar{}

\subsection{Hierarchical Imitation Learning With Experience Replay}
\label{sec:lotus:hierarchical-imitation}

\lotus{} uses the partitions $\{\tilde{\dataset}_{k}\}_{k=1}^{K_{\llstep}}$ obtained from continual skill discovery to train policies in the skill library $\{\skillpolicy{k}\}_{k=1}^{K}$. With the skill library, the meta-controller $\metacontroller{}$ can invoke individual skills.
In a lifelong learning setting, without full access to demonstrations from the previous tasks,~\lotus{} follows Experience Replay (\er{})~\cite{chaudhry2019tiny} and designates a memory buffer $B_\llstep$ at a lifelong learning step $\llstep$. This memory buffer stores the exemplar data from demonstrations~\cite{liu2023libero}. At lifelong learning step $\llstep$, \lotus{} designates a memory buffer $B_{\llstep}$ that stores the exemplar data from the previous demonstrations. 
Specifically,~\lotus{} trains the policy through behavior cloning using a combined dataset of new task demonstrations $D^{\tasknum}$ at step $\llstep$ and data from the memory buffer $B_{\llstep}$. After training,~\lotus{} updates the buffer by adding a subset of these demonstrations $B_{\llstep+1}=B_{\llstep}\cup D'^{\tasknum}$, where $D'^{\tasknum} \subset D^{\tasknum}$.

In the following paragraphs, we describe skill policy learning, where~\lotus{} keeps track of skill data partitions in the memory buffer and uses the partitioned data to train skill policies. Then, we describe the design of the meta-controller that invokes skills from the continually growing library.

\paragraph{Skill Policy Learning.}
The partitions $\{\tilde{\dataset}_{k}\}_{k=1}^{K_{\llstep}}$ from continual skill discovery provide the training datasets for each skill policy.
To retain the previous knowledge while finetuning the skill policies,~\lotus{} leverages exemplar data from the memory buffer $B_{\llstep}$ that also retains partition information of the saved demonstrations. The saved partition for a skill $k$ at a learning step $\llstep$ is denoted as $B_{k, \llstep}$ in buffer $B_{\llstep}$.
Every existing skill $\skillpolicy{k}$ is fine-tuned using $\tilde{\dataset}_{k}\cup B_{k, \llstep}$, while newly-created skills are directly trained on $\tilde{\dataset}_{k}$.
After the policy finishes training on task $\Task{\tasknum}$, the memory buffer updates the information of skill partitions with a subset of demonstrations $\tilde{\dataset}^{'}_{k} \subset \tilde{\dataset}_{k}$ for partition $k$ such that $B_{k, \llstep+1}=B_{k, \llstep} \cup \tilde{\dataset}^{'}_{k}$.

~\lotus{} models each low-level skill as a goal-conditioned visuomotor policy, which allows the meta-controller to specify which subgoal for the skill to achieve. The same as \buds{},
~\lotus{} encodes the look-ahead images within temporal segments using a subgoal encoder into the subgoal embedding $\param_t$ and is jointly trained with skill policies.
Representing the subgoals in latent space makes it computationally tractable for the meta-controller to predict $\param_t$ during inference.
The history of input images from the workspace and wrist cameras are encoded with ResNet-18 encoders~\cite{he2016deep} before passing it through the transformer encoder~\cite{vaswani2017attention, zhu2022viola} along with the subgoal embedding $\param_t$. For computing the actions, the output token from the transformer encoder corresponding to $\param_t$ is then passed into a Gaussian Mixture Model (GMM) neural network~\cite{mandlekar2021matters,bishop1994mixture} to model the multi-modal action distribution in demonstrations.

\paragraph{Skill Composition With The Meta-Controller.} 
To operationalize the learned library of skill policies, we use a meta-controller $\metacontroller{}$ to compose the skills. $\metacontroller{}$ is designed for two purposes: to select a skill $k$, and to specify subgoals $\param_t$ for the selected skill to achieve. Conditioning on task context $\context^{\tasknum}$, $\metacontroller{}$ decides which skill and subgoal to achieve given state observations. To understand the task progress, $\metacontroller{}$ perceives the current layouts of objects and the robot's states through workspace and eye-in-hand images and robot proprioception, respectively.
As the true state of the task is partially observable, the meta-controller takes a history of past observations as inputs and uses a transformer for temporal modeling of the underlying states. The observation inputs are converted to tokens with ResNet-18 encoders~\cite{he2016deep}, whereas $\context^{\tasknum}$ is encoded into a language token by a pretrained language model, Bert~\cite{devlin2018bert}, before passing it to the transformer encoder.
To capture the multi-modal distribution of skill indices and subgoals underlying demonstrations,~\lotus{} trains a conditional Variational Auto-Encoder (cVAE) by minimizing an ELBO loss over demonstrations~\cite{kingma2013auto}.

The meta-controller must handle a varying number of skills due to the ever-growing nature of the skill library. To address this challenge, we design the meta-controller with an output head that predicts a sufficiently large number of skills, noted $K_{max}$. Then, we introduce a binary mask whose first $K_{\llstep}$ entries are set to $1$ at a lifelong learning step $\llstep$. The mask limits the skill index prediction to the existing set of skills, and when new skills are added, the meta-controller can predict more skills based on modified masking. Note that the mask does not change back when the policy is evaluated on tasks prior to step $\llstep$, so that $\metacontroller{}$ can transfer new skills to the previously learned tasks.
Concretely, the meta-controller operates in three steps. First, it predicts logits $\mathbf{p} \in \mathbb{R}^{K_{max}}$. Next, it applies masking to generate modified logits $\mathbf{p}'$, where $\mathbf{p}'_k = \mathbf{p}_k$ if $k \leq K_{\llstep}$, and $\mathbf{p}'_{k}=-\infty$ otherwise. Finally, it computes the probability for each skill $k$ using $\text{Softmax}(\mathbf{p}'_k) = \frac{e^{\mathbf{p}'_k}}{\sum_{j=1}^{K_{\text{max}}} e^{\mathbf{p}'_j}}$. This masking approach prevents numerical instability that would occur if masking are applied directly to the output probabilities from logits $\mathbf{p}$, which would require probability reweighting.

% the meta-controller first predicts logits $\mathbf{p} \in \mathcal{R}^{K_{max}}$, then applies masking to $\mathbf{p}$ which returns modified logits denoted as $\mathbf{p}'$ where $\mathbf{p}'_k = \mathbf{p}_k$ if $k \leq K_{\llstep}$; otherwise, $\mathbf{p}'_{k}=-\infty$. The probability for a given skill $k$ is computed as $\text{Softmax}(\mathbf{p}'_k) = \frac{e^{\mathbf{p}'_k}}{\sum_{j=1}^{K{\text{max}}} e^{\mathbf{p}'_j}}$. This masking design avoids any numerical instability if applying masking directly on the output probability from logits $\mathbf{p}$, which would have required reweighting output probability.

\section{Experiments}
\label{sec:lotus:experiments}
We design the experiments to answer the following questions: 1) Does hierarchical policy design improve knowledge transfer in the lifelong learning setting? 2) Do the newly discovered skills facilitate knowledge transfer? 3) Is the hierarchical policy design of \lotus{} more sample-efficient than baselines that do not use skills? 4) Does the choice of large vision models affect knowledge transfer? 5) Is~\lotus{} practical for real-robot deployment?\loosepar{}

\subsection{Experimental Setup}
\label{sec:lotus:experimental-setup}
\paragraph{Simulation Experiments.} 
% We conduct evaluations in simulation using the task suites from the lifelong robot learning benchmark, \todo{\libero{}, which will be formally introduced in Chapter~\ref{chapter:libero}. We select three suites, namely \liberoobject{} (10 tasks), \liberogoal{} (10 tasks), and \liberofifty{} (50 kitchen tasks from LIBERO-100).} The three task suites evaluate the robot's ability to understand object concepts, execute various motions, and combine these two capabilities. Experiments over the three task suites emulate the complexity involved in lifelong learning that allows us to comprehensively evaluate the performance of \lotus{} and other baselines. 
We conduct evaluations in simulation using the task suites from \libero{}\footnote{The complete details of \libero{} task suites are presented in Chapter~\ref{chapter:libero}. We defer the description of \libero{} in the next chapter so that we can maintain the flow between Chapters~\ref{chapter:buds} and~\ref{chapter:lotus}.}, our lifelong robot learning benchmark which is designed for studying lifelong robot learning algorithms systematically. Specifically, we select three task suites from \libero{}, namely \liberoobject{} (10 tasks), \liberogoal{} (10 tasks), and \liberofifty{} (50 kitchen tasks from LIBERO-100). The three task suites evaluate the robot's ability to understand object concepts, execute various motions, and combine these two capabilities.
Experiments over the three task suites emulate the complexity involved in lifelong learning that allows us to comprehensively evaluate the performance of \lotus{} and other baselines.

We now describe the configurations for our lifelong learning experiments, building on the notation introduced in Section~\ref{sec:bg:open-world-formulation}. The variable $\tasknum_{\llstep}$ represents the number of tasks learned at a lifelong learning step $\llstep$. The total number of lifelong learning steps is denoted by $\LLTotalStep$. Both \liberoobject{} and \liberogoal{} undergo five lifelong learning steps ($\LLTotalStep=4$), while \liberofifty{} undergoes six steps ($\LLTotalStep=5$). During the base task stage, we initialize \liberoobject{} and \liberogoal{} with six tasks ($\tasknum_{\text{init}}=6$) and \liberofifty{} with 25 tasks ($\tasknum_{\text{init}}=25$). In the lifelong task stage, \liberoobject{} and \liberogoal{} learn one new task per step, while \liberofifty{} learns five new tasks per step. Each base task across all suites includes 50 demonstration trajectories. Each task introduced in the lifelong task stage has $10$ demonstrations.
Following the convention of ER~\cite{chaudhry2019tiny} algorithm design in \libero{},~\lotus{} stores $5$ demonstrations for each previous task when learning in subsequent lifelong learning steps.

\paragraph{Real Robot Tasks.} We evaluate~\lotus{} on a real robot manipulation task suite, \mutex{}~\cite{shah2023mutex} (50 tasks). They include a variety of tasks, such as ``open the air fryer and put the bowl with hot dogs in it.'' 
We choose $\LLTotalStep=6$ steps, choose $\tasknum_{\text{init}}=25$ for the base task stage and $5$ tasks per step in the lifelong task stage. 
In real-world experiments, each base task includes $30$ demonstrations, while tasks introduced during lifelong learning have $10$.~\lotus{} retains $5$ demonstrations for each previous task when learning in subsequent lifelong learning steps.

\subsection{Results}
\label{sec:lotus:results}
For all simulation experiments, we compare~\lotus{} against multiple
baselines in each task suite. To evaluate the performance of each model, we run 20 trials per task, repeated for three random seeds. We evaluate models with  FWT, NBT, and AUC, defined in Section~\ref{sec:bg:lifelong-imitation-learning}. 
We compare our method with the following baselines.

\begin{itemize}
    \item \mtft{}: A naive baseline with ResNet-Transformer architecture, fine-tuned on the new tasks in sequence~\cite{liu2023libero}.
    \item \mter{}~\cite{chaudhry2019tiny}: An Experience Replay baseline using the ResNet-Transformer without the inductive bias of skills.
    \item \buds{}~\cite{zhu2022buds}: The method developed in Chapter~\ref{chapter:buds}. We adopt \buds{} to lifelong learning by re-training it at every lifelong learning step but using the same policy architecture as \lotus{}.
    \item \lotus-ft{}: A \lotus{} variant that only fine-tunes neural networks using the new task demonstrations without memory buffers. %\weikang{do we need to mention SLIDE here?} 
\end{itemize}

\begin{table}[t]
  \centering
  \scriptsize 
  \setlength{\tabcolsep}{0.5em}
  \begin{minipage}{1.0\linewidth}
    \centering
    \resizebox{1\textwidth}{!}{%
    \begin{tabular}{ll|ccccc}
    \toprule
    Tasks & Evaluation Setting & \mtft{}~\cite{liu2023libero}& \mter{}~\cite{chaudhry2019tiny} & \buds{}~\cite{zhu2022buds} & \lotus-ft{} & \lotus{}\\
    \midrule
    \liberoobject{} & FWT $(\uparrow)$ & 62.0 $\pm$ 0.0 & 56.0 $\pm$ 1.0 & 52.0 $\pm$ 2.0 & 68.0 $\pm$ 4.0 & \highlight{74.0} $\pm$ 3.0  \\
    {} & NBT $(\downarrow)$ & 63.0 $\pm$ 2.0 & 24.0 $\pm$ 0.0 & 21.0 $\pm$ 1.0 & 60.0 $\pm$ 1.0 & \highlight{11.0} $\pm$ 1.0  \\
    {} & AUC $(\uparrow)$ & 30.0 $\pm$ 0.0 & 49.0 $\pm$ 1.0 & 47.0 $\pm$ 1.0 & 34.0 $\pm$ 2.0 & \highlight{65.0} $\pm$ 3.0  \\
    \rowcolor[gray]{0.9}\liberogoal{} & FWT $(\uparrow)$ & 55.0 $\pm$ 0.0 & 53.0 $\pm$ 1.0 & 50.0 $\pm$ 1.0 & 56.0 $\pm$ 0.0 & \highlight{61.0} $\pm$ 3.0  \\
    \rowcolor[gray]{0.9}{} & NBT $(\downarrow)$ & 70.0 $\pm$ 1.0 & 36.0 $\pm$ 1.0 & 39.0 $\pm$ 1.0 & 73.0 $\pm$ 1.0 & \highlight{30.0} $\pm$ 1.0  \\
    \rowcolor[gray]{0.9}{} & AUC $(\uparrow)$ & 23.0 $\pm$ 0.0 & 47.0 $\pm$ 2.0 & 42.0 $\pm$ 1.0 & 26.0 $\pm$ 1.0 & \highlight{56.0} $\pm$ 1.0  \\
    \liberofifty{} & FWT $(\uparrow)$ & 32.0 $\pm$ 1.0 & 35.0 $\pm$ 3.0 & 29.0 $\pm$ 3.0 & 32.0 $\pm$ 2.0 & \highlight{39.0} $\pm$ 2.0  \\
    {} & NBT $(\downarrow)$ & 90.0 $\pm$ 2.0 & 49.0 $\pm$ 1.0 & 50.0 $\pm$ 4.0 & 87.0 $\pm$ 2.0 & \highlight{43.0} $\pm$ 1.0  \\
    {} & AUC $(\uparrow)$ & 14.0 $\pm$ 2.0 & 36.0 $\pm$ 3.0 & 33.0 $\pm$ 3.0 & 16.0 $\pm$ 1.0 & \highlight{45.0} $\pm$ 2.0  \\
    \bottomrule
    \end{tabular}
    }
    \caption[Evaluation of \lotus{} and baseline comparison in \libero{} task suites.]{Comparison between \lotus{} and baselines over three simulation task suites. Results are averaged over three seeds and we report the mean and standard error. All metrics are computed based on success rates (\%).}
    \label{tab:lotus:main_results}
  \end{minipage}%
  \end{table}

Table~\ref{tab:lotus:main_results} provides a comprehensive evaluation of \lotus{} and the baselines in simulation. It answers question (1) by showing that \lotus{} consistently outperforms the best baseline, \mter{}, across all three metrics. While \mter{}'s forward transfer (FWT) performance is lower than its fine-tuning variant \mtft{} (consistent with findings in Liu et al.~\cite{liu2023libero}), \lotus{} achieves better FWT than its fine-tuning counterpart \lotus{}-ft. This result shows that the inductive bias of skills in policy architectures is important for a memory-based lifelong learning algorithm to effectively transfer the previous knowledge to new tasks.\loosepar{}

\begin{table}[ht!]
      \centering
      \begin{tabular}{@{}l|ccc}
        \toprule
        Metrics & CLIP & R3M & DINOv2 \\
        \midrule
        %\hline
        FWT $(\uparrow)$ & 12.0 $\pm$1.0  &  57.0 $\pm$4.0  &  61.0 $\pm$3.0 \\
        NBT $(\downarrow)$ &  43.0 $\pm$2.0  &  33.0 $\pm$2.0  &  30.0 $\pm$1.0 \\
        AUC $(\uparrow)$ &  29.0 $\pm$2.0  &  52.0 $\pm$3.0  &  56.0 $\pm$1.0 \\
        \bottomrule        
      \end{tabular}
      \caption{Ablation study on using different large vision models for continual skill discovery.\loosepar{}}
      \label{tab:lotus:ablation_model}
\end{table}

\paragraph{Ablation Studies.} We use \liberogoal{} for conducting all ablations. We answer question (2) by comparing \lotus{} against its variant that does not add new skills. This variant has a decrease in FWT, NBT, and AUC by $13.0$, $-3.0$, and $7.0$, respectively. The performance declines in all metrics when no new skills are introduced, highlighting the importance of continually growing the skill library to achieve better knowledge transfer. To address question (3), we increase the demonstrations per task for training~\mter{}. The AUC are $0.47$ (10 demos), $0.53$ (15 demos), $0.53$ (20 demos), and $0.57$ (25 demos), respectively. \mter{} only surpasses \lotus{} when using $25$ demonstrations, implying significantly better data efficiency of~\lotus{} compared to the baseline without skill factorization.

To answer question (4), we evaluate the effectiveness of our DINOv2-based method. We compare it against two other large vision models pretrained on Internet-scale and human activity datasets, namely CLIP~\cite{radford2021learning} and R3M~\cite{nair2022r3m}. The result in Table~\ref{tab:lotus:ablation_model} shows that our choice of DINOv2 is the best for continual skill discovery. At the same time, R3M is also significantly better than CLIP at skill discovery, even though R3M performs worse than DINOv2. Note that our open-world vision model choice is not limited to DINOv2 and can be replaced by superior models in the future.

\begin{figure}
    \centering
    \begin{minipage}[t]{1.0\linewidth}
    \includegraphics[width=\linewidth]{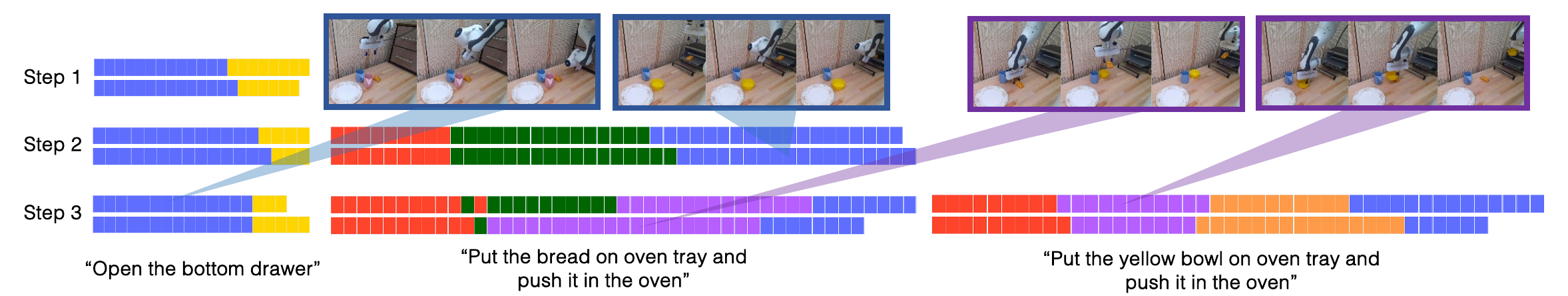}
    \end{minipage}
    \caption{Visualization of skill discovery results in \lotus{}.}
    \label{fig:lotus-segmentation-result}
\end{figure}

\paragraph{Real Robot Results.} We compare \lotus{} with the best baseline~\mter{} on~\mutex{} tasks. Our evaluation shows that \lotus{} achieves 50\% in FWT (+11\%), 21\% in NBT (+ 2\%), and 56\% in AUC (+9\%) in comparison to~\mter{}. The performance over the three metrics shows the efficacy of \lotus{} policies on real robot hardware, answering the question (5). This result also shows that our choice of DINOv2 features is general across both simulated and real-world images. Additionally, we visualize the skill compositions during real robot evaluation in Figure~\ref{fig:lotus-segmentation-result}, showing that \lotus{} not only transfers the previous skills to new tasks but also achieves promising results of transferring new skills to the previous tasks.

\section{Summary}
\label{sec:lotus:discussion}

In this chapter, we introduced~\lotus{}, a lifelong imitation learning method for building vision-based manipulation policies with skills. \lotus{} implements continual skill discovery through two key components. First, it uses an open-world vision model to identify recurring patterns in unsegmented demonstrations. Second, it employs an incremental clustering method to split demonstrations into an increasing number of data partitions. \lotus{} continually updates the skill library, avoiding catastrophic forgetting of the previous tasks and adding new skills to exhibit novel behaviors. \lotus{} uses hierarchical imitation learning with experience replay to train both the skill library and the meta-controller that adaptively composes various skills. Through the development of \lotus{}, we demonstrate how to exploit behavioral regularity from demonstration data and continually discover a library of skills for completing a wide range of tasks.

\paragraph{Limitations and Future Work.}~\lotus{} requires expert human demonstrations through teleoperation, which can often be costly. For future work, we intend to look into discovering skills from human videos building on top of \lotus{} and our methods from Chapters~\ref{chapter:orion} and \ref{chapter:okami}.

Moreover,~\lotus{} still requires storing demonstrations of the previously learned tasks in the experience replay to ensure effective forward and backward transfers. It would incur large memory burdens if the number of tasks were to increase to thousands.
For future work, we plan to investigate compressing data from prior tasks to improve the memory efficiency of memory-based algorithms.

In the following chapter, we introduce our lifelong robot learning benchmark on which \lotus{} simulation experiments are conducted. We describe our novel designs and insights into creating a simulation benchmark of lifelong robot learning benchmark that differs from common robot learning benchmarks.

\chapter{Lifelong Robot Learning Benchmark}
\label{chapter:libero}

In the previous chapter, we presented a continual imitation learning method through skill discovery and conducted simulation experiments for comparing \lotus{} against baselines. The simulation experiments are based on our lifelong robot learning benchmark, which we introduce in this chapter. Our simulation benchmark not only supports our study in Chapter~\ref{chapter:lotus}, but also supports general research in lifelong robot learning, for instance, prototyping lifelong robot learning algorithms. In the following, we first introduce the motivation to design a new benchmark for studying lifelong robot learning and then describe how the tasks are created in our benchmark. 

\section{Overview}
\label{sec:libero:overview}

The main body of the lifelong learning literature has focused on how agents transfer \emph{declarative} knowledge in visual or language tasks, which pertains to \emph{declarative knowledge} about entities and concepts~\citep{biesialska2020continual, mai2022online}. Yet it is understudied how agents transfer knowledge in decision-making tasks, which involves a mixture of both \emph{declarative} and \emph{procedural} knowledge (knowledge about how to do something). We refer to the problem of decision making in lifelong learning settings as  \textit{Lifelong Learning in Decision Making} (\lldm{}). Consider a scenario where a robot, initially trained to retrieve juice from a fridge, fails after learning new tasks. This failure might result from forgetting the \emph{location} of the juice or the fridge (declarative knowledge) or how to \emph{open} the fridge or \emph{grasp} the juice (procedural knowledge). Because the study of this complex knowledge transfer requires a testbed that supports reproducible evaluation of lifelong learning methods, we have yet to see methods that systematically and quantitatively analyze the knowledge transfer involved in \lldm{}. 

To bridge this research gap, this chapter introduces a new simulation benchmark, \libero{} (\emphasize{LI}felong learning \emphasize{BE}chmark on \emphasize{RO}bot manipulation tasks), to facilitate the systematic study of \lldm{}. An ideal \lldm{} testbed should enable continual learning across an expanding set of diverse tasks that share concepts and actions. \libero{} serves as such a testbed through a procedural generation pipeline, which supports endless task creation based on robot manipulation tasks that involve shared visual concepts (declarative knowledge) and interactions (procedural knowledge).

\begin{figure*}[ht!]
    \centering
    \includegraphics[width=\textwidth]{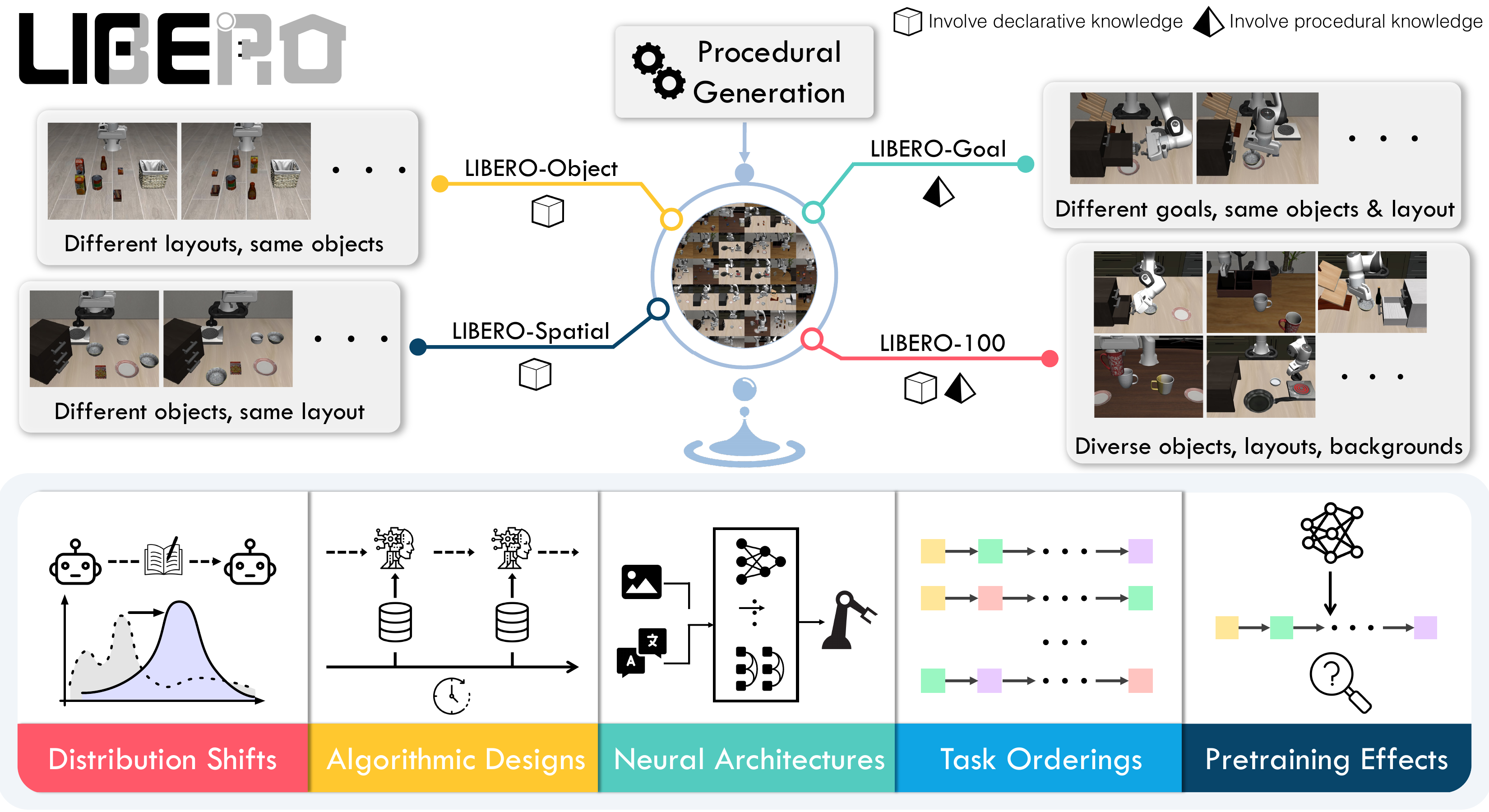}
    \caption[\libero{} overview.]{
    \textbf{Top}: \libero{} has four procedurally-generated task suites: \liberospatial{}, \liberoobject{}, and \liberogoal{} have 10 tasks each and require transferring knowledge about spatial relationships, objects, and task goals; \liberohundred{} has 100 tasks and requires the transfer of entangled knowledge.
    \textbf{Bottom}: The five key research topics \libero{} supports in studying \lldm{}. A comprehensive study on the five topics can be found in our published work~\cite{liu2023libero}. 
    }
    \label{fig:libero:libero-overview}
\end{figure*}

For benchmarking purposes, \libero{} includes 130 language-conditioned robot manipulation tasks inspired by human activities~\cite{grauman2022ego4d} and groups them into four suites. The four task suites are designed to examine distribution shifts in the object types, the spatial arrangement of objects, the task goals, or the mixture of the previous three (top row of Figure~~\ref{fig:libero:libero-overview}). \libero{} is scalable, extendable, and designed explicitly for studying lifelong learning in robot manipulation. To support efficient learning, we provide high-quality, human-teleoperated demonstration data for all 130 standardized tasks. At the bottom of Figure~\ref{fig:libero:libero-overview}, we also show the list of research topics that \libero{} can support. Note that the contribution of \libero{} in this dissertation only involves the task designs and task suites. For more details about the algorithmic and model designs, please see our published work at the Conference and Workshop on Neural Information Processing Systems Datasets and Benchmarks Track, 2023~\cite{liu2023libero}.

\section{LIBERO}
\label{sec:libero:libero}
This section introduces the task creation and standard task suites in \libero{}. We first describe the procedural generation pipeline that can support the never-ending creation of tasks (Section~\ref{sec:libero:procedural}). Then, we introduce the four standard task suites we generate for benchmarking purposes(Section~\ref{sec:libero:libero-suite}).

\subsection{Procedural Generation of Tasks}
\label{sec:libero:procedural}
\begin{figure*}[ht!]
    \centering
    \includegraphics[width=\textwidth]{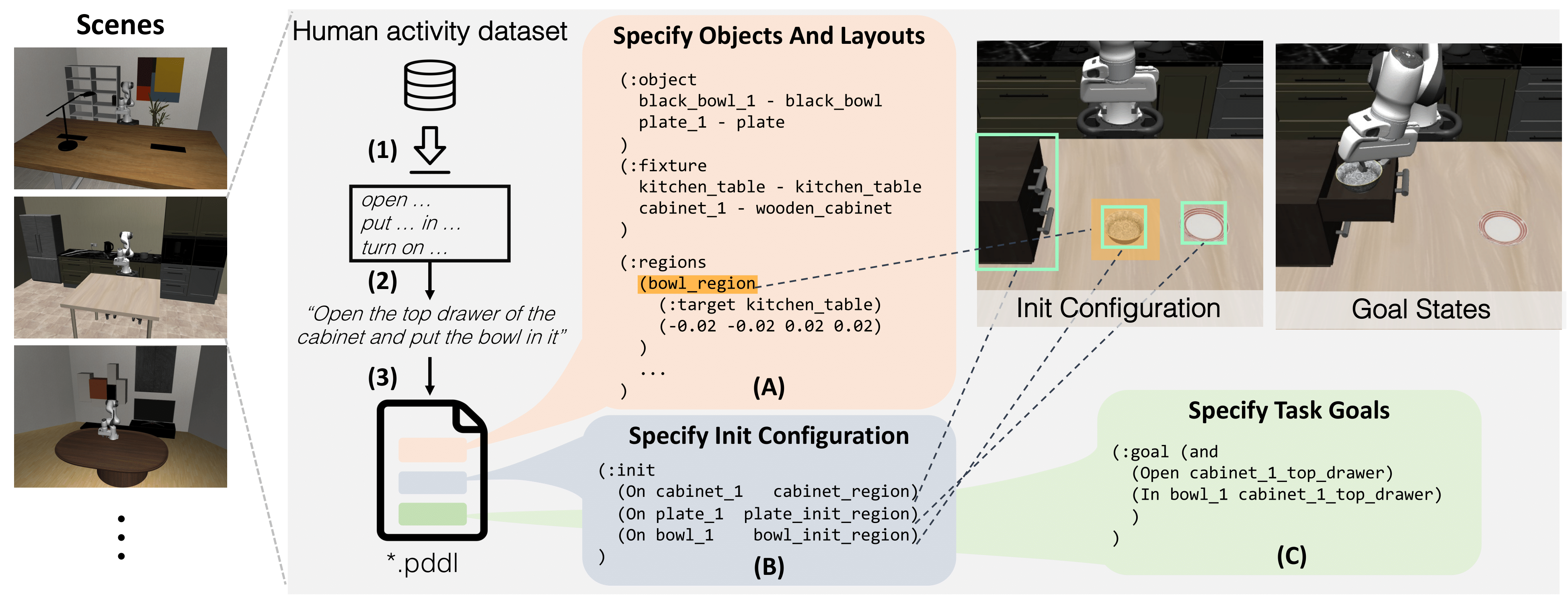}
    \caption[Procedural generation pipeline in \libero{}.]{\libero{}'s procedural generation pipeline:  Extracting behavioral templates from a large-scale human activity dataset \textbf{(1)}, Ego4D, for generating task instructions \textbf{(2)}; Based on the task description, selecting the scene and generating the PDDL description file \textbf{(3)} that specifies the objects and layouts \textbf{(A)}, the initial object configurations \textbf{(B)}, and the task goal \textbf{(C)}.
    }
    \label{fig:libero:libero-procedural-generation}
\end{figure*}

Research in \lldm{} requires a systematic way to create new tasks while maintaining task diversity and relevance to existing tasks. \libero{} procedurally generates new tasks in three steps, \textbf{1)} extract behavioral templates from language annotations of human activities and generate sampled tasks described in natural language based on such templates; \textbf{2)} specify an initial object distribution given a task description, and \textbf{3)} specify task goals using a propositional formula that aligns with the language instructions. Note that a task is uniquely defined by a tuple $(\initstate^{\context}, \mR^{\context})$ as defined in Section~\ref{sec:bg:open-world-formulation}.
Our generation pipeline is built on top of \robosuite{}~\citep{zhu2020robosuite}, a modular manipulation simulator that offers seamless integration. Figure~\ref{fig:libero:libero-procedural-generation} illustrates an example of task creation using this pipeline, and each component is explained in detail below.

\paragraph{Behavioral Templates and Instruction Generation.} Human activities serve as a fertile source of tasks that can inspire and generate a vast number of manipulation tasks. 
We choose a large-scale activity dataset, Ego4D~\citep{grauman2022ego4d}, which includes a large variety of everyday activities with language annotations. We pre-process the dataset by extracting the language descriptions and then summarize them into a large set of commonly used language templates. After this pre-processing step, we use the templates and select objects available in the simulator to generate a set of task descriptions in the form of language instructions. For example, we can generate an instruction ``Open the drawer of the cabinet'' from the template ``Open ... .'' A language instruction is essentially the context of a task, denoted as $\languagegoal$ following the convention established in Section~\ref{sec:bg:open-world-formulation}. The subscript $\text{lang}$ indicates that the context is a language description.

\paragraph{Initial State Distribution ($\initstate^{\context}$).} 
To specify $\initstate^{\context}$, we first sample a scene layout that matches the objects/behaviors in a provided instruction. For instance, a kitchen scene is selected for an instruction \textit{Open the top drawer of the cabinet and put the bowl in it}. Then, the details about $\initstate^{\context}$ are generated in the PDDL language~\citep{mcdermott1998pddl,srivastava2022behavior}.
Concretely, $\initstate^{\context}$ contains information about object categories and their placement (Figure~\ref{fig:libero:libero-procedural-generation}-\textbf{(A)}), and their initial status (Figure~\ref{fig:libero:libero-procedural-generation}-\textbf{(B)}).

\paragraph{Goal Specifications.} Based on $\initstate^{\context}$ and the language instruction $\languagegoal$, we specify the task goal using a conjunction of predicates. Predicates include \emph{unary predicates} that describe the properties of an object (such as \texttt{Open}(X) or \texttt{TurnOff}(X)) and \emph{binary predicates} that describe spatial relations between objects (such as \texttt{On}(A, B) or \texttt{In}(A, B)). An example of the goal specification using PDDL language can be found in Figure~\ref{fig:libero:libero-procedural-generation}-\textbf{(C)}. The simulation terminates when all predicates are verified true, and the corresponding reward function $\mR^{\context}$ returns $1$.

\paragraph{PDDL-based Scene Description File.} We provide the whole content of an example scene description file based on PDDL. This file corresponds to the task shown in Figure~\ref{fig:libero:libero-procedural-generation}.

\begin{center}
    \textbf{Example task:}\qquad \textcolor{purple}{\textit{Open the top drawer of the cabinet and put the bowl in it}}.
\end{center}

\begin{formal}
\begin{lstlisting}[caption={},basicstyle=\small, label=lst:pddl-example]
(define (problem LIBERO_Kitchen_Tabletop_Manipulation)
  (:domain robosuite)
  (:language open the top drawer of the cabinet and 
    put the bowl in it)
    (:regions
      (wooden_cabinet_init_region
          (:target kitchen_table)
          (:ranges (
              (-0.01 -0.31 0.01 -0.29)
            )
          )
          (:yaw_rotation (
              (3.141592653589793 3.141592653589793)
            )
          )
      )
      (akita_black_bowl_init_region
          (:target kitchen_table)
          (:ranges (
              (-0.025 -0.025 0.025 0.025)
            )
          )
          (:yaw_rotation (
              (0.0 0.0)
            )
          )
      )
      (plate_init_region
          (:target kitchen_table)
          (:ranges (
              (-0.025 0.225 0.025 0.275)
            )
          )
          (:yaw_rotation (
              (0.0 0.0)
            )
          )
      )
      (top_side
          (:target wooden_cabinet_1)
      )
      (top_region
          (:target wooden_cabinet_1)
      )
      (middle_region
          (:target wooden_cabinet_1)
      )
      (bottom_region
          (:target wooden_cabinet_1)
      )
    )

  (:fixtures
    kitchen_table - kitchen_table
    wooden_cabinet_1 - wooden_cabinet
  )

  (:objects
    akita_black_bowl_1 - akita_black_bowl
    plate_1 - plate
  )

  (:obj_of_interest
    wooden_cabinet_1
    akita_black_bowl_1
  )

  (:init
    (On akita_black_bowl_1 kitchen_table_akita_black_bowl
    _init_region)
    (On plate_1 kitchen_table_plate_init_region)
    (On wooden_cabinet_1 kitchen_table_wooden_cabinet_init
    _region)
  )

  (:goal
    (And (Open wooden_cabinet_1_top_region) 
         (In akita_black_bowl_1 wooden_cabinet_1_top_region)
    )
  )

)
\end{lstlisting}
\end{formal}

\subsection{Standard Task Suites}
\label{sec:libero:libero-suite}
While the pipeline in Section~\ref{sec:libero:procedural} supports the generation of an unlimited number of tasks, we offer fixed sets of tasks for benchmarking purposes.
\libero{} provides four standard task suites: \liberospatial, \liberoobject, \liberogoal, and \liberohundred. The first three task suites are curated to disentangle the transfer of \emph{declarative} and \emph{procedural} knowledge (as mentioned in~\ref{sec:libero:overview}), while \liberohundred{} is a suite of 100 tasks with entangled knowledge transfer. 

\liberospatial{}, \liberoobject{}, and \liberogoal{} each has 10 tasks and is designed to investigate the controlled transfer of knowledge about spatial information (declarative), objects (declarative), and task goals (procedural), respectively. Note that a suite of 10 tasks is enough to observe catastrophic forgetting while maintaining computation efficiency.
Specifically, all tasks in \liberospatial{} require the robot to place a bowl on the same plate. But there are two identical bowls that differ only in their location or spatial relationship to other objects. Hence, to successfully complete \liberospatial{}, the robot needs to continually learn and memorize new spatial relationships.
All tasks in \liberoobject{} require the robot to pick-place a unique object. 
Hence, to accomplish \liberoobject{}, the robot needs to continually learn and memorize new object types.
All tasks in \liberogoal{} share the same objects with fixed spatial relationships but differ only in the task goal. Hence, to accomplish \liberogoal{}, the robot needs to continually learn new knowledge about motions and behaviors.  \liberohundred{} contains 100 tasks that entail diverse object interactions and versatile motor skills. 

\begin{figure}[h!]
    \centering
    \includegraphics[width=0.8\linewidth]{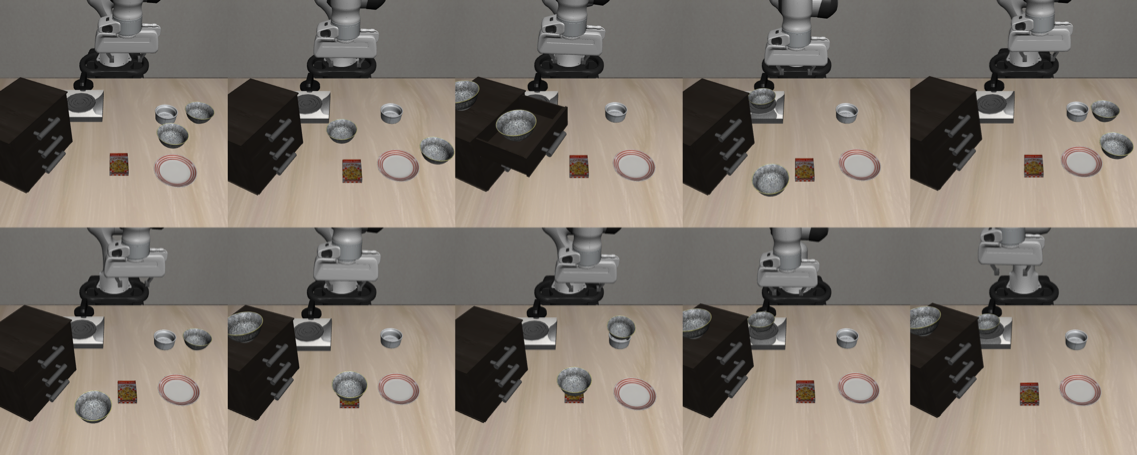}
    \caption[\liberospatial{} task suite visualization.]{\liberospatial{}.}
    \label{fig:libero:libero-spatial}
\end{figure}

\begin{figure}[h!]
    \centering
    \includegraphics[width=0.8\linewidth]{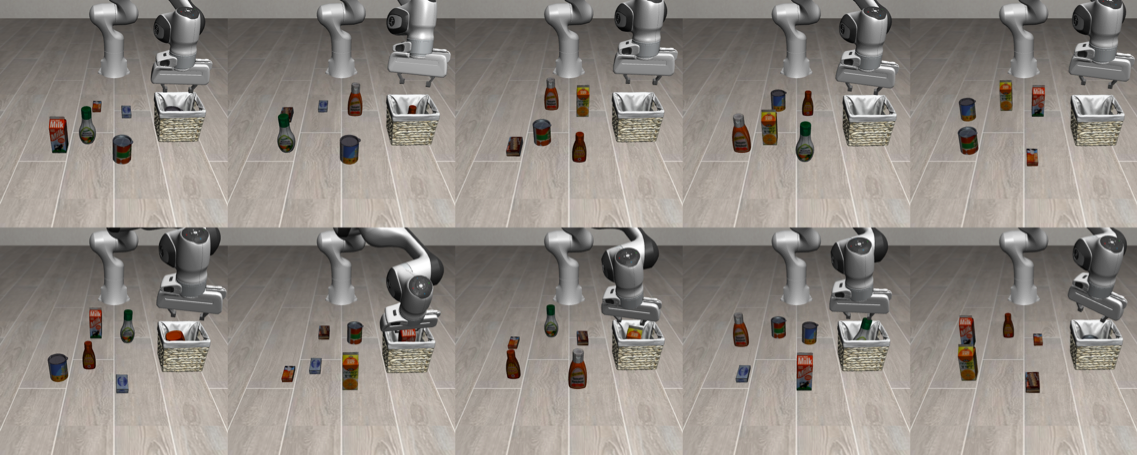}
    \caption[\liberoobject{} task suite visualization.]{\liberoobject{}.}
    \label{fig:libero:libero-object}
\end{figure}

\begin{figure}[h!]
    \centering
    \includegraphics[width=0.8\linewidth]{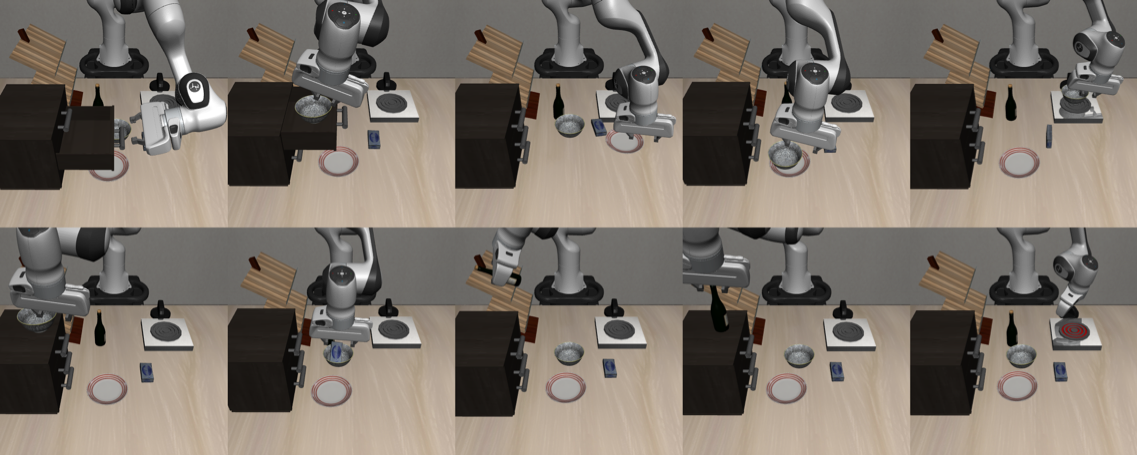}
    \caption[\liberogoal{} task suite visualization.]{\liberogoal{}.}
    \label{fig:libero:libero-goal}
\end{figure}

\begin{figure}[h!]
    \centering
    \includegraphics[width=0.9\linewidth]{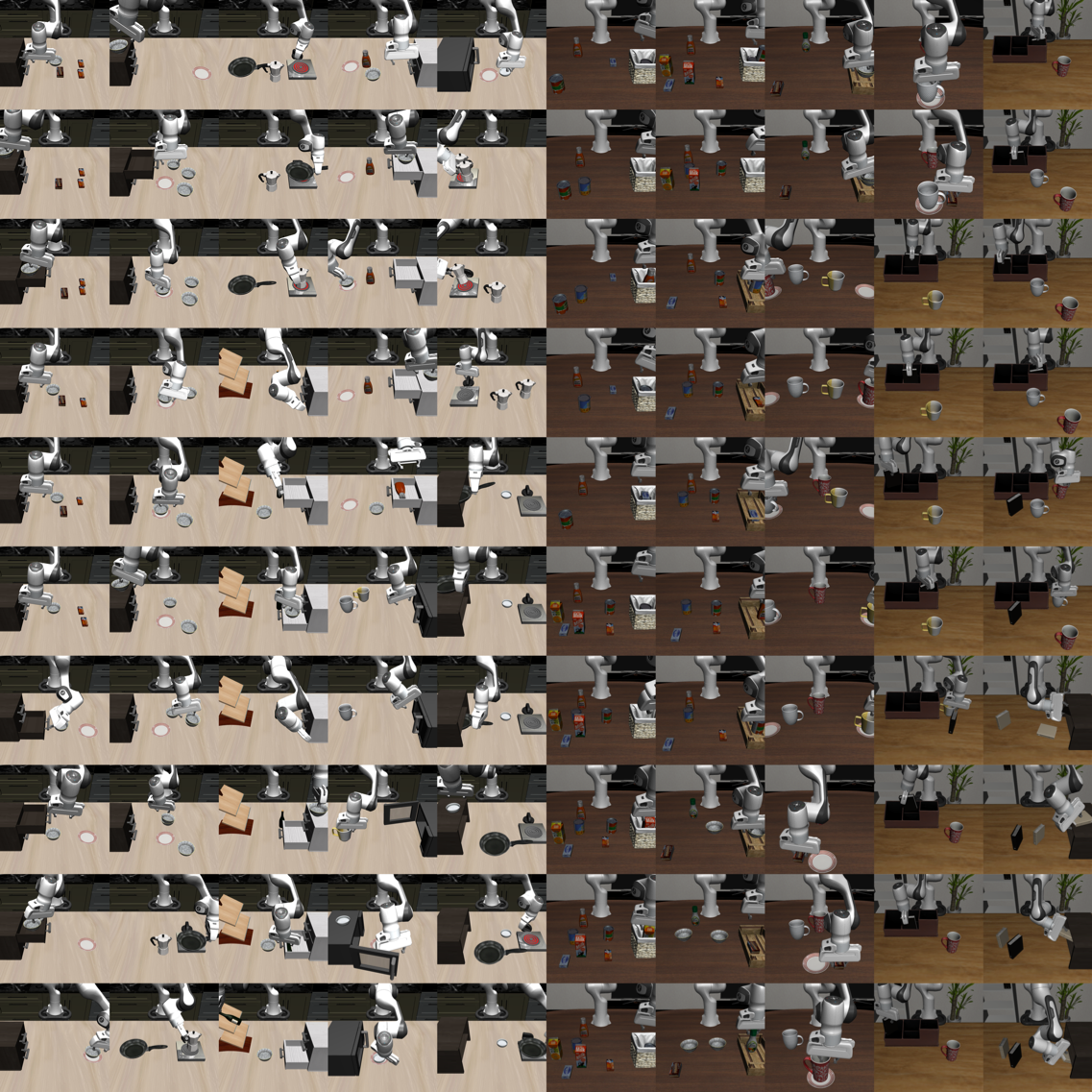}
    \caption[\liberohundred{} task suite visualization.]{\liberohundred{}.}
    \label{fig:libero:libero-100}
\end{figure}

\paragraph{Visualization of Task Suites.} We visualize all the tasks from the four task suites in Figure~\ref{fig:libero:libero-spatial}---~\ref{fig:libero:libero-100}. All the figures visualize the goal states of tasks except for Figure~\ref{fig:libero:libero-spatial}, which visualizes the initial states since the task goals are always the same.

\section{Summary}
\label{sec:libero:discussion}

In this chapter, we introduced~\libero{}, a new benchmark in the robot manipulation domain for supporting research in \lldm{}.~\libero{} includes a procedural generation pipeline that can support the creation of an unbounded number of manipulation tasks in simulation. We use this pipeline to create 130 standardized tasks and conduct a comprehensive set of experiments on policy and algorithm designs. This benchmark serves as a testbed for designing algorithms that exploit behavioral regularity, especially in the context of lifelong robot learning.

Beyond its contribution to this dissertation, our \libero{} benchmark is designed to support multiple general research directions: 1) designing better neural architectures to process spatial and temporal information; 2) developing improved algorithms to improve forward transfer ability; and 3) using pretraining to improve lifelong learning performance. For more details on our benchmarking results in~\libero{}, we refer readers to the full paper published at the 2023 Conference on Neural Information Processing Systems, Datasets and Benchmarks Track~\cite{liu2023libero}.

This chapter and the previous two chapters constitute Part~\ref{part:III} of this dissertation. In the following chapter, we will review the literature relevant to this dissertation.

% \part{Robot Learning Systems}
% \label{part:IV}
% \input{chapter_notes/xi_robot_learning_system}

\part{Related Work and Conclusion}

\chapter{Related Work}
\label{chapter:related-works}

This chapter provides an extensive literature review and explains the relation between prior work and the contributions of this dissertation. Section~\ref{related_work:il} reviews the literature on imitation learning for robot manipulation, which is relevant to our work in Chapters~\ref{chapter:viola}---\ref{chapter:lotus}. Section~\ref{related_work:representation} reviews the literature on representations in vision-based robot manipulation, which is relevant to our work in Chapters~\ref{chapter:viola}--\ref{chapter:orion}. Section~\ref{related_work:human_humanoid} reviews the literature on learning from human videos, which is relevant to our work in Chapters~\ref{chapter:orion} and \ref{chapter:okami}.
Section~\ref{related_work:skill_discovery} reviews the literature on skill discovery, which is relevant to our work in Chapters~\ref{chapter:buds} and \ref{chapter:lotus}. Section~\ref{related_work:lifelong-robot-learning} reviews the literature on methods and benchmarks for lifelong robot learning, which is relevant to our work in Chapter~\ref{chapter:libero}. Our literature review in this chapter was partially covered in some of the previous chapters (See the introductions in Chapters~\ref{chapter:viola},~\ref{chapter:orion},~\ref{chapter:okami}, and \ref{chapter:buds}). 

% \change{Section~\ref{related_work:il} reviews literature that is related to the topic of deep imitation learning and object-centric learning. Section~\ref{related_work:open-world-imitation} reviews literature that is related to the topic of imitation learning from human videos. Section~\ref{related_work:humanoid} reviews work on humanoids that are vital to \okami{} in Chapter~\ref{chapter:okami}. Section~\ref{related_work:skill-discovery} reviews literature that is related to the topic of skill discovery methods in robot manipulation. Section~\ref{related_work:lifelong-robot-learning} reviews literature that is related to the topic of lifelong robot learning and reviews the existing robot learning benchmarks. }  

\section{Imitation Learning for Robot Manipulation}
\label{related_work:il}

Throughout this dissertation, we focused on efficient sensorimotor learning for tackling Open-world Robot Manipulation. Specifically, we assumed demonstrations are given for tasks that we want robots to complete, and policies are learned from these demonstrations using Imitation Learning (IL) methods. This section discusses the related work on IL for robot manipulation. Even though we used an IL methodology, note that IL is not equivalent to efficient sensorimotor learning. We chose IL methods because they are sample-efficient and applicable to real robot hardware while being cost-effective. However, other algorithms can be used to achieve efficient sensorimotor learning if they are as effective as IL methods.

\subsection{Deep Imitation Learning for Visuomotor Policies}
\label{related_work:deep_il}

IL has been an established paradigm for acquiring manipulation policies for decades. Before the era of deep learning, non-parametric approaches such as DMP and PrMP were dominant for IL~\cite{schaal2006dynamic, kober2009learning, paraschos2013probabilistic, paraschos2018using}. These methods could effectively acquire manipulation behaviors through a small number of demonstrations. However, they fall short in handling high-dimensional observations, requiring ground-truth access to physical states that are not feasible in open-world scenarios.  

Deep imitation learning has emerged as a sample-efficient method for learning end-to-end manipulation policies that can handle high-dimensional observations in the real world~\cite{mandlekar2021matters, cui2022play,florence2019self, nasiriany2022learning, zhang2018deep, shridhar2022cliport,shridhar2023perceiver, mandlekar2020learning, wang2023mimicplay, wang2024dexcap, lin2024learning, ze20243d, chang2023one, yu2019one, valassakis2022demonstrate, johns2021coarse, di2022learning}. These methods can learn visuomotor policies to complete various tasks with just dozens of demonstrations, ranging from long-horizon manipulation~\cite{mandlekar2020learning, wang2023mimicplay, zhu2022viola} to dexterous manipulation~\cite{wang2024dexcap, lin2024learning, ze20243d}. Nevertheless, these approaches are susceptible to distributional shifts and observation noise~\cite{mandlekar2020iris, mandlekar2020learning, zhang2018deep, mandlekar2021matters}.

Object-centric priors have been previously explored in IL policies to overcome these issues~\cite{florence2019self, sieb2020graph}. However, prior works either focused manipulating single object instances or required costly annotations for pre-training object detections. In Chapter~\ref{chapter:viola}, based on the same conceptual idea as the previous IL methods using object-centric priors, we used a pre-trained Region Proposal Network (RPN) from an object detection foundation model~\cite{zhou2022detecting}. Our method, \viola{}, uses the pre-trained RPN to introduce general object proposals as object-centric priors into the end-to-end IL policies. The general object proposals can localize objects from in-the-wild image observations and allow the policies in \viola{} to exploit \textit{object regularity} in the real world, thus improving their robustness towards visual variations and solving tasks that involve complicated interactions with multiple objects. Later in Section~\ref{related_work:object_centric_representation}, we provide a comprehensive literature review on object-centric representations in manipulation.

Most prior IL approaches assume the test environments closely resemble the training environments in which the demonstrations are collected. This assumption hampers learned policies from being deployed in conditions different from their training. To improve the broader applicability of IL methods, we extended \viola{} to \groot{} in Chapter~\ref{chapter:groot}. \groot{} trains policies using demonstrations collected from a single environment, enabling them to generalize across diverse visual variations, including background changes, different camera viewpoints, and new object instances. Such intra-task systematic generalization of \groot{} policies is made possible because of the object-centric 3D representations that leverage object regularity, tracking the task-relevant objects despite large visual appearance changes.

\subsection{Hierarchical Imitation Learning for Robot Manipulation}
\label{related_work:hierarchical_il}

In Chapters~\ref{chapter:buds} and \ref{chapter:lotus}, we leveraged hierarchical imitation learning~\cite{le2018hierarchical} for training policies of sensorimotor skills~\cite{lynch2020learning, shiarlis2018taco}. Hierarchical imitation learning is a class of approaches that use temporal abstractions to tackle longer-horizon tasks than vanilla imitation models. In particular, we chose hierarchical behavior cloning, which has recently shown great promise in learning-based robot manipulation (For detailed formulation, see Section~\ref{sec:bg:hbc})~\cite{mandlekar2020iris,gupta2020relay,mandlekar2020learning,tung2020learning, wang2023mimicplay}. These methods learn a hierarchical policy, in which a high-level policy predicts subgoals and selects a low-level skill policy that computes actions to achieve the subgoals. The subgoals can be obtained through goal relabeling~\cite{andrychowicz2017hindsight}. 

\buds{} in Chapter~\ref{chapter:buds} differs from prior work in that we extracted a set of skills from multi-task demonstrations for task composition and directly handle raw sensory data. However, \buds{} assumes a fixed set of low-level skills in multitask settings, limiting its applicability to the lifelong learning setting where the number of skills might need to be updated over time. To overcome this limitation, we introduced \lotus{} in Chapter~\ref{chapter:lotus}, which uses hierarchical behavioral cloning with Experience Replay~\cite{chaudhry2019tiny}, enabling the learning of varying numbers of low-level policies.

\subsection{Learning Manipulation From a Single Demonstration}
\label{related_work:single_demo}
Collecting multiple demonstrations often requires domain expertise and incurs high costs. To reduce the costs, it is ideal to provide as few demonstrations as possible for learning a task. In this section, we discuss prior work on learning manipulation from single demonstrations, which constitutes the literature review for \orion{} and \okami{} in Chapters~\ref{chapter:orion} and \ref{chapter:okami}, focusing on the aspect of using a single demonstration for learning a task.

Prior works have explored learning manipulation policies from one demonstration using various approaches. A notable approach is one-shot imitation learning within the meta-learning framework proposed by Duan et al.~\cite{duan2017one}. While prior works on one-shot imitation learning have shown robots performing new tasks from one demonstration, they require extensive in-domain data and a well-curated set of meta-training tasks beforehand while leading to significant data collection costs but limited policy generalization at test time~\cite{chang2023one, yu2019one, valassakis2022demonstrate, johns2021coarse, di2022learning}.\loosepar{}

An alternative direction involves using a single demonstration for initial guidance, refining the policy through real-world self-play~\cite{wen2022you, di2022learning, haldar2023watch, haldar2023teach, johns2021coarse, valassakis2022demonstrate, jonnavittula2024view}. However, these prior methods mainly apply to reset-free tasks and struggle with scaling to multi-stage tasks, in which resetting to initial conditions is nontrivial. 

% Recently, foundation models have been used to directly imitate actions from a single demonstration, but the existing method requires ground-truth access to robot actions through kinesthetic teaching~\cite{di2024dinobot}.

In this dissertation, we introduced ``open-world imitation from observation'' in Section~\ref{sec:orion:open_world}, which refers to a new problem setting of imitating from a single video demonstration. With just one single human video, our method derives a robot policy that successfully completes the task while adapting to a wide range of visual and spatial variations different from the conditions in the video demonstration. Our methods from Chapter~\ref{chapter:orion} and \ref{chapter:okami} are similar to prior work in using a single demonstration for learning manipulation but stand out by not requiring prior data or self-play. Recent or concurrent works have also explored using a single video demonstration~\cite{heppert2024ditto, guo2023learning}, but they either assume known object instances \textit{a priori} or fail to achieve systematic generalization in an open-world setting described in Section~\ref{sec:bg:open-world-formulation}. The key to our methods is our exploitation of spatial regularity, as explained in Chapters~\ref{chapter:orion} and \ref{chapter:okami}, which endows a robot policy with spatial understanding to extract useful information from video observations. Notably, unlike other works that abstract away embodiment motions due to kinematic differences between the robot and the human, \okami{} in Chapter~\ref{chapter:okami} exploits embodiment motion information based on the kinematic similarity between humans and humanoids. Specifically, we introduce \emph{object-aware retargeting} which adapts human motions to humanoid robots.\loosepar{}

\section{Representations for Vision-Based Robot Manipulation}
\label{related_work:representation}
We introduced methods in Chapters~\ref{chapter:viola} and \ref{chapter:groot} for learning generalizable visuomotor policies using behavioral cloning. The key to our methods is to construct object-centric representations from visual observations. In this section, we first review the related work over general visual representations and then discuss the related work using or learning object-centric representations for manipulation.

\subsection{Visual Representations for Visuomotor Policies}
\label{related_work:visual_representation}

Various types of visual representations have been studied in vision-based manipulation. Early work commonly used intermediate visual representations of known objects like bounding boxes~\cite{mulling2013learning, duan2017one, sieb2020graph}, overfitting to specific object instances and cannot localize any new object instances. End-to-end learning methods seek to directly map pixel images to actions with neural networks~\cite{zhang2018deep, finn2016deep}, but they are susceptible to covariate shifts and causal confusion~\cite{ross2011reduction, park2021object}, resulting in poor generalization performance. Much literature has investigated inductive biases and learning techniques to overcome issues of end-to-end learning, including pretraining on large datasets~\cite{nair2022r3m, xiao2022masked}, generative augmentation~\cite{mandi2022cacti, yu2023scaling, chen2023genaug}, spatial attention~\cite{zeng2020transporter, shridhar2022cliport}, and affordances~\cite{shridhar2023perceiver, jiang2021synergies}. However, these representations are purposefully designed for specific motion primitives, such as pick-and-place, or the representations require fine-tuning yet have shown limited performance in downstream tasks~\cite{hansen2022pre}.

Our work in Chapters~\ref{chapter:viola} and \ref{chapter:groot} focused on the design of object-centric representations. In the following subsections, we explain the relationship between our work and other related work that focuses on object-centric representations.

\subsection{Object-Centric Representation for Robot Manipulation}
\label{related_work:object_centric_representation}

The robotics and vision communities have extensively explored using object-centric representations for reasoning about visual scenes in a modular way. Prior work has shown the effectiveness of object-centric representations in downstream manipulation tasks by factorizing visual scenes into disentangled object concepts~\cite{tremblay2018deep, tyree20226, migimatsu2020object, wang2019deep, devin2018deep}. In robotics, researchers commonly use poses~\cite{tremblay2018deep, tyree20226, migimatsu2020object} and bounding boxes~\cite{wang2019deep, devin2018deep} to represent objects present in a scene. However, these representations are confined to known object categories or instances. Prior work leveraged unsupervised learning approaches~\cite{locatello2020object, burgess2019monet} to endow manipulation policies with object awareness~\cite{wang2021generalization, heravi2022visuomotor}. However, these approaches are limited to simulation domains and fall short in generalizing to real-world visual observations.

Recent developments in foundation models allow robots to access open-world object concepts through pre-trained vision models~\cite{caron2021emerging, kirillov2023segment, oquab2023dinov2, cheng2022xmem}. These models enable a wide range of abilities, such as imitation of long-horizon tabletop manipulation~\cite{shiplug}, in-context learning of tabletop manipulation~\cite{di2024keypoint}, and mobile manipulation in the wild~\cite{stone2023open}. These models can perceive objects beyond specific categories, demonstrating strong generalization abilities that are vital to vision-based manipulation tasks.

In Chapters~\ref{chapter:viola} and~\ref{chapter:groot}, we used off-the-shelf vision foundation models to localize objects despite changes in visual backgrounds and camera angles and to identify unseen task-relevant objects that share semantic similarity with the training objects.
% (new instances from the same category).
In Chapter~\ref{chapter:viola}, we used object proposals to construct our object-centric representations for robot manipulation tasks. This approach has been motivated by the effectiveness of region proposal networks (RPNs) on out-of-distribution images~\cite{zhou2022detecting}. Such effectiveness is made possible due to significant progress in generating object proposals for various downstream tasks, such as object detection~\cite{cai2018cascade, zhou2019objects, zhou2022detecting} and visual-language reasoning~\cite{su2019vl, chen2020uniter}.
In Chapter~\ref{chapter:groot}, we showed that \groot{} leverages these pretrained models to construct object-centric 3D representations, which are key to learning more generalizable policies than prior methods do.

Aside from exploiting object regularity in \viola{} and \groot{}, using vision foundation models can also help exploit \textit{spatial regularity} by constructing structured representations on top of object localization results. Our work in Chapter~\ref{chapter:orion} also uses vision foundation models to leverage open-world, object-centric concepts. We introduced a graph-based representation called Open-world Object Graph (OOG) using the object-centric concepts identified by vision foundation models. OOG is critical to constructing a manipulation plan that serves as the spatiotemporal abstraction of actionless human videos. The OOG representation shares similarities with prior methods that factorized scene or task-relevant visual concepts into scene graphs~\cite{kumar2023graph, mo2019structurenet, huang2023planning, qureshi2021nerp, zhu2021hierarchical, sieb2020graph}. But unlike these methods, our OOG representation is tailored to capture both open-world object concepts and enable generalization across different embodiments, specifically between a human and a robot.

\section{Learning from Human Videos}
\label{related_work:human_humanoid}

We introduced methods in Chapters~\ref{chapter:orion} and \ref{chapter:okami} to learn manipulation skills from human videos. In this section, we discuss the literature relevant to these chapters. Notably, \okami{} in Chapter~\ref{chapter:okami} focuses on controlling humanoid robots based on the concept of motion retargeting. We also review the literature on humanoid control and motion retargeting. 

\paragraph{Learning Manipulation From Human Video Demonstrations.} Human videos offer a rich repertoire of behaviors interacting with objects, making them an invaluable data source for manipulation. A large body of work has explored how to leverage human video data for learning robot manipulation~\cite{wang2023mimicplay, xiong2021learning, bahl2022human, kumar2023graph, liu2018imitation, sharma2019third, smith2019avid, xu2023xskill}, either through pre-training latent representations~\cite{nair2022r3m, wang2023mimicplay, xu2023xskill}, learning explicit trajectory priors~\cite{lee2022learning, shaw2024learning}, learning implicit reward functions~\cite{chen2021learning, ma2022vip}, learning 6D representations of actions~\cite{bahety2024screwmimic}, or learning generative models that in-paint human morphologies~\cite{ko2023learning, bharadhwaj2023zero, bahl2022human, bharadhwaj2023towards}. However, they either require additional robot data from the target tasks or paired data between humans and robots. In this dissertation, we took a novel direction by tackling how a robot can imitate or learn from a single human video only: the robot does not rely on pre-existing data, models, or ground-truth annotations \textit{in scenes} where video recording and robot evaluation take place. We refer to such a setting as \textit{open-world imitation from observation}, where the robot is not programmed or trained to interact with the objects in the video \textit{a priori}, and the video data does not include any robot actions. Our setting is closely related to the problem of ``Imitation Learning from Observation''~\cite{torabi2021imitation}, where state-only demonstrations are used to derive policies for physical interaction. However, this line of work assumes that 1) a simulation replica of the tasks exists and 2) physical states of the agents or objects are known~\cite{pavse2020ridm, karnan2022adversarial, torabi2019imitation, torabi2018generative, torabi2018behavioral}. In contrast, our setting does not assume the digital replica of real-world tasks, and all the object information is only perceived through visual observations. In Chapters~\ref{chapter:orion} and \ref{chapter:okami}, we introduced \orion{} and \okami{}, which tackle the aforementioned problem on tabletop manipulators and humanoid robots, respectively. Our methods also demonstrate how we can exploit the \textit{spatial regularity} such that robots can learn from actionless video observations.

\paragraph{Humanoid Robot Control.} In this paragraph, we review the literature on humanoid robot control to provide readers with contexts in Chapter~\ref{chapter:okami}. Methods like motion planning and optimal control have been developed for humanoid locomotion and manipulation~\cite{cheng2024expressive, hu2014online, he2024learning}. These model-based approaches rely on precise physical modeling and expensive computation~\cite{nakaoka2005task, hu2014online, escande2014hierarchical}. To mitigate the strict requirements of precise models, researchers have explored reinforcement learning of policies and sim-to-real transfer~\cite{cheng2024expressive,liao2024berkeley}. However, these methods still require significant labor and expertise in designing simulation tasks and reward functions, limiting their successes to locomotion domains. In parallel to automated methods, a variety of human control mechanisms and devices have been developed for humanoid teleoperation using motion capture suits~\cite{darvish2019whole, hu2014online, penco2019multimode, kim2013whole, di2016multi, arduengo2021human, cisneros2022team}, telexistence cockpits~\cite{tachi2003telexistence, ramos2018humanoid, ishiguro2020bilateral, 8593521, schwarz2021nimbro}, VR devices~\cite{seo2023trill, hirschmanner2019virtual, Lim2022OnlineTF}, and videos that track human bodies~\cite{he2024learning, fu2024humanplus}. While these systems can control the robots to generate diverse behaviors, real-time human input is required, posing significant cognitive and physical burdens on human teleoperators. Unlike prior work, \okami{} only requires a single RGB-D human video to teach the humanoid robot a new skill for a task, significantly reducing the labor cost.\loosepar{}

\paragraph{Motion Retargeting.} In this paragraph, we review the literature on motion retargeting techniques that are relevant to the ``object-aware retargeting'' process in Chapter~\ref{chapter:okami}. Motion retargeting has wide applications in computer graphics and 3D vision~\cite{Gleicher1998RetargettingMT}, where extensive literature studies how to adapt human motions to digital avatars~\cite{luo2023perpetual, peng2021amp, jiang2024motiongpt}. This technique has been adopted in robotics for recreating human-like motions on humanoid or anthropomorphic robots. Prior work has explored various retargeting methods, including optimization-based approaches~\cite{nakaoka2005task, hu2014online, penco2019multimode, kuindersma2016optimization}, geometric-based methods~\cite{liang2021dynamic}, and learning-based techniques~\cite{cheng2024expressive, choi2020nonparametric, he2024learning}. However, in the field of robot manipulation, these retargeting methods have only been used within teleoperation systems, lacking the integration of a vision pipeline for automatic adaptation to object locations. \okami{} integrates open-world vision into the retargeting process, endowing motion retargeting with object awareness so that the robot at test time can mimic human motions from video demonstrations while adapting to object locations.\loosepar{}

\section{Skill Discovery}
\label{related_work:skill_discovery}

A large body of work has explored skill discovery, which studies how a robot identifies recurring segments of sensorimotor experiences, often termed skills. A major line of work focuses on acquiring skills from self-exploration in environments. Many works fall into the options framework~\cite{sutton1999between}, discovering skills through hierarchical reinforcement learning~\cite{vezhnevets2017feudal,fox2017multi, gregor2016variational,konidaris2009skill,bagaria2019option,kumar2018expanding}. Another line of work uses information-theoretic metrics to discover skills from unsupervised interaction~\cite{eysenbach2018diversity,hausman2018learning,sharma2019dynamics}. These methods typically demand high sample complexity and often rely on ground-truth physical states. High sample complexity hinders them from applying to real robot hardware, and the assumption of ground-truth physical states does not hold when we consider an Open-world Robot Manipulation task. 

\paragraph{Skill Discovery from Demonstrations.} An alternative to self-exploration is to segment skills from demonstrations using methods such as Bayesian inference~\cite{niekum2012learning, konidaris2010constructing, niekum2015online} and trajectory reconstruction~\cite{shankar2020learning, tanneberg2021skid}. These approaches produce temporal segmentation on low-dimensional physical states, making them hard to scale to raw sensor data. Weakly supervised learning methods discover skill segments in raw sensory data through temporal alignment on demonstrations but require manual human annotations of task sketch~\cite{shiarlis2018taco, pirk2020modeling}. 

In Chapter~\ref{chapter:buds}, we presented \buds{}, which resonates with prior work on skill discovery from demonstrations. However, \buds{} directly operates on raw sensor data and does not require manual labeling on execution stages in demonstrations. Similar to prior work~\cite{su2018learning, chu2019real}, we took advantage of multi-sensory cues in demonstrations. An important difference is that our method produces closed-loop sensorimotor policies, while the others focus primarily on learning task structures.

Note that \buds{} assumes fixed state-action distributions in multitask settings, preventing it from tackling continually changing situations during the robot's lifespan. In Chapter~\ref{chapter:lotus}, we extended \buds{} to \lotus{}, which discovers skills from demonstrations in lifelong imitation learning (See Section~\ref{sec:bg:generalization-il-policy} for formulation). In lifelong imitation learning settings, demonstrations are collected in a sequence of tasks, where the data distribution is constantly changing. \lotus{} holds the promise of scaling up robot deployment where it can learn new tasks through its deployment lifespan.

% All methods of skill discovery touch upon leveraging behavioral regularity in the physical world. In this dissertation, we formally introduce the perspective of behavioral regularity for viewing skill discovery methods.

\paragraph{Bottom-up Methods in Perception and Control.} The hierarchical clustering used in \buds{} follows a bottom-up principle. In this paragraph, we provide an overview of bottom-up methods in perception and control. Bottom-up processing of sensory information traces back to Gibson's theory of direct perception~\cite{gibson1966senses}, of which the basic idea is that a higher level of information is built up on the retrieval of direct sensory information in a lower level. Bottom-up methods have been successfully employed in various perception tasks. These methods construct hierarchical representations by grouping fine-grained visual elements, such as pixels/superpixels for image segmentation~\cite{farag2016bottom} and spatiotemporal volumes for activity understanding~\cite{lan2015action,sarfraz2019efficient,sarfraz2021temporally}. Recently, bottom-up deep visual recognition models~\cite{newell2016stacked, law2018cornernet, zhou2019bottom} achieve competitive performance compared to the mainstream top-down methods.
The bottom-up design principles have also been studied for robot control. A notable example is the subsumption architecture developed in behavior-based robotics, which decomposes a complex robot behavior into hierarchical layers of sub-behaviors~\cite{brooks1986robust,brooks1991intelligence, nicolescu2002hierarchical}. In Chapter~\ref{chapter:buds}, our proposed method, \buds{}, leverages a similar bottom-up principle to discover hierarchical representations of demonstration trajectories. Furthermore, our experiments in \buds{} demonstrate how imitation learners can exploit such hierarchies to scale to long-horizon manipulation behaviors.

\section{Lifelong Robot Learning}
\label{related_work:lifelong-robot-learning}

Chapter~\ref{chapter:lotus} addresses the problem of lifelong imitation learning for robot manipulation. The problem is fundamentally related to a general problem of Lifelong Learning for Decision Making (\lldm{}), and benchmarks designed for lifelong learning are needed to study the \lldm{} problem. In this section, we review the literature on both topics. 

\subsection{Lifelong Learning for Decision Making (\lldm{})} 
\label{related_work:lldm}
Lifelong learning aims to develop generalist agents that adapt to new tasks in ever-changing environments~\cite{wang2023voyager, tessler2017deep, mendez2023embodied, powers2023evaluating, kirkpatrick2017overcoming}. Prior work proposes three main approaches to address the problem of catastrophic forgetting in deep learning: Dynamic Architecture approaches, Regularization-based approaches, and Rehearsal approaches. 

% Although some recent methods explore the combination of different approaches~\citep{ayub2022few,kang2022class,rios2020lifelong} or new strategies~\citep{zhou2022forward,saha2021space,cheung2019superposition}, our benchmark aims to provide an in-depth analysis of these three basic lifelong learning directions to reveal their pros and cons on robot learning tasks.

The dynamic architecture approach gradually expands the learning model to incorporate new knowledge~\citep{rusu2016progressive,yoon2017lifelong,mallya2018packnet,hung2019compacting,wu2020firefly,ben2022lifelong}. Regularization-based methods, on the other hand, regularize the learner to a previous checkpoint when it learns a new task~\citep{kirkpatrick2017overcoming,chaudhry2018riemannian,schwarz2018progress,liu2022continual}. Rehearsal methods save exemplar data from prior tasks and replay them with new data to consolidate the agent's memory~\citep{chaudhry2019tiny,lopez2017gradient,chaudhry2018efficient,buzzega2020dark}. 

% Our benchmark~\libero{} (Chapter~\ref{chapter:libero}) supports the evaluation of all three kinds of lifelong learning algorithms. For a comprehensive review of lifelong learning methods, we refer readers to surveys \citep{de2021continual, parisi2019continual}.

Prior Regularization-based work has trained monolithic policies~\cite{xie2020deep, xie2022lifelong, chaudhry2019tiny,haldar2023polytask,liu2023continual}, but such a methodology has shown limited performance in knowledge transfer for robot manipulation~\cite{liu2023libero}. Alternatively, another line of research attempts to leverage skills through compositional modeling of lifelong learning tasks~\cite{mendez2022modular,ben2022lifelong,xie2022lifelong,chen2023fast}. They attempt to enable more efficient knowledge transfer than their monolithic policy counterparts. These methods, primarily based on hierarchical reinforcement learning, induce high sample complexity. They fail to scale to complex domains such as vision-based manipulation. 

Unlike prior work, our method in Chapter~\ref{chapter:lotus},~\lotus{}, follows a hierarchical imitation learning framework with Rehearsal-based algorithmic design. Specifically, \lotus{} learns a hierarchical policy through hierarchical behavioral cloning with Experience Replay~\cite{rolnick2019experience}, allowing sample-efficient learning while effectively transferring (backward and forward) knowledge in vision-based manipulation domains.

\subsection{Lifelong Robot Learning Benchmarks}
\label{related_work:benchmark}

\paragraph{Lifelong Learning Benchmarks} Pioneering work has adapted standard vision or language datasets for studying lifelong learning. This line of work includes image classification datasets like MNIST~\citep{deng2012mnist}, CIFAR~\citep{krizhevsky2009learning}, and ImageNet~\citep{deng2009imagenet}; segmentation datasets like Core50~\citep{lomonaco2017core50}; and natural language understanding datasets like GLUE~\citep{wang2018glue} and SuperGLUE~\citep{sarlin2020superglue}. Besides supervised learning datasets, video game benchmarks in Reinforcement Learning (RL) (e.g., Atari~\citep{mnih2013playing}, XLand~\citep{team2021open}, and VisDoom~\citep{kempka2016vizdoom}) have also been used for studying lifelong learning. However, lifelong learning in standard supervised learning settings does not involve procedural knowledge transfer, while RL problems in games do not represent real-world distributions of daily activities. 

We also review a list of benchmarks that have been proposed to tackle \lldm{}, but none of them can support the scale and diversity of lifelong robot learning problems as \libero{} support. ContinualWorld~\citep{Woczyk2021ContinualWA} modifies the 50 simple manipulation tasks in MetaWorld for lifelong learning. CORA~\citep{powers2021cora} builds four lifelong RL benchmarks based on Atari, Procgen~\citep{cobbe2020leveraging}, MiniHack~\citep{samvelyan2021minihack}, and ALFRED~\citep{shridhar2020alfred}. 
F-SIOL-310~\citep{ayub2021f} and OpenLORIS~\citep{she2020openloris} are challenging real-world lifelong object learning datasets that are captured from robotic vision systems but do not support policy evaluation. Prior work has also analyzed different components in a lifelong learning agent~\citep{mirzadeh2022architecture,Woczyk2022DisentanglingTI,ermis2022memory}, but they do not focus on robot manipulation problems.

\paragraph{Robot Learning Benchmarks.} A variety of robot learning benchmarks have been proposed to address challenges in meta-learning (MetaWorld~\cite{yu2020meta}), causality learning (CausalWorld~\cite{ahmed2020causalworld}), multi-task learning~\cite{james2020rlbench, li2023behavior}, policy generalization to unseen objects~\cite{mu2021maniskill, gu2023maniskill2}, and compositional learning~\cite{mendez2022composuite}. Compared to existing benchmarks in lifelong learning and robot learning, the task suites in \libero{} are curated to address the research topics of \lldm{}. \libero{} includes a large number of tasks based on everyday human activities that feature rich interactive behaviors with a diverse range of objects. Additionally, the tasks in \libero{} are procedurally generated, making the benchmark scalable and adaptable. Moreover, the provided high-quality human demonstrations for \libero{} task suites support and encourage learning efficiency, making the benchmark resource-friendly for academic institutions.

\section{Summary}
\label{related_work:summary}
In this chapter, we provided an extensive review of robot learning literature that our methods build upon or compare with. By the completion of this dissertation, this field has been evolving quickly. Therefore, we mainly focused on related work published prior to or concurrent with our methods. In the following chapter, we will conclude this dissertation and introduce future directions that can be explored based on the contributions of this dissertation. 
\chapter{Conclusion}
\label{chapter:conclusion}

In this dissertation, we studied Open-world Robot Manipulation, a problem that is formulated to emulate the real-world complexity that robots might encounter in everyday deployment. While various methodologies can be used to derive solutions to this problem, we focused on efficient sensorimotor learning, which is a cost-effective methodology for robots to acquire new skills. This methodology is particularly amenable for real robot deployment as it aims to minimize the inherent cost of operating real robot hardware.

Throughout this dissertation, we explored how to achieve efficient sensorimotor learning from the perspective of regularity. Regularity, which refers to regular, consistent patterns and structures, is predominant in the physical world and reveals invariant features across various situations. The existence of regularity makes it possible for robots to generalize from a few examples instead of learning from a mass collection of data. Based on our insights, we focused on answering the following research question in this dissertation:

\begin{tcolorbox}[colback=white, colframe=black, boxrule=0.2mm, arc=0.2mm, boxsep=0.5mm]
How can robots exploit regularities in the physical world to efficiently learn generalizable manipulation policies?
\end{tcolorbox}

To answer this research question, we answered three subquestions:

\begin{enumerate}

\item \textit{How can a robot generalize across different visual variations when objects remain the same?}  In Chapter~\ref{chapter:viola}, we focused on developing an object-centric imitation learning framework, \viola{}, that integrates object-centric priors from vision foundation models. In Chapter~\ref{chapter:groot}, we focused on extending \viola{} to \groot{}, an object-centric 3D imitation learning method that allows intra-task generalization of policies. With our methods, we showed how to use vision foundation models to exploit object regularity (introduced in Section~\ref{sec:bg:object_regularity}) and help learn generalizable policies.

\item \textit{How can a robot generalize from a single example through observation?}  In Chapter~\ref{chapter:orion}, we posed a new problem, ``open-world imitation from observation,'' and proposed \orion{} that tackles the problem using tabletop manipulators. In Chapter~\ref{chapter:okami}, we focused on investigating the same problem but considering humanoids as robot hardware. We developed \okami{}, which is an object-aware retargeting method that allows humanoids to imitate dexterous, bimanual manipulation from a single-video demonstration. \orion{} and \okami{} show how we can exploit spatial regularity (introduced in Section~\ref{sec:bg:spatial_regularity}), capturing the spatial relations of task-relevant objects and transferring the manipulation skills from actionless observations to robots.

\item \textit{How can a robot remember the previous tasks while learning new tasks faster?} In Chapter~\ref{chapter:buds}, we laid the groundwork for identifying recurring segments from demonstrations by proposing a bottom-up skill discovery approach, \buds{}. Then, in Chapter~\ref{chapter:lotus}, we proposed \lotus{}, which builds upon \buds{} and maintains a growing library of low-level skills in a lifelong imitation learning setting. The skill library of \lotus{} allows a hierarchical policy to ``remember'' how to accomplish the previous tasks while reusing existing skills to learn new tasks faster. To systematically evaluate \lotus{} policies along with other baselines, we proposed \libero{} in Chapter~\ref{chapter:libero}, a simulation benchmark designed for lifelong robot learning research. In summary, these three chapters together showcase how we can exploit behavioral regularity (introduced in Section~\ref{sec:bg:behavioral_regularity}) and develop robot autonomy that continually learns.

\end{enumerate}

Overall, our answers to the main research question involve developing robot learning methods that exploit object, spatial, and behavioral regularities. Our methods allow robots to learn from a small amount of demonstration data and generalize to new situations or new tasks.

\paragraph{Summary of Contributions.}  In Section~\ref{sec:intro-contributions}, we introduced the contributions of each chapter. Here, we highlight this dissertation's key algorithmic contributions and innovative results.
To summarize, this dissertation presents algorithmic contributions to the robot learning literature, including:

\begin{itemize}
    \item Behavioral cloning methods for learning generalizable policies from a small number of teleoperation demonstrations (Chapters~\ref{chapter:viola} and \ref{chapter:viola}).
    \item Methods for tabletop manipulators and humanoid robots to learn from in-the-wild, single-video observations (Chapters~\ref{chapter:orion} and \ref{chapter:okami}).
    \item Lifelong learning of policies with skill discovery for effective backward and forward transfer (Chapters~\ref{chapter:buds} and \ref{chapter:lotus}).
\end{itemize}

Our contributions to the robot learning community are not limited to algorithmic innovation. Our contributions also include the results that are among the first in the robot learning community to demonstrate:

\begin{itemize}
    \item The first deployment of an end-to-end closed-loop visuomotor policy that makes coffee autonomously (Chapter~\ref{chapter:viola}).
    \item The first deployment of behavioral cloning policies that allow robots to learn from demonstrations collected in a single setting while generalizing to unseen conditions with varying visual backgrounds, camera perspectives, and target object instances (Chapter~\ref{chapter:groot}).
    \item The first deployment of policies that learn from single-video demonstrations collected using daily recording devices such as iPhones/iPads (Chapter~\ref{chapter:orion}).
    \item The first policy enabling a humanoid robot to imitate bimanual, dexterous manipulation from a single-video human demonstration (Chapter~\ref{chapter:okami}).
    \item The first lifelong robot learning benchmark with innate functionality to create new tasks programmatically (Chapter~\ref{chapter:libero}).
\end{itemize}

% \todo{Potential impact of regularity-based approaches on open-world robot manipulation - Paving the way for affordable, widespread, and user-friendly personal robots.}

\section{Future Work}
\label{sec:conclusion:future_works}

In this dissertation, we delivered contributions that aim to answer the main research question of how a robot can leverage regularities, enabling efficient sensorimotor learning for Open-world Robot Manipulation. By answering the research question, we worked towards developing autonomous robots that can be deployed in the open world.

Eventually, when robots operate ubiquitously in the open world, they will need to co-exist with humans. In my opinion, we should focus on building robots not to replace humans, but to assist humans as intelligent assistants or companions. This collaborative relationship requires robots to understand human needs, preferences, and behaviors while complementing human capabilities in ways that enhance overall productivity and well-being.

These requirements hint at a research theme with a scope larger than Open-world Robot Manipulation---\textbf{Human-Robot Coevolution}. While a few existing efforts have used the term human-robot coevolution~\cite{UTokyoAIProject}, its objective has not been clearly defined, and there is no systematic guideline on what the important research directions are to work on.

\begin{figure}
    \centering
    \includegraphics[width=1.0\linewidth]{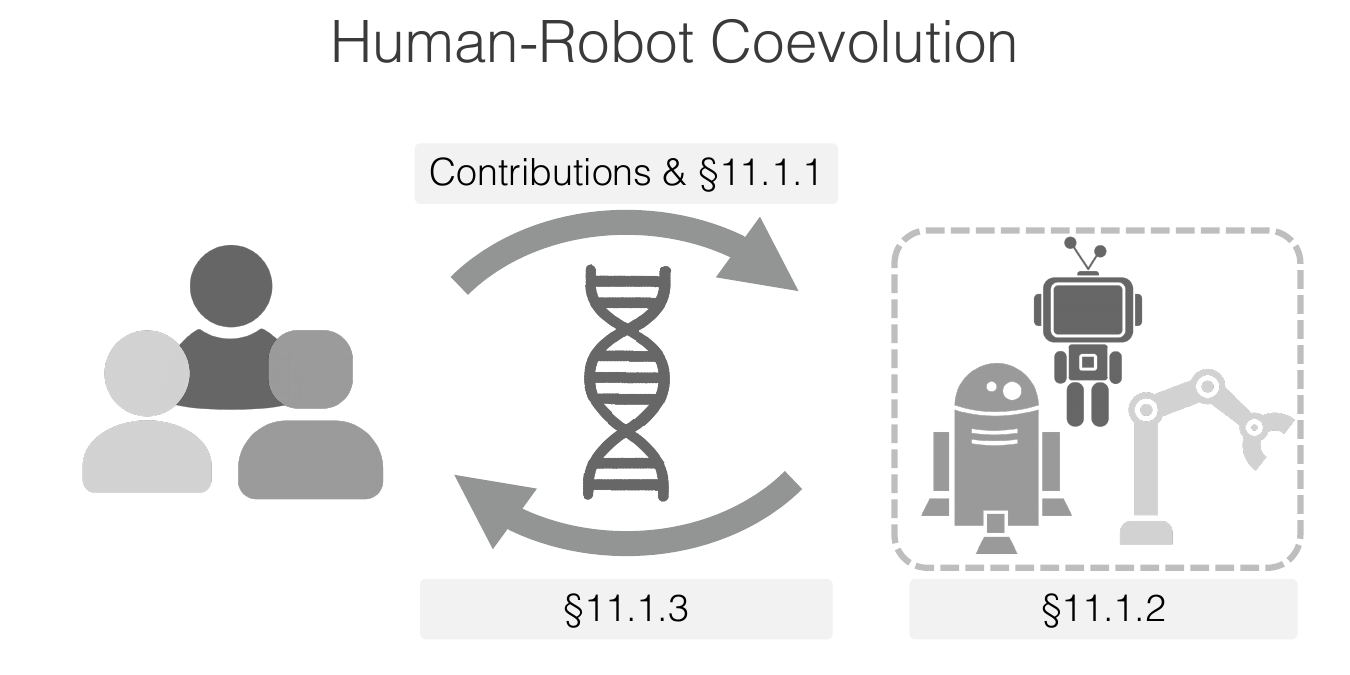}
    \caption{Human-Robot Coevolution presents a long-term research theme based on the technical problem of Open-world Robot manipulation we tackled in this dissertation. }
    \label{fig:conclusion:hrc}
\end{figure}

In this section, we provide a concrete description of the concept and describe important research aspects to work on based on the concept of open-world robot manipulation we have developed in this dissertation. We aim to establish a foundation upon which future research can build, creating a cohesive body of knowledge that addresses both technical and social dimensions of human-robot interaction.

We provide a concrete description of Human-Robot Coevolution as follows: Human-Robot Coevolution is a research topic that examines how humans and robots will evolve together in our daily environments, mutually influencing or determining each other's development. Note that the coevolutionary relationship between humans and technology is not something new. There have been many kinds of technology that have played a vital role in human history: Writing systems (parchment, paper), communications (telephones, radios), transportation (automobiles, aircraft), computation (personal computers, smartphones), and so on. These technologies not only have improved life quality, but also have fundamentally changed how we perceive and understand the world.

We believe that similar transformative changes will take place when robots truly live among us, and robotic technology will reshape human societies, behaviors, and cognitive processes in ways that we are just beginning to understand. Therefore, human-robot coevolution is an interdisciplinary study that consists of many research components. It bridges robotics, artificial intelligence, psychology, sociology, ethics, and design, requiring researchers to consider both technical capabilities and human factors when developing next-generation robotic systems.

In this subsection, we specifically focus on three future directions based on our contributions related to human-robot coevolution: Learning from internet videos, long-term robot autonomy, and personalized interactive robots. In each future direction, we explain key challenges that need to be addressed. Figure~\ref{fig:conclusion:hrc} shows how the dissertation contributions and each future direction fit into the scheme of Human-Robot Coevolution.

\subsection{Learning from Internet Videos}
\label{sec:conclusion:internet_video}

In Figure~\ref{fig:conclusion:hrc}, the arrow from humans to robots refers to the general research direction of how humans can teach robots new skills efficiently. This general research direction includes our contributions to efficient sensorimotor learning in this dissertation. Here, we introduce a new direction to work on---learning from internet videos. 

In Chapter~\ref{chapter:orion} and \ref{chapter:okami}, we studied how robots can imitate manipulation skills from in-the-wild video observations, primarily focusing on using videos captured through personal recording devices such as smartphones and cameras. Building upon our contributions, a promising direction for future research lies in leveraging the large repository of Internet human videos. Internet human videos are readily available on platforms like YouTube, which hosts an enormous collection of videos, a significant portion of which capture human activities across diverse environments. These activity videos contain humans’ daily movements and physical interaction in diverse real-world scenarios, serving as a useful resource for teaching robots complex sensorimotor skills. Moreover, Internet videos serve as a continually growing source of data for robots to learn from, with millions of new videos being uploaded daily. 

Learning from internet videos, however, presents several challenges that need to be addressed. In \orion{} and \okami{}, we used RGB-D videos as inputs. RGB-D videos provide metric depth information that enables robots to exploit spatial regularity in the physical world and develop accurate spatial understanding. Such spatial understanding is critical for robots to imitate videos without action labels. In contrast, internet videos are predominantly available in monocular RGB format, which lacks the metric depth information for robots to understand the spatial relations of objects from videos. The missing depth modality in Internet videos necessitates the development of new approaches for inferring and using spatial information from RGB video data. One possible approach to tackle the challenge is to use recent vision foundation models that estimate depth from monocular RGB videos~\cite{yang2024depth, wang2024dust3r}. However, it remains an open question whether these depth estimation foundation models are applicable to manipulation where the estimation of object geometries has to be accurate. 

Another challenge stems from the frequent occurrence of hand-object occlusions that are common in videos. This issue becomes particularly predominant when we consider dexterous and fine-grained manipulation tasks, especially those involving complex in-hand manipulation behaviors. To effectively learn from videos that involve hand-object occlusions, we need to develop more sophisticated hand-object interaction models that can reliably reconstruct and track both hand poses and the objects. Addressing this challenge requires advances in computer vision models that incorporate priors such as physical understanding, temporal consistency, or regularity in human hands' interaction with objects. 

Furthermore, our work primarily focuses on imitating hand movements or upper body actions from single-vide human demonstrations. Internet videos, on the other hand, involve a lot of behaviors that require whole-body motions. It remains a challenge to enable a humanoid robot to imitate whole-body motions from single-video demonstrations. In the setting of ``open-world imitation from observation,'' whole-body imitation introduces new complexity in terms of balancing, whole-body coordination, and spatial awareness of a large workspace going beyond the tabletop. Tackling the challenge involves simultaneously imitating both locomotion and manipulation, commonly referred to as loco-manipulation. It will be a promising direction to imitate loco-manipulation from single video demonstrations, pushing the frontier of efficient sensorimotor learning.

\subsection{Long-term Robot Autonomy}
\label{sec:conclusion:long_term_robot_autonomy}

Human-Robot Coevolution opens up a wide range of possibilities for future research. However, studying such a research theme can be made possible only if we have robots that actually operate continually in our daily lives. In this dissertation, we focused on the problem of Open-world Robot Manipulation and developed methods for tackling this problem. While we studied the problem by emulating the complexity of manipulation tasks in the real world (intra-task and inter-task generalization), it remains a long quest to have a robot that operates continually in our daily lives. 

All the tasks in experiments of this dissertation and most experiments in the robot learning community assume manipulation that lasts only minutes or long-term duration that involves simple, repetitive primitives. We envision long-term autonomy for robot manipulation, where robots should operate continually for hours, days, or even 24/7 while interacting with all kinds of objects that might be encountered in our everyday environments. 

Recently, my colleague and I introduced the framework of Building-wide Mobile Manipulation~\cite{shah2024bumble}, which enables us to study manipulation directly within daily building environments, using real deployment scenarios as our testbed. This framework requires robots to develop efficient sensorimotor learning capabilities and advanced spatial understanding that extends from table-level to room-level comprehension, encompassing large scenes. 

Existing work tackles mobile manipulation by implementing hand-crafted motor programs or motion planning with the assumption of perfect world modeling. However, current practice falls short of developing versatile manipulation behaviors and is hard to scale to diverse scenarios without a pre-built map. Therefore, robots must be equipped with the ability to perform sensorimotor skills based on their visual observations, necessitating the integration of learning-based visuomotor policies within a mobile manipulation framework. Moreover, robots deployed in buildings also require the ability to interact with humans. In the case of human-robot interaction, manipulation behaviors should extend from engaging with static or articulated objects to include social interactions, such as handshaking and handover, making it crucial for robots to integrate seamlessly into our daily environments.

To efficiently acquire mobile manipulation policies that meet the aforementioned requirements, a robot not only needs to exploit the three regularities identified in this dissertation but also leverage additional regularities to scaffold efficient sensorimotor learning building-wide. For example, we can exploit physical regularity that allows robots to predict the future physical states of environments. Human regularity is also another promising one to leverage, which allows robots to understand the patterns of how humans behave and subsequently enables robots to interact with humans effectively.

\subsection{Personalized Interactive Robots}
\label{sec:conclusion:lrl}

Another important research direction matches the arrow pointing from robots to humans in the figure, which is to build personalized interactive robots. Not only do we want robots to do what humans teach them to do, but we also want robots to adapt to individual preferences of users. Building personalized interactive robots is key to unlocking the deployment of robots around our environments, where individual users will experience satisfying interaction with robots that are operating in our buildings. In Chapter~\ref{chapter:lotus}, we introduced \lotus{}, a lifelong robot learning method that enables robots to continually learn over a sequence of tasks. Lifelong robot learning is a framework that studies how robots continually learn over their lifespan once they are deployed. Therefore, it refers to a class of algorithms that can support the development of building personalized interactive robots. We introduce two concrete downstream applications that future research can work on.

One important application is the personalization of robot behaviors. This application scenario considers robots that need to adapt to individual user preferences. For example, when setting up a table, a robot can rearrange plates and utensils in arbitrary configurations. However, during deployment, the robot might need to arrange the objects in a specific pattern that aligns with user preferences. Such preferences can be specified through contexts in our CMDP formulation of Open-world Robot Manipulation (Section~\ref{sec:bg:open-world-formulation}). The personalization of robot behaviors requires robots to adapt to new contexts, where each new context of the manipulation can be considered a new task. Tackling this problem is related to a recent advancement under the topic of in-context robot learning~\cite{fu2024context}. However, current policy architectures are predominantly based on transformers~\cite{vaswani2017attention}, which inherently face limitations due to the quadratic memory complexity of the self-attention mechanisms and the restricted context length. Designing new architectures for implementing policies is vital to achieving the personalization we are interested in. One possible approach is to leverage new architectures based on the concept of test-time training, a model that updates its internal weights on the fly without explicitly saving all the history information in the context, mitigating the transformer architecture's limitation with respect to context length~\cite{liu2024longhorn}.  

Another important application is how to make robots intentionally forget certain behaviors. We term this application a Skill Unlearning problem, in which we want robots in a lifelong learning process to forget certain behaviors that are potentially dangerous or undesirable. Prior robot policies that are rule-based programs can modify behaviors through simple rule additions or heuristic adjustments, but such a practice does not apply to future robot policies that are most likely to be implemented with neural networks. It remains an open challenge to make sure a neural-network-based robot policy does not generate undesirable behaviors it has learned before, especially in the process of continually learning. A possible way to tackle Skill Unlearning is to develop methods based on prior work, Continual Learning and Private Unlearning (CLPU)~\cite{Liu2022ContinualLA}, which presents a paradigm of how to unlearn declarative knowledge. Extending such methods to unlearn procedural knowledge, such as sensorimotor skills, is an interesting and unexplored direction.

\subsection{Summary}
We proposed Human-Robot Coevolution, an interdisciplinary research theme focusing on how open-world robot manipulation in human-centric environments will transform our society. We introduced three future research directions that are relevant to the contributions of this dissertation. Studying this research theme requires not only technological improvements based on open-world robot manipulation but also careful consideration of its ethical implications for society. We are responsible for deploying these technologies thoughtfully, considering their impact on everyday lives and existing job opportunities.

Just as the internet has fundamentally changed our psychological and social dynamics, personal robots will likely introduce new psychological phenomena. Having intelligent companions and assistants around us will reshape how people understand the world and interact with their surroundings. These intelligent systems may influence human development, social relationships, and even our concepts of agency and intelligence in ways we cannot fully predict.

We need to anticipate these changes and ensure that Human-Robot Coevolution benefits humanity while minimizing potential harms. Achieving this goal requires not just technical innovation, but also interdisciplinary collaboration with experts in ethics, psychology, and social sciences. By bringing together diverse perspectives, we can develop frameworks for responsible innovation that maximize the benefits of robotic technology while preserving human agency and well-being.

\section{Concluding Remarks}

In this dissertation, we tackled the problem of Open-world Robot Manipulation via a methodology of efficient sensorimotor learning. We approached the problem through a perspective of regularity. Understanding and exploiting regularity in the physical world is critical to developing robot intelligence that effectively operates in our real-world environments. Specifically, we identified three major regularities: object regularity, spatial regularity, and behavioral regularity. Then, we proposed methods that can exploit the three regularities.

The pursuit of building general-purpose personal robots transcends the immediate goal of creating autonomous assistants and companions. Studying how to develop such generalist robot autonomy serves as an intellectual expedition that deepens our understanding of both the world and ourselves. The quest to build robot intelligence not only advances technological capabilities but also sheds light on the fundamental nature of intelligence itself---as Richard Feynman said, ``What we cannot create, we do not truly understand.''

\appendices
\chapter{Notation Summary}
\label{chapter:notation}

The following list includes all the notations introduced in Section~\ref{sec:bg:open-world-formulation}.
\begin{itemize}
    \item $\mathcal{S}$ is the state space of all robot's observations of an MDP.
    \item $\mathcal{A}$ is the action space of robot commands of an MDP. 
    \item $\mathcal{P}$ is the stochastic transition probability of an MDP. 
    \item $\mathcal{R}$ is the reward function of an MDP.
    \item $\maxhorizon$ is the maximal horizon of an MDP. 
    \item $\initstate$ is the initial state distribution of an MDP.
    \item $s_t$: State observed at time $t$.
    \item $a_t$: Action taken at time $t$.
    \item $s_{t+1}$: Next state at time $t+1$ after $s_t$.
    \item $\pi$ is a policy as the solution to an MDP.
    \item $\mathcal{J}(\pi)$ is the expected return of a policy $\pi$.
    \item $\context$ is a context variable that corresponds to a task specification from a human. A context can take various forms, including language descriptions, demonstrations, etc. 
    \item $\ContextSpace$: is the context space of human specifications.
    \item $\CMDPMapping$ is a function that maps any context $\context \in \ContextSpace$ to a finite-horizon MDP.
    \item $\Task{\context}$ is a task specified by a context $\context$.
    \item $\pi(\cdot;\Task{\context})$ is the solution to a CMDP, i.e., a policy that conditions on a specified task $\Task{\context}$. 
    \item $\feature_{\bg}$: Features that represent the background component of states.
    \item $\Feature_{\bg}$: Space of features $\feature_{\bg} \in \Feature_{\bg}$.
    \item $\feature_{\cam}$: Features that represent the camera viewpoint component of states.
    \item $\Feature_{\cam}$: Space of features $\feature_{\cam} \in \Feature_{\cam}$.
    \item $\feature_{\obj}$: Features that represent the object instance component of states.
    \item $\Feature_{\obj}$: Space of features $\feature_{\obj} \in \Feature_{\obj}$.
    \item $\feature_{\spatial}$: Features that represent the spatial layout of states.
    \item $\Feature_{\spatial}$: Space of features $\feature_{\spatial} \in \Feature_{\spatial}$.
    \item $\rho(\cdot)$: A function that maps some feature variables to the initial state distribution.
    \item $|\tau|$: Length of a demonstration trajectory $\tau$ (total timesteps).
    \item $\mathcal{J}_\text{intra}(\pi;\Task{})$ is the expected return of a policy $\pi$ over a task $\Task{}$.
    \item $\maxtasknum$ is the maximal number of tasks in a multitask or lifelong learning setting.
    \item $\tasknum$ is the index of a task from all the $\maxtasknum$ tasks. 
    \item $\LLTotalStep$ is the total number of steps a robot sequentially learns over the $\maxtasknum$ tasks in the lifelong learning process.
    \item $\llstep$ is the index of a step in the lifelong learning process.
    \item $\mathcal{J}_{\text{inter}}(\pi)$ is the expected return of a policy $\pi$ across all the previously learned tasks in the lifelong learning setting.
    \item $r_{i,j}$: Success rate on task $j$ after learning $i$ tasks. Both $i$ and $j$ are indexing variables without specific semantics.
    \item $\bar{r}_i$: Success rate averaged over all intermediate checkpoints when training on task $i$.
\end{itemize}

The following list includes all the notations introduced in Section~\ref{sec:bg:policy-skills}.
\begin{itemize}
    \item $\metacontroller{}$ is the high-level policy in a hierarchical policy.
    \item $\skillpolicy{i}$ is a low-level skill policy $i$ in a hierarchical policy.
    \item $K$ is the total number of low-level policies in a hierarchical policy.
    \item $\param$: Skill parameters predicted by a meta-controller $\metacontroller{}$.
    \item $k$ is the index of a low-level policy sampled from a categorical distribution. 
    \item $\mathbf{p}(s_t, \context)$ is the probability of selecting a low-level policy given a state $s_t$ and the context $\context$.  
    \item $p_{i}(s_t, \context)$ is a logit in $\mathbf{p}(s_t, \context)$.
\end{itemize}

The following list includes all the notations introduced in Section~\ref{sec:bg:imitation-learning}.
\begin{itemize}
    \item $D^c$ is the demonstration dataset for task $T^c$.
    \item $o_t$ is the sensory input of a robot at time $t$.
    \item $\tau^c_i$ is a demonstration trajectory from a demonstration dataset $D^c$.
    \item $\mathcal{J}^{\text{BC}}_{\text{intra}}$ is the objective function of behavioral cloning over a task. 
    \item $\mathcal{L}_{\text{BC}}$ is a supervised learning loss in a behavioral cloning algorithm.
    \item $K_{\text{mix}}$ is the of modes in Gaussian Mixture Models. Note that this is unrelated to $K$, which represents the number of skills. 
    \item $\theta$ is the learnable parameters of a Gaussian Mixture Model(GMM) policy model.
    \item $\eta_k$ is the mixture weights in a GMM.
    \item $\mu_k$ is the mean of a mode in a GMM.
    \item $\sigma_k$ is the standard deviation of a mode in a GMM.
    \item $\hcbvar(t)$ represents all the intermediate variables predicted by a meta-controller at time $t$ in the hierarchical behavioral cloning algorithm. 
    \item $\bar{D}^{\context}$ is an augmented demonstration data from $D^{\context}$.
    \item $\bar{\tau}^{\context}_{i}$ is an augmented demonstration trajectory from $\tau^{\context}_{i}$.
    \item $\mathcal{J}^{\text{BC}}_{\text{inter}}$: Objective function of lifelong imitation learning.
    \item $\mathcal{L}_{\text{BC}}$: Loss function of behavioral cloning.
    \item $\mathcal{L}_{\text{GMM}}$: Loss function of behavioral cloning when policy outputs are modeled as Gaussian Mixture Models.
    \item $\bar{\tau}$: An augmented demonstration trajectory based on $\tau$. This augmented trajectory provides labels for the hierarchical behavioral cloning algorithm.
    \item $\bar{D}$: An augmented demonstration dataset that consists of augmented demonstration trajectories. 
    \item $q \in \mathbb{R}^{n}$: Joint configuration of a robot.
\end{itemize}

The following list includes all the notations introduced in Chapters~\ref{chapter:viola}---\ref{chapter:lotus}.

\begin{itemize}
    \item $h_t$: A feature vector encoded from the state observation $o_t$. 
    \item $z_t$: Object-centric representation at time $t$, derived from the state observation $s_t$.
    \item $\objectnum$: Number of object proposals from an image in a \viola{} policy.
    \item $\hat{a}_t$: Latent action tokens outputted from the transformer in \viola{}.
    \item $\tokenin$: Input latent vector (token) into a transformer model.
    \item $\tokenout$: Output latent vector (token) from a transformer model.
    \item $\Tokenin$: The tensor representation of all input tokens.
    \item $\Tokenout$: The tensor representation of all output tokens.
    \item $\hat{a}$: Latent vector output from a neural network further decoded into a policy action $a$.
    \item $\historytime$: The maximal history timestep at which the sensory input of a robot is included in the state $s_t$. 
    \item $PE$: Positional Encoding. 
\end{itemize}

\begin{itemize}
    \item $V$: A video.
    \item $\mG$: An Open-world Object Graph (OOG).
    \item $\kfindex$: The index of a keyframe from the video (i.e., a step in the manipulation plan). 
    \item $\okamiplan$: A step in the reference plan in \okami{}.
    \item $\KF$: The total number of subgoals from in the reference plan ($F+1$ is the total number of keyframes, as the first frame is also a keyframe but not a subgoal).
    \item $\mG.vo_i$: An object node in an OOG. 
    \item $\mG.vh$: A hand node in an OOG.
    \item  $\mG.vp_{ij}$: A point node in an OOG.
    \item $\mG.eo_{ik}$: An object-object edge in an OOG.
    \item $\mG.eh_{i}$: An object-hand edge in an OOG.
    \item $\mG.ep_{ij}$: An object-point edge in an OOG.
    \item $\mathcal{V}$: The set of all vertices in a graph data structure. 
    \item $\mathcal{E}$: The set of all edges in a graph data structure. 
    \item $\widehat{\objectnode}_{\target}$: Point cloud of a target object at a keyframe from the human video.
    \item $\widehat{\objectnode}_{\reference}$: Point cloud of a reference object at a keyframe from the human video.  
    \item  $\hat{\kptraj}_{\target}$: The keypoint trajectory of a target object between two keyframes, estimated from the human video. 
    \item $\objectnode_{\target}$: Point cloud of a target object in the current robot observation.
    \item $\objectnode_{\reference}$: Point cloud of a reference object in the current robot observation.
    \item  $\kptraj_{\target}$: The predicted keypoint trajectory of a target object between two keyframes.  
    \item $\tcp$: An end-effector pose in both position and rotation. 
\end{itemize}

\begin{itemize}
    \item $\tilde{\dataset}_{k}$: The partitioned dataset for training the $k$-th low-level skill policy.  
    \item $\subgoaltime$: The lookahead time window to determine a subgoal. 
    \item $\segmentfinal$: The last timestep of the end of a segment. 
    \item $\tg$: The timestep of a determined subgoal state for $t$. 
    \item $\text{Encoder}_{k}$: The encoder to encode a subgoal state $s_t$ into skill parameters $\param_{t}$, which a low-level policy $\skillpolicy{k}$ takes as inputs. This encoder is used to generate pseudo labels for the low-level policy to predict the parameters.   

    % \item $W$: The minimal length of the initial temporal chunk for subsequent hierarchical clustering. 
    \item $\languagegoal$: The language description of a task goal. 
    \item $K_{\llstep}$: The total number of skills discovered at the lifelong learning step $\llstep$.
    \item $B_{k, \llstep}$: The memory buffer for the $k$-th skill at the lifelong learning step $\llstep$. 
\end{itemize}
\chapter{Acronym Summary}
\label{chapter:acronym}

\section*{Chapter~\ref{chapter:intro}}
\begin{itemize}
    \item PCs: Personal Computers. %, general-purpose electronic devices for individual use.
    \item LLMs: Large Language Models.
    \item VLMs: Vision-Language Models.
\end{itemize}

\section*{Chapter~\ref{chapter:bg}}

\begin{itemize}
    \item MDP: Markov Decision Process. %, a mathematical framework for modeling decision-making in scenarios with probabilistic outcomes.
    \item CMDP: Contextual Markov Decision Process. %, an extension of MDP that incorporates contextual information.
    \item GMM: Gaussian Mixture Model. %, a probabilistic model representing multiple Gaussian distributions.
    \item BC: Behavioral Cloning. %, a method for learning behavior policies directly from demonstration data.
    \item HBC: Hierarchical Behavioral Cloning. %, an extension of BC that incorporates hierarchical structure.
    \item OSC: Operational Space Controller. %, a framework for controlling robot motion in task-relevant coordinates.
    \item DoF: Degrees-of-Freedom. %, the number of independent parameters defining a system's configuration.
    \item FWT: Forward Transfer metric. %, measures the impact of prior learning on new task performance.
    \item NBT: Negative Backward Transfer metric. %, quantifies performance degradation on previously learned tasks.
    \item AUC: Area Under the success rate Curve. %, a metric for evaluating performance over time.
    \item RGB: Red, Green, Blue color channels used in digital image representation.
    \item RGB-D: RGB and Depth information combined for richer scene understanding.
    \item IR: Infrared, electromagnetic radiation with wavelengths longer than visible light.
    \item SLAM: Simultaneous Localization And Mapping. %, a technique for building maps while tracking location.
\end{itemize}

\section*{Chapter~\ref{chapter:viola}}

\begin{itemize}
    \item VIOLA: Visuomotor Imitation via Object-centric Learning, proposed in Chapter~\ref{chapter:viola}.
    \item RPN: Region Proposal Network. %, a neural network module that predicts object bounding boxes from images.
    \item RGB: Red, Green, Blue color channels used in digital image representation.
    \item FPN: Feature Pyramid Network. % $, a neural network architecture that creates multi-scale feature representations for object detection.
    \item ROI: Region Of Interest. % , a selected portion of an image for focused analysis or processing.
    \item MHSA: Multi-Head Self-Attention mechanism inside a transformer architecture.
    \item FFN:  Feed-Forward Network of fully connected layers. % , a neural network where information flows in one direction without loops.
    \item MLP: Multi-Layer Perceptron. % , a class of feedforward neural networks with multiple layers of nodes.
    \item LayerNorm: Layer Normalization. % , a technique to normalize the input tensor over the channel dimension.
    \item GMM: Gaussian Mixture Model. % , a probabilistic model that represents data distribution as a weighted sum of Gaussian distributions.
    \item SOTA: State-Of-The-Art. % , referring to the best model developed so far in a specific field or research direction.
    \item BC: Behavioral Cloning.% , an imitation learning algorithm based on supervised training.
    \item BC-RNN: Behavioral Cloning with Recurrent Neural Network. 
    \item OREO: The Object-aware
Regularization baseline. % , which is proposed by Park et al.~\cite{park2021object}.
    \item VQ-VAE: Vector-Quantized Variational AutoEncoder. % , a type of generative model that learns discrete representations of data.
    \item IoU: Intersection over Union. % , a metric measuring the overlap between predicted and ground truth bounding boxes.
    % \item PE: Positional Encoding. % , a technique to incorporate position information in transformer-based models.
\end{itemize}

\section*{Chapter~\ref{chapter:groot}}

\begin{itemize}
    \item GROOT: Generalizable Rbot Manipulation Policies for Visuomotor Control, proposed in Chapter~\ref{chapter:groot}.
    \item SCM: Segmentation Correspondence Model.
    \item VOS: Video Object Segmentation.
    \item S2M: The pre-trained interactive segmentation model named Scribble-to-Mask. % ~\cite{cheng2021mivos}.
    \item SAM: The pre-trained segmentation model named Segment-Anything. % ~\cite{kirillov2023segment}.
    \item Point-MAE: Masked Autoencoders for Point Cloud. 
    \item MLP: Multi-Layer Perceptron. % , a class of feedforward neural networks with multiple layers of nodes.
    \item DINOv2: The pre-trained vision transformer model via self-supervised learning. % , proposed by Oquab et al.~\cite{oquab2023dinov2}.
\end{itemize}

\section*{Chapter~\ref{chapter:orion}}

\begin{itemize}
    \item ORION: Open-world Video Imitation proposed in Chapter~\ref{chapter:orion}.
    \item OOG: Open-world Object Graph.
    \item TAP: Track-Any-Point. 
    \item SE(3): The Special Euclidean Group in three dimensions. 
    \item IK: Inverse Kinematics.
    \item DoF: Degrees-of-Freedom.
    \item HaMeR: The hand reconstruction model. %  proposed by Pavlakos et al.~\cite{pavlakos2023reconstructing}.
    \item RANSAC: Random sample consensus.
    \item FPFH: Fast-Point Feature Histograms.
    \item iOS: iPhone Operating System.   
\end{itemize}

\section*{Chapter~\ref{chapter:okami}}

\begin{itemize}
    \item OKAMI: Object-aware Kinematic Retargeting for Manipulation Imitation proposed in Chapter~\ref{chapter:okami}.
    \item GPT4V: GPT for Vision.
    \item SMPL-H: The Skinned Multi-Person Linear Model with Hands.
    \item ACT: The Action-Chunking Transformer model. 
    \item SLAHMR: Simultaneous Localization And Human Mesh
Recovery. %  proposed by Ye et al.~\cite{ye2023decoupling}. 
\end{itemize}

\section*{Chapter~\ref{chapter:buds}}

\begin{itemize}
    \item BUDS: Bottom-Up Discovery of Sensorimotor Skills, proposed in Chapter~\ref{chapter:buds}.
    \item cVAE: Conditional Variational Autoencoder.
    \item ELBO: Evidence Lower Bound.
    \item CP: The baseline model based on Changepoint Detection.
    \item RBF: Radial Basis Functions.
    \item GTI: The baseline model named Generalization
Through Imitation. % ~\cite{mandlekar2020learning}.
    % \item IRIS: The baseline model named Implicit Reinforcement
    % without Interaction at Scale.% proposed by Mandlkar et al.~\cite{mandlekar2020iris}.
    % \item RPL: The baseline model named Replay Policy Learning. % ~\cite{gupta2020relay}.
\end{itemize}

\section*{Chapter~\ref{chapter:lotus}}

\begin{itemize}
    \item LOTUS: Lifelong knowledge Transfer using skills, proposed in Chapter~\ref{chapter:lotus}.
    \item ER: Experience Replay.
    \item GMM: Gaussian Mixture Model.
    \item ELBO: Evidence Lower Bound.
    \item LIBERO: The lifelong learning benchmark on robot manipulation tasks proposed in Chapter~\ref{chapter:libero}.
    % \item FWT: Forward transfer metric.
    % \item NBT: Negative backward transfer metric.
    % \item AUC: Area under the success rate curve.
    \item MTFT: The multitask fine-tuning baseline.
    \item CLIP: The pre-trained vision model named Contrastive Language-Image Pre-training.
    \item R3M: The pre-trained vision model named Reusable Representations for Robot Manipulation.
\end{itemize}

\section*{Chapter~\ref{chapter:libero}}

\begin{itemize}
    \item LIBERO: Lifelong Learning Benchmark on Robot Manipulation Tasks, proposed in Chapter~\ref{chapter:libero}.
    \item PDDL: Planning Domain Definition Language.
\end{itemize}

\section*{Chapter~\ref{chapter:related-works}}
\begin{itemize}
    \item IL: Imitation Learning.
    \item RL: Reinforcement Learning. 
    \item DMP: Dynamic Movement Primitives.
    \item PrMP: Probabilistic Movement Primitives.
    \item CLPU: Continual Learning and Private Unlearning.
\end{itemize}

\newpage

\chapter{Additional Details of Part~\ref{part:I}}
\label{chapter:appendix_chapter_I}

\section{\viola{} Implementation Details}
\label{ablation_sec:viola:implementation}
We describe all the details of our model implementation aside from the ones mentioned in the main text.

\subsection{Model Details}
\label{ablation_sec:viola:model}

\paragraph{Region and Context Features Computation.}
For encoding images, we use ResNet-18~\cite{he2016deep} as the backbone. To obtain a spatial feature map with sufficient resolution for feature pooling, we remove the last four layer of ResNet-18, producing a spatial feature map with the size of $16\times 16$. Then we perform ROI Align~\cite{he2017mask} to get the pooled features of size $6 \times 6$. In order to map the pooled features into the same size as input tokens, we add a linear layer to project the pooled feature. We have already mentioned in Section~\ref{sec:viola:method} that we obtain \textit{global context featurex} by passing the spatial feature maps through a Spatial Softmax layer.
For eye-in-hand images, we use a Resnet-18 backbone followed by Spatial Softmax to get \textit{eye-in-hand features}. 
We use one linear layer to map robots' states to \textit{proprioceptive features}. 

\paragraph{Sensor Modalities.} Both \viola{} and the baslines use the same set of sensor modalities, namely the RGB images from the workspace camera, RGB images from the eye-in-hand camera, joint configuration, and binary gripper state. We use the eye-in-hand camera following the same setup as in the previous work on imitation learning for manipulation~\cite{mandlekar2021matters}. This camera view has been empirically shown to contribute significantly to the performance of visuomotor policies ~\cite{mandlekar2021matters, hsu2022vision, burgess2022eyes}.

\paragraph{Neural Network Details.} We use a standard Transformer~\cite{vaswani2017attention} architecture. We use $4$ layers of transformer encoder layers and $6$ heads of the multi-head self-attention modules. For the two-layered fully connected networks, we use 1024 hidden units for each layer. For the GMM output head, we choose the number of modes to be $5$, the same as in Mandlekar et al.~\cite{mandlekar2021matters}.

\paragraph{Positional Encoding.} We provide details about both positional features for regions and the temporal positional encoding. For all the encodings, they have the same dimensionality $D$ as the input tokens. For a region's positional feature, we compute it based on its bounding box positions $bbox_{pos}=(x_0, y_0, x_1, y_1)$:
\begin{align*}
    PE(bbox_{pos}, 4i) &= \psi(\frac{x_0}{10^{4i/D}}) \\
    PE(bbox_{pos}, 4i+1) &= \psi(\frac{y_0}{10^{(4i+1)/D}}) \\
    PE(bbox_{pos}, 4i+2) &= \psi(\frac{x_1}{10^{(4i+2)/D}})\\
    PE(bbox_{pos}, 4i+3) &= \psi(\frac{x_2}{10^{(4i+3)/D}})
\end{align*}
and $\psi$ is the sine function when i is even and cosine function when i is odd.

For computing temporal positional encoding, we use the following equation for each dimension $i$ in the encoding vector at a temporal position $pos$:
\begin{align*}
    PE(pos, 2i) &= \sin{(\frac{pos}{10^{2i/D}})} \\
    PE(pos, 2i+1) &= \cos{(\frac{pos}{10^{(2i+1)/D}})}
\end{align*}

We choose the frequency of positional encoding to be $10$, which is different from the one in the original transformer paper. We make this design choice because our input sequence is much shorter than those in natural language tasks, hence we choose a smaller value to have sufficiently distinguishable positional features for input tokens. 

\paragraph{Data Augmentation.} 
We apply both color jittering and pixel shifting to~\viola{} and all the baselines. Random erasing is used for~\viola{} and its variants. We applied random erasing to~\viola{} to prevent the policy from overfitting to region proposals. We did not apply this augmentation for baseline behavioral cloning models such as \bcrnn{} because adding random erasing yields lower performance than without using random erasing. For implementing random erasing, we use the open-sourced Torchvision function whose parameters are: $p=0.5$, $\text{scale}=(0.02, 0.05)$, $\text{ratio}=(0.5, 1.5), \text{value}=\text{random}$.
% We randomly apply random erasing during training with the probability of $0.5$. Random erasing will randomly fill in random gaussian noise into the region that is size of 2$\%$ to 5$\%$ of the image, with aspect ratio ranging from $0.5$ to $1.5$ (in terms of Torchvision function, the parameters are: $p=0.5$, $\text{scale}=(0.02, 0.05)$, $\text{ratio}=(0.5, 1.5), \text{value}=\text{random}$).

We have also applied color jittering to the demonstration data. We perform color jittering as follows: brightness=0.3, contrast=0.3, saturation=0.3, hue=0.05 on 90\% of the trajectories. We keep 10\% unchanged so that we still keep the main visual cue patterns from demonstrations. We do 4 pixel shifting as in prior work from Mandlekar et al.~\cite{mandlekar2021matters}.

Aside from data augmentation on images, we also add very small gaussian noise on the proposal positions to add more variety on pixel locations of proposals.

\paragraph{Training Details.} In all our experiments of \viola{}, we train for $50$ epochs. We use a batch size of $16$ and a learning rate of $10^{-4}$. We use negative log likelihood as the loss function for action supervision loss since we use a GMM output head. As we notice that validation loss doesn't correlate with policy performance~\cite{mandlekar2021matters}, we use a pragmatic way of saving model checkpoint, which is to save the checkpoint that has the lowest loss over all the demonstration data at the end of training. As we use a Transformer as a model backbone, we apply several optimization technique that is suitable for training transformer. We use an AdamW optimizer~\cite{kingma2014adam} along with cosine annealing scheduler of learning~\cite{loshchilov2016sgdr}. We also apply gradient clip~\cite{vaswani2017attention}: $0.1$ for the two long-horizon tasks, namely \kitchen{} and \coffee{}, and $10$ for the rest of the tasks. For training baselines, we take the general configs from Mandlekar et al., which is $500$ epochs for BC and $600$ epochs for BC-RNN~\cite{mandlekar2021matters}. We follow the same model saving criteria for BC-RNN as \viola{}.

% The Robomimic benchmark study reports the highest performance of all checkpoints across training, as the loss values do not correlate with policy performance~\cite{mandlekar2021matters}. We recognize that this criterion can inflate the model performance, and it's not pragmatic to evaluate all model checkpoints during real robot experiments. Therefore, we adopt the saving criteria that is same as Zhu et al.~\cite{zhu2022bottom}, where the policies that have lowest training loss on demonstration datasets are taken. As we show in our experiments, our training procedure of \viola{} can easily find a good-performance checkpoint consistently across simulation and real-world.

\subsection{Visual Feature Design}
\label{ablation_sec:viola:visual-feature}
In \viola{}, we learn a spatial feature map from scratch for extracting visual features, aiming to learn actionable visual features that are informative for continuous control. 
We coondcut an ablation study to demonstrate two key points: 1) the importance of learning actionable visual features from scratch, and 2) that pre-trained visual features from visual tasks alone are insufficient for control tasks. Our experiment compared our model against two variants that use feature maps from Feature Pyramid Network (FPN) in RPN---one with fine-tuning and one without. The result is presented in Table~\ref{tab:viola:visual-features}. The table suggests that directly using pre-trained spatial feature maps without fine-tuning is overall worse in performance than learning from scratch. Fine-tuning the feature map on the downstream tasks doesn't give a matching performance as learning from scratch. This result shows the importance of optimizing actionable visual features for manipulation tasks. Fine-tuning FPN gives an increase in performance to almost match our original model, but doesn't outperform it. This result shows that pre-trained visual features are not critical to the model performance, not to mention that the trainable parameters in FPN take about $200$ MB while our convolutional encoder for spatial feature maps only takes $14$ MB trainable parameters.

\begin{figure}[t]
\centering
  \resizebox{1.0\linewidth}{!}{
  \begin{tabular}{lcccc}
    \toprule
    \textbf{Models} & \canonical & \placement & \distracting & \camera \\
    \midrule
\viola{}(From Scratch) & {87.6 $\pm$ 1.1} & {68.3 $\pm$ 1.5} & {74.4 $\pm$ 5.7} & {50.7 $\pm$ 0.6} \\
w/o Fine-tuned FPN & {79.7 $\pm$ 1.5} & {52.7 $\pm$ 1.2} & {61.8 $\pm$ 3.3} & {53.7 $\pm$ 1.8} \\
w/ Fine-tuned FPN & {84.9 $\pm$ 1.1} & {60.4 $\pm$ 78.4} & {76.7 $\pm$ 5.1} & {46.7 $\pm$ 1.7} \\
    \bottomrule
    \end{tabular}
    }
    \caption{\label{tab:viola:visual-features} Comparison among models that use spatial feature maps learned from scratch, pre-trained spatial feature masks without or with fine-tuning. }
\end{figure}

\paragraph{Evaluation Horizons.} Based on the collected demonstrations, we determine the evaluation horizons of our simulation tasks in the case of \canonical{}, which are: 1) $1000$ for \sort{}, $800$ for \stack{}, and $1500$ for \kitchen{}. And for evaluating in testing variants, we have $200$ steps more than the evaluation steps of \canonical{} for each task.

\section{\groot{} Implementation Details}
\label{ablation_sec:groot:implementation}
We describe all the details of our model implementation aside from the ones mentioned in the main text.

\paragraph{Neural Network Details.} We use a standard Transformer~\cite{vaswani2017attention} architecture in \groot{}. We use $4$ layers of transformer encoder layers and $6$ heads of the multi-head self-attention modules. For the two-layered fully connected networks, we use 1024 hidden units for each layer. For the GMM output head, we choose the number of modes to be $5$, which is the same as in ~\citet{mandlekar2021matters}.

Additionally, Table~\ref{tab:groot:component-list} summarizes what components are pre-trained and frozen, and what components are trained from scratch.

\begin{table}[h!]
    \centering
    \resizebox{0.6\textwidth}{!}{%
        \begin{tabular}{l|c}
        \toprule
        Components &  Pre-trained or From Scratch\\
        \midrule
        S2M~\cite{cheng2021modular} & Pre-trained\\
        VOS~\cite{cheng2022xmem} & Pre-trained\\
        DinoV2~\cite{oquab2023dinov2} & Pre-trained \\
        SAM~\cite{kirillov2023segment} & Pre-trained \\
        PointNet++~\cite{qi2017pointnet++} &  From Scratch\\
        Transformer~\cite{vaswani2017attention} & From Scratch \\
        GMM-MLP~\cite{mandlekar2021matters,bishop1994mixture} & From Scratch \\
        \bottomrule
        \end{tabular}
    }
    \vspace{5pt}
    \caption[List of model components in \groot{} policies, categorized by whether they are pre-trained or trained from scratch.]{This table presents which component is pre-trained and then frozen in our experiments, and which is trained from scratch. We denote them as Pre-trained and From Scratch, respectively.}
    \label{tab:groot:component-list}
    \vspace{-0.2in}
\end{table}

\paragraph{Temporal Positional Encoding.}
For computing temporal positional encoding, we follow the equation for each dimension $i$ in the encoding vector at a temporal position $pos$:
\begin{align*}
    PE(pos, 2i) &= \sin{(\frac{pos}{10^{2i/D}})} \\
    PE(pos, 2i+1) &= \cos{(\frac{pos}{10^{(2i+1)/D}})}
\end{align*}

We choose the frequency of positional encoding to be $10$ which is different from the one in the original transformer paper. This is because our input sequence is much shorter than those in natural language tasks, hence we choose a smaller value to have sufficiently distinguishable positional features for input tokens. Note that all the input tokens at the same timestep are added with the same positional encoding to inform the transformer about the temporal order of received.

\paragraph{Number of Clusters in Point-MAE.} Across all experiments, we choose $10$ for the number of clusters in Point-MAE. Here, we also denote the dimensions of inputs and outputs in the process of Point-MAE. Suppose that we have a 3D point cloud that consists of $P$ points. The input to Point-MAE is a vector in $\mathbb{R}^{P\times 3}$. Point-MAE groups the point cloud into $n$ clusters, with $\bar{p}$ points in each cluster. This operation gives us grouped point clouds that are in $\mathbb{R}^{n\times \bar{p} \times 3}$. The grouped point clouds are passed through a PointNet++ encoder, giving us latent vectors in $\mathbb{R}^{n\times \bar{p} \times h}$, where h is the dimension of latent features. Finally, given a masking ratio $\alpha$, the actual latent vectors that are included in the transformer input are vectors in $\mathbb{R}^{(1-\alpha)n\times \bar{p} \times h}$.

\paragraph{Training Details.}  For point masking, we use a masking ratio of $0.6$ in simulation, and $0.75$ for real world. Because of the limited field of view, the occlusion of objects in simulation is severe. To properly evaluate policies in simulation, we add an eye-in-hand camera that only captures close-distance depth (the depth observation is clipped to the range of gripper tips). This design choice allows the policies to learn while preventing policies from relying entirely on eye-in-hand cameras. 

In all our experiments of \groot{}, we train for $100$ epochs. We use a batch size of $16$ and a learning rate of $10^{-4}$. We use negative log-likelihood as the loss function for action supervision loss since we use a GMM output head. As we notice that validation loss doesn't correlate with policy performance~\cite{mandlekar2021matters}, we adopt a pragmatic way of saving model checkpoint as in \viola{}, which is to save the checkpoint that has the lowest loss over all the demonstration data at the end of training (See Appendix~\ref{ablation_sec:viola:implementation}).  We apply a gradient clip at $100$ across all the experiments to prevent training from gradient explosion.

\paragraph{Baseline Implementations.}  As we mainly focus on comparing the effectiveness of representations, all the transformer-based baselines (\viola{} and \mae{}) use the same transformer architure for a fair comparison. Since none of the baselines were proposed for learning with RGB-D observations, we implement them with minimal changes to accommodate the RGB-D observations. For \bcrnn{}, we encode depth images with an additional resnet encoder, and concatenate the features along with the other features as inputs to the RNN backbone. For \viola{}, we extract the task-agnostic proposals and back-project each proposal into point clouds, giving \viola{} a fair comparison with our approach. As for \mae{}, we patchify both RGB and depth images and pass the unmasked patches into the transformer architecture.

\paragraph{Point Cloud Encoders.} We conduct an ablation experiment to study the importance of point net encoders in \groot{}. We use the simulation task ``Put the moka pot on the stove'' to compare the PointNet++ architecture used with two other common architectures for processing point clouds, namely DGCNN~\cite{wang2019dynamic} and Point Transformer~\cite{zhao2021point}. We show the ablation study in Table~\ref{tab:groot:pointnet}. We see that Point Transformer-based encoder generally performs better than the DGCNN-based one, but neither of them brings significant improvements to the policy. Therefore, we stick to our choice of PointNet++ for encoding point clouds.

\begin{table}[h]
\centering
    \resizebox{1.0\textwidth}{!}{%
        \begin{tabular}{l|c|c|c}
        \toprule
        \textbf{Point cloud encoder} & \textbf{PointNet++} & \textbf{DGCNN (GNN-based)} & \textbf{Point Transformer} \\
        \midrule
        \canonical{} & $80.0 \pm 8.2$ & $65.0 \pm 4.1$ & $67.5 \pm 7.5$ \\
        \distractionseasy{} & $63.3 \pm 13.1$ & $63.3 \pm 6.2$ & $77.5 \pm 7.5$ \\
        \distractionshard{} & $63.3 \pm 8.5$ & $63.3 \pm 4.7$ & $75.0 \pm 5.0$ \\
        \cameraeasy{} & $81.7 \pm 9.4$ & $65.0 \pm 0.0$ & $60.0 \pm 0.0$ \\
        \camerahard{} & $61.7 \pm 9.4$ & $63.3 \pm 8.5$ & $75.0 \pm 5.0$ \\
        \bottomrule
        \end{tabular}
}
\caption{Ablation study on the choice of point cloud encoders.}
\label{tab:groot:pointnet}
\end{table}

\begin{figure}[h]
    \centering
    \includegraphics[width=0.4\linewidth, trim=0cm 0cm 0cm 0cm,clip]{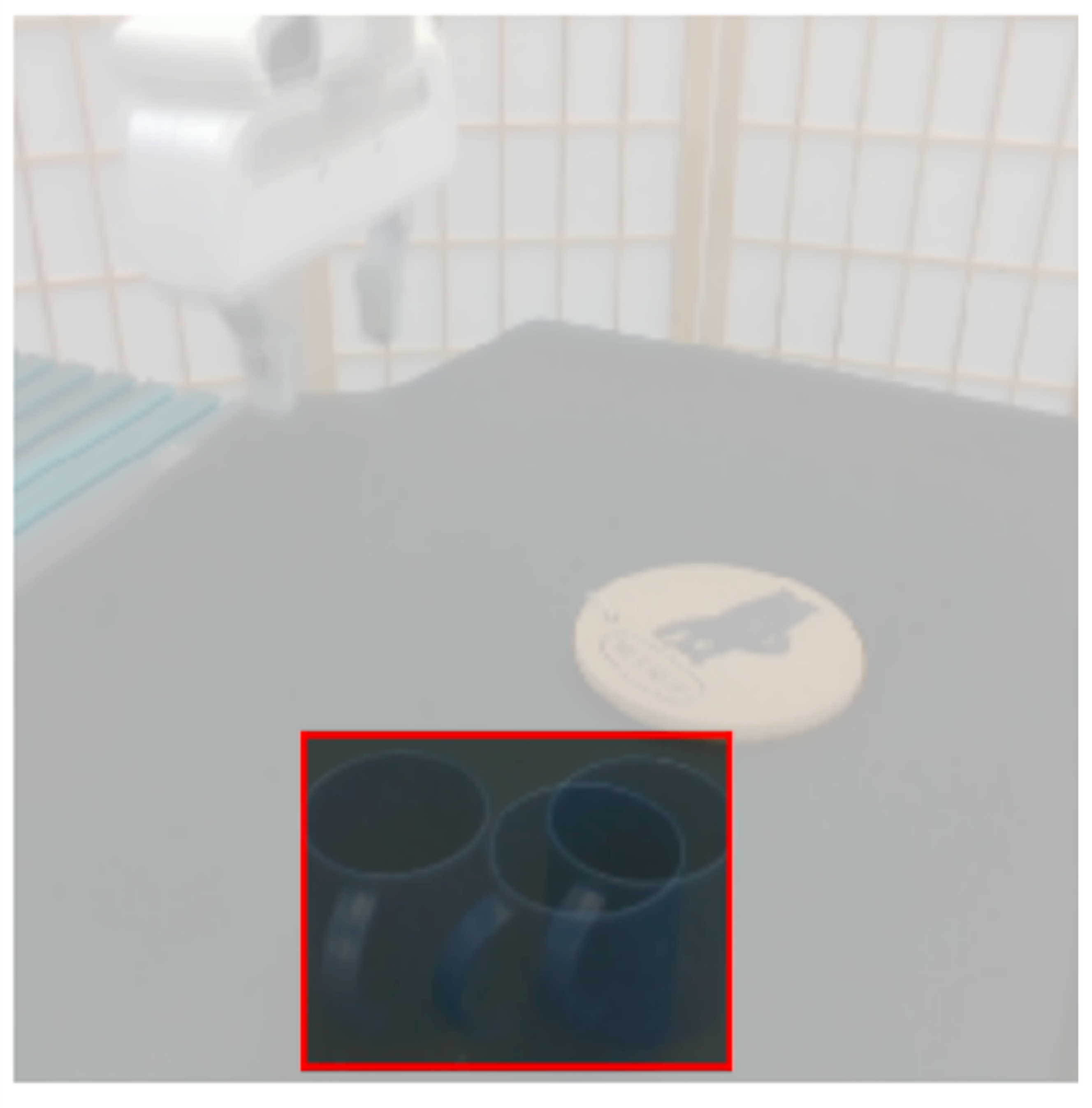}
    \caption[Overlayed image of initial frames from different rollouts of an evaluation task.]{Overlayed image of initial frames from three representative rollouts of the task ``Put The Mug On The Coaster'' during camera generalization tests.}
    \label{fig:groot:initial}
\end{figure}

\paragraph{Initial Conditions.} To systematically evaluate policies in diverse visual conditions, we pre-specify a region and randomly place the objects inside the region. This region is used for object placements during demonstration collection as well as policy evaluation. Here, Figure~\ref{fig:groot:initial} shows an overlay image of initial frames from three representative rollouts of the task ``Put The Mug On The Coaster'' during camera generalization tests. As the figure shows, the mug is placed at different places with different orientations at the beginning of rollouts. The objects will be placed at locations that do not overlap with other locations. In all, the initial conditions are randomized such that policies really need to go toward the correct location of an object in order to achieve high success rates.

\newpage

\chapter{Additional Details of Part~\ref{part:II}}
\label{chapter:appendix_chapter_II}

\section{\orion{} Implementation Details}
\label{ablation_sec:orion:implementation}

\subsection{Additional Details of Tasks}
\label{ablation_sec:orion:tasks}

\paragraph{Success Conditions.} We describe the success conditions for each of the seven tasks in detail. 
\begin{itemize}
    \item \mugcoaster{}: A mug is placed upright on the coaster.
    \item \simpleboat{}: A red block is placed in the slot closest to the back of the boat. The block needs to be upright in the slot.  
    \item \chip{}: A bag of chips is placed on the plate, and the bag does not touch the table. 
    \item \llama{}: A pot of succulents is inserted into a white vase in the shape of a llama. 
    \item \rearrange{}: The mug is placed upright on the coaster, and the cream cheese box is placed on the plate.
    \item \complexboat{}: The chimney-like part is placed in the slot closest to the front of the boat. The red block is placed in the slot closest to the back of the boat. Both blocks need to be upright in the slots. 
    \item \breakfast{}: The mug is placed on top of a coaster, the cream cheese box is placed in the large area of the plate, and the food can is placed on the small area as shown in the video demonstration. 
\end{itemize}

\subsection{Additional Details in Methods}
\label{ablation_sec:orion:model}

\paragraph{Changepoint Detections.} 
We use changepoint detection to identify changes in velocity statistics of TAP keypoints. Specifically, we use a kernel-based changepoint detection method and choose radial basis function~\cite{killick2012optimal}. The implementation of this function is directly based on an existing library Ruptures~\cite{truong2020selective}. 

\paragraph{Plane Estimation.} In Section~\ref{sec:orion:plan-generation}, we have mentioned using the prior knowledge of tabletop manipulation scenarios and transforming the point clouds by estimating the table plane. Here, we explain how the plane estimation is computed. Concretely, we rely on the plane estimation function from Open3D~\cite{zhou2018open3d}, which gives an equation in the form of $ax + by + cz = d$. From this estimated plane equation, we can infer a normal vector of the estimated table plane, $(a, b, c)$, in the camera coordinate frame. Then, we align this plane with xy plane in the world coordinate frame, where we compute a transformation matrix that displaces the normal vector $(a, b, c)$ to the normalized vector $(0, 0, 1)$ along the z-axis of the world coordinate frame. This transformation matrix is used to transform point clouds in every frame so that the plane of the table always aligns with the xy plane of the world coordinate.

\paragraph{Object Localization at Test Time.} When we localize objects at test time, there could be some false positive segmentation of distracting objects. Such vision failures will prevent the robot policy from successfully executing actions. To exclude such false positive object segmentaiton, we use Segmentation Correspondence Model (SCM) from GROOT~\cite{zhu2023groot}, where SCM filters out the false positive segmentation of the objects by computing the affinity scores between masks using DINOv2 features. 

\paragraph{Global Registration.} In \orion{}, we use global registration to compute the transformation between observed object point clouds from videos and those from testing environments. We implement this part using a RANSAC-based registration function from Open3D~\cite{zhou2018open3d}. Specifically, given two object point clouds, we first compute their features using Fast-Point Feature Histograms (FPFH)~\cite{rusu2009fast}, and then perform a global RANSAC registration on the FPFH features of the point clouds~\cite{choi2015robust}. 

\paragraph{Implementation of SE(3) Optimization.} 
We parameterize each end-effector pose into a translation variable and a rotation variable and randomly initialize each variable using the normal distribution. We choose quaternions as the representation for rotation variables, and we normalize the randomly initialized vectors for rotation so that they remain unit quaternions. 
With such parameterization, we optimize the SE(3) end-effector trajectories over the object described in Equation~\ref{eq:orion:optim}. However, jointly optimizing both translation and rotation from scratch typically results in trivial solutions, where the rotation variables do not change much from the initialization due to the vanishing gradients. To avoid trivial solutions, we implement a two-stage process. In the first stage, we only optimize the rotation variables with 200 gradient steps. Then, the optimization proceeds to the second stage, where we optimize both the rotation and translation variables for another 200 gradient steps. In this case, we prevent the optimization process from getting stuck in trivial solutions for rotation variables. We implement the optimization process using Lietorch~\cite{teed2021tangent}. 

\subsection{System Setup}
\label{ablation_sec:orion:setup}
\paragraph{Details of Camera Observations.} As mentioned in Section~\ref{sec:orion:experiments}, we use an iPad with a TrueDepth camera for collecting human video demonstrations. We use an iOS app, Record3D, that allows us to access the depth images from the TrueDepth camera. We record RGB and depth image frames in sizes $1920\times 1080$ and $640\times 480$, respectively. To align the RGB images with the depth data, we resize the RGB frames to the size $640\times 480$. The app also automatically records the camera intrinsics of the iPhone camera so that the back-projection of point clouds is made possible.

To stream images at test time, we use an Intel Realsense D435i. In our robot experiments, we use RGB and depth images in the size $640\times 480$ or $1280\times720$ in varied scenarios, all covered in our evaluations. Evaluating on different image sizes showcases that our method is not tailored to specific camera configurations, supporting the wide applicability of constructed policy.

\paragraph{Real Robot Control.} In our evaluation, we reset the robot to a default joint position before object interaction every time. Then, we use a reaching primitive for the robot to reach the interaction points. Resetting to the default joint position enables an unoccluded observation of task-relevant objects at the start of each decision-making step. Note that the execution of object interaction does not necessarily require resetting. To command the robot to interact with objects, we convert the optimized SE(3) action sequence to a sequence of joint configurations using inverse kinematics and control the robot using joint impedance control. We use the implementation of Deoxys~\cite{zhu2022viola} for the joint impedance controller that operates at $500$ Hz. To avoid abrupt motion and make sure the actions are smooth, we further interpolate the joint sequence from the result of inverse kinematics. Specifically, we choose the interpolation so that the maximal displacement for each joint does not exceed $0.5$ radian between two adjacent waypoints.

\section{\okami{} Implementation Details}
\label{ablation_sec:okami:implementation}

\subsection{Additional Details of Tasks}
\label{ablation_sec:okami:tasks}

\paragraph{Success Conditions.} We describe the success conditions we use to evaluate if a task rollout is successful or not. 
\begin{itemize}
    \item \salt{}: The salt bottle reaches a position where the salt is poured out into the bowl.
    \item \toy{}: The plush toy is put inside the container, with more than 50\% of the toy inside the container.
    \item \laptop{}: The display is lowered towards the base until the two parts meet at the hinge (i.e., the laptop is closed).
    \item \drawer{}: The drawer is pushed back to the containing region, either it's a drawer or a layer of a cabinet. 
    \item \snacks{}: The snack is placed on top of the plate, with more than 50\% of the snack package on the plate.
    \item \bagging{}: The chip bag is put into the shopping bag, which is initially closed. 
\end{itemize}

\subsection{Human Reconstruction From Videos}
\label{supp:okami:recon}
\paragraph{Improved Models with SLAHMR and HaMeR.} 
For the 3D human reconstruction, we start by tracking the person in the video and getting an initial estimate of their 3D body pose using 4D Humans~\cite{goel2023humans}.
This body reconstruction cannot capture the hand pose details (i.e., the hands are flat).
Therefore, for each detection of the person in the video, we detect the two hands using ViTPose~\cite{xu2023vitpose++}, and for each hand, we apply HaMeR~\cite{pavlakos2024reconstructing} to get an estimate of the 3D hand pose.
However, the hands reconstructed by HaMeR can be inconsistent with the arms from the body reconstruction (e.g., different wrist orientation and location).
To address this, we apply an optimization refinement to make the body and the hands consistent in each frame, and encourage that the holistic body and hands motion is smooth over time.
This optimization is similar to SLAHMR~\cite{ye2023decoupling}, with the difference that besides the body pose and location of the SMPL+H model~\cite{romero2017embodied}, we also optimize the hand poses.
We initialize the procedure using the 3D body pose estimate from 4D Humans and the 3D hand poses from HaMeR.
Moreover, we use the 2D projection of the 3D hands predicted by HaMeR to constrain the projection of the 3D hand keypoints of the holistic model using a reprojection loss.
Finally, we can jointly optimize all the parameters (body location, body pose, hand poses) over the duration of the video, as described in SLAHMR~\cite{ye2023decoupling}.

Our modified SLAHMR incorporates the SMPL-H model~\cite{romero2017embodied} to include hand poses in the human motion reconstruction. We initialize hand poses in each frame using 3D hand estimates from HaMeR~\cite{pavlakos2024reconstructing}. The optimization process then jointly refines body locations, body poses, and hand poses over the video sequence. This joint optimization allows for accurate modeling of how hands interact with objects, which is crucial for manipulation tasks.

The optimization minimizes the error between the following two terms: 2D projections of the 3D joints from the SMPL-H model \textit{and} the detected 2D joint locations from the video. We use standard parameters and settings as described in SLAHMR~\cite{ye2023decoupling}, while using the SMPL-H model instead of the SMPL model.

\paragraph{Inference Requirements.} 
The model of human reconstruction is large and needs to be run on a computer with sufficient computation. Here we provide details about the runtime performance of the human reconstruction model. We use a desktop that comes with a GPU RTX3090 that has the size of the memory 24 GB. 
For a 10-second video with fps 30, it processes 10 minutes. \loosepar{}

\subsection{Prompts of Using GPT4V}
\label{supp:okami:prompt}
In order to use GPT4V in \okami{}, we need GPT4V's output to be in a typed format so that the rest of the programs can parse the result. Moreover, in order for the prompts to be general across a diverse set of tasks, our prompt does not leak any task information to the model. Here we describe the three different prompts in \okami{} for using GPT4V. 

\paragraph{Identify Task-relevant Objects. } We design the following prompt to invoke GPT4V so that \okami{} can identify the task-relevant objects from a provided human video: 

\begin{formal}
\textbf{Prompt:} You need to analyze what the human is doing in the images, then tell me:
1. All the objects in front scene (mostly on the table). You should ignore the background objects.
2. The objects of interest. They should be a subset of your answer to the first question. 
They are likely the objects manipulated by human or near human. Note that there are irrelevant objects in the scene, such as objects that does not move at all. You should ignore the irelevant objects.

Your output format is:

\begin{verbatim}
The human is xxx. 
All objects are xxx.
The objects of interest are:
```json
{
    "objects": ["OBJECT1", "OBJECT2", ...],
}
```
\end{verbatim}

Ensure the response can be parsed by Python `json.loads', e.g.: no trailing commas, no single quotes, etc. 
You should output the names of objects of interest in a list [``OBJECT1'', ``OBJECT2'', ...] that can be easily parsed by Python. The name is a string, e.g., ``apple'', ``pen'', ``keyboard'', etc.
\end{formal}

\paragraph{Identify Target Objects.} \okami{} uses the following prompt to identify the target object of each step in the reference plan.

\begin{formal}
\textbf{Prompt:}
The following images shows a manipulation motion, where the human is manipulating an object.

Your task is to determine which object is being manipulated in the images below. You need to choose from the following objects: \{a list of task-relevant objects\}.

Tips: the manipulated object is the object that the human is interacting with, such as picking up, moving, or pressing, and it is in contact with the human's \{the major moving arm in this step\} hand.

Your output format is:

\begin{verbatim}
```json
{
    "manipulate_object_name": "MANIPULATE_OBJECT_NAME",
}
```
\end{verbatim}

Ensure the response can be parsed by Python `json.loads', e.g.: no trailing commas, no single quotes, etc.
\end{formal}

\paragraph{Identify Reference Objects.} Here is the prompt that asks GPT4V to identify the reference object of each step in the reference plan:

\begin{formal}
\textbf{Prompt:}
The following images shows a manipulation motion, where the human is manipulating the object \{manipulate\_object\_name\}.

Please identify the reference object in the image below, which could be an object on which to place \{manipulate\_object\_name\}, or an object that \{manipulate\_object\_name\} is interacting with.
Note that there may not necessarily have an reference object, as sometimes human may just playing with the object itself, like throwing it, or spinning it around.
You need to first identify whether there is a reference object. If so, you need to output the reference object's name chosen from the following objects: \{a list of task-relevant objects\}.

Your output format is:

\begin{verbatim}
```json
{
    "reference_object_name": "REFERENCE_OBJECT_NAME" or "None",
}
```  
\end{verbatim}
Ensure the response can be parsed by Python `json.loads', e.g.: no trailing commas, no single quotes, etc.
\end{formal}

\subsection{Details on Factorized Process for Retargeting}
\label{appendix:okami:retarget}
\paragraph{Body Motion Retarget.} To retarget body motions from the SMPL-H representation to the humanoid, we extract the shoulder, elbow, and wrist poses from the SMPL-H models. We then use inverse kinematics to solve the body joints on the humanoid, ensuring they produce similar shoulder and elbow orientations and similar wrist poses. The inverse kinematics is implemented using an open-sourced library Pink~\cite{pink2024}. The IK weights we use for shoulder orientation, elbow orientation, wrist orientation, and wrist position are $0.04$, $0.04$, $0.08$, and $1.0$, respectively.

\paragraph{Hand Pose Mapping.} 
As we describe in the method section, we first retarget the hands from SMPL-H models to the humanoid's dexterous hands using a hybrid implementation of inverse kinematics and angle mapping. Here are the details of how this mapping is performed. Once we obtain the SMPL-H models from a video demonstration, we can obtain the locations of 3D joints from the hand mesh models based on SMPL-H representation. Subsequently, we can compute the rotating angles of each joint that correspond to certain hand poses. Then, we apply the computed joint angles to the hand meshes of a canonical SMPL-H model, which is pre-defined to have the same size as the humanoid robot hardware. From this canonical SMPL-H model, we can get the 3D keypoints of hand joints and use an off-the-shelf optimization package, dex-retarget, to directly compute the hand joint angles of the robot~\cite{qin2023anyteleop}.

\paragraph{Inverse Kinematics. } After warping the arm trajectory, we use inverse kinematics to compute the robot's joint configurations. We assign weights of $1.0$ to hand position and $0.08$ to hand rotation, prioritizing accurate hand placement while allowing the arms to maintain natural postures.

For retargeting human hand poses to the robot, we map the human hand joint angles to the corresponding joints in the robot's hand. This mapping enables the robot to replicate fine-grained manipulations demonstrated by the human, such as grasping and object interaction. Our implementation ensures that the retargeted motions are physically feasible for the robot and that overall execution appears natural and effective for the task at hand.

\subsection{Additional Details of Plan Generation} \label{appendix:okami:plan-generation}
For temporal segmentation, we sample keypoints from the segmented objects in the first frame and track them across the video using CoTracker~\cite{karaev2023cotracker}. We compute the average velocity of these keypoints at each frame and apply an unsupervised changepoint detection algorithm~\cite{killick2012optimal} to detect significant changes in motion, identifying keyframes that correspond to subgoal states.

To determine contact between objects, we compute the relative spatial locations and distances between the point clouds of objects. If the distance between objects falls below a predefined threshold, we consider them to be in contact. For non-contact relations that are difficult to infer geometrically---such as a cup in a pouring task---we use GPT4V to predict semantic relations based on the visual context. GPT4V can infer that the cup is the reference object in a pouring action, even without direct contact.

\subsection{Trajectory Warping}
\label{appendix:okami:warping}

Here, we mathematically describe the process of trajectory warping. We denote the trajectory for robot as $\tau^{\text{robot}}$ retargeted from $\tau_{t_i:t_{i+1}}^{\text{SMPL}}$ in the generated plan. We also denote the starting point and end point of $\tau^{\text{robot}}$ as $p_{\text{start}}$, $p_{\text{end}}$, respectively. Note that all points along the trajectory are represented in SE(3) space.

Each point $p_t$ on the original retargetd trajectory can be described by the following function:
\begin{equation}
    p_t = p_{\text{start}} + (\tau^{\text{robot}}(t) - p_{\text{start}})
    \label{eq:okami:point-warping}
\end{equation}
where $t\in \{t_i, \dots, t_{i+1}\}$, $\tau^{\text{robot}}(t_i)=p_{\text{start}}$, $\tau^{\text{robot}}(t_{i+1})=p_{\text{end}}$.

When warping the trajectory, we either only need to adapt the trajectory to the new target object location, or adapt the trajectory to the new locations of both the target and the reference objects, as described in Section~\ref{sec:okami:oar}. Without loss of generality, we denote the SE(3) transformation for the starting point is $\tcp_{\text{start}}$, and the SE(3) transformation for the ending point is $\tcp_{\text{end}}$. Now, the warped trajectory can be described by the following function:
\begin{equation}
    p_t = \tcp_{\text{start}} \cdot p_{\text{start}} + (\hat{\tau}^{\text{robot}}(t) - \tcp_{start}\cdot p_{\text{start}})
\end{equation}
where $\hat{\tau}^{\text{robot}}(t)=\frac{\tau^{\text{robot}}(t)-p_{\text{start}}}{p_{\text{end}} - p_{\text{start}}}(
\tcp_{\text{end}}\cdot p_{\text{end}} - \tcp_{\text{start}}\cdot p_{\text{start}}) + \tcp_{\text{start}}\cdot p_{\text{start}}$, $\forall t\in \{t_i, \dots, t_{i+1}\}$. In this way, we have $\hat{\tau}^{\text{robot}}(t_i)=\tcp_{\text{start}}\cdot p_{\text{start}}$, $\hat{\tau}^{\text{robot}}(t_{i+1})=\tcp_{\text{end}}\cdot p_{\text{end}}$. 
Note that this trajectory warping assumes the end point of a trajectory is not the same as the starting point, which is a common assumption for most of the manipulation behaviors.

\subsection{Additional Details of Simulation}
\label{ablation_sec:okami:simulation}

\begin{figure}[ht]
    \centering
    \includegraphics[width=\linewidth]{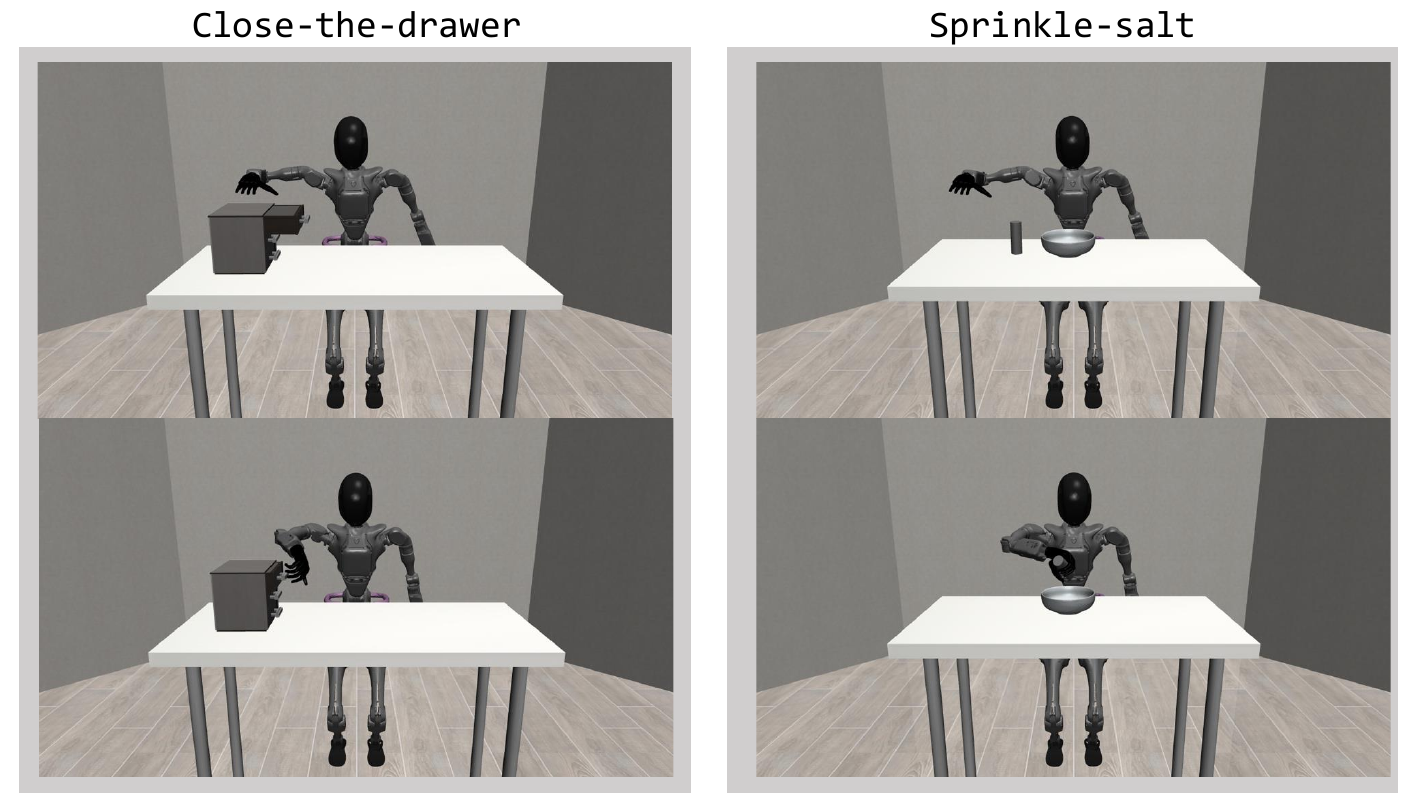}
    \caption[Visualization of simulation tasks.]{The screenshots of the starting and ending frames of the two simulation tasks, \drawer{} and \salt{}. }
    \label{fig:okami:simulation}
\end{figure}

For easy reproducibility, we create \salt{} and \drawer{} in simulation that replicate their real-world counterparts. The simulation tasks are visualized in Figure~\ref{fig:okami:simulation}. We implement these tasks using Robosuite~\cite{zhu2020robosuite}, with the option for robot embodiment as ``GR1FixedLowerBody.''

Note that for the policy of each task, we use the same human video as the ones used in real robot experiments. We compare three methods in simulation, namely \okami{}(w/vision), \okami{}(w/o vision), and \orion{}.~\okami{}(w/vision) is the same method we use in our real robot experiments. \okami{} (w/o vision) is the simplified version of \okami{} where we assume the model directly gets the ground-truth poses of objects. The evaluation results are shown in Table~\ref{tab:okami:sim_results}, where each reported number is the success rate averaged over $50$ rollouts.

\begin{table}[h]
    \centering
    \begin{tabular}{ccc}
    \hline
        Method & \salt{} & \drawer{} \\
    \hline
        \okami{} (w/ vision) & 82 & 84 \\
        \okami{} (w/o vision) & 100 & 100\\
        \orion{} & 0 & 10 \\
    \hline
    \end{tabular}
    \caption[Evaluation of \okami{} policies in simulation.]{The average success rates (\%) of different methods over two simulated tasks, \salt{} and \drawer{}.}
    \label{tab:okami:sim_results}
\end{table}

We notice that the simulation results are generally better than the real robot experiments. The performance difference comes from the easy physical interaction between dexterous hands and objects compared to the real robot hardware. Also, \okami{} without vision can achieve a much higher success rate than \okami{} with vision because the noise and uncertainty of perception are abstracted away. Specifically, a large portion of uncertainties come from the partial observation of object point clouds and the estimation of the object location that deviates from the ground-truth locations. The success of \okami{} highly depends on the quality of trajectory warping, which is dependent on the correct estimation of object locations. This simulation result also indicates that the performance of \okami{} is expected to improve if more powerful vision models with higher accuracy are available.

\newpage

\chapter{Additional Details of Part~\ref{part:III}}
\label{chapter:appendix_chapter_III}
\section{\buds{} Implementation Details}
\label{sec:buds:implementation}

\subsection{Multi-sensory Fusion}
Figure 2 of the cited work~\cite{lee2020making} shows a detailed diagram of the fusion process. The fused representation is optimized over an adapted ELBO (evidence lower bound) loss from Equations (3) and (4) from Lee et al.~\cite{lee2020making}. A major difference is that we optimize the sensory fusion model over reconstructing the current states while Lee et al. optimized over reconstructing the next states. This different design choice comes from the fact that their work has focused on using the latent representation as inputs for policy learning which needs to encode future state information, while we focus on learning the statistical patterns of multi-sensory data at the current state.

 We use the same convolutional network structures as the models from \citet{lee2020making}, except that we do not have skip connections from the encoder to the decoder parts.  We choose $32$ for The latent dimension of the fused representation. For training the fusion model for each task, we use $1000$ training epochs, with a learning rate of $0.001$ and a batch size of $128$.

\subsection{Skill Segmentation}

\paragraph{Hyperparameters.} We present the hyperparameters for the unsupervised clustering step. The maximum number of clusters for each task is $6$ for \tooluse{} and \hammer{}, $8$ for \kitchen{} and \budsrealrobot{}, and $10$ for \multitask{}. The breadth-first search stops when the number of mid-level temporal segments exceeds twice the maximum number of clusters. We also use a minimum length threshold to reject a small cluster. The threshold we set for each task is: $30$ for \tooluse{}, \budsrealrobot{}, $35$ for \hammer{}, $20$ for \multitask{}. In \buds{}, these values of hyperparameters are tuned heuristically, and how to extend to an end-to-end method is one future direction. 

\subsection{Sensorimotor Policies}
\paragraph{Implementation Details.}

\buds{} focuses on learning closed-loop sensorimotor skills, taking observations from robot sensors and the latent subgoal vector $\param{t}$ as input. The observations include two RGB images ($128 \times 128$) from the workspace camera and eye-in-hand camera, along with proprioception of joint and gripper states. For visual input processing, we employ ResNet-18~\cite{he2016deep} as the visual encoder, followed by Spatial Softmax~\cite{finn2016deep} to extract feature map keypoints. These keypoints are concatenated with proprioception data (joint angles and past five frames of gripper states~\cite{zhang2018deep}), and the resulting vectors pass through fully connected layers with LeakyReLU activation to output end-effector motor commands. A subgoal encoder $\text{Encoder}_k$, which is a ResNet-18 module with spatial softmax, processes only the workspace camera image of the subgoal state from demonstration data. Similarly, the meta controller $\metacontroller{}$ processes the current workspace camera image using a ResNet-18 module. All ResNet-18 modules in our framework use a modified architecture with the last two layers removed, producing $4\times4$ feature maps. For robot control, we implement a position-based Operational Space Controller (OSC)~\cite{khatib1987unified} with a binary controller for the parallel-jaw gripper, operating at $20$ Hz. During evaluation, the meta controller runs at $4$ Hz while sensorimotor skills maintain $20$ Hz operation.

We choose the dimension for subgoal vector $\omega_t$ to be $32$, and the number of 2D keypoints from the output of Spatial Softmax layer to be $64$. We choose $\subgoaltime=30$ for all single-task environments (Both simulation and real robots). We choose $\subgoaltime=20$ for the multitask environment  \multitask{}. We have different parameters for these variables because skills are relatively short in each task in \multitask{} domain compared to all single-task environments. 

\paragraph{Training Details.}
 To increase the generalization ability of the model, we apply data augmentation~\cite{kostrikov2020image} to images for both training skills and meta controllers. To further increase the robustness of policies $\skillpolicy{k}$, we also add some noise from Gaussian distribution with a standard deviation of $0.1$. 
 
 For all skills, we train for $2001$ epochs with a learning rate of $0.0001$, and the loss function we use is $\ell_{2}$ loss. We use two layers ($300$, $400$ hidden units for each layer) for the fully connected layers in all sing-task environments, while three layers ($300$, $300$, $400$) hidden units for each layer for fully connected layers in \multitask{} domain. For meta controllers, we train $1001$ epochs in all simulated single-task environments, $2001$ epochs in \multitask{}, and $3001$ epochs in \budsrealrobot{}. For KL coefficients during cVAE training, we choose $0.005$ for \tooluse{}, \hammer{}, and $0.01$ for all other environments. 
 
The training of cVAE follows the convention as in prior work~\cite{mandlekar2020iris, mandlekar2020learning}, which minimizes an ELBO (evidence lower bound loss) on training data. To obtain the training data for predicting latent subgoal vectors and skill indices, we augmented the demonstrations from the clustering and skill learning step: 1) The training labels of skill indices came from our clustering step. 2) The latent subgoal vectors were computed per state, and the encoders for computing the vectors were jointly trained with skill policies.

\paragraph{Impact of different clustering algorithms.} To evaluate different skill classification methods, we compare K-means and spectral clustering algorithms using the \kitchen{} task as our testbed. Table~\ref{tab:buds:clustering-comparison} shows the task success rates when applying these two clustering approaches while keeping all other \buds{} algorithms unchanged. Our results indicate that switching from spectral clustering to K-means yields comparable performance, though the spectral clustering variant demonstrates marginally better results. 

\begin{table}
\centering
  \begin{tabular}{lccc}
    \toprule
    \textbf{} & Spectral Clustering & K-means Clustering \\
    \midrule
    \kitchen~ & $72.0\pm4.0$ & $70.6\pm3.7$\\
    \bottomrule
  \end{tabular}
\caption[Comparison of different clustering algorithms.]{Success rates (\%) in \kitchen{} with different clustering algorithms. }
  \label{tab:buds:clustering-comparison}
\end{table}

\subsection{Additional Details of Tasks}
\label{ablation_sec:buds:tasks}
\paragraph{Task Descriptions} We describe the execution stages that the robot needs to accomplish before reaching task goals in each environment.

\begin{figure}[ht!]
     \includegraphics[width=\linewidth]{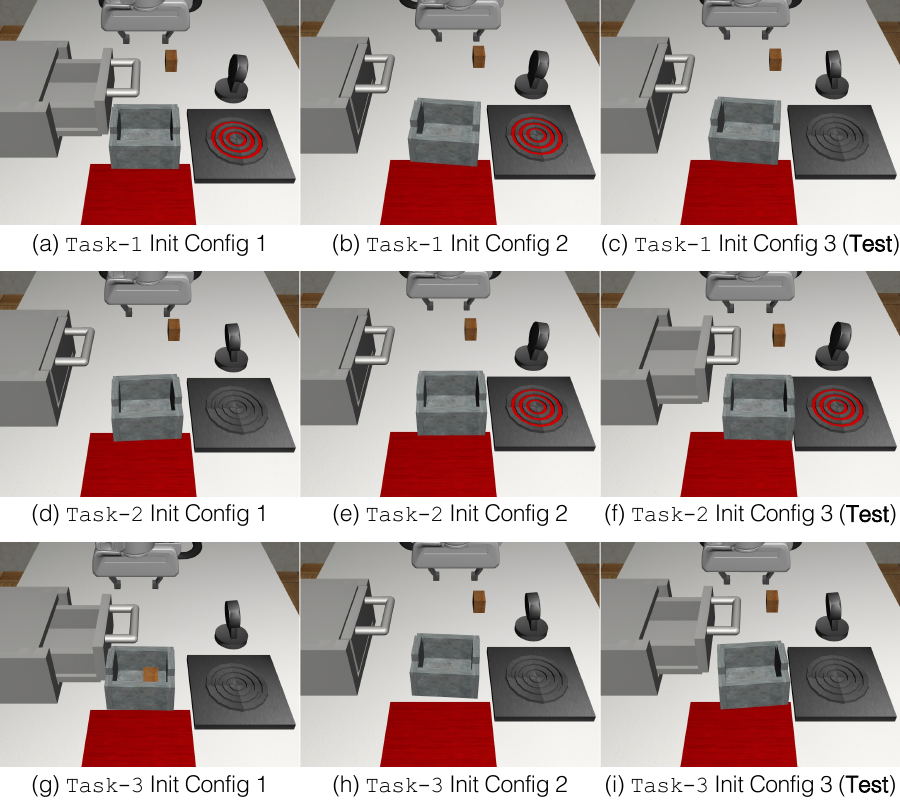}
     \vspace{-1cm}
    \caption[Task configurations for multitask evaluation in \buds{}.]{Screenshots of example initial configurations for three tasks in \multitask{}. Each row corresponds to a task, and left two figures in each row represent initial configurations are covered in \budstrainmulti{}, \budstrainsingle{}. Every set of initial configurations shown in the right figure of each row is covered in \budstest{}. (a): The stove is on, the object and the pot are on the table, the drawer is not closed and blocked the arm; (b): The stove is on, the object and the pot on the table, the drawer is closed; (c): The stove is off in the beginning, the object and the pot are on the table; (d): The stove is off, the drawer is closed, the object and the pot are on the table, the drawer is closed; (e): The stove is on, the drawer is closed, the object and the pot are on the table; (f): The stove is on but the drawer is not closed, the object and the pot are on the table; (g): The drawer is open, the stove is off,  the pot is on the table and the object is already in the pot; (h): The drawer is closed, the stove is off, the object and the pot are on the table; (i): The drawer is open, the stove is off, the object and the pot are both on the table. }
    \label{fig:buds:multitask-configs}
\end{figure}

\begin{itemize}
\item \textbf{\tooluse{} Task:} The task goal is to put the cube into the metal pot. The robot needs to use the tool to fetch a cube, which is not reachable for the robot because the configuration for directly picking it up is almost at the singularity of the robot's configuration. After fetching, it needs to put the tool aside and put the cube into the pot.

\item \textbf{\hammer{} Task:} The task goal is to put the hammer in the drawer and close the drawer. To achieve this goal, the robot needs to open up the drawer, place the hammer into the drawer, and close the drawer. The hammer is small and hard to grasp stably. 

\item \textbf{\kitchen{} Task:} The robot's goal is to cook and serve a simple dish in the serving region and turn off the stove. For this goal, the robot has to complete multiple steps in a sequence: turn on the stove, place the pot on the stove, put the ingredient into the pot, put the pot on the table, push it to the serving region (red region on the table), and turn off the stove at the end. This contains more execution stages of contact-rich motion than \tooluse{} and \hammer{} do.

\item \textbf{\multitask{} Domain:} We design three tasks in this domain, each with a different manipulation goal. The manipulation goals are 1) \task{1}: The drawer is closed, the stove is turned on, the object is in the pot and the pot is placed on the stove; 2) \task{2}: The drawer is closed, the stove is turned on, the object is in the pot and the pot is in the serving region. 3) \task{3}: The drawer is closed, the stove cannot be turned on, the object is in the pot and the pot is in the serving region. And there are three sets of initial configurations for each task, which we visualize in Figure~\ref{fig:buds:multitask-configs}. We collect $120$ demonstrations for each task. We collect the same number of demonstrations for all different sets of initial configurations. The demonstration sequences of tasks for \budstest{} present two characteristics. 1) For \task{1} and \task{2}, the sequences start from initial configurations in the other task of training data, thus having intersections in execution stages; 2) For \task{3}, the sequences are longer than any of the sequences for the task during training, so the approach needs to compose skills for longer execution stages. As we have seen from Table~\ref{tab:buds:multitask}, the testing performance in \task{3} is lower than \task{1} and \task{2} due to this more challenging characteristic than the other two tasks.

\item \textbf{\budsrealrobot:} The robot needs to take away the lid of the pot, place the pot to the plate, pick up the tool, use the tool to push the pot along with the plate to the margin of the table, and place down the tool in the the end. This task covers versatile motions of prehensile grasping motion, non-prehensile pushing motion, and tool using on the real hardware. 
For capturing visual images, we use Kinect Azure as the workspace camera, and Intel Realsense D435i as the eye-in-hand camera. We capture RGB images from the two cameras, and then scale images down to $128\times128$ as input to our models.

\end{itemize}

\bibliographystyle{unsrtnat} % Defines bibliography style to unsrtnat for unsorted order
\bibliography{references} % Points to the file named references.bib

\end{document}